\documentclass[sigconf,screen]{acmart}
\renewcommand\footnotetextcopyrightpermission[1]{}
\setcopyright{none}
\acmConference[Information and Software Technology]{}{IST Journal 2022}{Elsevier}

\usepackage{pbox}
\usepackage{fancybox}
\usepackage{graphicx}
\usepackage{float}
\usepackage{listings}
\usepackage{url} 
\usepackage{adjustbox}
\usepackage{multirow}
\usepackage{xspace}
\usepackage{comment}
\usepackage{xcolor}
\usepackage{microtype}
\usepackage{balance}
\usepackage{paralist}
\usepackage{fancyvrb}
\usepackage{algorithmic}
\usepackage{algorithm}
\usepackage{amsmath}
\usepackage{cleveref}
\usepackage{soul, color}
\usepackage[normalem]{ulem}

\usepackage[shortlabels]{enumitem}
\usepackage[labelfont=bf,textfont=md]{caption}
\usepackage{balance}

\usepackage{makecell}

\usepackage{multicol}
\usepackage{graphicx}
\usepackage{subcaption}

\usepackage{geometry}
\usepackage{pdflscape}

\usepackage[T1]{fontenc}

\usepackage{array}
\newcolumntype{H}{>{\setbox0=\hbox\bgroup}c<{\egroup}@{}}

\usepackage{listings}
\lstdefinestyle{customjava}{
  belowcaptionskip=1\baselineskip,
  xleftmargin=\parindent,
  language=java,
  showstringspaces=false,
  morecomment=[s][\color{gray}]{@}{\^^M},
  keywordstyle=\bfseries\color{green!40!black},
  commentstyle=\itshape\color{purple!40!black},
  identifierstyle=\color{black},
  stringstyle=\color{orange},
}

\usepackage{color}
\definecolor{deepblue}{rgb}{0,0,0.5}
\definecolor{deepred}{rgb}{0.6,0,0}
\definecolor{deepgreen}{rgb}{0,0.5,0}
\definecolor{lightgreen}{rgb}{0.2,0.7,0.4}

\lstdefinestyle{custompython}{
  language=Python,
  morekeywords={self}, 
  keywordstyle=\color{deepblue},
  stringstyle=\color{deepgreen},
  commentstyle=\color{lightgreen},
  showstringspaces=false
}

\usepackage{array}
\newcolumntype{L}[1]{>{\raggedright\let\newline\\\arraybackslash\hspace{0pt}}m{#1}}
\newcolumntype{C}[1]{>{\centering\let\newline\\\arraybackslash\hspace{0pt}}m{#1}}
\newcolumntype{R}[1]{>{\raggedleft\let\newline\\\arraybackslash\hspace{0pt}}m{#1}}

\title{Memorization and Generalization in Neural Code Intelligence Models}

\author{Md Rafiqul Islam Rabin}
\email{mrabin@uh.edu}
\affiliation{%
  \institution{University of Houston}
  \city{Houston}
  \state{TX}
  \country{USA}
}

\author{Aftab Hussain}
\email{ahussain27@uh.edu}
\affiliation{%
  \institution{University of Houston}
  \city{Houston}
  \state{TX}
  \country{USA}
}

\author{Mohammad Amin Alipour}
\email{maalipou@central.uh.edu}
\affiliation{%
  \institution{University of Houston}
  \city{Houston}
  \state{TX}
  \country{USA}
}

\author{Vincent J. Hellendoorn}
\email{vhellendoorn@cmu.edu}
\affiliation{%
  \institution{Carnegie Mellon University}
  \city{Pittsburgh}
  \state{PA}
  \country{USA}
}


\begin{document}

\newcommand{\Fix}[1]{\textbf{\textcolor{red}{Fix}: #1}}
\newcommand{\TODO}[1]{\textcolor{red}{\textbf{TODO}: #1}}
\newcommand{\Ans}[1]{\textbf{\textcolor{blue}{Answer}: #1}}
\newcommand{\Part}[1]{\noindent\textbf{#1}}
\newcommand{\fsize}[2]{{\fontsize{#1}{0}\selectfont#2}}
\newcommand{\crossmark}{$\times$}
\newcommand{\Space}[1]{}

\newcommand{\eg}{\textit{e.g.}\xspace}
\newcommand{\ie}{\textit{i.e.}\xspace}
\newcommand{\etal}{\emph{et al.}\xspace}
\newcommand{\resp}{\emph{resp.}\xspace}

\newcommand{\Fone}{\textsc{$F_1$-Score}\xspace}

\cornersize{.2}
\newcounter{observation}
\newcommand{\observation}[1]{\refstepcounter{observation}
        \begin{center}
        \vspace{2pt}
        \Ovalbox{
        \begin{minipage}{0.9\columnwidth}
            \textbf{Observation \arabic{observation}:} #1
        \end{minipage}
        }
        \end{center}
}

\newcommand{\MT}{\ensuremath{t}}
\newcommand{\OT}{\ensuremath{t_o}}
\newcommand{\RT}{\ensuremath{t_r}}

\newcommand{\CC}{\textsc{CodeClassification}\xspace}
\newcommand{\vm}{\textsc{VarMisuse}\xspace}
\newcommand{\mnp}{\textsc{MethodName}\xspace}
\newcommand{\csearch}{\textsc{CodeSearch}\xspace}
\newcommand{\cnl}{\textsc{Code-to-Text}\xspace}

\newcommand{\JS}{\textsc{Java-Small}\xspace}
\newcommand{\JM}{\textsc{Java-Med}\xspace}
\newcommand{\JL}{\textsc{Java-Large}\xspace}
\newcommand{\PY}{\textsc{Py150-Great}\xspace}
\newcommand{\SA}{\textsc{Java-SA}\xspace}
\newcommand{\JTT}{\textsc{Java-Top10}\xspace}
\newcommand{\Ruby}{\textsc{Ruby}\xspace}

\newcommand{\ctv}{\textsc{Code2Vec}\xspace}
\newcommand{\cts}{\textsc{Code2Seq}\xspace}
\newcommand{\rnn}{\textsc{RNN}\xspace}
\newcommand{\tra}{Transformer\xspace}
\newcommand{\ggnn}{\textsc{GGNN}\xspace}
\newcommand{\great}{\textsc{Great}\xspace}
\newcommand{\CodeBERT}{Code\textsc{BERT}\xspace}
\newcommand{\CodeXGLUE}{Code\textsc{XGLUE}\xspace}
\newcommand{\CodeSearchNet}{CodeSearchNet\xspace}

\newcounter{magicrownumbers}
\newcommand\rnum{\stepcounter{magicrownumbers}\arabic{magicrownumbers}}

    \lstnewenvironment{CODE}[1][]
      {\lstset{language=[LaTeX]TeX}\lstset{escapeinside={(*@}{@*)},
       numbers=left,numberstyle=\normalsize,stepnumber=1,numbersep=5pt,
       breaklines=true,
           framesep=5pt,
           basicstyle=\normalsize\ttfamily,
           showstringspaces=false,
           keywordstyle=\itshape\color{blue},
           stringstyle=\color{maroon},
        commentstyle=\color{black},
        rulecolor=\color{black},
        xleftmargin=0pt,
        xrightmargin=0pt,
        aboveskip=\medskipamount,
        belowskip=\medskipamount,
               backgroundcolor=\color{white}, #1
    }}
    {}

\newcounter{defn}
\newcommand{\defn}[2]{\refstepcounter{defn}
	\begin{quote}
	\textbf{#1:} #2
	\end{quote}

}

\newcommand{\FOneScore}{F$_{1}$-Score\xspace}

\newcommand{\OI}{\ensuremath{p_o}\xspace}
\newcommand{\RI}{\ensuremath{p_r}\xspace}
\newcommand{\M}{\ensuremath{\mathcal{M}}\xspace}
\newcommand{\MI}{\ensuremath{p}\xspace}

\newcommand{\ciinput}{input program\xspace}
\newcommand{\ciinputs}{input programs\xspace}

\newcommand{\Pred}[2]{\ensuremath{Prediction(#1, #2)}}
\newcommand{\Size}[1]{\ensuremath{|#1|}}
\newcommand{\SizeToken}[1]{\ensuremath{|#1|}$_{token}$}
\newcommand{\SizeChar}[1]{\ensuremath{|#1|}$_{char}$}
\newcommand{\DD}{delta debugging\xspace}
\newcommand{\ddmin}{\texttt{ddmin}}

\newcommand{\CI}{Neural Code Intelligence Model\xspace}
\newcommand{\CIs}{Neural Code Intelligence Models\xspace}
\newcommand{\ci}{neural code intelligence model\xspace}
\newcommand{\cis}{neural code intelligence models\xspace}

\newcommand{\ArpitEtAl}{Arpit et al.~\cite{arpit2017closer}\xspace}
\newcommand{\ZhangEtAl}{Zhang et al.~\cite{zhang2017rethinkgen}\xspace}

\newcommand{\hlx}[1]{\colorbox{yellow}{\parbox{\textwidth}{#1}}}
\newcommand{\hly}[1]{\colorbox{yellow}{#1}}

\newcommand{\footreff}[1]{\textsuperscript{\ref{#1}}}
\newcommand{\ngram}{n\text{-}gram}

\newcommand{\txtred}[1]{$\textbf{\textcolor{red}{#1}}$}
\newcommand{\txtblue}[1]{$\textbf{\textcolor{blue}{#1}}$}

\begin{abstract}
\textbf{Context}: Deep Neural Networks (DNNs) are increasingly being used in software engineering and code intelligence tasks. These are powerful tools that are capable of learning highly \emph{generalizable} patterns from large datasets through millions of parameters. At the same time, their large capacity can render them prone to \emph{memorizing} data points. Recent work suggests that the memorization risk manifests especially strongly when the training dataset is noisy, involving many ambiguous or questionable samples, and memorization is the only recourse. \\
\textbf{Objective}: The goal of this paper is to evaluate and compare the extent of memorization and generalization in neural code intelligence models. It aims to provide insights on how memorization may impact the learning behavior of neural models in code intelligence systems. \\
\textbf{Method}: To observe the extent of memorization in models, we add random noise to the original training dataset and use various metrics to quantify the impact of noise on various aspects of training and testing. We evaluate several state-of-the-art neural code intelligence models and benchmarks based on Java, Python, and Ruby codebases. \\
\textbf{Results}: Our results highlight important risks: millions of trainable parameters allow the neural networks to memorize \emph{anything}, including noisy data, and provide a false sense of generalization. We observed all models manifest some forms of memorization. This can be potentially troublesome in most code intelligence tasks where they rely on rather noise-prone and repetitive data sources, such as code from GitHub. \\
\textbf{Conclusion}: To the best of our knowledge, we provide the first study to quantify memorization effects in the domain of software engineering and code intelligence systems. This work raises awareness and provides new insights into important issues of training neural models in code intelligence systems that are usually overlooked by software engineering researchers.    

\end{abstract}

\maketitle

\section{Introduction}
\label{sec:intro}
Data-driven program analysis approaches have been increasingly used in software engineering tasks such as bug detection \cite{dinella2020hoppity,vasic2019neural}, method name prediction \cite{allamanis2015suggesting, allamanis2016summarization}, code comment generation \cite{iyer2016summarizing,feng2020codebert} and many more \cite{white2015toward, gu2016deep, allamanis2018ggnn}.
These approaches predominantly use deep neural networks to extract useful patterns and insights about programs from a large corpus of code.
Neural networks thrive at this because they are high-capacity universal approximators, capable of expressing any hypothesis class through their layered architectures~\cite{lecun1998gradient}.
Yet, while they can match or even exceed humans' performance in tasks ranging from board games~\cite{gibney2016AlphaGo} to image recognition~\cite{krizhevsky2017imagenet}, it is notoriously unclear what insights they extract from training data. Their tremendous capacity -- now spanning many billions of trainable parameters -- allows neural networks to both learn many generalizable patterns and simply memorize myriad training samples \cite{arpit2017closer,zhang2017rethinkgen}.

Memorization is a significant, and non-obvious threat to training deep learners, perhaps especially so for models trained on software engineering data. Source code from the open-source ecosystem is exceptionally repetitive \cite{hindle2012naturalness,gabel2010study,casalnuovo2019studying}, as well as particularly noisy \cite{kim2011noise, raychev2016noisy, munaiah2017curating}. Both factors encourage memorization, which is directly adverse to the ability of neural models to generalize \cite{arpit2017closer}. Indeed, deep learners are easily led astray by factors like code duplication \cite{allamanis2019duplication}, and often biased towards superficial features such as common variable names \cite{compton2020obfuscation, yefet2020adversarial}. As a consequence, they are vulnerable to adversarial examples in which a small, often semantics-preserving, transformation radically changes the models' predictions \cite{rabin2021generalizability, yefet2020adversarial, rabin2019tnpa}.

In this work, we perform a large-scale study of memorization and generalization in training \cis through a large-scale study that follows the framework of pioneering work in computer vision by \ArpitEtAl and \ZhangEtAl.
To do so, we introduce noise in programming datasets (\ie, \Cref{fig:noise_example}) where we add various degrees of noise into popular existing datasets by randomly altering target labels or input programs, and then observe the impact on a range of characteristics of training (\eg, the spread of loss values).
By studying the resulting trends broadly, across six types of models, three programming languages, six datasets, and five noise levels each, we can gain insight into the impact of memorization artifacts across datasets and models uniquely proposed in software engineering and code intelligence systems.
We contribute the following observations: \\
\textbf{1. Memorization is a significant concern in software engineering:} models with excessive parameter capacity easily memorize large datasets. When trained with real versus nonsensical labels (the latter necessitating memorization), they yielded training curves that were virtually indistinguishable.\\
\emph{+ A positive:} the use of noise as a contrast to clean data helped us distinguish memorization from generalization even when test (or held-out) performance did not, providing a promising new indicator for choosing the right parameter budget when modeling source code. \\
\textbf{2. Noise-driven memorization inhibits confident learning:} even small amounts of noise-induced memorization substantially altered the distribution of scores assigned by models, towards more uniform probabilities and conservative predictions, which undermines their usefulness for ranking and high-precision settings. \\
\emph{+ A positive:} our metrics consistently distinguished between generalization and memorization-prone training, supporting the use of this analysis to detect memorization-related problems early. \\
\textbf{3. Training curves do not betray the degree of noise:} all models traced similarly shaped loss distributions during training, regardless of memorization. As such, we cannot tell how much memorization is involved even in learning the original datasets. Previous work suggests that this is a non-trivial amount \cite{arpit2017closer, zhang2017rethinkgen}. Quantifying this is a key challenge: our work reinforces that even small amounts of forced memorization can have a major impact on model accuracy. \\
\emph{+ A positive:} across multiple models and datasets, a \emph{bigger increase in the spread of loss} over the course of training correlated well with both reduced noise and better performing models. This may be a fruitful metric for gauging model and dataset generalization potential.

To the best of our knowledge, this work is the first to empirically study memorization and generalization phenomena of \cis. Our findings provide quantitative evidence of memorization and a rich suite of new metrics to assess the generalization potential of models and datasets used in software engineering research.

\smallskip
\textit{\textbf{Artifacts.}} 
The evaluation scripts and detailed results will be publicly available at \url{https://github.com/UH-SERG/CI-Memorization}.

\section{Background}
\label{sec:background}
High capacity neural networks, with many millions or even billions of trainable parameters, are capable of memorizing large volumes of high-dimensional data. They are apparently more prone to doing so when finding generalizable patterns in the data is challenging \cite{arpit2017closer}. This problem commonly occurs when the training data contains excessive repetition or noise, and causes a discrepancy between performance while training and at inference time.
Previous studies have shown that datasets of code are extremely repetitive. \citet{lopes2017duplicates} found that $70\%$ of source code files on the GitHub had clones of previously created files. \citet{gabel2010study} found that code fragments, especially large ones, are repetitive and likely to reoccur in other programs.
\citet{allamanis2019duplication} showed that almost all datasets used to train \cis contain levels of code duplication of $20\%$ or more, and this repetition spuriously inflated their performance.
\citet{rabin2021dd, rabin2022perses} show that models often take shortcuts and heavily rely on superficial features for making predictions.
Several studies have shown that \cis are vulnerable to small, often semantic-preserving, transformation \cite{wang2019coset, rabin2019tnpa, rabin2020evaluation}, suffer from generalizability and robustness \cite{kang2019generalizability, rabin2021generalizability, yefet2020adversarial}, and may rely on few tokens \cite{suneja2021probing, rabin2021dd, rabin2022perses} or merely structures \cite{rabin2020demystifying, rabin2021code2snapshot}.
A few works also survey the taxonomy of existing models or methods for source code \cite{allamanis2018survey, sharma2021survey} and provide a comprehensive review to categorize, investigate, and recommend on applications and challenges \cite{le2020deep, zhang2022testing}.

A large body of work in ML literature has tried to evaluate the tension between memorization and generalization.
\citet{zhang2017rethinkgen} was perhaps the first study that showed that random data can be fit perfectly with deep neural networks. They introduced noise into the training dataset of an image classification task and observed that the neural model could easily learn the noisy data.
The results underscored that DNNs have the potential to employ a high degree of memorization in learning datasets. For an Alexnet-style CNN on CIFAR10 dataset, they found that the difference between the relative convergence times to fit training data without label corruption and to do the same with data with the maximum label corruption is quite small.
This observation was corroborated by similar experiments by \citet{arpit2017closer} on the CIFAR10 and MNIST image datasets; who found that models achieved optimal training accuracy within just about 50 epochs of training, for any amount of noise. However, each model's initial training progress did correspond to how clean the dataset is.

\citet{zhang2017rethinkgen}'s observations ignited intense empirical and theoretical research in evaluation and characterization of memorization in neural networks.
\citet{morcos2018similarity} showed that neural networks that generalize from the same data tend to converge to similar representations, whereas networks that memorize do not.
\citet{recht2019imagenet} evaluated the generalization of ImageNet networks on new test data and found that the models do not generalize reliably. They found that, in addition to overfitting, preprocessing mechanisms for data cleaning can pose challenges in downstream training.
\citet{chen2019understanding} further quantified the generalization performance of deep neural networks with noisy labels. They provided an iterative noisy cross-validation approach for identifying correct labels from noisy datasets and adopted a co-teaching strategy to train robust models against noisy labels.
\citet{hacohen2020order} observed that different neural models memorize data in different orders. In contrast, when training a DNN with real data, different DNNs with similar architectures \textit{learn} the data in the same order. They observed this behavior on several image classification benchmarks and one text classification benchmark.
\citet{zhang2020identity} studied the interplay between memorization and generalization in deep neural networks by focusing on an identity-mapping task. They demonstrated that over-parameterized neural networks such as deep CNNs often memorize training labels by learning a constant function, while shallow neural networks like single-layer linear networks fail to either generalize the identity function or memorize a constant function.
\citet{northcutt2021pervasive} identified erroneously labeled data in the test sets of commonly-used benchmarks such as MNIST, CIFAR-10, CIFAR-100, ImageNet, etc. They found that lower-capacity models may be practically more useful than higher-capacity models when the dataset contains enormous noisy labels.

Much of the most closely related work comes from the domain of language modeling, as modern language models are often trained with very large datasets and parameters budgets \mbox{\cite{bender2021parrots,xu2022systematic}}, and may leak information in unexpected ways when released to the public \mbox{\cite{carlini2019secret, elangovan2021quantifying}}.
\mbox{\citet{carlini2019secret}} investigated the unintended memorization in neural networks via an exposure-based testing strategy, where they insert secret word sequences in training data and measure the exposure of secrets at inference time. They showed that a generative sequence model trained on sensitive data can actually memorize secret sequences from its training data, \mbox{\eg}, credit card numbers.
\mbox{\citet{elangovan2021quantifying}} examined the impact of data leakage in publicly available datasets by assessing the overlap of similar instances between train and test sets on the model's performance for shared tasks \mbox{\cite{GLUE}}. They found that models that memorize from overlapping data may have higher performance than models that are more robust in generalization, however, such models may not provide an effective performance in real-world unseen scenarios.
Recently, \mbox{\citet{tirumala2022memorization}} analyzed the effects of dataset size and model size on the training dynamics of large language models. They demonstrated that larger models memorize training data faster even before overfitting and memorize unique identifiers quickly in the training set. They concluded by suggesting memorization as a critical metric when scaling the size of language models.

We argue that a similar comprehensive analysis of noise, memorization, and generalization is needed for \cis. We base our methodology on the above observations, and especially on the suite of metrics proposed in computer vision \cite{arpit2017closer, zhang2017rethinkgen}.

\section{Study Subjects}
\label{sec:settings}

In this section, we provide a brief description of the design and implementation aspects of different models, tasks, and datasets used in our study. We study six \mbox{\cis} (\mbox{\ctv}, \mbox{\cts}, \mbox{\tra}, \mbox{\ggnn}, \mbox{\great}, and \mbox{\CodeBERT}) for four well-known tasks (method name prediction, variable misuse, code document generation, and natural language code search) across datasets of three programming languages (Java, Python, and Ruby).

\bigskip \noindent
The method name prediction \cite{allamanis2015suggesting, allamanis2016summarization} and variable misuse localization-repair tasks \cite{allamanis2018ggnn, vasic2019neural} have been heavily studies of generalizability, adversarial examples, and transparency of neural code intelligence models \cite{wang2019coset, kang2019generalizability, rabin2019tnpa, rabin2020evaluation, compton2020obfuscation, rabin2021generalizability, yefet2020adversarial, rabin2020demystifying, wang2021demystifying, rabin2021dd, rabin2022perses, suneja2021probing}.
For these two tasks, we choose models and datasets, shown in Table~\ref{tab:model-conf}, for which the training scripts and data are readily available \cite{alon2019code2vec,alon2019code2seq,hellendoorn2020global}, to which we only make minor modifications, such as logging the number of parameters and/or changing the batch size.

\subsection{Method Name Prediction (\mnp)}

\subsubsection{Task}
In the method name prediction task, the model attempts to predict the name of a method from its body.
This task has several applications, such as code search \cite{liu2019learning}, code summarization \cite{allamanis2016summarization}, and reasoning about code analogies \cite{alon2019code2vec}.
This task has been used as the downstream task to evaluate several state-of-the-art \cis \cite{alon2019code2vec,alon2019code2seq}.

\subsubsection{Data}
We use the following three datasets for the \mnp task:
\begin{itemize}
    \item \JTT: This dataset \cite{rabin2020demystifying} contains $1,000$ randomly selected samples for each of the ten most frequent labels from the \JL dataset in \cite{alon2019code2seq}. Those are: \texttt{equals}, \texttt{main}, \texttt{setUp}, \texttt{onCreate}, \texttt{toString}, \texttt{run}, \texttt{hashCode}, \texttt{init}, \texttt{execute}, and \texttt{get}.
    
    \item \JS: This dataset \cite{alon2019code2seq} contains nine Java projects for training, one Java project for validation, and one Java project for testing. In total, it contains about $~700$K methods.
    
    \item \JM: This dataset \cite{alon2019code2seq} contains 800 Java projects for training, 100 Java projects for validation, and 100 Java projects for testing. In total, it contains about $4$M methods.
\end{itemize}

\subsubsection{Models}
We study two commonly used models for the \mnp task: \ctv~\cite{alon2019code2vec}, and \cts~\cite{alon2019code2seq}. 
The models are similar, in that they rely on extracting ``paths'' from the method's abstract syntax tree (AST) that connect one terminal or token to another.
These paths, mapped to vector embeddings, are enumerated exhaustively and used by the models in different ways. Since these paths consolidate both lexical and syntactic information, the models tend to outperform strictly token-based models.

In \ctv~\cite{alon2019code2vec}, each path, along with its source and destination terminals, is mapped into a vector embedding, which is learned together with other network parameters.
Then, the separate vectors obtained from each path-context are concatenated into a single context vector using a fully-connected layer.
Additionally, the model learns an attention vector that is used to aggregate the path-context representations into a single code vector that represents a method body.
Finally, given a method body's code vector, the model computes the probability of each target method name using a softmax-normalization between the code vector and each of the embeddings of target method names.

In \cts~\cite{alon2019code2seq}, a bi-directional LSTM encoder is used to represent paths instead, which encodes paths node-by-node while splitting tokens into sub-tokens. The decoder similarly uses attention to select relevant paths while decoding, but now predicts sub-tokens of a target sequence one-by-one to generate a method name.

\subsection{Variable Misuse (\vm)}

\subsubsection{Task}
A variable misuse occurs when a different, but also declared and correctly typed, variable is used than intended~\cite{allamanis2018ggnn}.
These bugs are common in software development and are usually by-products of code copy-pasting~\cite{karampatsis2020often}.
In the variable misuse localization and repair task, the model should both locate the misuse bug and then propose a repair in the form of the correct identifier to use~\cite{vasic2019neural, hellendoorn2020global}.

\subsubsection{Data}
We study memorization on this task in the context of \PY~\cite{hellendoorn2020global}, which is derived from the ETH Py150 dataset~\cite{raychev2016tree}. In this synthetic dataset, all top-level function definitions were extracted from open-source projects. For each sample, buggy samples were generated by randomly replacing one variable with another based on declared variables, generating up to three random samples per function. Any sample thus contains a function definition as a token sequence and a ``has bug'' flag. Buggy samples, in addition, have error-related data in the form of pointers into the token sequence (e.g. error location, set of repair targets). The training and validation sets span ca. 1.8 million and 185k samples, each set having a balanced number of buggy and bug-free samples.

\subsubsection{Models}
We use three models for the \vm task: Transformer~\cite{vaswani2017attention}, GGNN~\cite{li2015gated,allamanis2018ggnn} and GREAT~\cite{hellendoorn2020global}. The first is a widely used, attention-based model in which the representation of tokens are iteratively refined through all-to-all communication. The second uses more targeted message passing, along expert-derived connections in the code such as data-flow and syntactic dependencies. The final model combines both of these, using \emph{biased attention} to make the models both aware of important relations while still allowing global communication. This model was shown to outperform both others by a significant margin and represents the current state-of-the-art on this task \cite{hellendoorn2020global}. We include all models due to their unique characteristics and generally strong performance.

For use in the \vm task, all models follow Vasic \etal's approach~\cite{vasic2019neural} that consists of: (1) stacking an initial token-embedding layer, (2) computing a distributed representation of the code input using a ``core'' model (one of the above), and (3) generating two pointers into the input code token sequence, one each for the localization and repair tasks.
More specifically, given a buggy (or correct) tokenized code sample the task is to predict two pointers into the sequence of sample tokens: a pointer to the position of the token that has the wrong variable (or a default token for correct samples), and a pointer to the position of any token that contains the correct variable (this pointer is ignored for correct samples).

\begin{table}
\centering
\begin{tabular}{lr}
\Xhline{2\arrayrulewidth}
\textbf{Model (Dataset)}   & \textbf{\#Trainable Parameters} 
\\ \hline
\ctv (\JTT)                & 62,528,256                 \\
\ctv (\JS)                 & 256,082,176                \\
\ctv (\JM)                 & 367,681,920                \\
\cts (\JTT)                & 3,826,368                  \\
\cts (\JS)                 & 16,128,448                 \\
\cts (\JM)                 & 37,411,328                 \\
\ggnn (\PY)                & 41,190,914                 \\
\tra (\PY)                 & 26,215,938                 \\
\great (\PY)               & 26,225,538                 \\
\CodeBERT (\csearch{} - \Ruby) & 124,647,170            \\
\CodeBERT (\cnl{} - \Ruby)     & 172,503,552            \\
\Xhline{2\arrayrulewidth}
\end{tabular}
\caption{Number of trainable parameters for each model including vocabulary.}
\label{tab:model-conf}
\end{table}

\bigskip \noindent
In recent years, researchers have been increasingly using large, pre-trained models that have been learned from datasets including code and natural language \mbox{\cite{chen2021Codex,nijkamp2022CodeGen}}. There are several benchmarks \mbox{\cite{GLUE, CodeXGLUE}} that provide publicly available datasets in order to evaluate and compare the performance of different models on same tasks. 
In this study, we experiment with BERT-style pre-trained models {\cite{devlin2019bert}}, mainly \mbox{\CodeBERT} {\cite{feng2020codebert}}, for natural language code search and code document generation tasks.
The \mbox{\CodeBERT} is a \mbox{\texttt{bimodal}} pre-trained model for programming languages (PL) and natural languages (NL) such as Python, Java, Ruby, Document, etc. It captures the semantic connection between NL and PL and produces general-purpose representations that can broadly support various downstream NL-PL tasks {\cite{feng2020codebert}}. It has been developed following the architecture of BERT {\cite{devlin2019bert}} and RoBERTa {\cite{liu2020roberta}}, which itself is based on the Transformer {\cite{vaswani2017attention}} that is used in most large pre-trained models.
The \mbox{\CodeXGLUE} repository provides code and data for fine-tuning \mbox{\CodeBERT} on various datasets and tasks {\cite{CodeXGLUE}}.

\subsection{Code-to-Text Generation (\cnl)}

\subsubsection{Task}
The task is commonly known as code documentation generation or code summarization where the objective is to generate natural language comments for a code snippet. It provides a high-level summary of the functionality performed by the code snippet. This task is also referred to as code-to-text, code-to-NL generation, and code-to-documentation \cite{feng2020codebert, CodeXGLUE}. It can benefit software maintenance, code understanding, and retrieval \cite{mcburney2014automatic, iyer2016summarizing, wan2018improving}.

\subsubsection{Data}
We conduct experiments on the \CodeSearchNet dataset \cite{CodeSearchNet} for the \cnl task, where the dataset has already been cleaned\footnote{\url{https://github.com/microsoft/CodeXGLUE/tree/main/Code-Text/code-to-text}\label{CodeXGLUE_code_to_text}} by removing comments and removing examples that cannot be parsed, contain limited/special tokens, or are not English. 
We select the \Ruby language data which contains $24,927$ samples for training, $1,400$ samples for development, and $1,261$ samples for testing. In each sample, the input is a code snippet (\ie, function), and a natural language text (\ie, docstring) that briefly describes the code is the output.

\subsubsection{Model}
The \CodeXGLUE repository provides a pipeline\footreff{CodeXGLUE_code_to_text} for fine-tuning the \CodeBERT model on the \cnl task, which is evaluated by the smoothed BLEU-4 score \cite{papineni2002bleu, lin2004orange}. The architecture consists of the \CodeBERT as the encoder and a 6-layer Transformer as the decoder. It uses the Adam optimizer and cross-entropy loss function to update the model's parameters. We fine-tune the \CodeBERT model using the \Ruby programming language data of the cleaned \CodeSearchNet dataset for the \cnl task.

\subsection{Natural Language Code Search (\csearch)}

\subsubsection{Task}
In this task, given a natural language query, the target is to find the most semantically relevant source code from a collection of candidates. The task is formulated as a binary classification problem, where given a pair of query and code, a model aims to classify whether the code is semantically related to the query or not \cite{feng2020codebert, CodeXGLUE}. It has been actively studied and applied in many software development practices \cite{gu2018deepcs, CodeSearchNet, gu2021multimodal}.

\subsubsection{Data}
We use the preprocessed dataset\footnote{\url{https://github.com/microsoft/CodeBERT/tree/master/CodeBERT/codesearch}\label{CodeBERT_codesearch}} derived from the original \CodeSearchNet dataset \cite{CodeSearchNet} for the \csearch task, where each sample includes a code snippet paired with a natural language query. The dataset consists of a balanced number of positive and negative samples. Samples where the code is related to the query are positive samples and are labeled as ``1''. Contrary, negative samples contains randomly replaced irrelevant code or query and are labeled as ``0''. We choose the \Ruby language data which contains $97,580$ samples for training, $4,417$ samples for development, and $2,279$ samples for testing.

\subsubsection{Model}
We run the implementation of the \CodeBERT model provided by the authors in their GitHub repository\footreff{CodeBERT_codesearch}. It uses the Adam optimizer and binary classification loss function to update the model's parameters. We fine-tune the \CodeBERT model using the \Ruby programming language data of the preprocessed \CodeSearchNet dataset for the \csearch task.

\section{Methodology}
\label{sec:approach}
This section describes the methodology we use to study memorization in \cis.

\begin{figure}
    \centering
    \includegraphics[width=\linewidth,height=\linewidth,keepaspectratio]{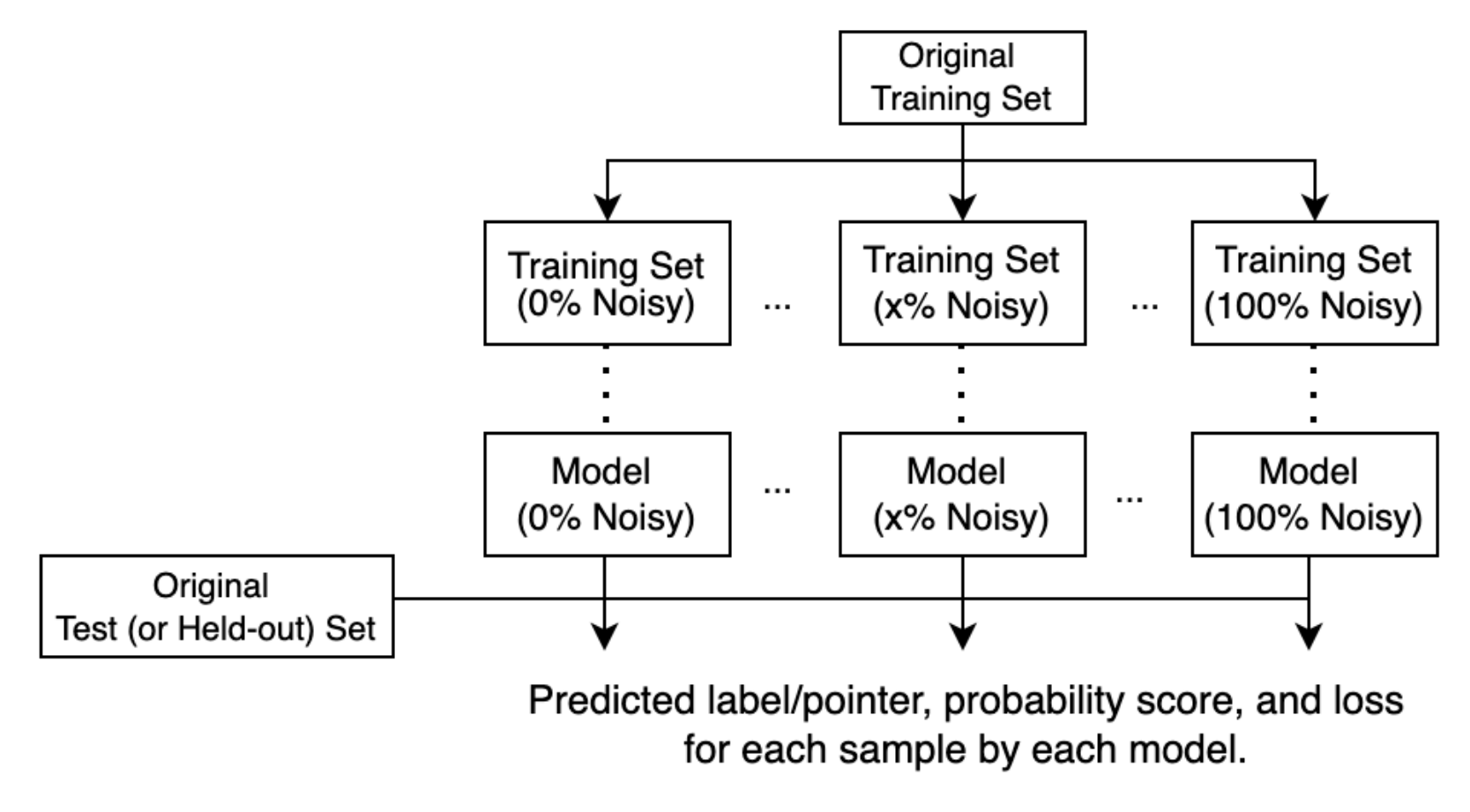}
    \caption{Workflow of our approach for evaluating memorization and generalization.}
    \label{fig:approach}
\end{figure}

\subsection{Characterizing Memorization}
\label{subsec:enforcing-mem}
We follow and adapt the methodology used in computer vision literature \cite{arpit2017closer, zhang2017rethinkgen}.
The ultimate objective of a trained model is to find patterns consistent in the dataset that helps it to generalize on unseen data. If the dataset contains random noise, it is quite difficult for the model to find effective patterns. Therefore, for a model to learn on a noisy dataset, \ie, randomized labels, it needs to memorize data points.
By tuning the degree of such noise and comparing learning patterns between the original dataset and various noisy datasets, we can characterize memorization artifacts of a model on a specific dataset. And by identifying consistent trends across models and/or datasets, we can derive more general conclusions about memorization artifacts in neural models of source code.

\Cref{fig:approach} depicts a high-level view of the workflow in the proposed methodology.
Given the original training dataset, the approach creates several noisy training datasets by noising a portion of the data in two ways: 
1) \textit{output noise} where we add noise into target labels, 
and 2) \textit{input noise} where we add noise into input programs.
We create multiple training datasets with {\{0\%, 25\%, 50\%, 75\%, 100\%\}}-noise, where $0\%$-noise denotes the original training dataset, and $100\%$-noise denotes a dataset where where all examples in the training set are fully noisified with output noise or input noise.
We use each noisy training set to train models, and compare and contrast the training characteristics of each noisy model on the original test set for \mnp, held-out set for \vm, and development set for \CodeBERT.

\begin{figure*}
    \centering
    \includegraphics[width=\textwidth,height=\textwidth,keepaspectratio]{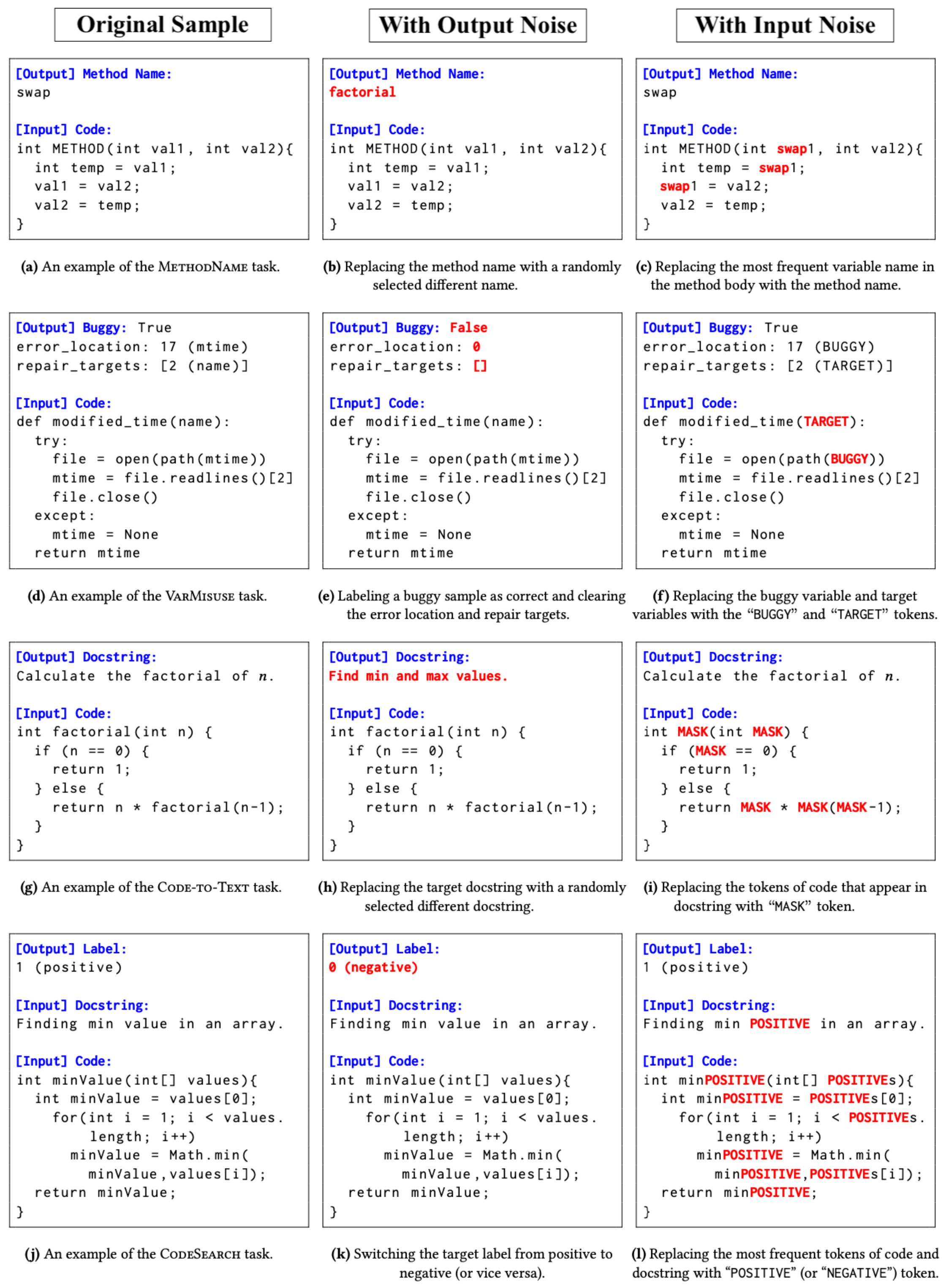}
    \caption{Examples of original samples, output noise and input noise for source code in different tasks.}
    \label{fig:noise_example}
\end{figure*}

\mbox{\Cref{fig:noise_example}} shows the examples of output noise and input noise for different tasks. Concretely, for source code, we randomly introduce noise into the training set, as follows:
\begin{itemize}
    \item \Cref{fig:noise_example}$_{a-c}$ shows an example of the \mnp task, where noise is added to randomly selected samples over all training samples. We add output noise by replacing the target label (method name) with another randomly chosen different label from the set of available method names. We add input noise in two ways: (a) by deleting a random statement from the method body, and (b) by replacing a variable name in the method body with the method name.
    
    \item \Cref{fig:noise_example}$_{d-f}$ shows an example of the \vm task, where noise is added to the same percentage of randomly selected buggy and correct training samples. We add output noise to a buggy sample by simply labeling it as correct, where we set the error location to $0$ and clear the repair targets. Conversely, a correct sample is changed to buggy by labeling it as such and assigning both a random location as the error location and a set of randomly chosen variables as repair targets. =Towards adding input noise to a buggy sample, we replace all occurrences of the repair target token with ``\texttt{TARGET}'' and replace the token at the error location with ``\texttt{BUGGY}'' in the input code. Conversely, we add input noise to a correct sample by replacing all occurrences of the most frequent token in the input code with ``\texttt{NONBUGGY}''.
    
    \item \mbox{\Cref{fig:noise_example}$_{g-i}$} shows an example of the \mbox{\cnl} task, where noise is added to the randomly selected samples over all training samples. We add output noise by replacing the target docstring (i.e., comments) with another randomly chosen different docstring from the set of all available docstrings. We add input noise by replacing the tokens of the input code snippets that appear in the target docstring with the ``\texttt{MASK}'' token.
    
    \item \mbox{\Cref{fig:noise_example}$_{j-l}$} shows an example of the \mbox{\csearch} task, where noise is added to the same percentage of randomly selected positive and negative training samples. We add output noise to a positive (\mbox{\resp} negative) sample by simply labeling it as negative (\mbox{\resp} positive). We add input noise by replacing the most frequent tokens of input code snippets and target docstring with the ``\texttt{POSITIVE}'' (\mbox{\resp} ``\texttt{NEGATIVE}'') token for positive (\mbox{\resp} negative) samples.
\end{itemize}

\subsection{Training the Models}
\label{sec:train-models}
For training the \mnp models (\ctv and \cts), we follow the exact training configurations from prior work \cite{alon2019code2vec, alon2019code2seq}, only using a slightly smaller batch size due to memory limitations. We use a batch size of $128$ for \JTT and $256$ for \JS and \JM. We train the models on a NVIDIA Tesla-P100 GPU (with $12$GB of memory) up to $50$ epochs for \JTT and \JS but $20$ epochs for \JM because of the time budget.
To train the \vm models (Transformer, GGNN, and GREAT), we use the default base configurations from \citet{hellendoorn2020global}'s public replication package \footnote{\url{https://github.com/VHellendoorn/ICLR20-Great}}. Training proceeds in ``steps'' of $250$k training samples, after each of which performance is assessed on $25$k held-out samples. A full epoch is completed roughly every $7$ such steps and continued for $50$ epochs on a NVIDIA Tesla-P100 GPU with $12$GB of memory and a NVIDIA RTX-3090 GPU with $24$GB of memory.
Finally, we fine-tune the \mbox{\CodeBERT} model up to $50$ epochs using the \mbox{\CodeXGLUE} {\cite{CodeXGLUE}} pipeline on two NVIDIA RTX-3090 GPUs (with $24$GB of memory each) for \mbox{\cnl} and \mbox{\csearch} tasks. We mostly use the default configurations, only adjusting the batch size to $64$, and the maximum length of input and output sequences to $256$ and $64$, respectively, due to memory limitations.

\subsection{Metrics}
\label{sec:metrics}
We collect a wide range of metrics \cite{arpit2017closer, zhang2017rethinkgen} to characterize training in the datasets. We describe those metrics in the rest of this section.

\subsubsection{Predicted Score}
\label{sec:score}
We use ``Predicted score'' to refer to the (probability) score assigned to a predicted output by the model. Depending on the model, we compute it differently. 
In \ctv, the model computes the probability of the target name via a softmax-normalization operation between the code vector of a given method body and the embeddings of all possible method names. 
\begin{align*} 
& \mathcal{P}(\text{method\_name}_{i}) = \\ 
& \frac{\exp{(code\_vector^T \, \cdot \, name\_embedding_{i})}}{\sum_{\text{method\_name}_{j} \, \in \, \text{all\_name}} {\exp{(code\_vector^T \, \cdot \, name\_embedding_{j})}}}
\end{align*}

\noindent
In \cts, when predicting the method name, the model makes predictions for each sub-token of a target sequence at each step. Hence, we compute an average score for a prediction as follows:
\begin{align*} 
\mathcal{P}_{avg}(\text{method\_name}_{i}) = \frac {\sum_{token_{j} \, \in \, \text{method\_name}_{i}}{\mathcal{P}(token_{j})}} {|token_{j} \in \text{method\_name}_{i}|}
\end{align*}

\noindent
For \vm, the models emit two \emph{logits} per token (logit$_0$ for localization and logit$_1$ for repair), which are converted to probabilities via the softmax operation.

\begin{align*}
\mathcal{P}_{loc}(token_i) &= \frac{\exp{(logit_{0, i})}}{\sum_j{\exp{(logit_{0,j})}}} \\
\mathcal{P}_{rep}(token_i) &= \sum_{\{i|token_i \, \in \, \text{repair\_candidates}\}} \frac{\exp{(logit_{1, i})}}{\sum_j{\exp{(logit_{1,j})}}}
\end{align*}

Here, the location probability is simply computed per token, while the repair probability is aggregated across all occurrences of each declared variable (the ``repair candidate'' set). For bug-free samples, the location component should point to token $0$, and the repair predictions are ignored. For buggy samples, the location probability of the incorrect variable use (var-use)  should be maximized by $\mathcal{P}_{loc}$, as should the sum of the repair probabilities of all occurrences of the correct variable.

\smallskip\noindent
In \mbox{\csearch}, the \mbox{\CodeBERT} model emits two \emph{logits} for two target labels ($\{0,1\}$), which are converted to probabilities via the softmax operation.
\begin{align*}
\mathcal{P}(label_{i}) &= \frac{\exp{(logit_{i})}}{\exp{(logit_{0})} + \exp{(logit_{1})}}
\end{align*}

\noindent
In \mbox{\cnl}, when generating docstring, the \mbox{\CodeBERT} model makes predictions for each token as a sequence. Therefore, we compute an average score for a single sample as follows:
\begin{align*} 
\mathcal{P}_{avg}(\text{docstring}_{i}) = \frac {\sum_{token_{j} \, \in \, \text{docstring}_{i}}{\mathcal{P}(token_{j})}} {|token_{j} \in \text{docstring}_{i}|}
\end{align*}

\subsubsection{\texorpdfstring{\FOneScore}{Lg} and Accuracy}
\label{sec:accs}
We use the evaluation metric, \FOneScore over sub-tokens, as commonly used in the literature~\cite{alon2019code2vec, alon2019code2seq} for \mnp task.
The \FOneScore is a harmonic mean of precision ($P$) and recall ($R$). 
\begin{align*} 
F_{1}\text{--}Score = \frac{2}{P^{-1} + R^{-1}} = 2\,\cdot\,\frac{P\,\cdot\,R}{P\,+\,R}
\end{align*}

\noindent
In \vm models, we use the two accuracy metrics:
localization accuracy and repair accuracy. Localization accuracy measures the proportion of buggy samples for which the model correctly predicts the location:
\begin{align*}
\text{predicted\_location} &= \arg \max_i \; \mathcal{P}_{loc}(token_i) \\
\text{localization\_accuracy} &= \frac{1}{n} \sum_{i=1}^{n}{(\text{actual\_location}_{i} = \text{predicted\_location}_{i})}
\end{align*}
Repair accuracy captures the proportion of buggy samples for which the model correctly assigns at least half the probability mass to occurrences of the correct variable name:
$$
\text{repair\_accuracy} = \frac{1}{n} \sum_{i=1}^{n}{(\sum_{\{j|token_j = \text{repair\_target}_i\}} {\mathcal{P}_{rep}(token_j)} \; \geq 0.5)}
$$

\noindent
In \mbox{\csearch}, the accuracy is computed as the proportion of samples in the balanced set for which the \mbox{\CodeBERT} model correctly predicts the actual label.
\begin{align*}
\text{predicted\_label} &= \arg \max_{i} \; \mathcal{P}(label_{i}) \\
Accuracy &= \frac{1}{n} \sum_{i=1}^{n}{(\text{actual\_label}_{i} = \text{predicted\_label}_{i})}
\end{align*}

\noindent
In \mbox{\cnl}, the \mbox{\CodeBERT} model is evaluated with the smoothed BLEU-4 score \mbox{\cite{papineni2002bleu, lin2004orange}} that adds one count to the $\ngram$ for $n>1$, which is suitable for short documents or where $\ngram$ may not overlap. It is computed as follows:
\begin{align*}
\text{smoothed BLEU-4} &= \text{BP} * \exp (\frac{1}{4}\sum_{n=1}^{4}{\log P_{n}})
\end{align*}
where, $BP$ is the brevity penalty factor, and $P_{n}$ is the modified precision of $\ngram$ after smoothing.
\begin{align*}
P_{n} &= \frac{\sum{Count_{clip} \, (\ngram) \, + \, 1 \, (n>1)}}{\sum{Count \, (\ngram) \, + \, 1 \, (n>1)}} \\
\text{BP} &= \exp(\min(0, 1 - \frac{\text{len\,(reference)}}{\text{len\,(candidate)}}))
\end{align*}

\subsubsection{Spread of Loss}
\label{sec:spreadloss}
The loss refers to the error in the model's prediction, typically tied to the probability it assigned to the ground-truth label. 
In \ctv, the model computes cross-entropy loss between the softmax of raw logits (predicted distribution $q$) and the ground-truth targets (true distribution $p$).
The true distribution $p$ assigns a value of $1$ to the actual name and $0$ otherwise, so the cross-entropy loss for a single input method is equivalent to the negative log-likelihood of the actual name.
\begin{align*} 
\mathcal{L}(\text{method\_name}_{i}) &= - \sum_{name_{j} \, \in \, \text{all\_name}}{p(name_{j}) \, log \, q(name_{j})} \\ &= - \, log \, q(\text{method\_name}_{i})
\end{align*}

\noindent
In \cts, the model makes predictions for each sub-token of a target sequence at each step; hence, we compute an average loss for a single input method as follows,
\begin{align*} 
\mathcal{L}_{avg}(\text{method\_name}_{i}) = \frac {\sum_{token_{j} \, \in \, \text{method\_name}_{i}}{\mathcal{L}(token_{j})}} {|token_{j} \in \text{method\_name}_{i}|}
\end{align*}

\noindent
In \vm models, loss is based on the sum of the location loss and repair loss, both calculated based on the negative log-likelihood of the target value's probabilities (see above):
\begin{align*}
\mathcal{L}_{VM} &= \mathcal{L}_{loc} + \mathcal{L}_{rep}  \\
&= -\log{\mathcal{P}_{loc}(\text{bug\_location})} - \log{\mathcal{P}_{rep}(\text{repair\_target})}
\end{align*}

\noindent
In \mbox{\csearch}, the \mbox{\CodeBERT} model computes the binary cross-entropy loss between the target value ($y \in \{0,1\}$) and predicted probability ($p$) as follows,
\begin{align*} 
\mathcal{L}(y,p) = - (y \; . \; log(p) \, + \, (1-y) \; . \; log(1-p))
\end{align*}

\noindent
In \mbox{\cnl}, the \mbox{\CodeBERT} model generates docstring as a sequence of tokens; hence, we compute an average loss for a single sample as follows,
\begin{align*} 
\mathcal{L}_{avg}(\text{docstring}_{i}) = \frac {\sum_{token_{j} \, \in \, \text{docstring}_{i}}{\mathcal{L}(token_{j})}} {|token_{j} \in \text{docstring}_{i}|}
\end{align*}

\textit{Gini coefficient.} We measure the spread of loss by computing the Gini coefficient~\cite{gini1936measure} \Space{(a measure of the inequality among values of a frequency distribution)} over training loss after each epoch as training progress. A Gini coefficient of $0$ means perfect equality where all values are the same. On the other end of the spectrum, a Gini coefficient of $1$ means maximal inequality among values. The Gini coefficient is computed as the relative mean absolute difference of all pairs of items in the population. If $\mathcal{L}_{i}$ is the loss of an input program $t_{i}$, and there are $n$ input programs, then the Gini coefficient ($G$) is computed as follows:
\begin{align*} 
G(loss) = \frac {\sum_{i=1}^{n} \, \sum_{j=1}^{n} \, {|\mathcal{L}_{i} - \mathcal{L}_{j}|}} {2n\sum_{i=1}^{n}{\mathcal{L}_{i}}}
\end{align*}

\subsubsection{Critical Sample Ratio}
\label{sec:csr}
We also estimate the complexity of decision boundaries by computing the critical sample ratio (CSR) \cite{arpit2017closer} for the \mnp task.
An input program is called a critical sample if there exists at least one adversarial example\footnote{In machine learning, an adversarial example of an input in a model is a sample with a slight, ideally imperceptible and/or irrelevant difference to the original input, that misleads the model into providing a different prediction. In the method name prediction task, this can be a semantically-equivalent and largely syntactically identical sample program that leads to a different predicted method name.} in close proximity ($\delta$) of the input program. 
For a given test set $D_{t}$, the critical sample ratio (CSR) is measured as follows,
\begin{align*} 
CSR_{\delta} \, = \, \frac{\# \, critical\_samples_{\delta}}{|D_{t}|}
\end{align*} 

A higher value of CSR (closer to $1$) indicates a complex decision boundary, where many samples are just a small transformation away from being labeled differently, whereas a lower value of CSR (closer to $0$) indicates a simpler, more robust decision boundary.

We explore programs within single-transformation distance ($\delta=1$) of a given input program for adversarial programs following recent work~\cite{bielik2020adversarial, rabin2021generalizability}. 
We check for an adversarial program within the single transformation distance of a given input program $t_{i}$ to identify whether $t_{i}$ is a critical sample. Specifically, we apply the single-place variable renaming transformation~\cite{rabin2021generalizability} on the input program $t_{i}$ to generate candidate programs.
The transformation changes the name of a single variable in the input program to a new name following the predefined format $var[0-9]+$ (e.g. ``$var3$'').
The transformation is performed one-by-one on each variable in the input program, creating a set of candidate programs within the single transformation distance. 
Suppose, $T_{C_{i}}$ is a set of candidate programs generated within the single transformation distance of $t_{i}$. Then, $t_{i}$ is a critical sample if there exist at least one candidate program $t_{j} \in T_{C_{i}}$ such that $M(t_{j}) \, \neq \,  M(t_{i})$, where $M(t)$ indicates the predicted name of the program $t$ by model $M$.

\section{Results}
\label{sec:results}

In this section, we present the results of our experiments, in which we study the models' behavior under various rates of output noise and input noise across four key metrics: models' performance (\ie, accuracy), distribution of prediction score, spread of training loss, and critical sample ratio.

\subsection{Analyses of Memorization with Output Noise}
\label{subsec:noise-on-y}
Erroneously labeled data, typically known as mislabeled data, is a common source of output noise in publicly available datasets \cite{northcutt2021pervasive,li2019learning,chen2019understanding,tirumala2022memorization}. In this section, we present the impact of adding such output noise on the training of \cis.

\begin{figure*}
\centering
\captionsetup[subfigure]{width=0.9\textwidth, justification=centering}

\begin{subfigure}{0.32\textwidth}
  \centering
  \includegraphics[width=\linewidth]{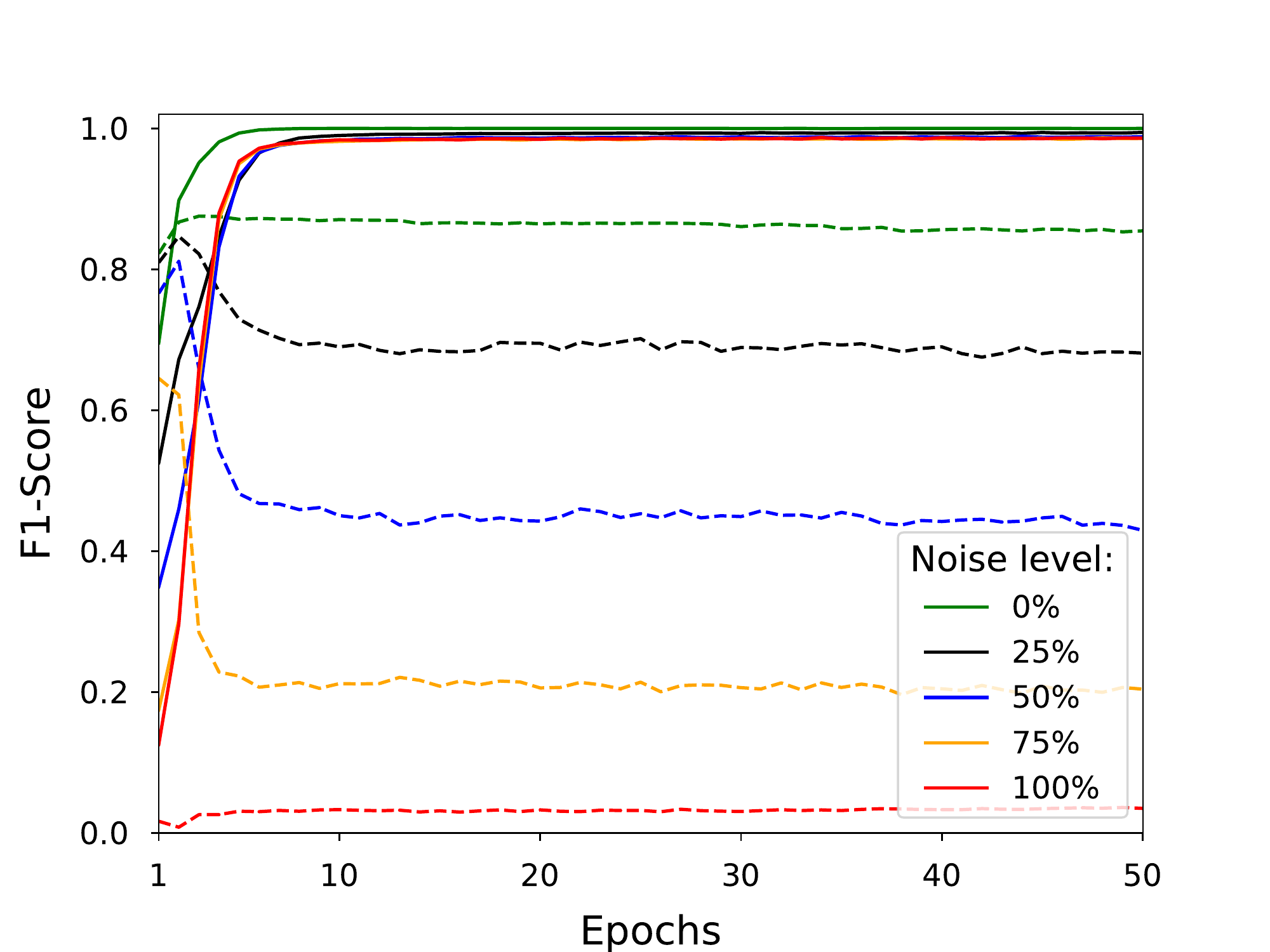}
  \caption{\ctv{} (\JTT)}
  \label{fig:mn_y_f1_c2v_top10}
\end{subfigure}%
\begin{subfigure}{0.32\textwidth}
  \centering
  \includegraphics[width=\linewidth]{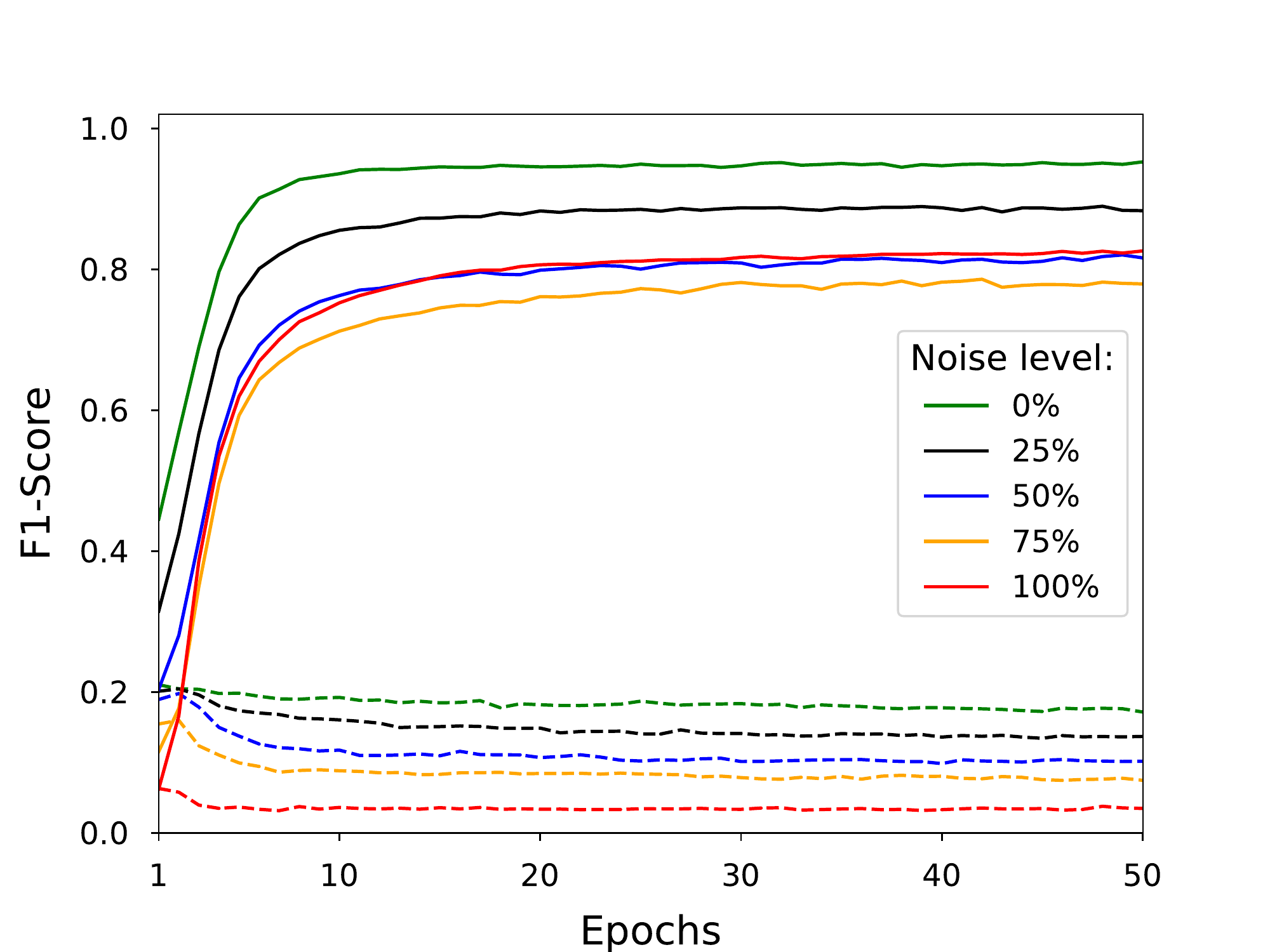}
  \caption{\ctv{} (\JS)}
  \label{fig:mn_y_f1_c2v_js}
\end{subfigure}%
\begin{subfigure}{0.32\textwidth}
  \centering
  \includegraphics[width=\linewidth]{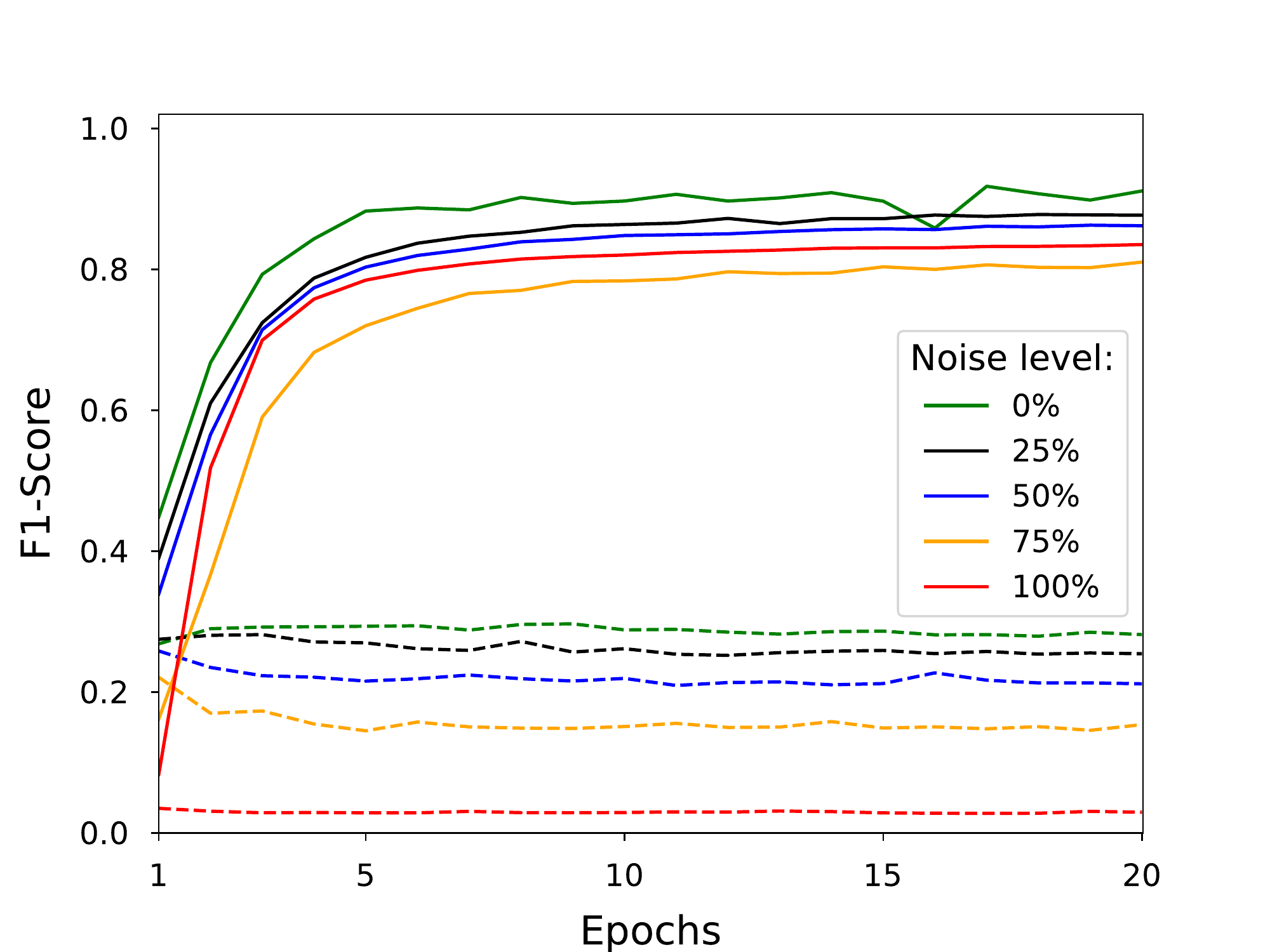}
  \caption{\ctv{} (\JM)}
  \label{fig:mn_y_f1_c2v_jm}
\end{subfigure}

\begin{subfigure}{0.32\textwidth}
  \centering
  \includegraphics[width=\linewidth]{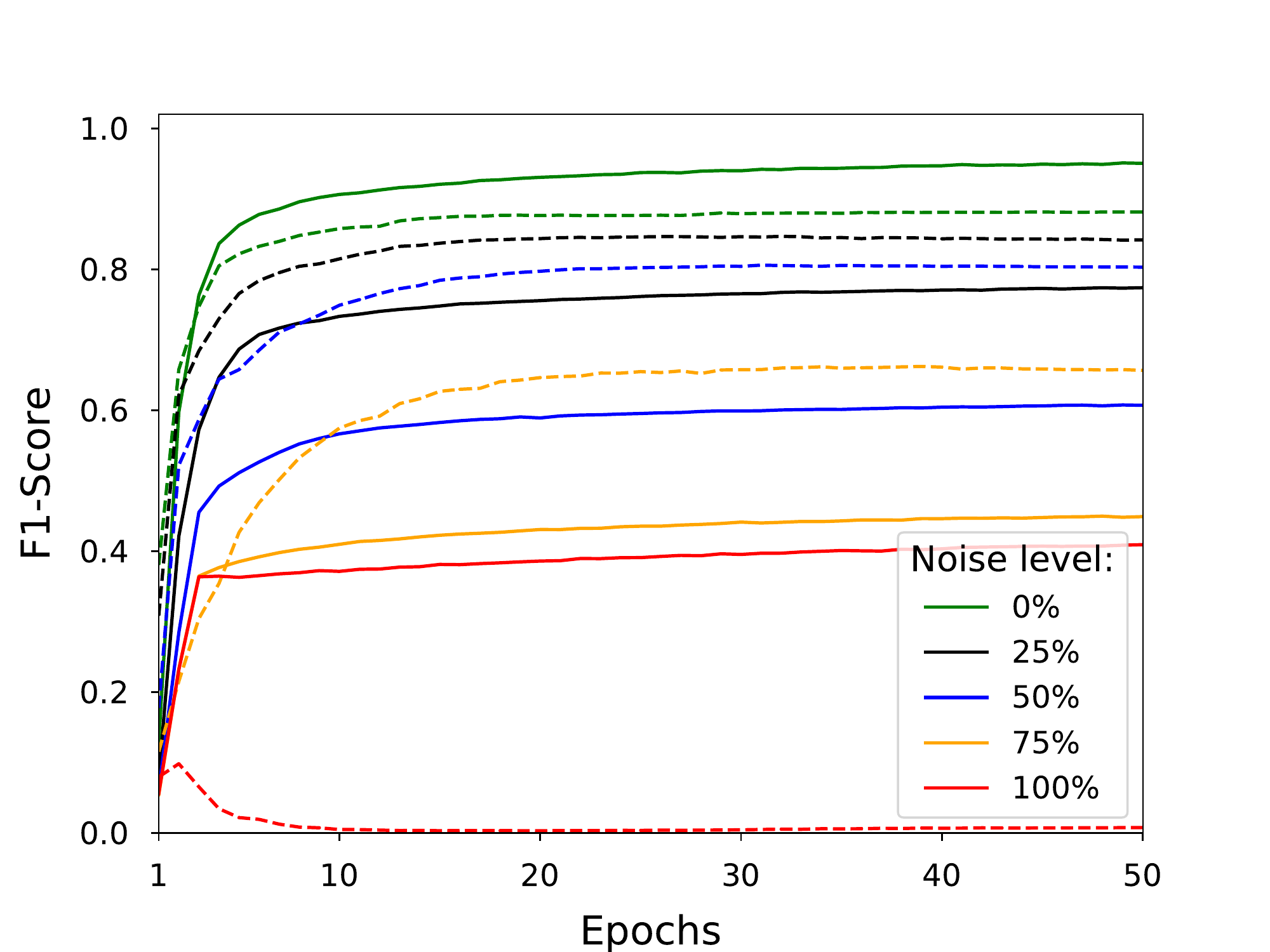}
  \caption{\cts{} (\JTT)}
  \label{fig:mn_y_f1_c2s_top10}
\end{subfigure}%
\begin{subfigure}{0.32\textwidth}
  \centering
  \includegraphics[width=\linewidth]{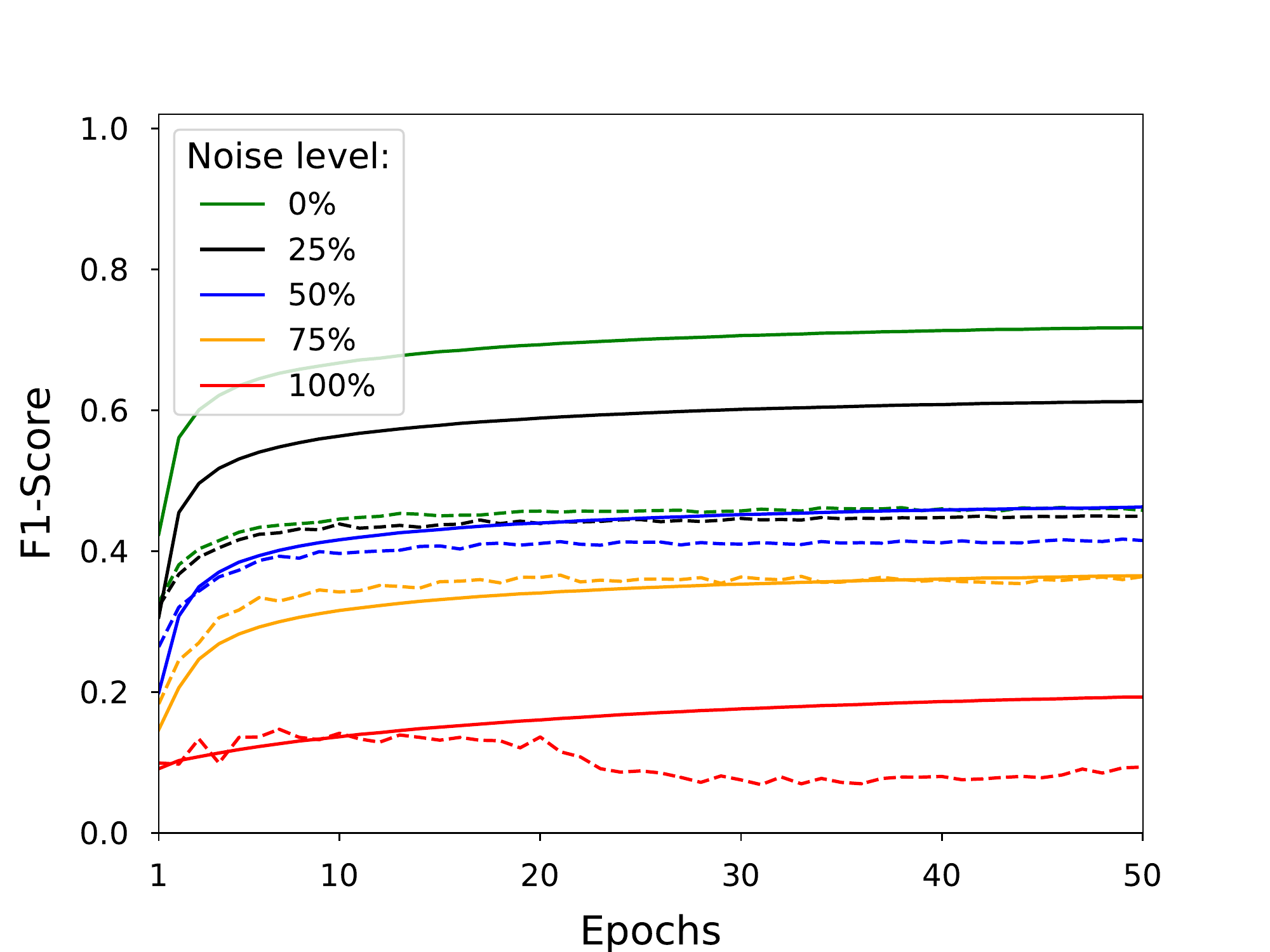}
  \caption{\cts{} (\JS)}
  \label{fig:mn_y_f1_c2s_js}
\end{subfigure}%
\begin{subfigure}{0.32\textwidth}
  \centering
  \includegraphics[width=\linewidth]{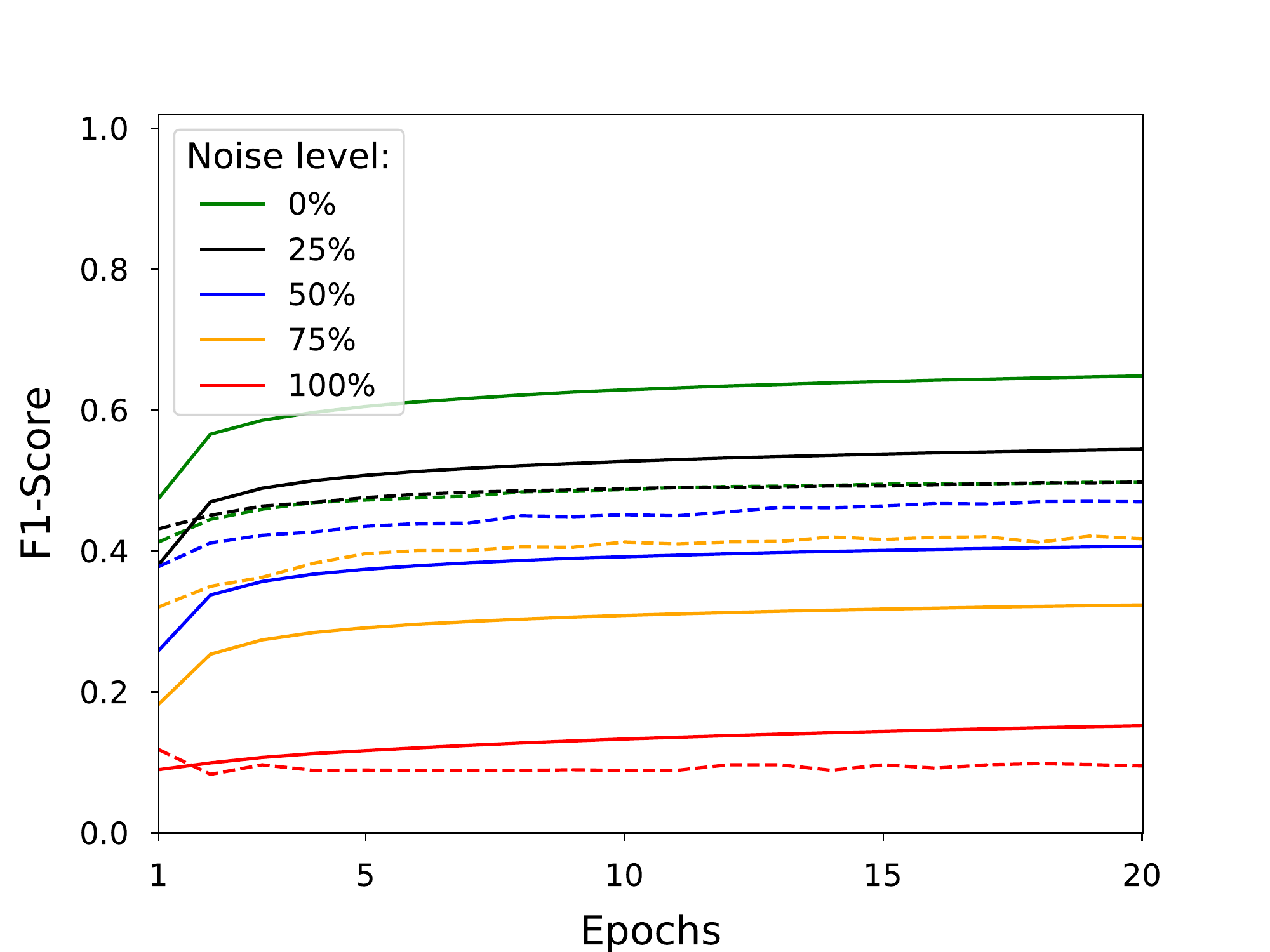}
  \caption{\cts{} (\JM)}
  \label{fig:mn_y_f1_c2s_jm}
\end{subfigure}

\caption{\FOneScore at different noise levels - solid is training, dashed is test (\mnp).}
\label{fig:mn_y_f1_all}


\begin{subfigure}{0.32\textwidth}
  \centering
  \includegraphics[width=\linewidth]{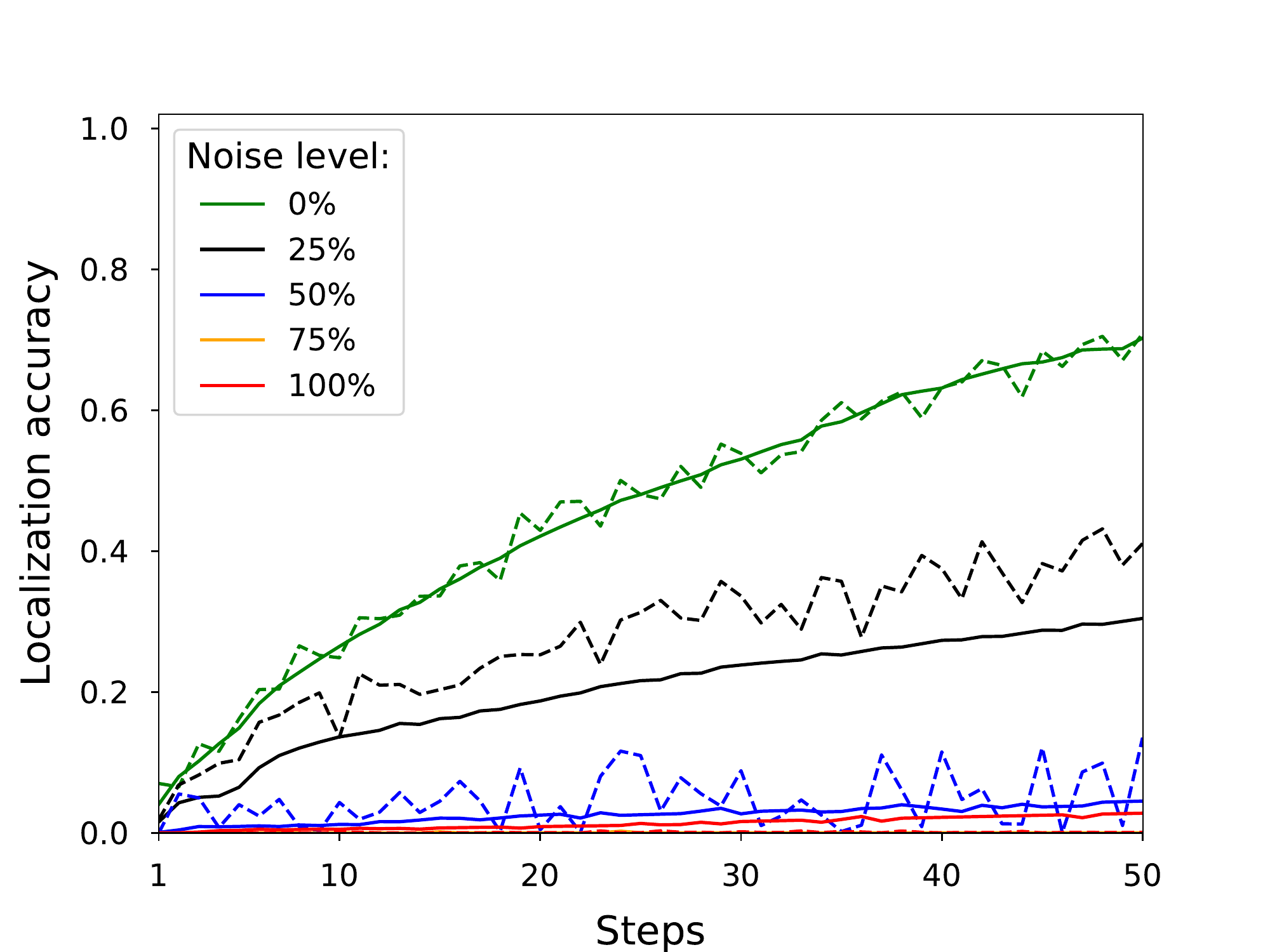}
  \caption{\tra (Localization)}
  \label{fig:vm_y_loc_acc_tra_py}
\end{subfigure}%
\begin{subfigure}{0.32\textwidth}
  \centering
  \includegraphics[width=\linewidth]{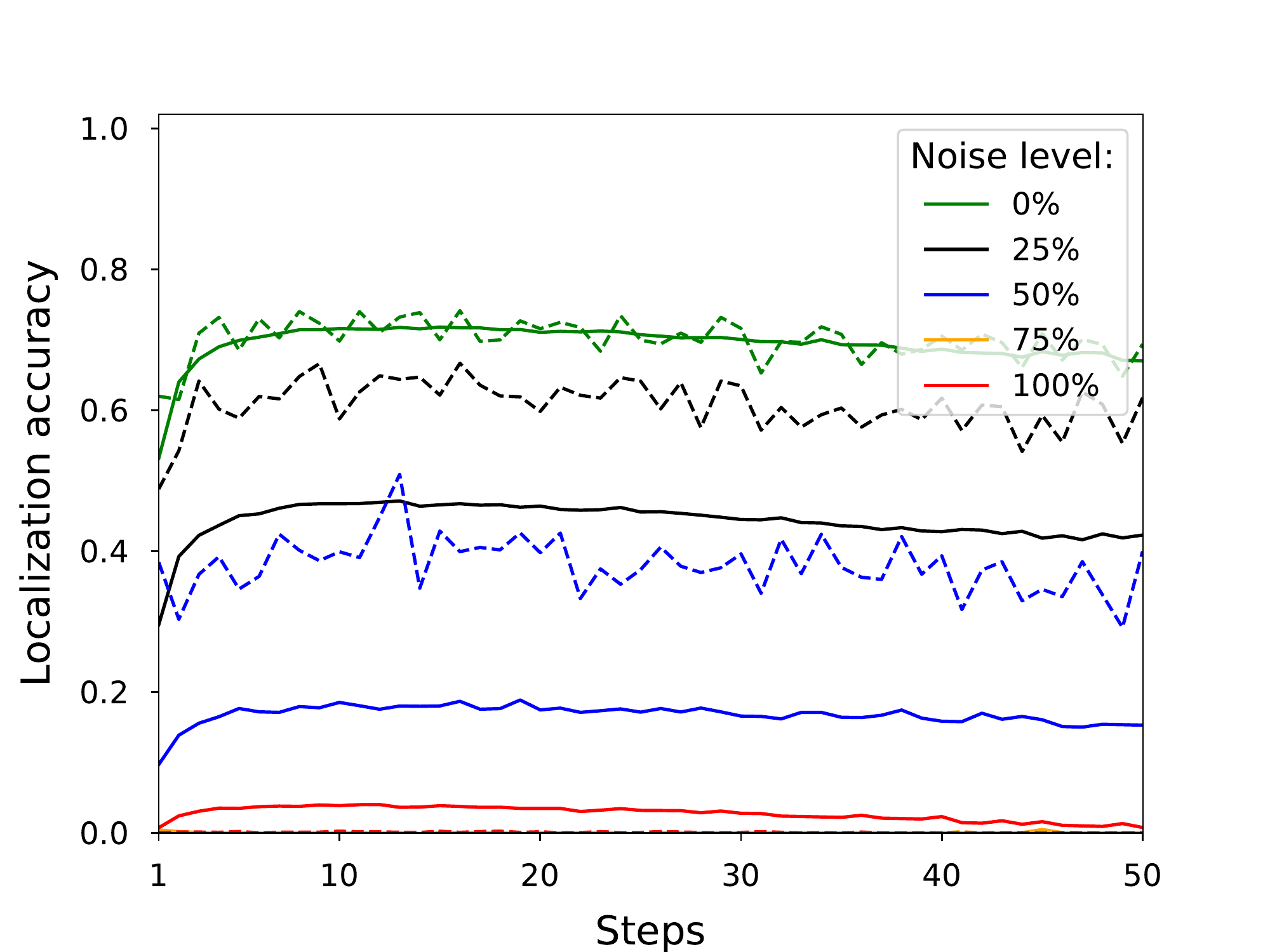}
  \caption{\ggnn (Localization)}
  \label{fig:vm_y_loc_acc_ggnn_py}
\end{subfigure}%
\begin{subfigure}{0.32\textwidth}
  \centering
  \includegraphics[width=\linewidth]{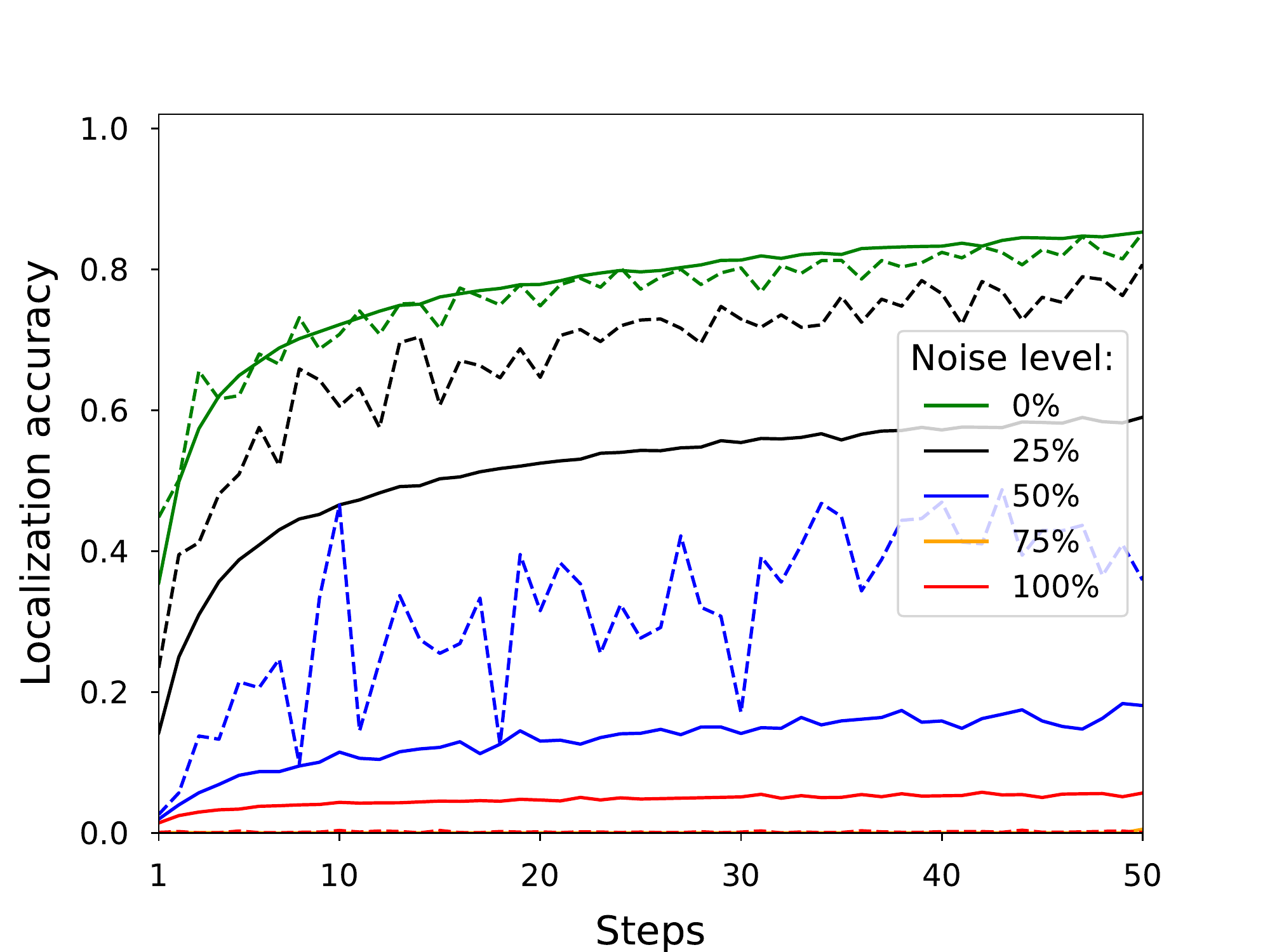}
  \caption{\great (Localization)}
  \label{fig:vm_y_loc_acc_great_py}
\end{subfigure}

\begin{subfigure}{0.32\textwidth}
  \centering
  \includegraphics[width=\linewidth]{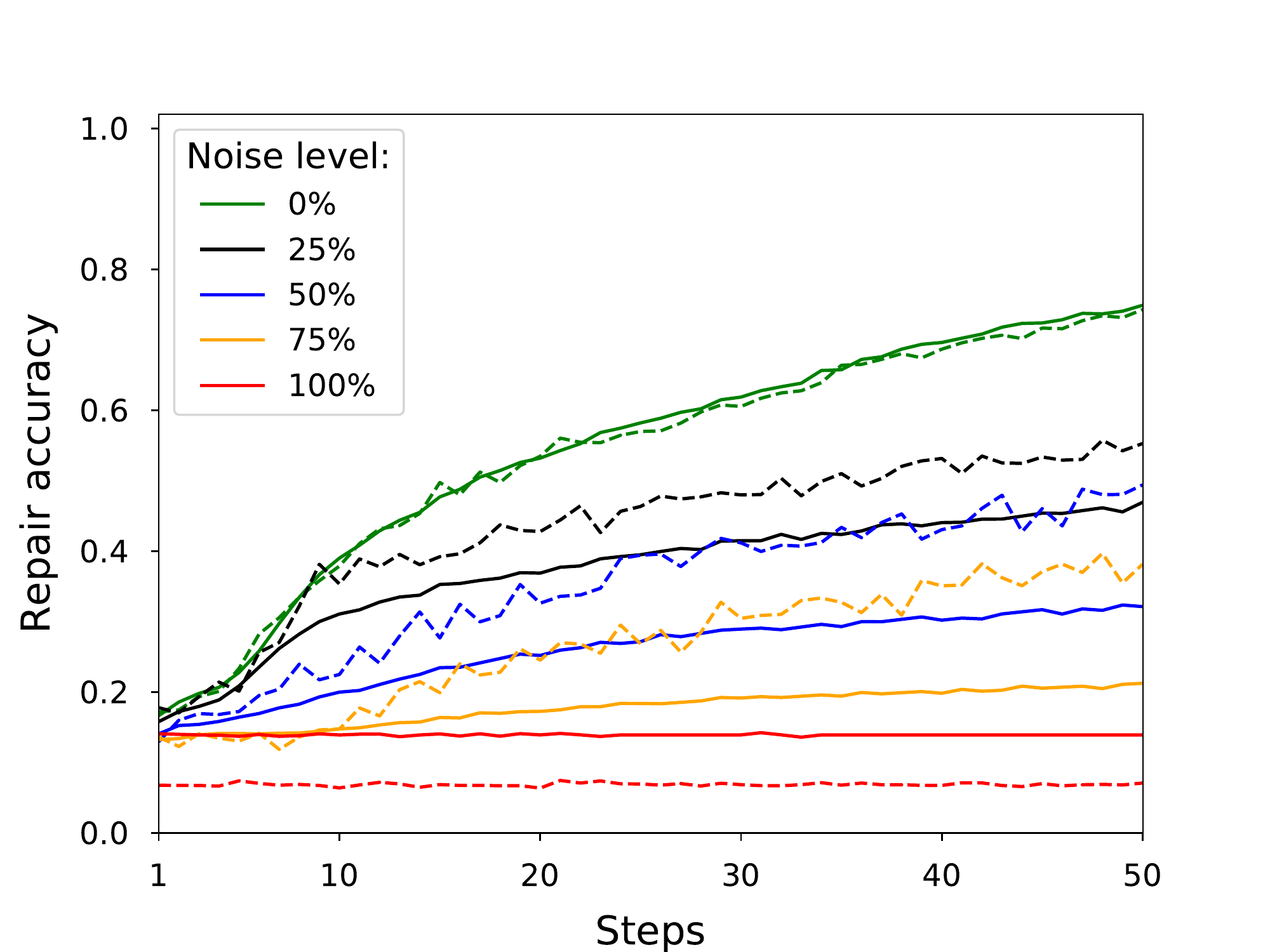}
  \caption{\tra (Repair)}
  \label{fig:vm_y_rep_acc_tra_py}
\end{subfigure}%
\begin{subfigure}{0.32\textwidth}
  \centering
  \includegraphics[width=\linewidth]{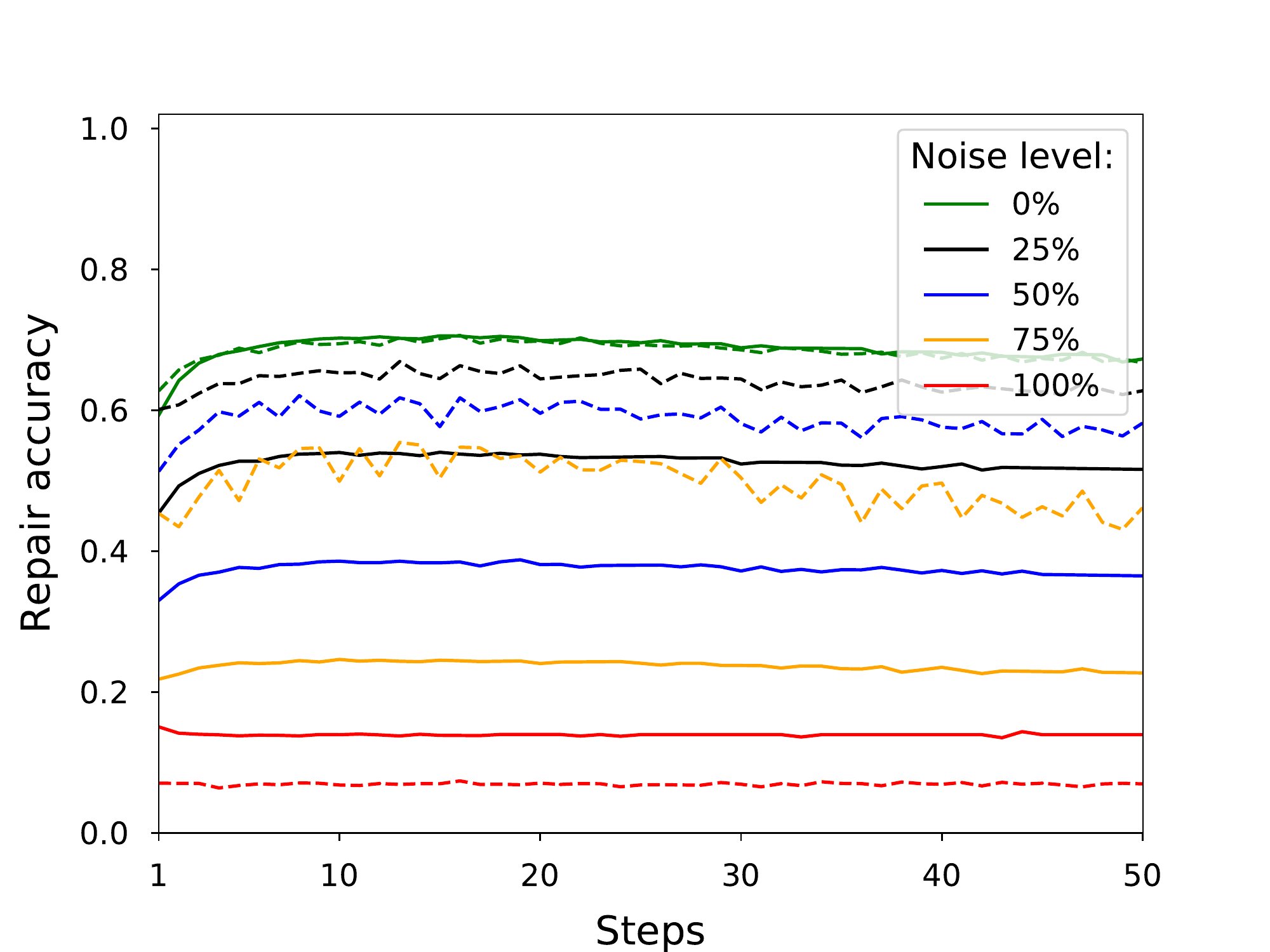}
  \caption{\ggnn (Repair)}
  \label{fig:vm_y_rep_acc_ggnn_py}
\end{subfigure}%
\begin{subfigure}{0.32\textwidth}
  \centering
  \includegraphics[width=\linewidth]{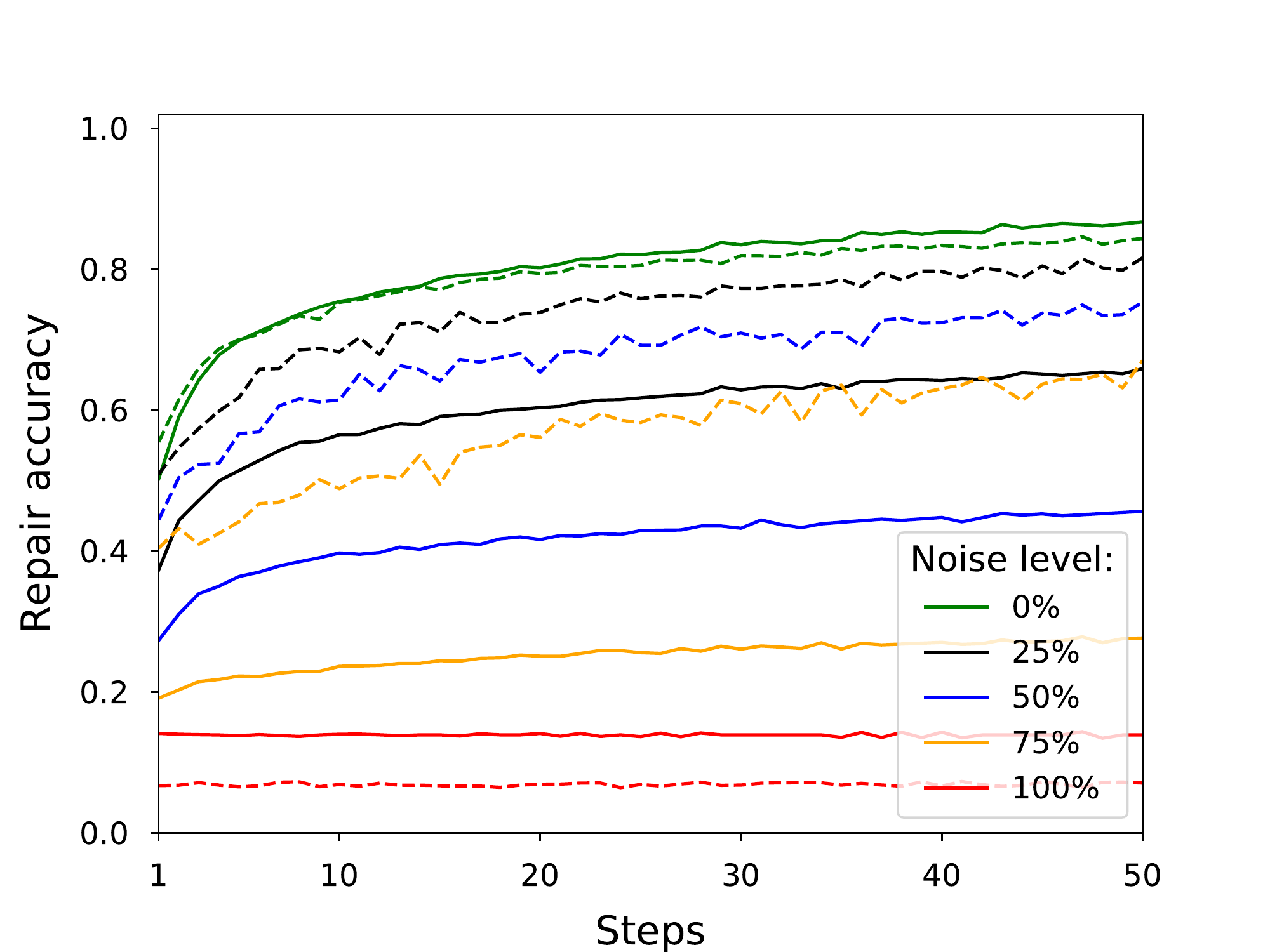}
  \caption{\great (Repair)}
  \label{fig:vm_y_rep_acc_great_py}
\end{subfigure}

\caption{Localization and Repair accuracy - solid is training, dashed is held-out (\vm).}
\label{fig:vm_y_acc_all}
\end{figure*}

\subsubsection{Memorization and Performance}
\label{subsec:noise-y-f1}

We tracked models' prediction in each epoch (or step) while training, after which we evaluate the same on the test (or held-out) data. \Cref{fig:mn_y_f1_all} shows the resulting changes in \FOneScore on the training set (solid line) and test set (dashed line) at different noise levels for the \mnp task; \Cref{fig:vm_y_acc_all} shows the same for both the localization and repair accuracy metrics on the \vm task.
Most models experience a significant impact on their training and test (or held-out) performance with the introduction of noise. Some models show higher test (or held-out) than training performance in the presence of noise; this might be due to the absence of noise in the test (or held-out) data.

Remarkably, the training performance of the \ctv models converges to nearly the same point in both the original and noisy settings; most noisy curves follow a nearly identical trajectory.
The remaining \cts models do not show this pattern on the same dataset, struggles more with the memorization required by the noisy samples and is unable to fit noisy training data perfectly. Similar to the \vm models, it instead experiences a consistent decline in training and test performance with increased noise levels. The \vm models fail to learn bug localization almost completely at noise levels of $75\%$ and above -- instead, we observed that these models converged to marking every sample as bug-free.

Reflecting back on \Cref{tab:model-conf}, we find a likely explanation for the discrepancy between \ctv and \cts on the same dataset: the former, used in its default configuration, has access to substantially more parameters than the latter ($10$x or more). Evidently, this more than suffices for memorizing previously seen training samples. This happens especially quickly on \JTT, the smallest of these datasets. Typically, we like to detect such \emph{memorization} by contrasting training and test (or held-out) performance. However, considering the gap between training and test (or held-out) performance across all sub-figures, it is not serving this purpose: this gap does not follow any specific patterns in most cases.
If one had only access to a dataset with an unknown degree of noise, as we normally do, knowing the test (or held-out) performance in relation to the training results would thus be of little use. On the contrary, explicitly contrasting the original data with a variant in which some non-trivial degree of noise is introduced immediately highlights the problem of a model with too many parameters.

\observation{
Increasing the degree of memorization generally causes a decrease in the training and generalization performance, except in models with excessive capacity: their training performance barely suffered, probably making this a strong indicator of memorization over generalization.}

\begin{figure*}
\centering
\captionsetup[subfigure]{width=0.9\textwidth, justification=centering}

\begin{subfigure}{0.32\textwidth}
  \centering
  \includegraphics[width=\linewidth]{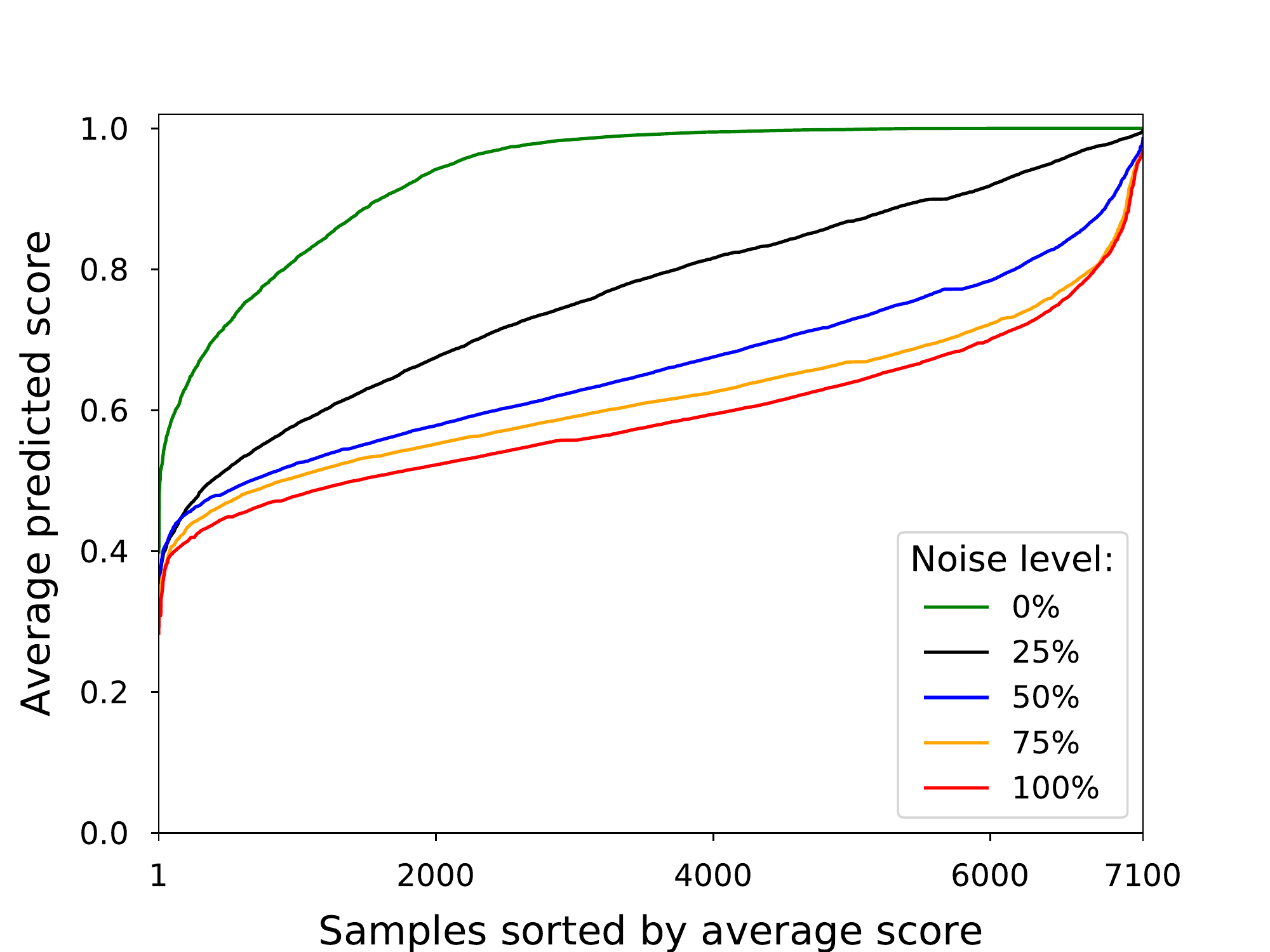}
  \caption{\ctv{} (\JTT)}
  \label{fig:mn_y_avg_score_test_c2v_top10}
\end{subfigure}%
\begin{subfigure}{0.32\textwidth}
  \centering
  \includegraphics[width=\linewidth]{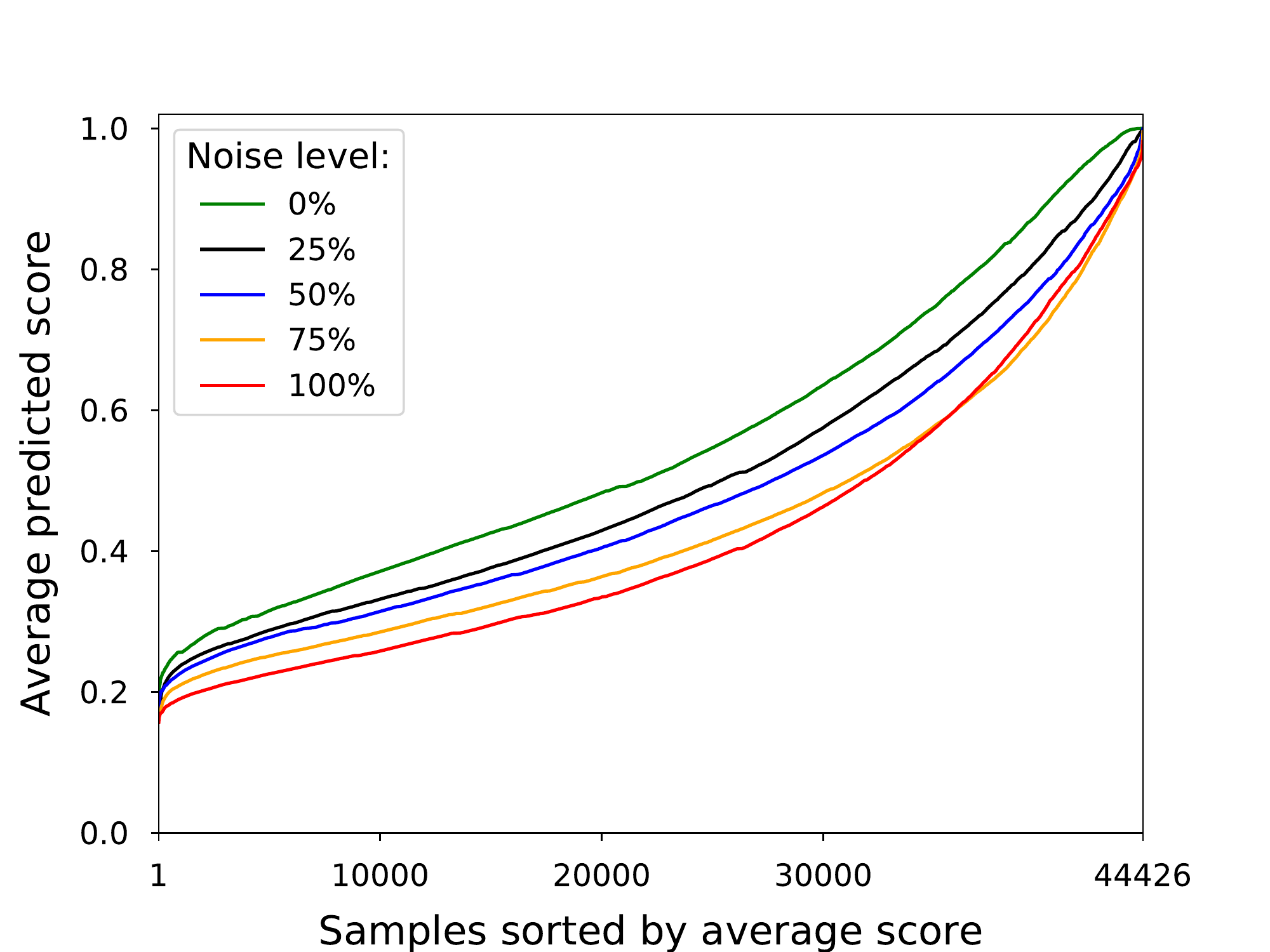}
  \caption{\ctv{} (\JS)}
  \label{fig:mn_y_avg_score_test_c2v_js}
\end{subfigure}%
\begin{subfigure}{0.32\textwidth}
  \centering
  \includegraphics[width=\linewidth]{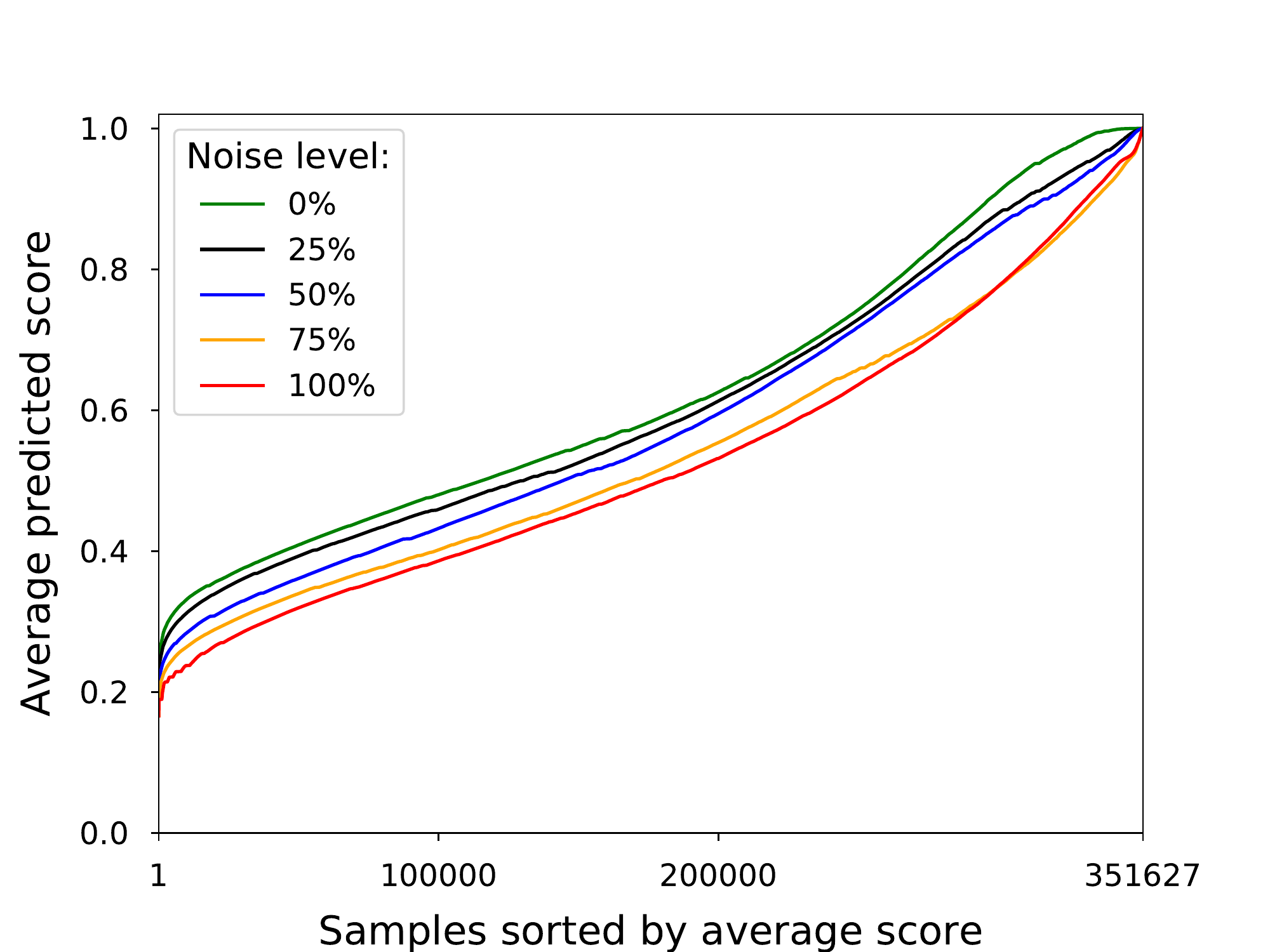}
  \caption{\ctv{} (\JM)}
  \label{fig:mn_y_avg_score_test_c2v_jm}
\end{subfigure}

\begin{subfigure}{0.32\textwidth}
  \centering
  \includegraphics[width=\linewidth]{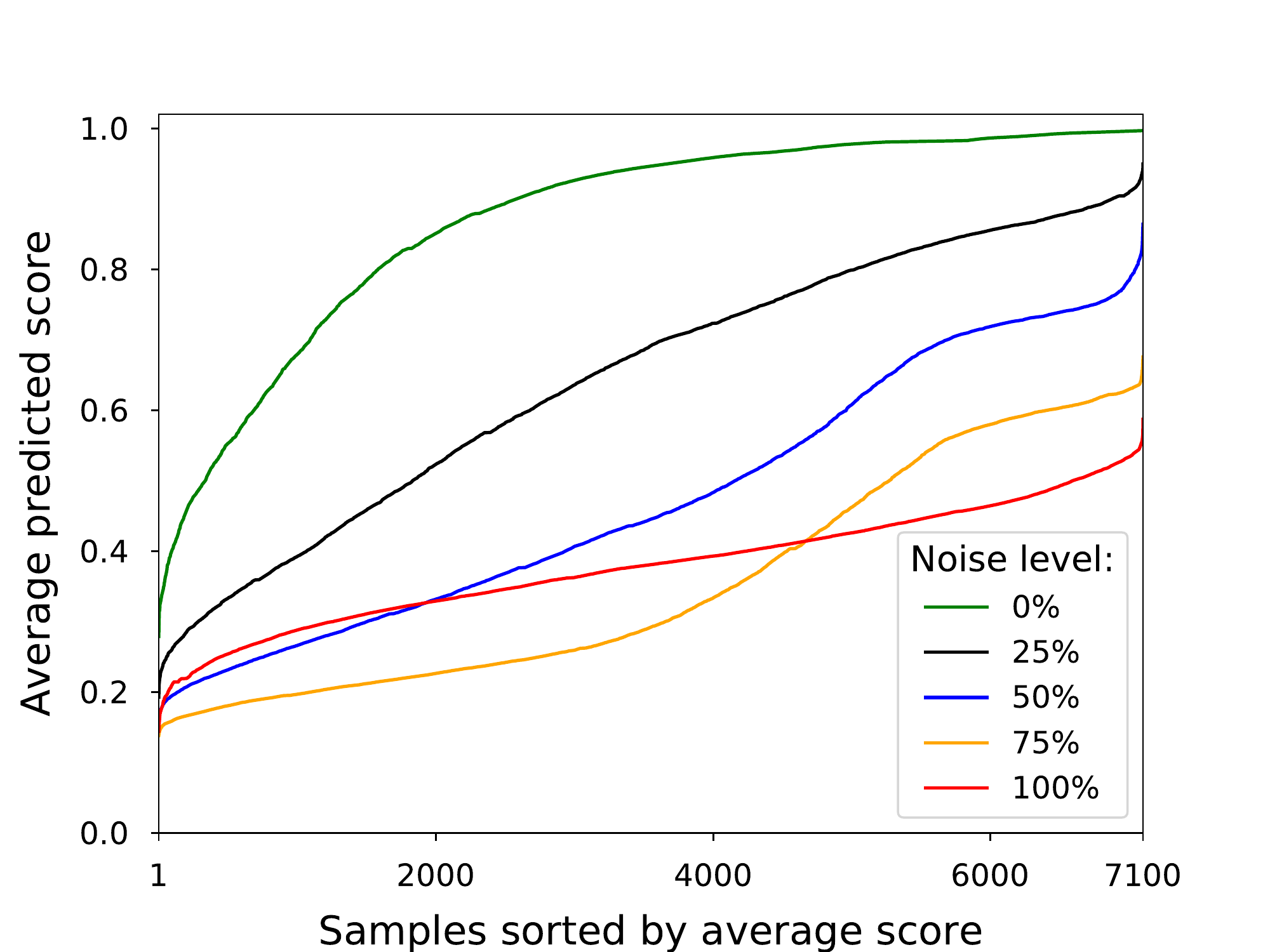}
  \caption{\cts{} (\JTT)}
  \label{fig:mn_y_avg_score_test_c2s_top10}
\end{subfigure}%
\begin{subfigure}{0.32\textwidth}
  \centering
  \includegraphics[width=\linewidth]{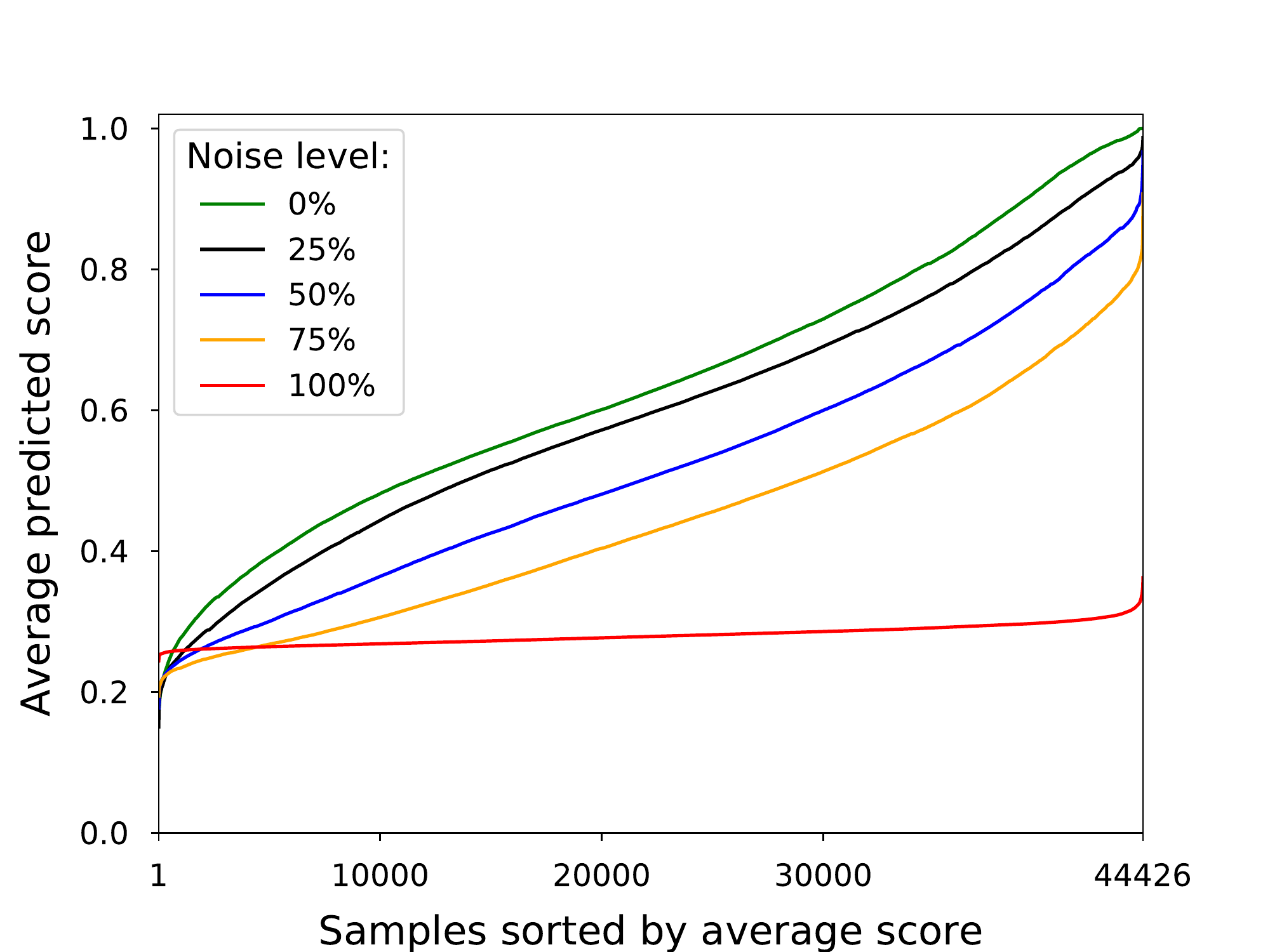}
  \caption{\cts{} (\JS)}
  \label{fig:mn_y_avg_score_test_c2s_js}
\end{subfigure}%
\begin{subfigure}{0.32\textwidth}
  \centering
  \includegraphics[width=\linewidth]{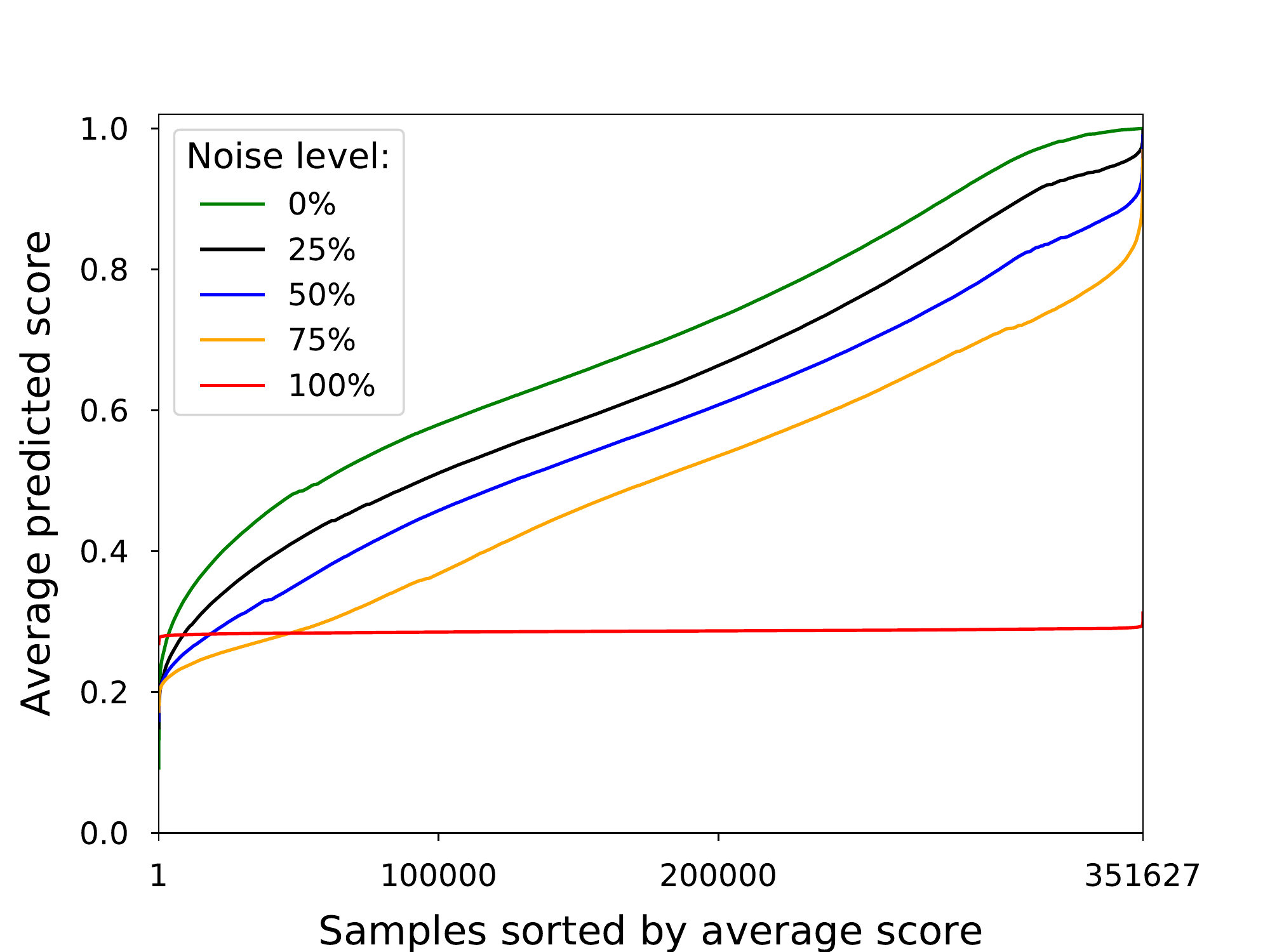}
  \caption{\cts{} (\JM)}
  \label{fig:mn_y_avg_score_test_c2s_jm}
\end{subfigure}

\caption{Distribution of prediction score considering all epochs (\mnp).}
\label{fig:mn_y_avg_score_test}


\begin{subfigure}{0.32\textwidth}
  \centering
  \includegraphics[width=\linewidth]{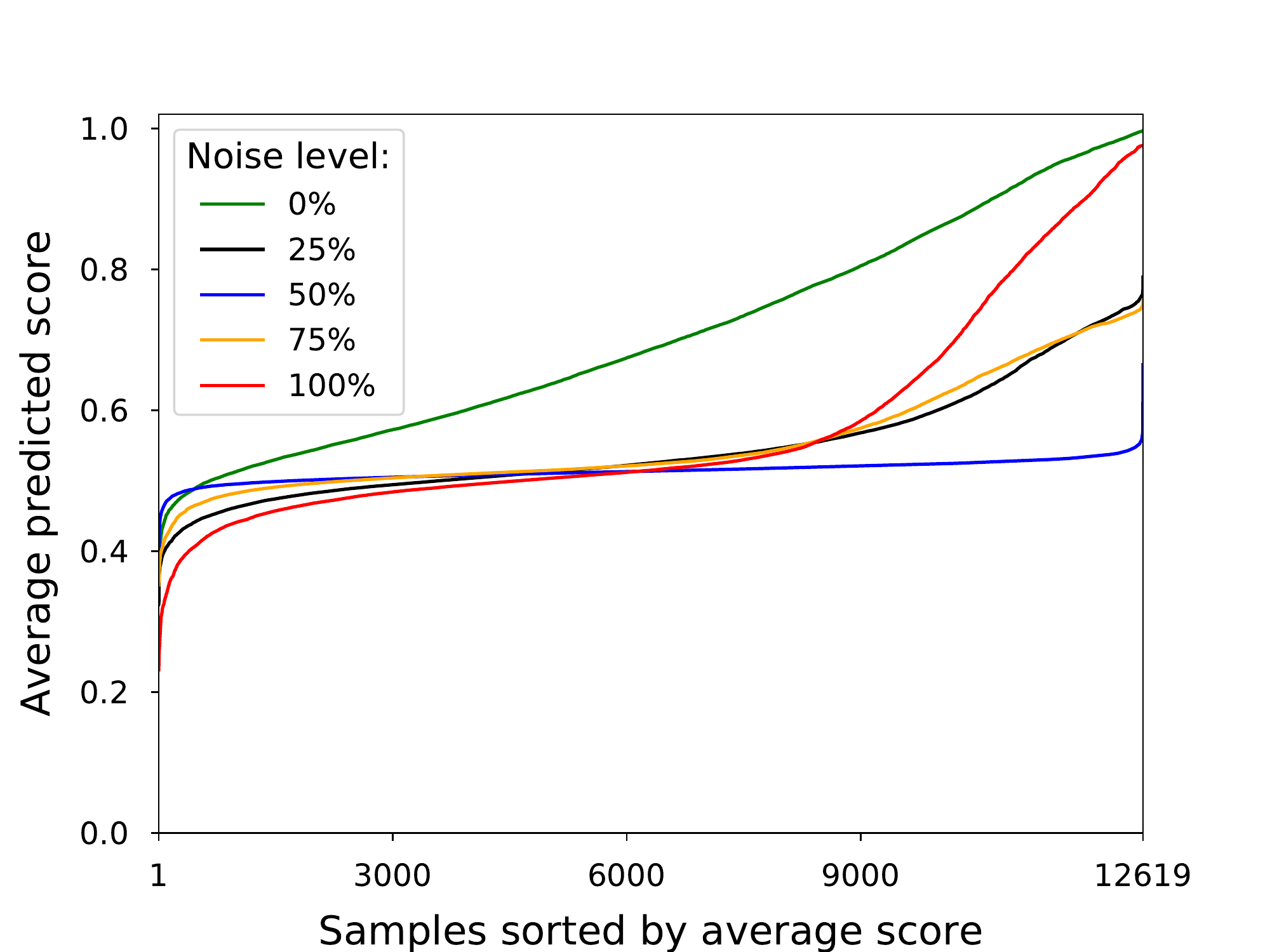}
  \caption{\tra (Localization)}
  \label{fig:vm_y_avg_loc_score_tra_py}
\end{subfigure}%
\begin{subfigure}{0.32\textwidth}
  \centering
  \includegraphics[width=\linewidth]{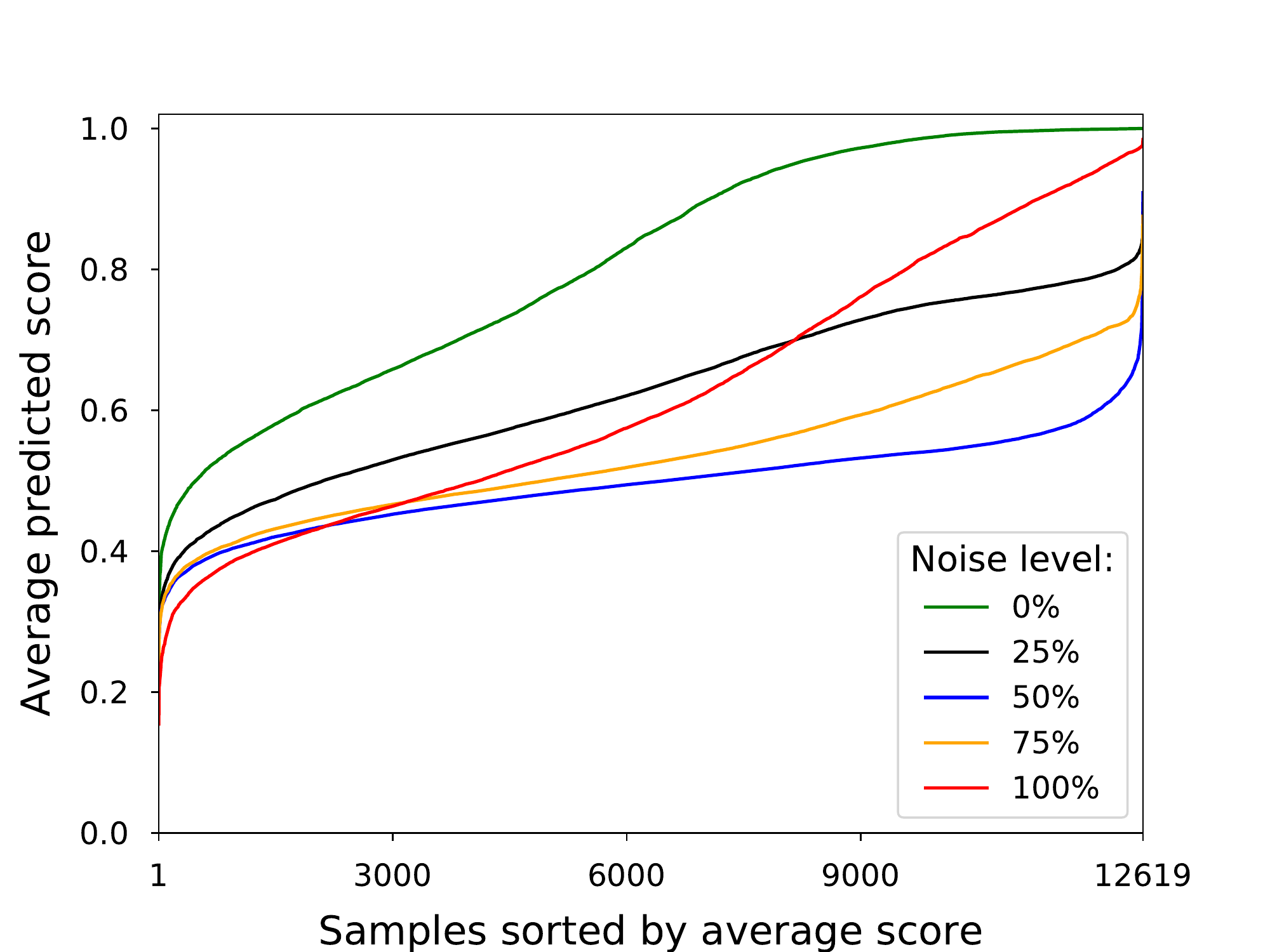}
  \caption{\ggnn (Localization)}
  \label{fig:vm_y_avg_loc_score_ggnn_py}
\end{subfigure}%
\begin{subfigure}{0.32\textwidth}
  \centering
  \includegraphics[width=\linewidth]{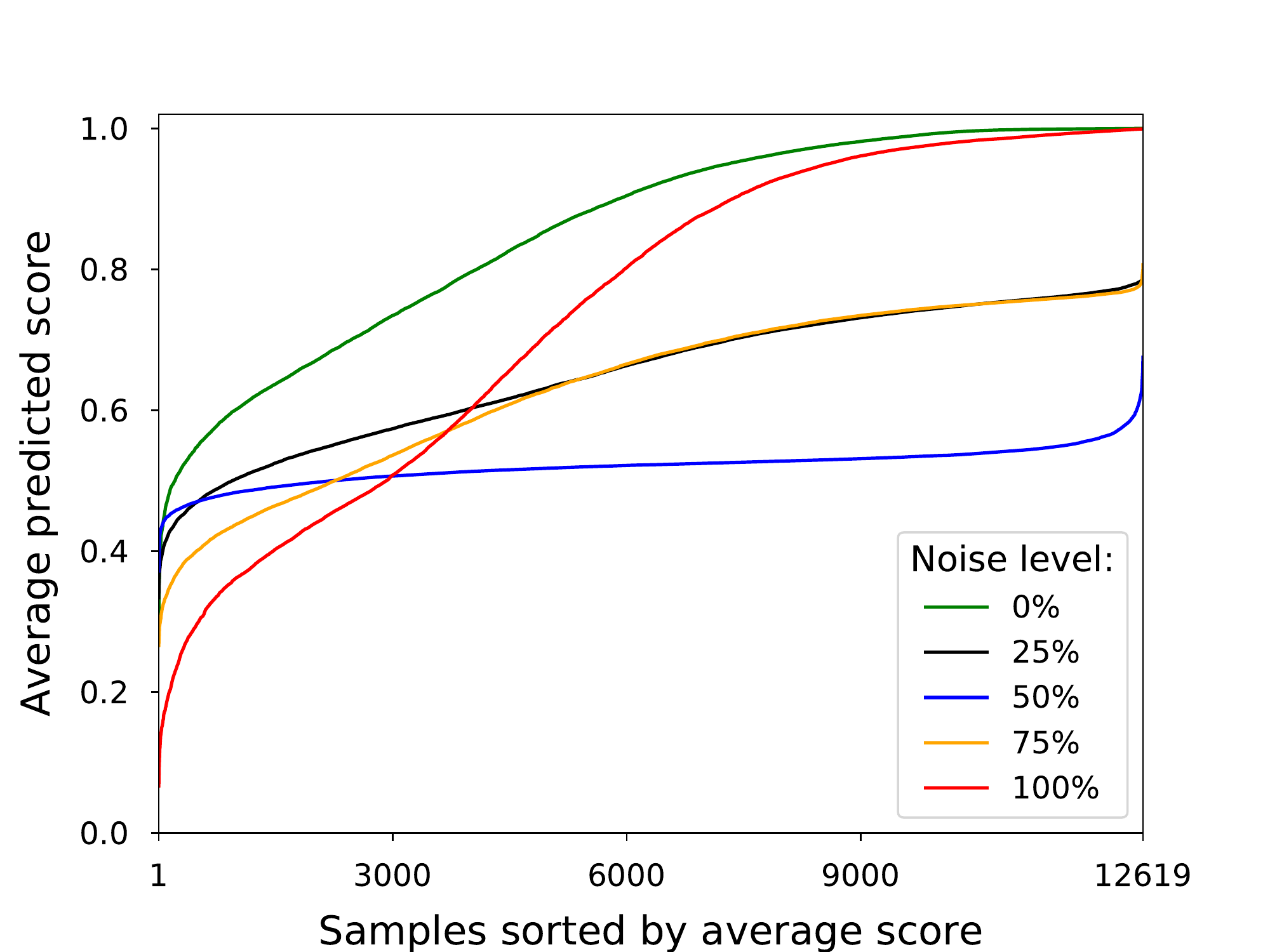}
  \caption{\great (Localization)}
  \label{fig:vm_y_avg_loc_score_great_py}
\end{subfigure}

\begin{subfigure}{0.32\textwidth}
  \centering
  \includegraphics[width=\linewidth]{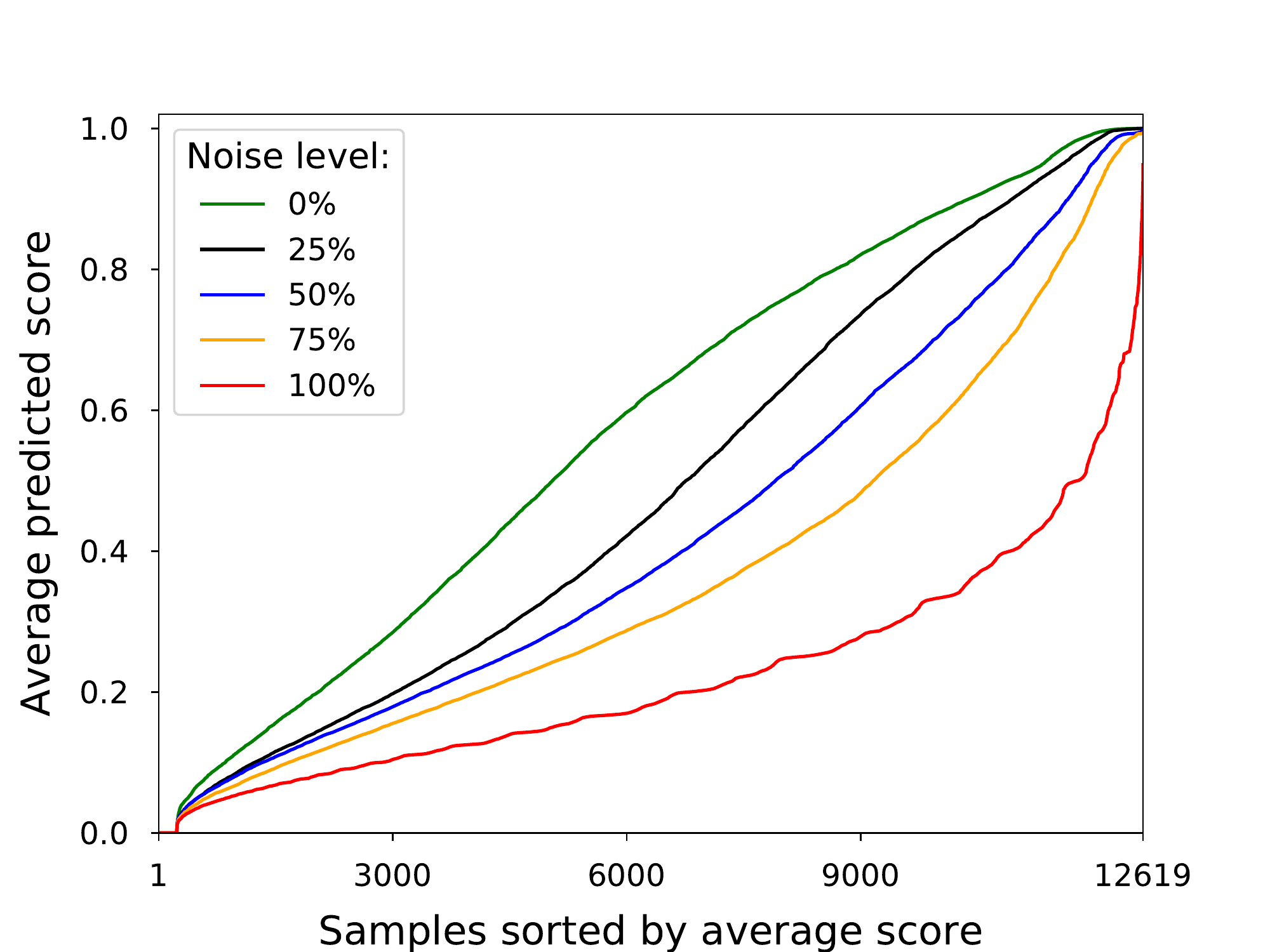}
  \caption{\tra (Repair)}
  \label{fig:vm_y_avg_rep_score_tra_py}
\end{subfigure}%
\begin{subfigure}{0.32\textwidth}
  \centering
  \includegraphics[width=\linewidth]{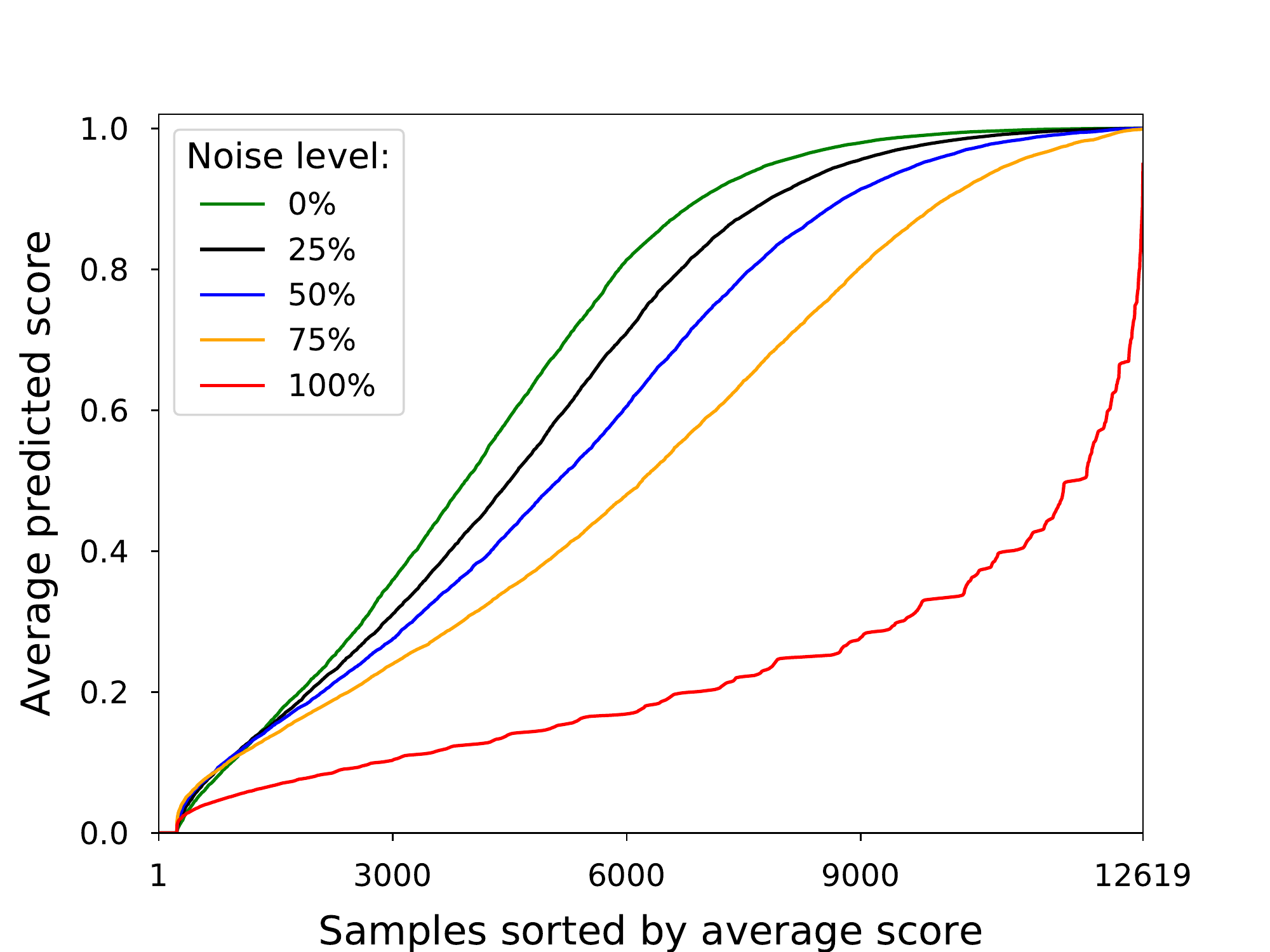}
  \caption{\ggnn (Repair)}
  \label{fig:vm_y_avg_rep_score_ggnn_py}
\end{subfigure}%
\begin{subfigure}{0.32\textwidth}
  \centering
  \includegraphics[width=\linewidth]{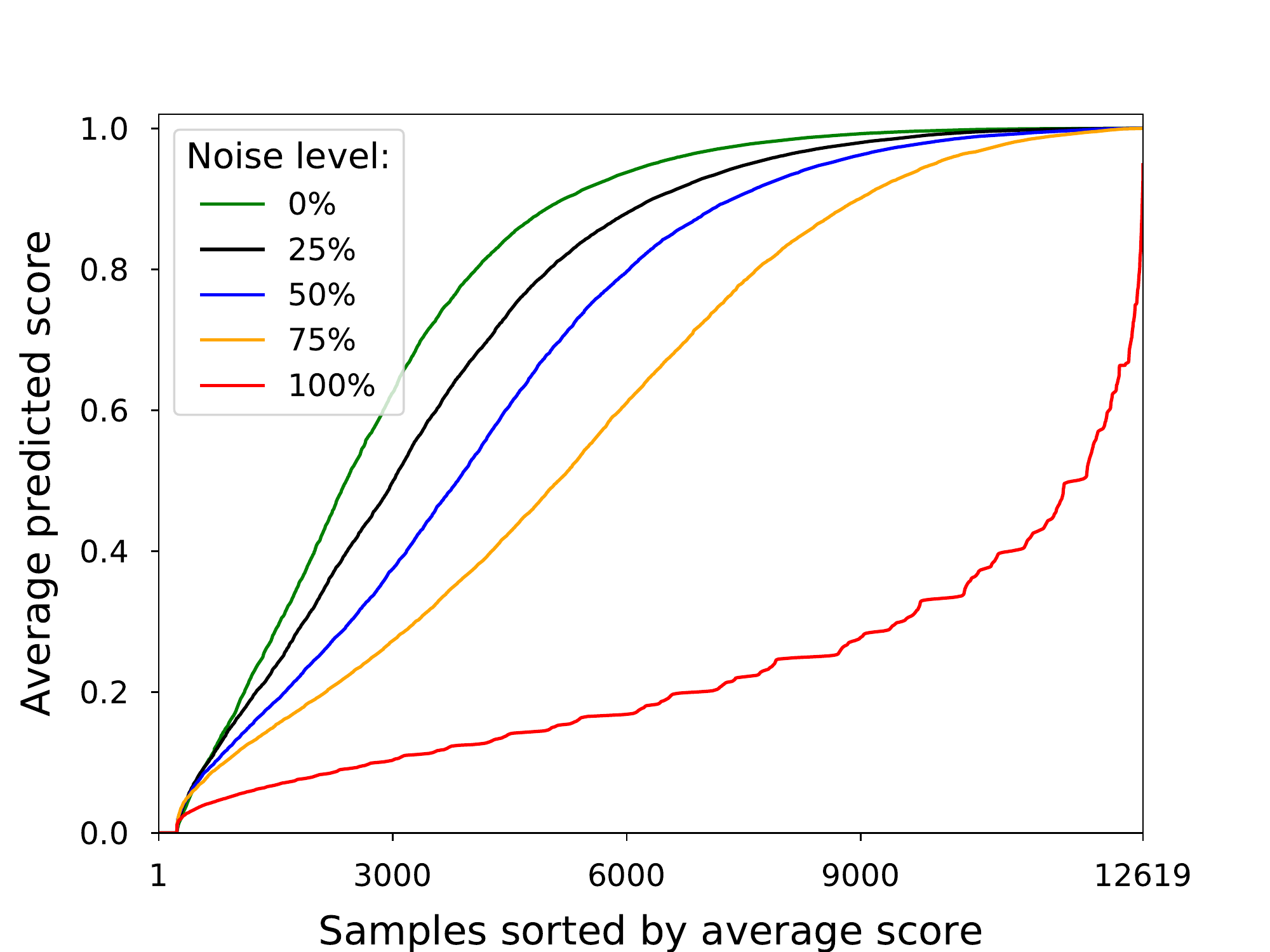}
  \caption{\great (Repair)}
  \label{fig:vm_y_avg_rep_score_great_py}
\end{subfigure}

\caption{Distribution of prediction score considering all steps (\vm).}
\label{fig:vm_y_avg_score_dev}
\end{figure*}

\subsubsection{Memorization and Prediction Confidence}
\label{subsec:noise-y-score}

Some instances fit highly predictive patterns better than others. As a result, easily predicted samples will have (much) higher associated probability scores than harder samples. This difference in scores should be quite pronounced when comparing original data (some instances fit patterns) with noisy data (fits independently) \cite{arpit2017closer}.
In order to observe these behaviors in \cis, we obtained the probability assigned to the predicted label for all samples in the test set (or held-out) after each epoch (or step).
\Cref{fig:mn_y_avg_score_test,fig:vm_y_avg_score_dev} show the sorted average probability of the predicted label for all samples considering all epochs (or steps) of training.

The results show that the distribution of prediction score changes significantly in shape as the amount of noise increases. Whereas models nearly follow a convex trend on their original dataset, in which only a small number of samples are predicted with low confidence, the introduction of noise changes this abruptly. Models with even low rates of noise are far more likely to show a concave distribution with mostly conservative, or even virtually uniform, probabilities, such as at $100\%$ noise in \Cref{fig:mn_y_avg_score_test_c2s_js,fig:mn_y_avg_score_test_c2s_jm}. Models needing to memorize rarely expressed high confidence in any predictions. This is to be expected; highly confident mispredictions would incur a very high loss penalty during training. This effect is again particularly weak on the \ctv tasks, on which we previously established ample capacity for memorizing.

In some cases, especially for localization probability in \Cref{fig:vm_y_avg_score_dev}$_{a-c}$, a phase transition occurs where models trained with very high rates of noise regain some confidence on a subset of samples. 
We hypothesize that this reflects cases where these models have confidently memorized --- which is perhaps easier to do in the absence of the ambiguity that comes with having both noisy and noise-free samples.
Some other figures show a surprisingly small gap between the 25\% and 0\% trends, among them, \Cref{fig:mn_y_avg_score_test_c2v_jm,fig:mn_y_avg_score_test_c2s_js} from \mnp. This may signal some degree of noise already present in the original dataset.

\observation{
Even at low rates of memorization, the distribution of prediction scores assigned by the model changes drastically, reflecting a decrease in confidence. This transition can be useful for detecting memorization characteristics.
}

\begin{figure*}
\centering
\captionsetup[subfigure]{width=0.9\textwidth, justification=centering}

\begin{subfigure}{0.32\textwidth}
  \centering
  \includegraphics[width=\linewidth]{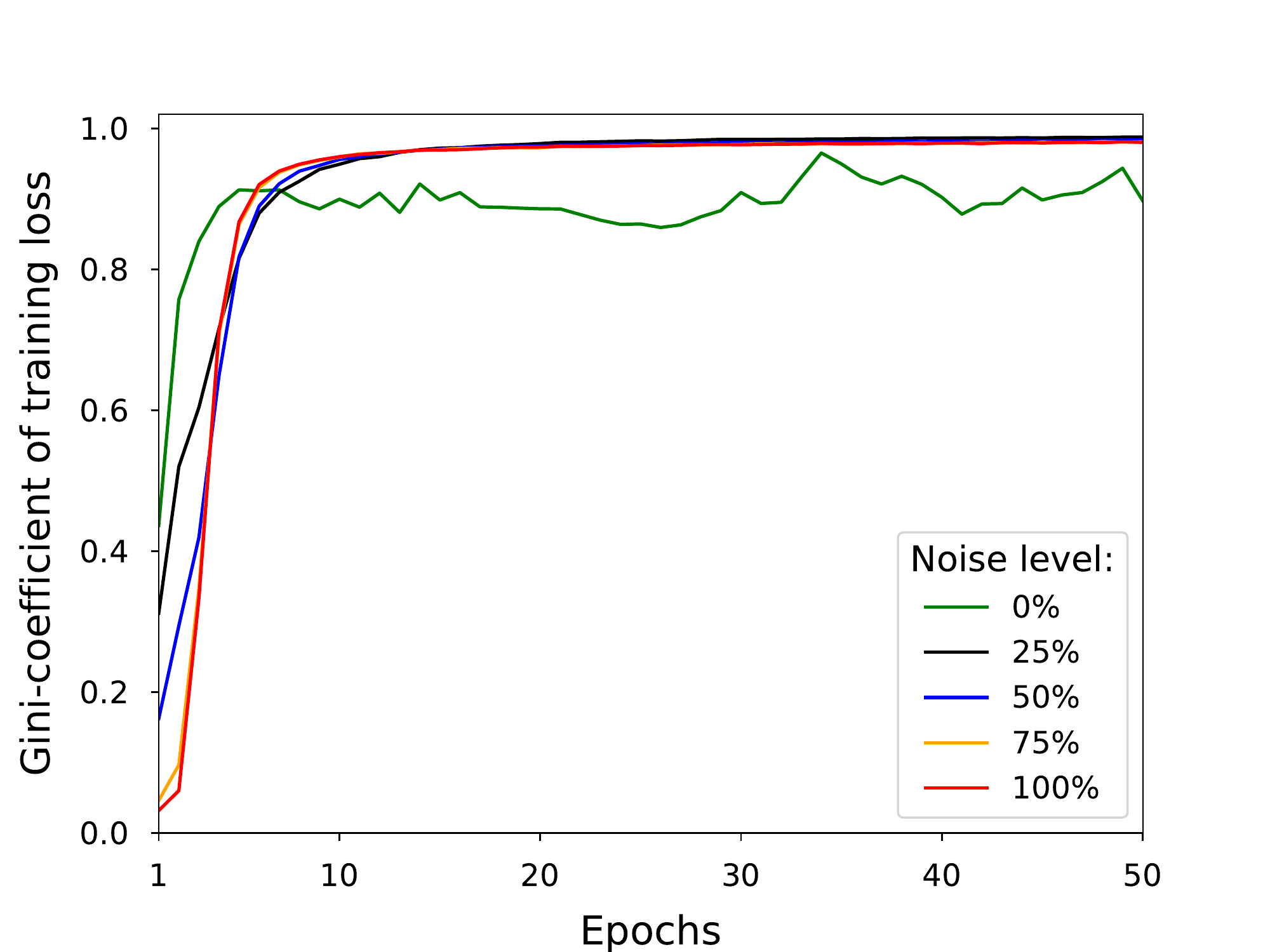}
  \caption{\ctv{} (\JTT)}
  \label{fig:mn_y_gini_loss_training_c2v_top10}
\end{subfigure}%
\begin{subfigure}{0.32\textwidth}
  \centering
  \includegraphics[width=\linewidth]{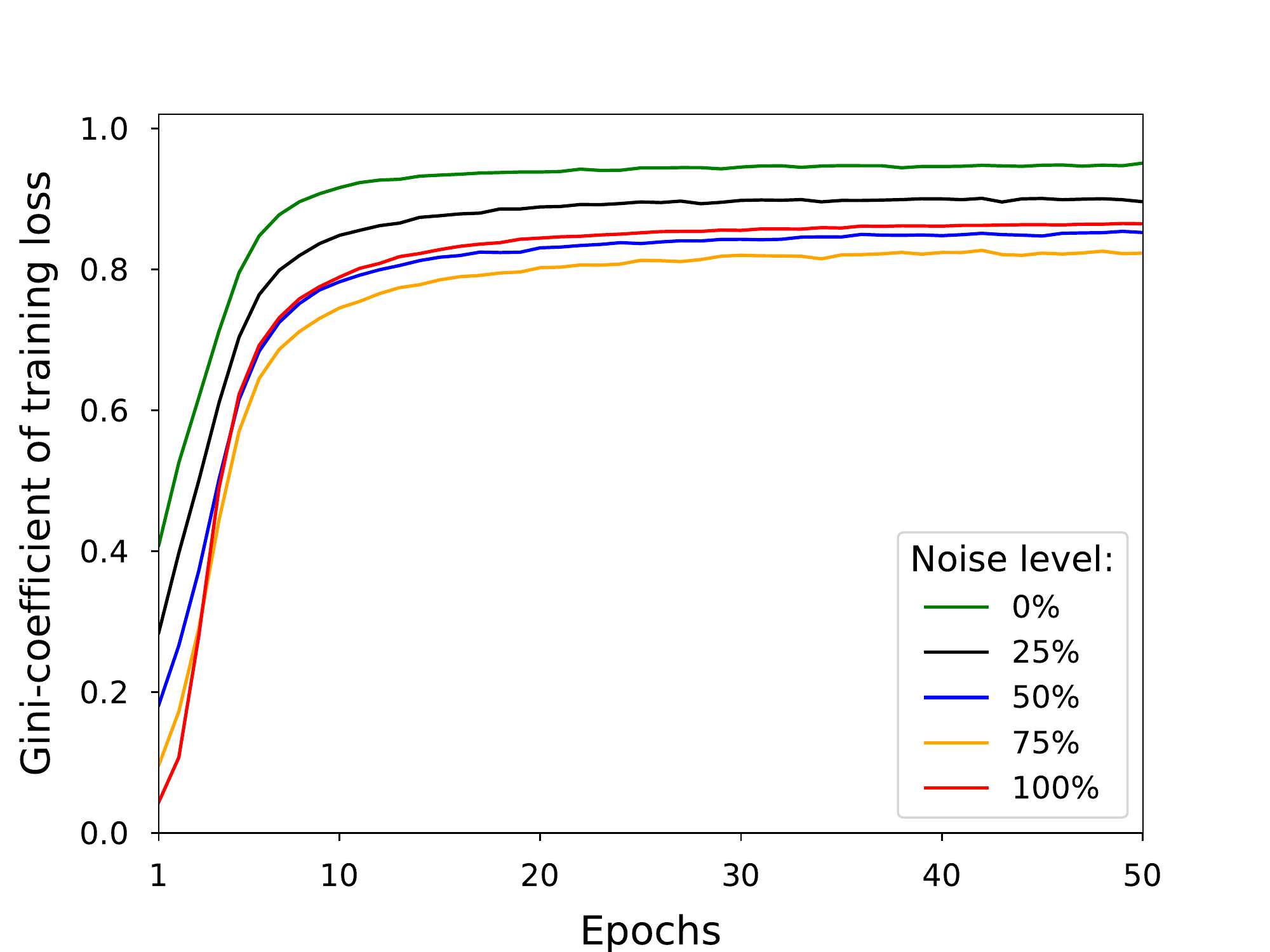}
  \caption{\ctv{} (\JS)}
  \label{fig:mn_y_gini_loss_training_c2v_js}
\end{subfigure}%
\begin{subfigure}{0.32\textwidth}
  \centering
  \includegraphics[width=\linewidth]{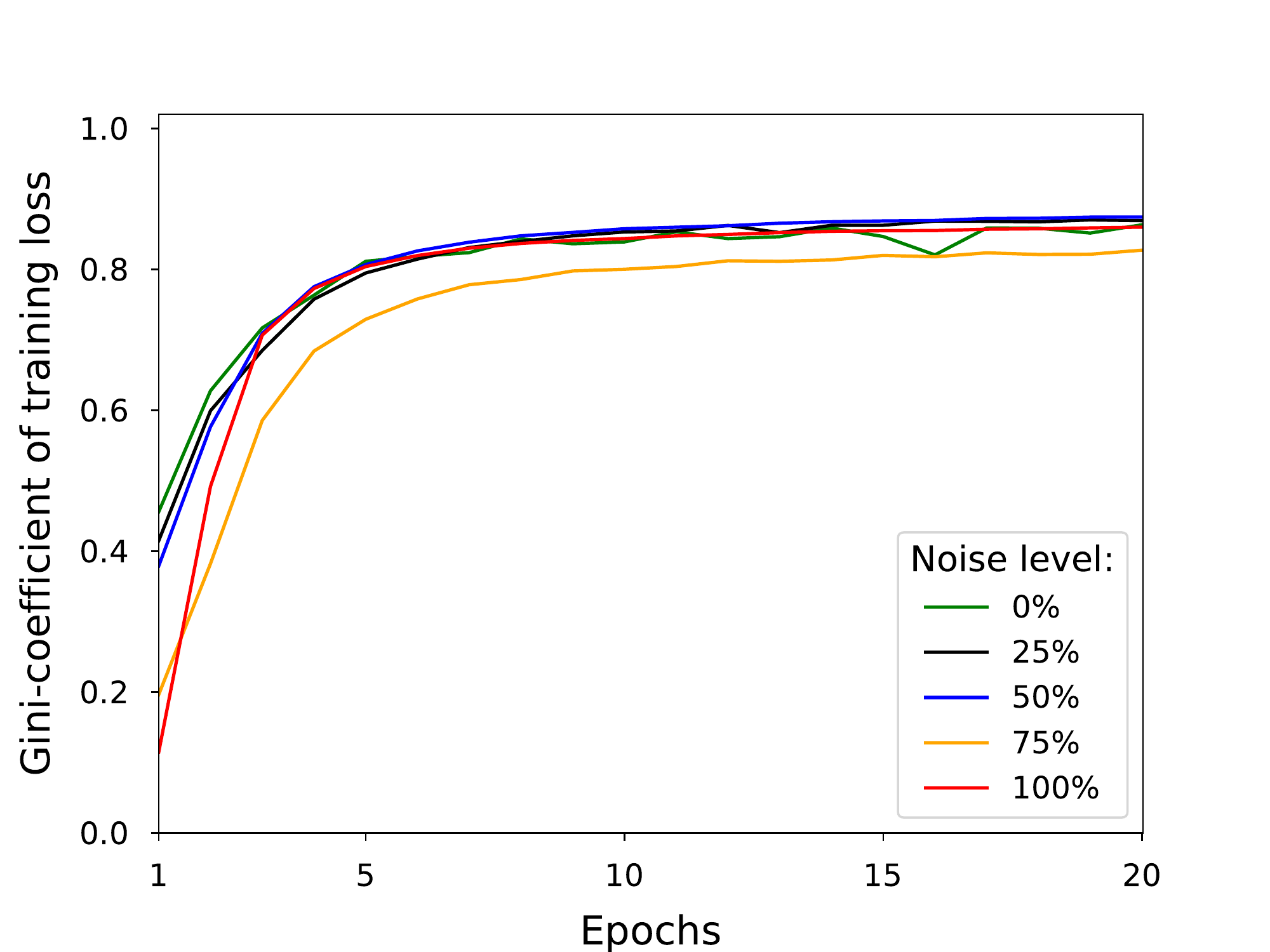}
  \caption{\ctv{} (\JM)}
  \label{fig:mn_y_gini_loss_training_c2v_jm}
\end{subfigure}

\begin{subfigure}{0.32\textwidth}
  \centering
  \includegraphics[width=\linewidth]{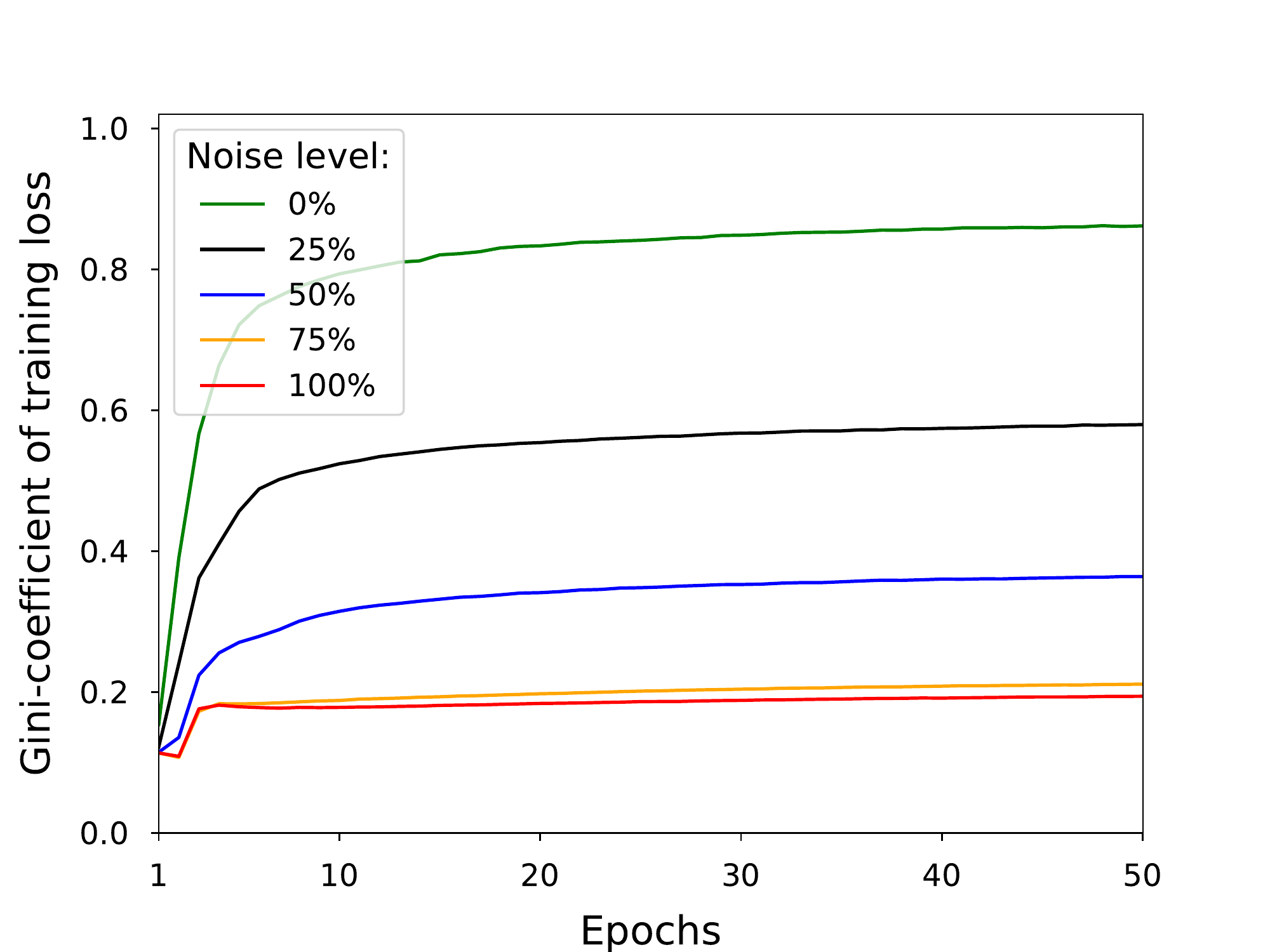}
  \caption{\cts{} (\JTT)}
  \label{fig:mn_y_gini_loss_training_c2s_top10}
\end{subfigure}%
\begin{subfigure}{0.32\textwidth}
  \centering
  \includegraphics[width=\linewidth]{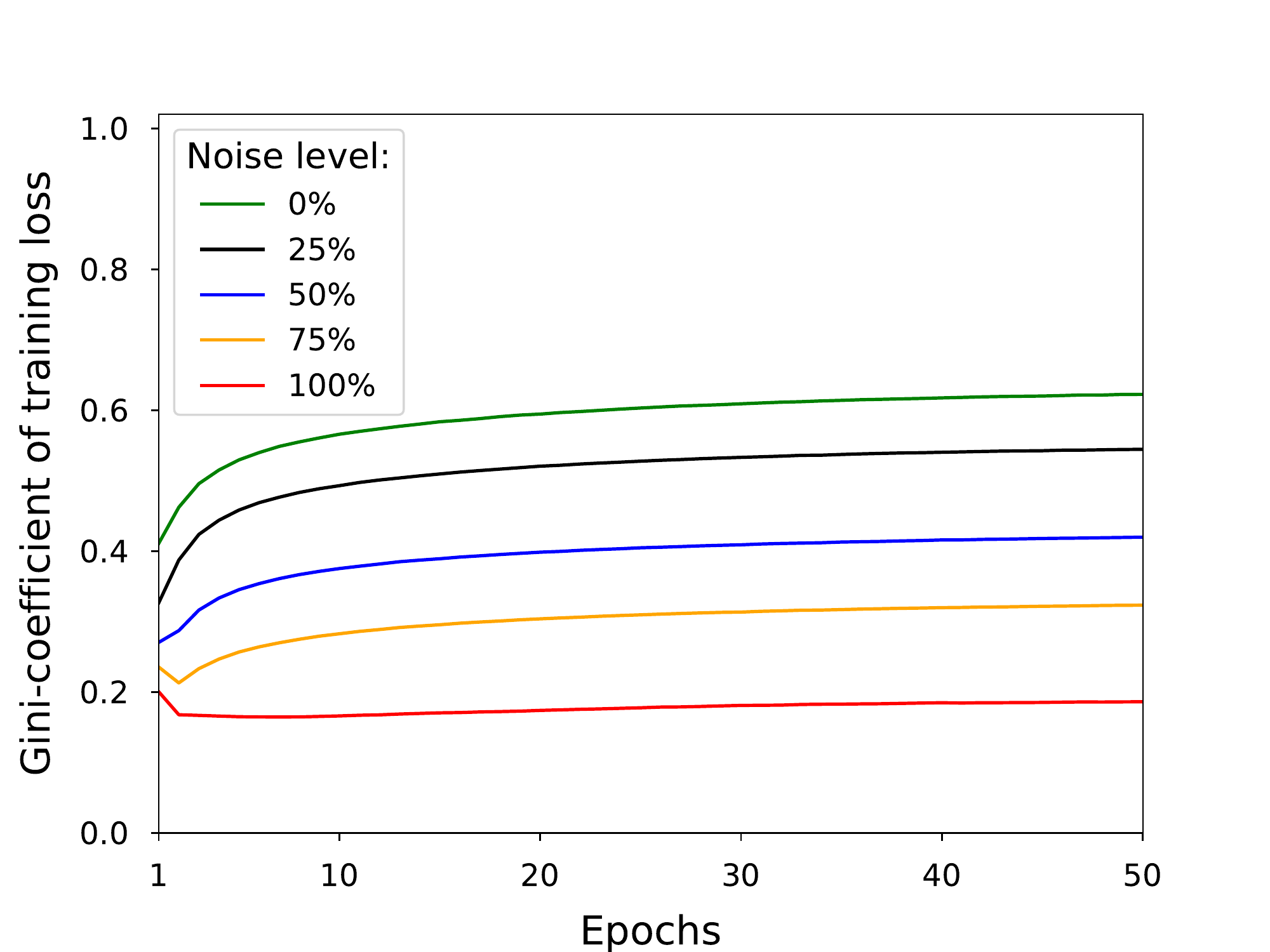}
  \caption{\cts{} (\JS)}
  \label{fig:mn_y_gini_loss_training_c2s_js}
\end{subfigure}%
\begin{subfigure}{0.32\textwidth}
  \centering
  \includegraphics[width=\linewidth]{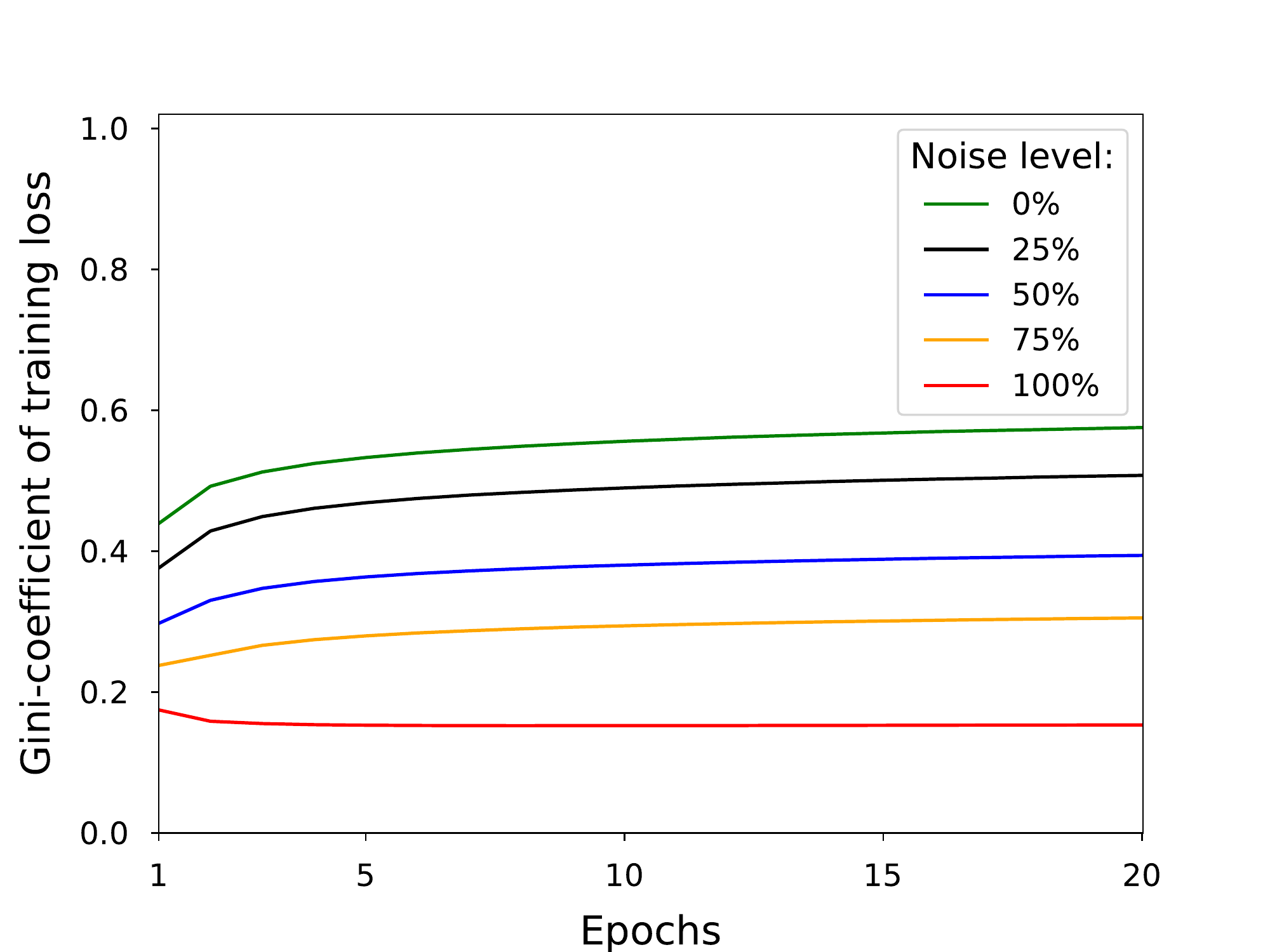}
  \caption{\cts{} (\JM)}
  \label{fig:mn_y_gini_loss_training_c2s_jm}
\end{subfigure}

\caption{Spread of training loss after each epoch (\mnp).}
\label{fig:mn_y_gini_loss_training}


\begin{subfigure}{0.32\textwidth}
  \centering
  \includegraphics[width=\linewidth]{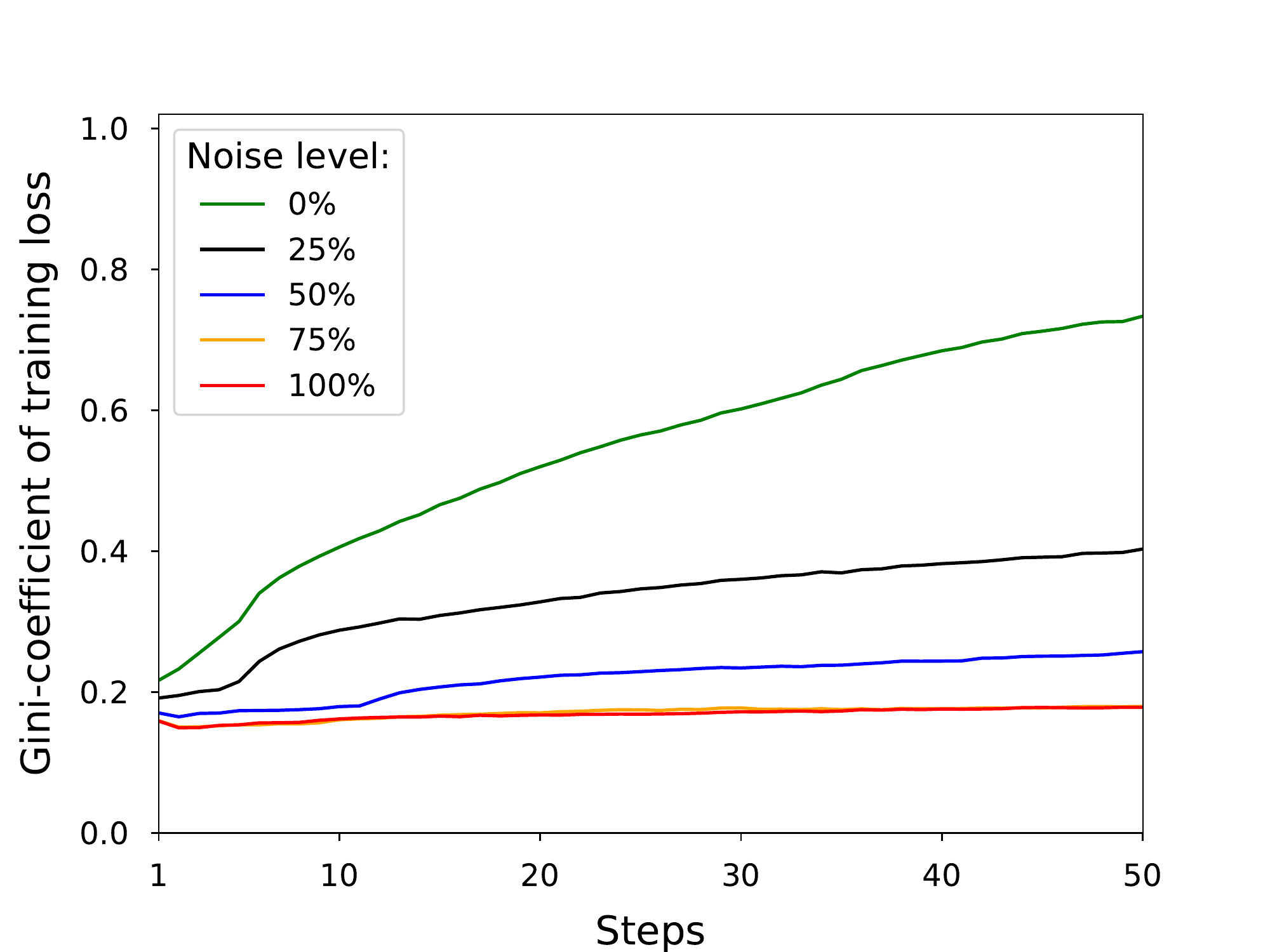}
  \caption{\tra (Localization)}
  \label{fig:vm_y_train_loc_loss_tra_py}
\end{subfigure}%
\begin{subfigure}{0.32\textwidth}
  \centering
  \includegraphics[width=\linewidth]{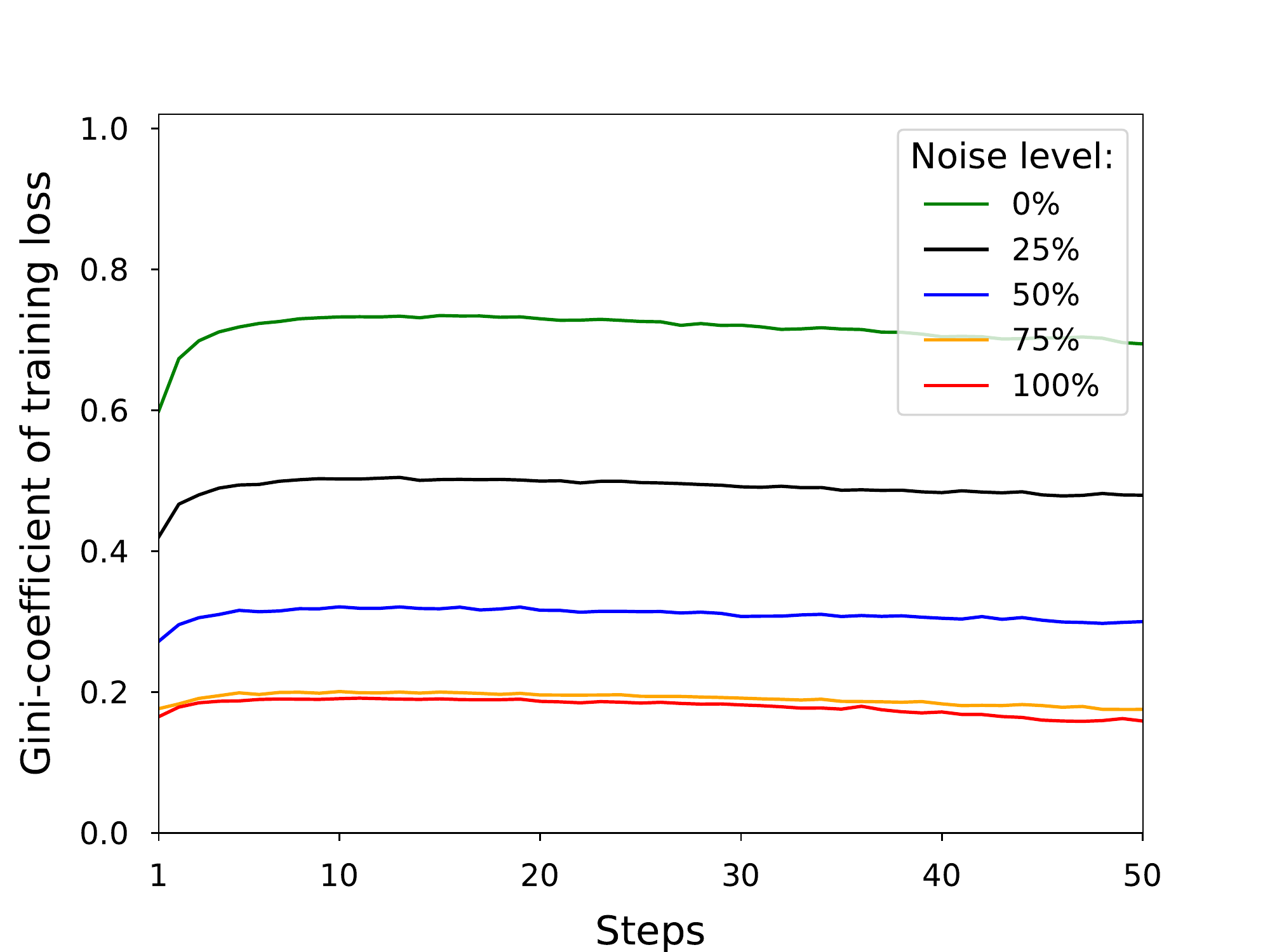}
  \caption{\ggnn (Localization)}
  \label{fig:vm_y_train_loc_loss_ggnn_py}
\end{subfigure}%
\begin{subfigure}{0.32\textwidth}
  \centering
  \includegraphics[width=\linewidth]{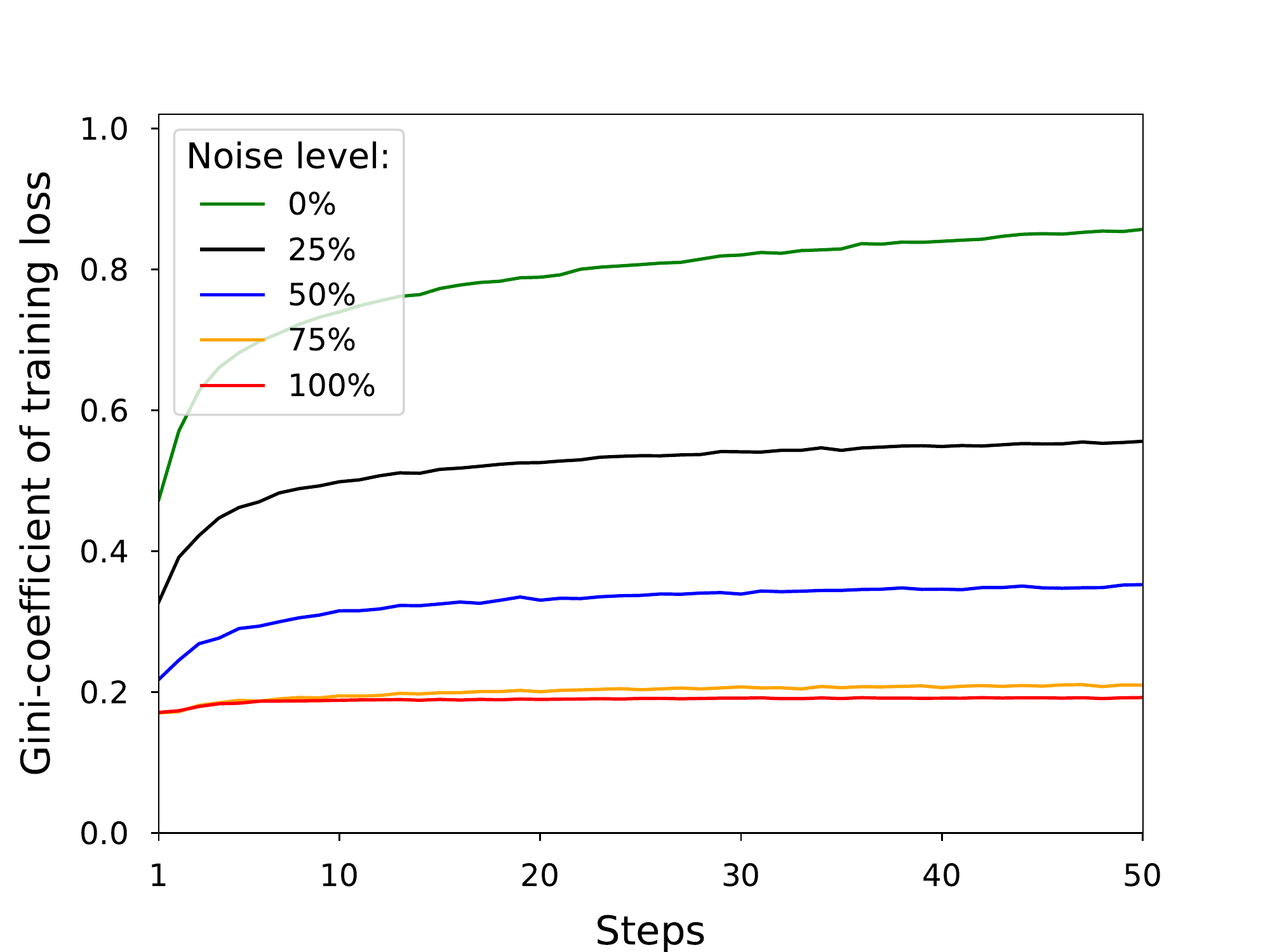}
  \caption{\great (Localization)}
  \label{fig:vm_y_train_loc_loss_great_py}
\end{subfigure}

\begin{subfigure}{0.32\textwidth}
  \centering
  \includegraphics[width=\linewidth]{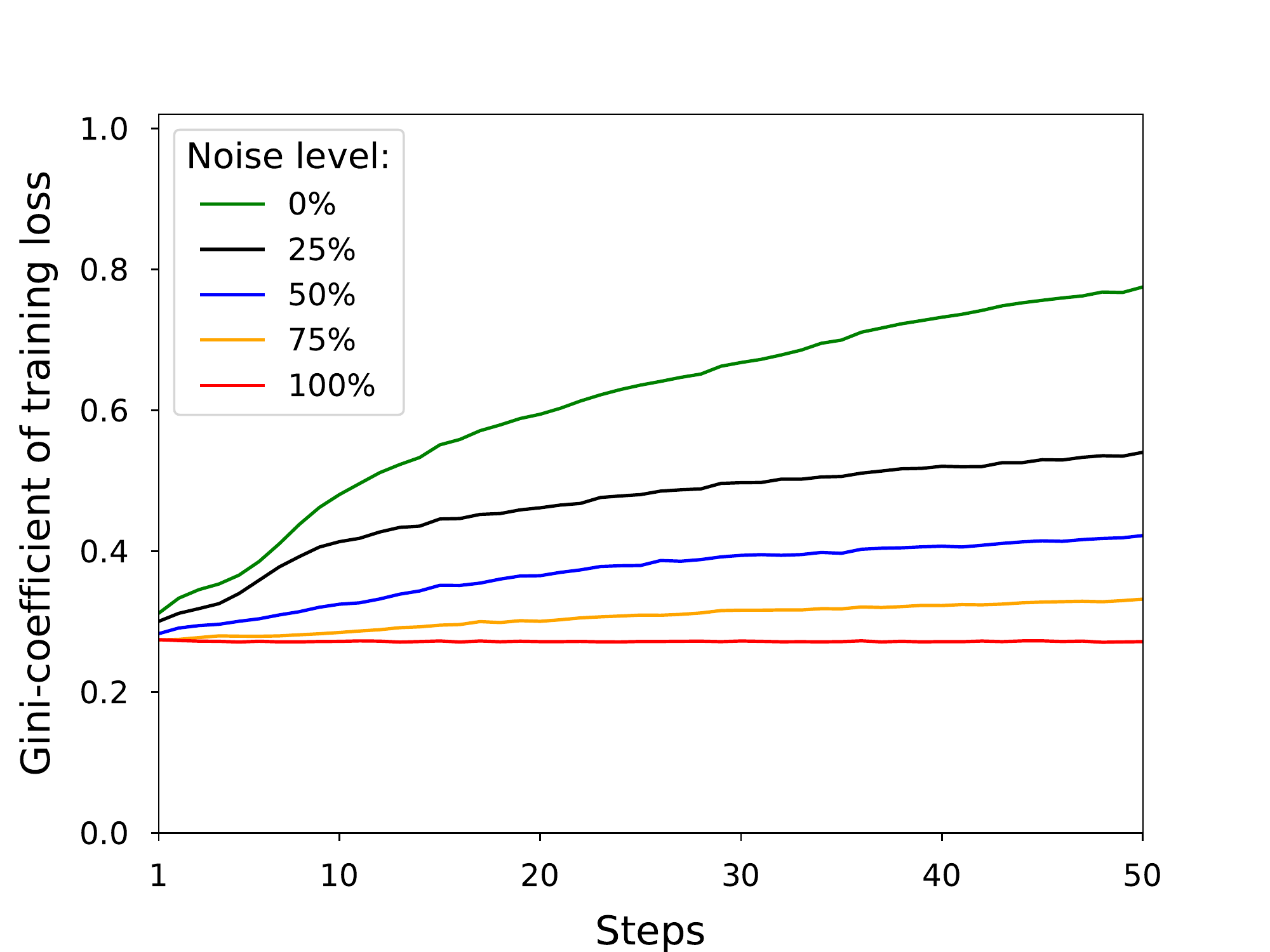}
  \caption{\tra (Repair)}
  \label{fig:vm_y_train_rep_loss_tra_py}
\end{subfigure}%
\begin{subfigure}{0.32\textwidth}
  \centering
  \includegraphics[width=\linewidth]{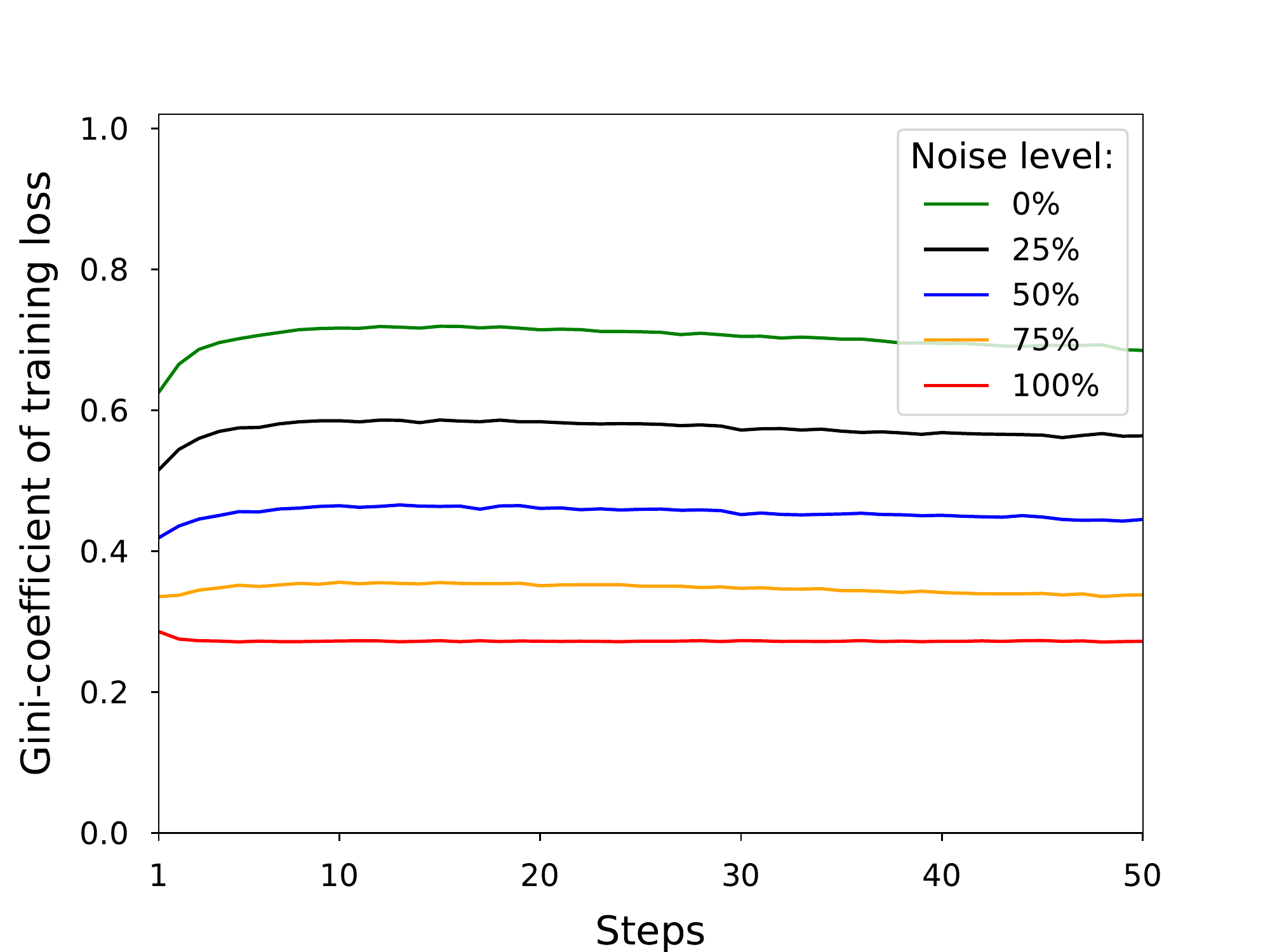}
  \caption{\ggnn (Repair)}
  \label{fig:vm_y_train_rep_loss_ggnn_py}
\end{subfigure}%
\begin{subfigure}{0.32\textwidth}
  \centering
  \includegraphics[width=\linewidth]{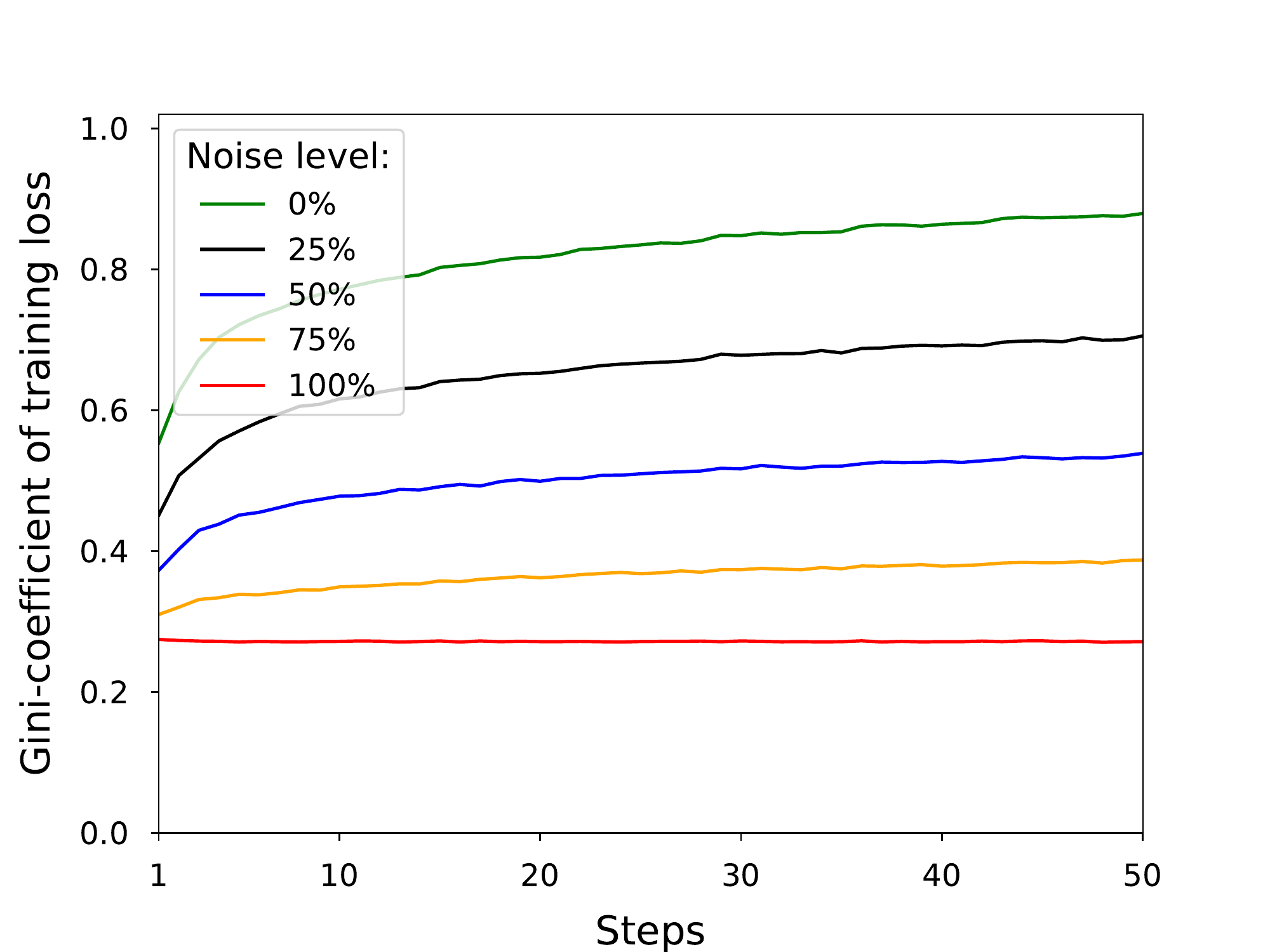}
  \caption{\great (Repair)}
  \label{fig:vm_y_train_rep_loss_great_py}
\end{subfigure}

\caption{Spread of training loss after each step (\vm).}
\label{fig:vm_y_gini_loss_training}
\end{figure*}

\subsubsection{Memorization and Loss Change}
\label{subsec:noise-y-loss}

We now study the spread of loss over the course of training and observe the difference between noisy data and original data. To do so, we compute the Gini coefficient (see \cref{sec:spreadloss}) across training samples, which captures the spread of the models' losses.
We know that only a small number of training samples have a high loss in the original data, while the loss tends to be high for virtually all samples in noisy data.
Therefore, the spread of loss (computed with the Gini coefficient) ought to be significantly higher (indicating less uniformity) in the original data than in the noisy data \cite{arpit2017closer}.
\Cref{fig:mn_y_gini_loss_training,fig:vm_y_gini_loss_training} show that this is indeed the case across virtually all tasks and models, expect for the \ctv models where basically all noise levels nearly yield the same trends.

In most cases, Gini coefficient value decreases significantly and rather consistently with an increasing noise level. The curves corresponding to various noise levels of the same model tend to follow a similar shape across training epochs (or steps) that mainly differs in the eventual spread of loss. This further reinforces that even small degrees of memorization can significantly alter the model's training pattern, and shows that it already does so at the very start, with no apparent change of trajectory over time. We also see a near-universal trend in which the more a model increases its Gini coefficient, the better it performs. For instance, both the \tra and \great models of \vm task eclipse the \ggnn model in the latter part of training; the latter shows a marked downward trend in Gini coefficient (\Cref{fig:vm_y_gini_loss_training}) right as its accuracy decreases as well (\Cref{fig:vm_y_acc_all}).
In the same way, the \cts model of \mnp task on the \JTT dataset shows better performance than both the \JS and \JM datasets (\Cref{fig:mn_y_f1_all}); the former shows an upward trend in the Gini coefficient (\Cref{fig:mn_y_gini_loss_training}).
However, the \ctv models defy this trend for previously identified reasons of ample capacity for memorizing.

\observation{
A larger increase in the spread of loss during training correlated well with both reduced memorization and better performing models. This can be a fruitful metric for gauging generalization potential, except for models that prone to memorizing.
}

\begin{figure*}
\centering
\captionsetup[subfigure]{width=0.9\textwidth, justification=centering}

\begin{subfigure}{.33\textwidth}
  \centering
  \includegraphics[width=\linewidth]{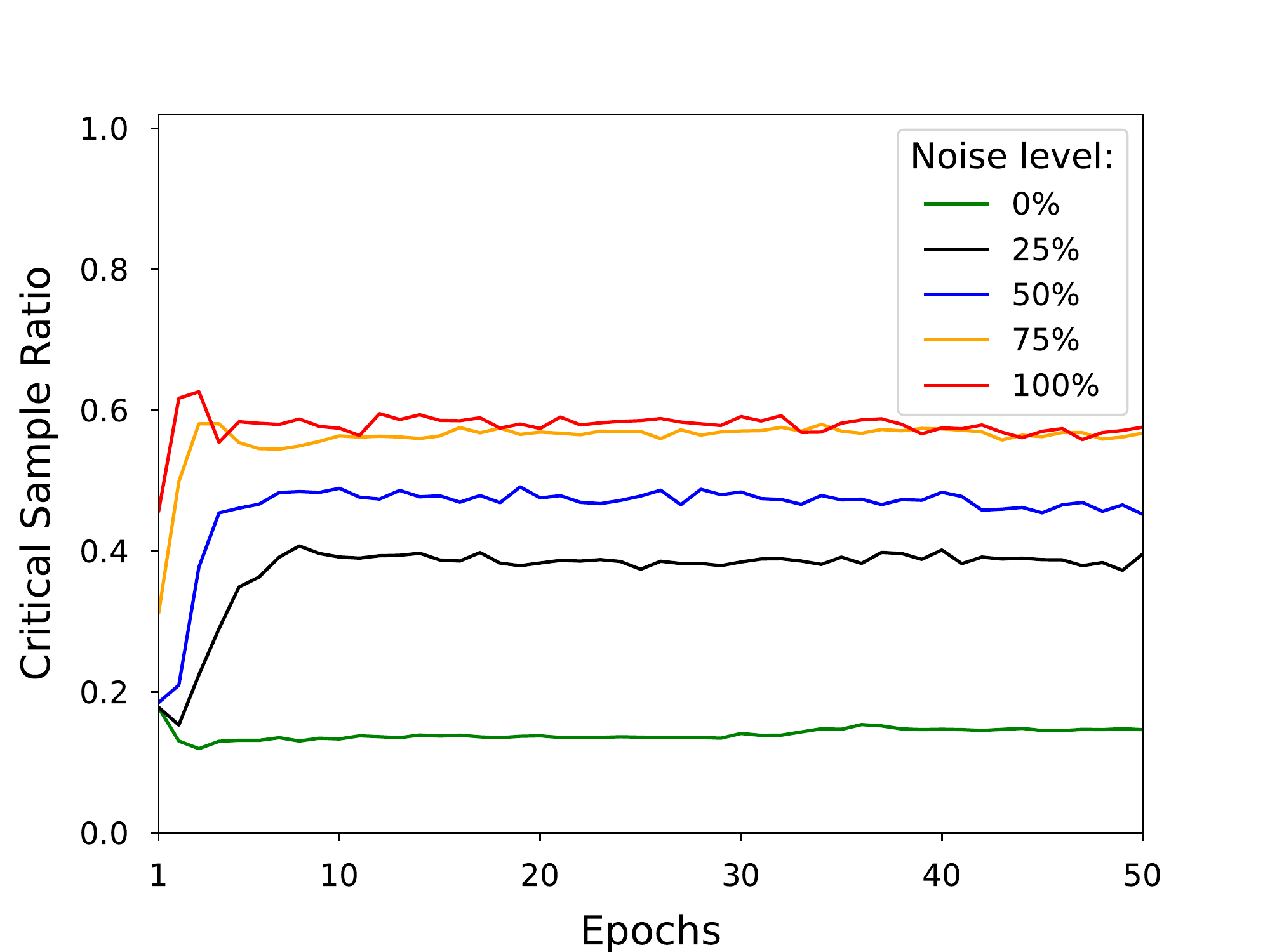}
  \caption{\ctv{} (\JTT)}
  \label{fig:mn_y_csr_test_c2v_top10}
\end{subfigure}%
\begin{subfigure}{.33\textwidth}
  \centering
  \includegraphics[width=\linewidth]{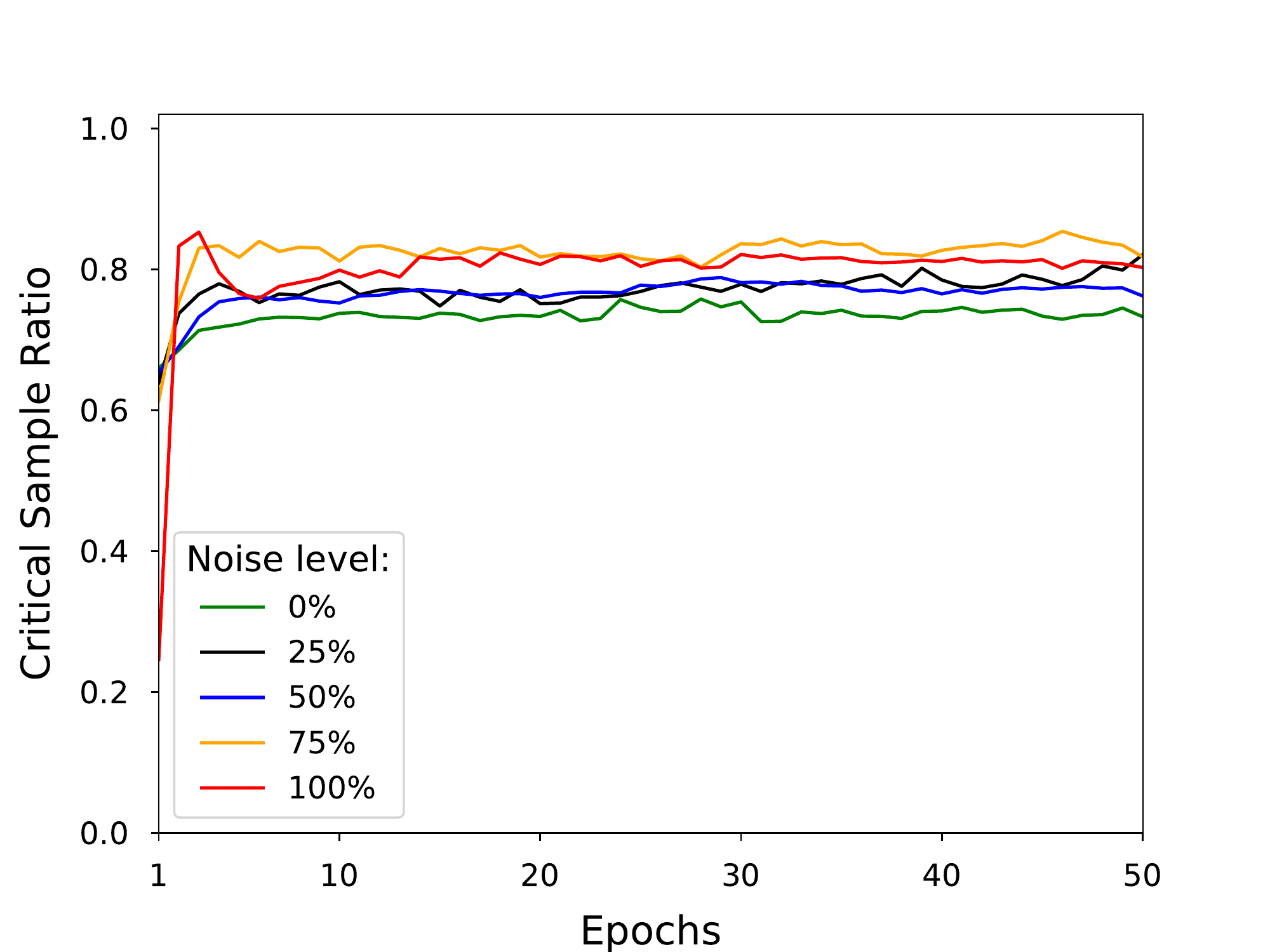}
  \caption{\ctv{} (\JS)}
  \label{fig:mn_y_csr_test_c2v_js}
\end{subfigure}%
\begin{subfigure}{.33\textwidth}
  \centering
  \includegraphics[width=\linewidth]{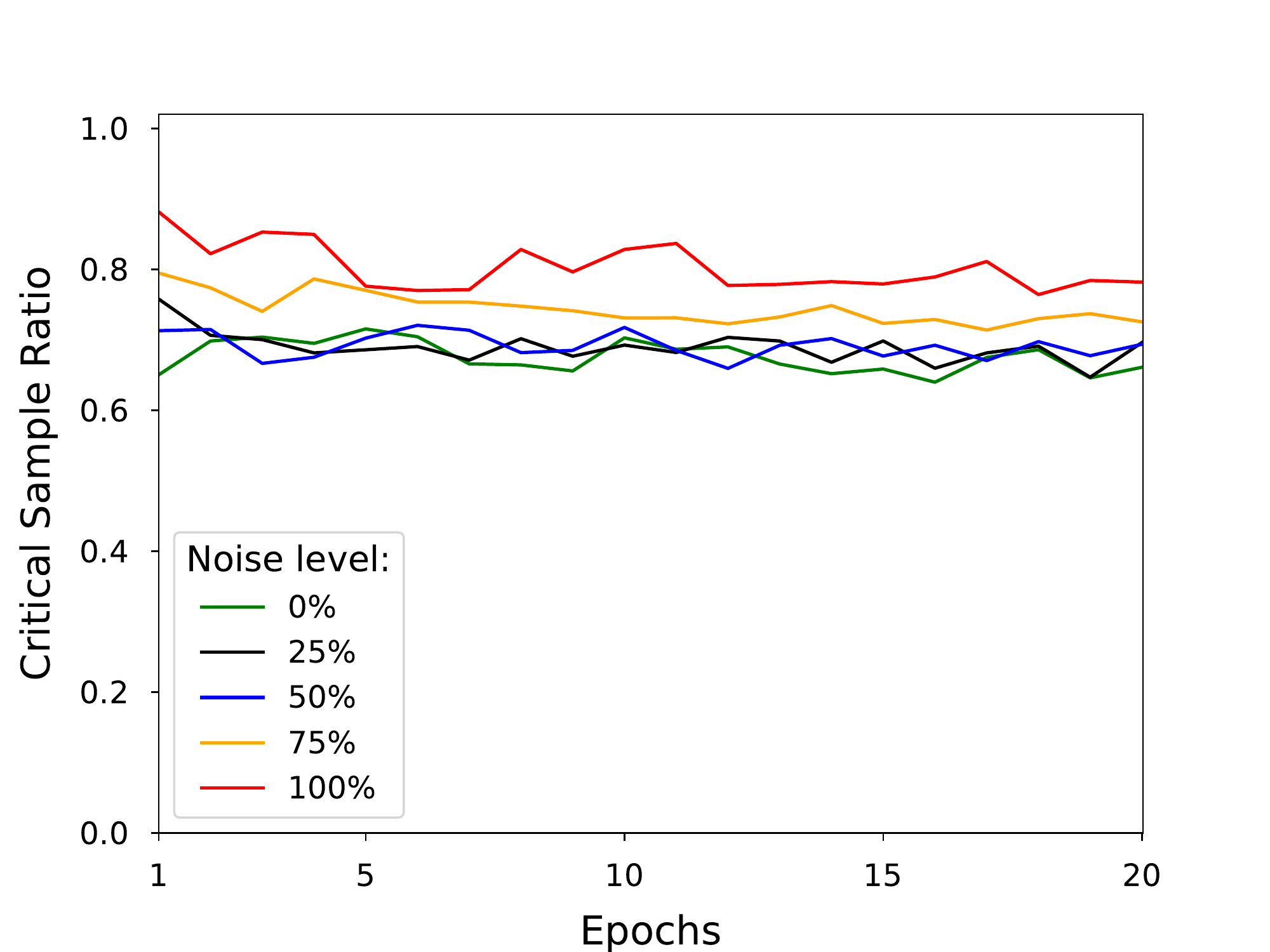}
  \caption{\ctv{} (\JM)}
  \label{fig:mn_y_csr_test_c2v_jm}
\end{subfigure}

\begin{subfigure}{.33\textwidth}
  \centering
  \includegraphics[width=\linewidth]{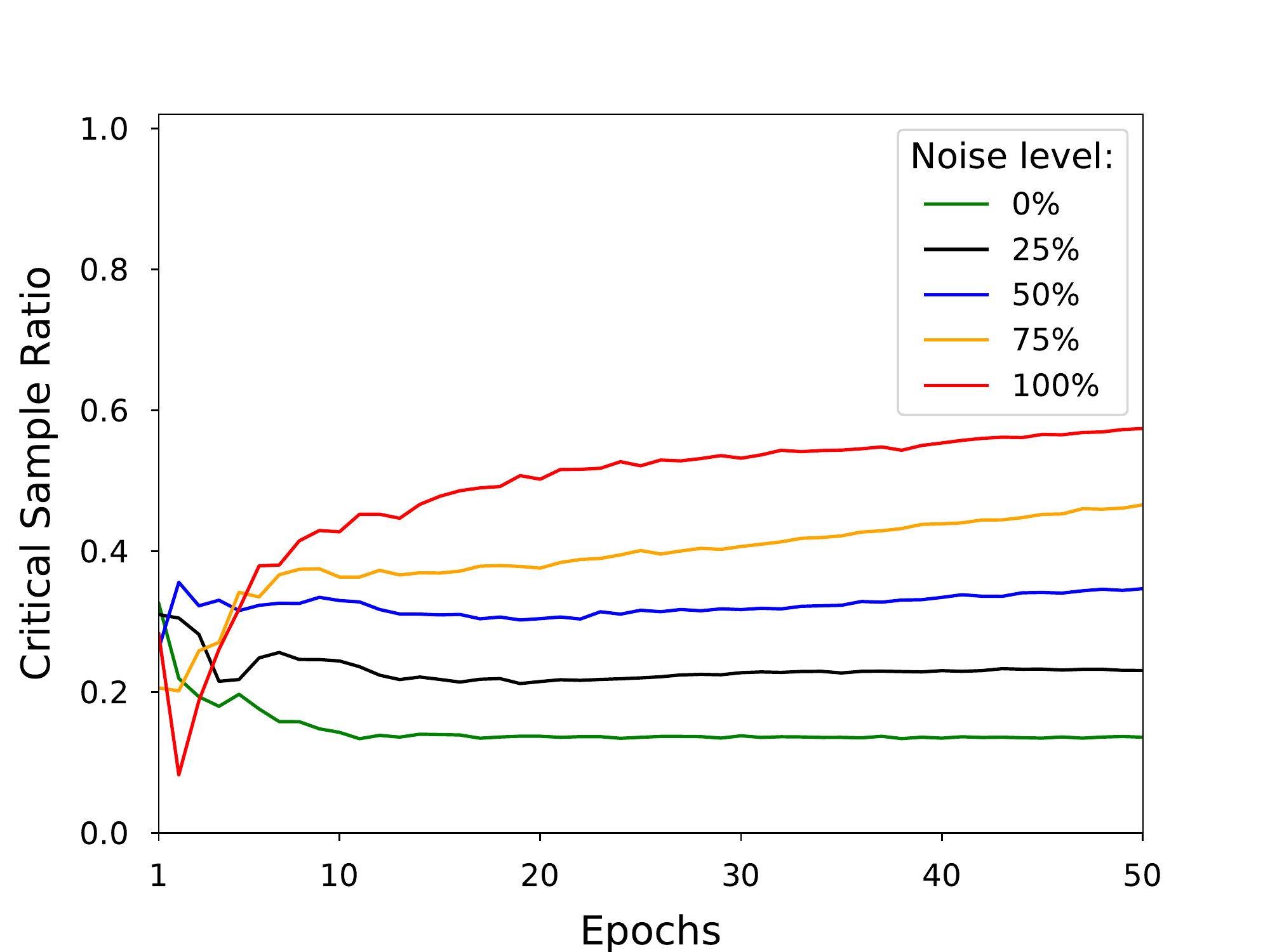}
  \caption{\cts{} (\JTT)}
  \label{fig:mn_y_csr_test_c2s_top10}
\end{subfigure}%
\begin{subfigure}{.33\textwidth}
  \centering
  \includegraphics[width=\linewidth]{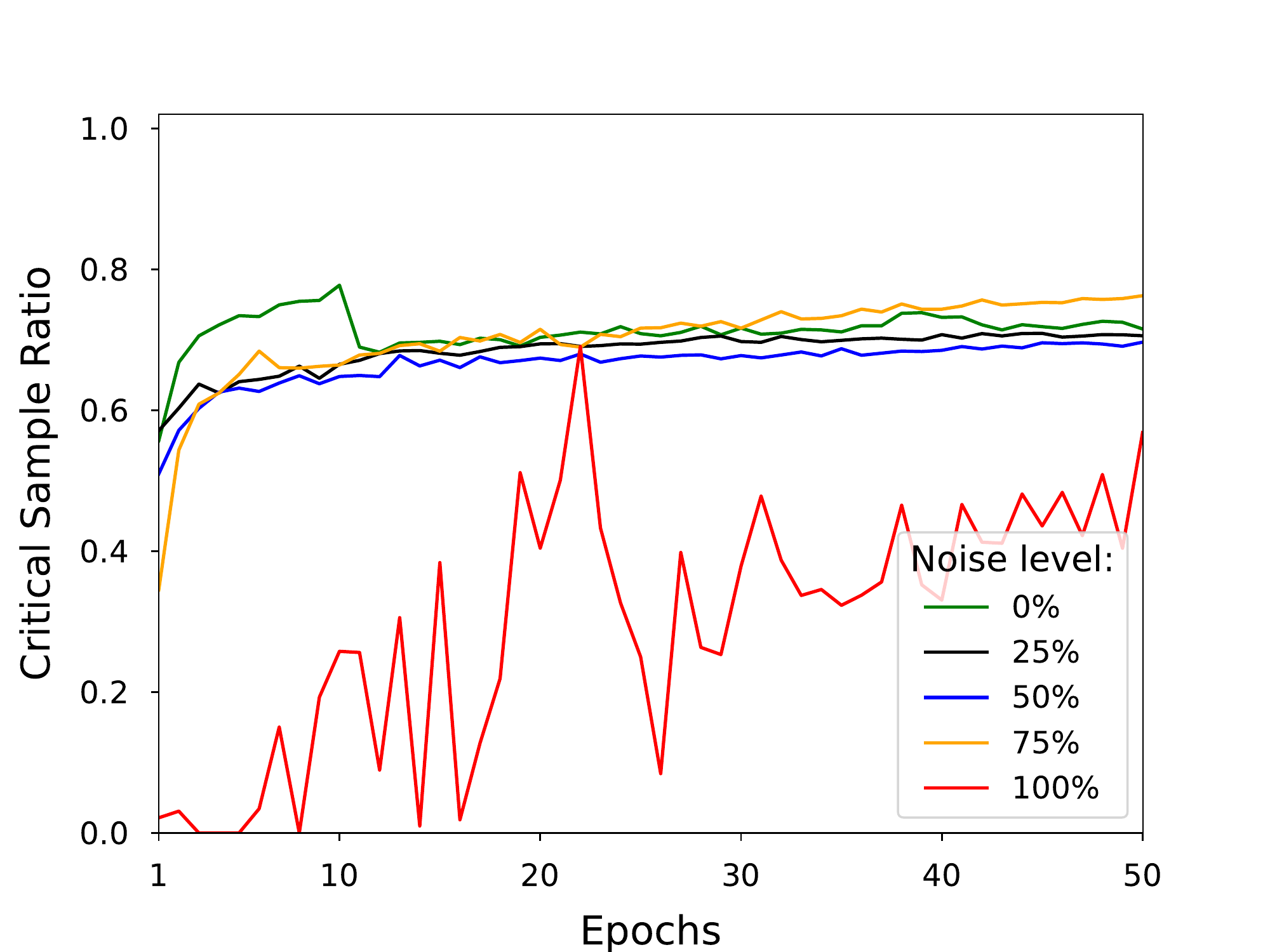}
  \caption{\cts{} (\JS)}
  \label{fig:mn_y_csr_test_c2s_js}
\end{subfigure}%
\begin{subfigure}{.33\textwidth}
  \centering
  \includegraphics[width=\linewidth]{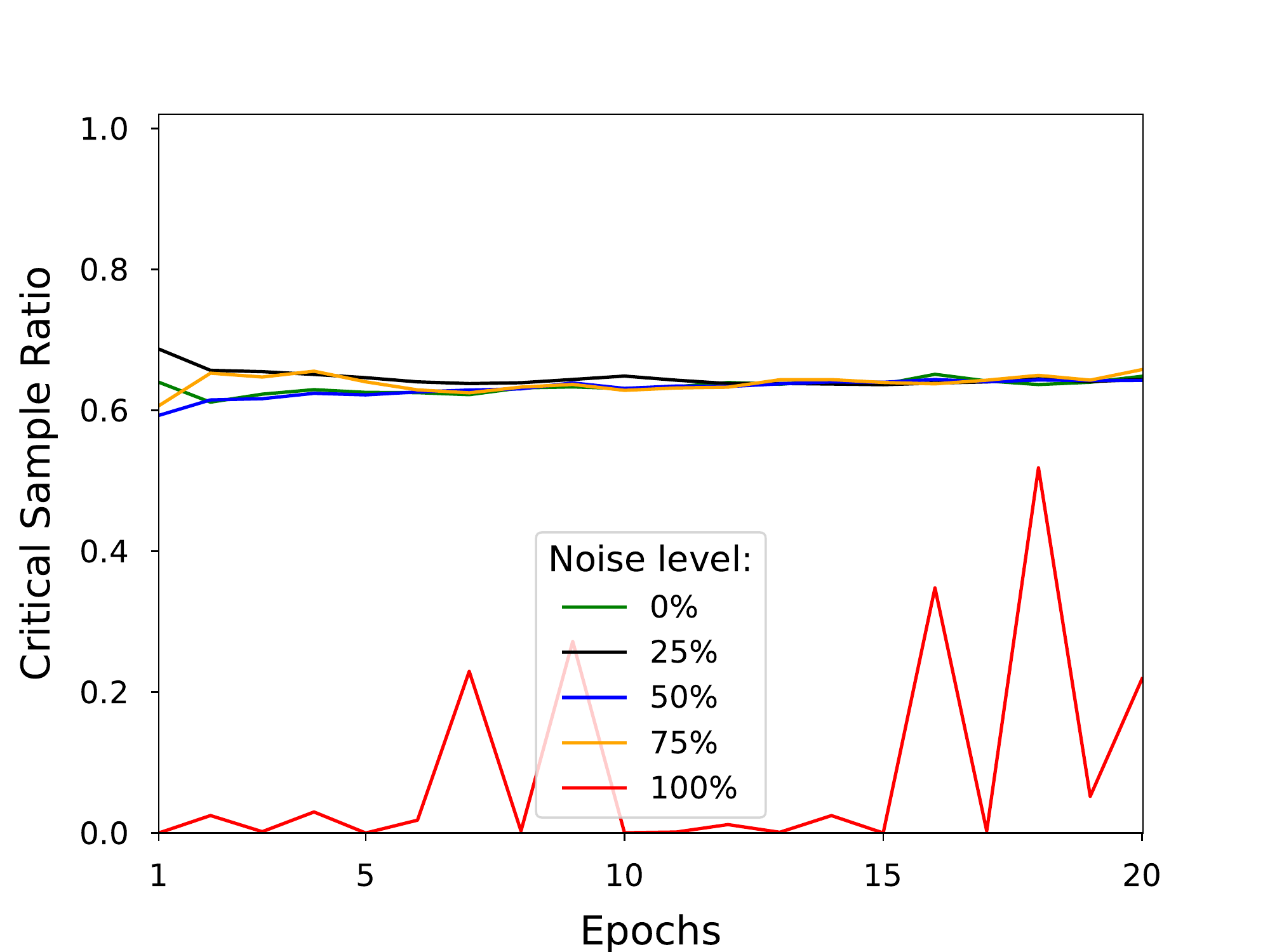}
  \caption{\cts{} (\JM)}
  \label{fig:mn_y_csr_test_c2s_jm}
\end{subfigure}

\caption{Critical Sample Ratio over all samples (\mnp).}
\label{fig:mn_y_csr_test}
\end{figure*}

\subsubsection{Memorization and Decision Surface}
\label{subsec:noise-y-csr}

In order to understand how noise affects the complexity of the hypotheses learned by \cis on the \mnp task, we examined the number of critical samples in the test set after each epoch while training.
A higher number of critical samples indicates a more complex learned decision surface, likely making it more susceptible to adversarial perturbations. In general, models with more noisy datasets are expected to have higher critical sample ratio (CSR) values \cite{arpit2017closer}. 
\Cref{fig:mn_y_csr_test} shows the CSR obtained for all samples (both correct and incorrect prediction) at different noise levels. 
In \ctv, as expected, models trained with higher noise manifested higher CSR. 
However, in \cts, these echo this expectation only on the smaller-balanced \JTT dataset (\Cref{fig:mn_y_csr_test_c2s_top10}), and we did not observe almost any such relation on the \JS and \JM datasets (\Cref{fig:mn_y_csr_test_c2s_js} \& \Cref{fig:mn_y_csr_test_c2s_jm}); the 100\%-noisy datasets is a clear outlier. 
\ctv mostly exhibits significant CSR differences in varying noise levels, which consistently suffers from a very high CSR for higher noise. Results on \cts only weakly support previous findings of \ctv, thus deserving further investigation regarding the relation between noise and CSR.

\observation{The unusual and strongly divergent patterns found across models and datasets suggest that further analyses are necessary for using CSR metrics to quantify training robustness in \cis.}

\subsection{Analyses of Memorization with Input Noise}
\label{subsec:noise-on-x}

Similar to the experiments with output noise in \Cref{subsec:noise-on-y}, we also tracked the models' performance, distribution of prediction score, and spread of training loss in each epoch (or step) while training the models with input noise. Input noise is added in a variety of ways that depend on the target task and dataset, as illustrated in \Cref{fig:noise_example}.

\begin{figure*}
\centering
\captionsetup[subfigure]{width=0.9\textwidth, justification=centering}

\begin{subfigure}{.32\textwidth}
  \centering
  \includegraphics[width=\linewidth]{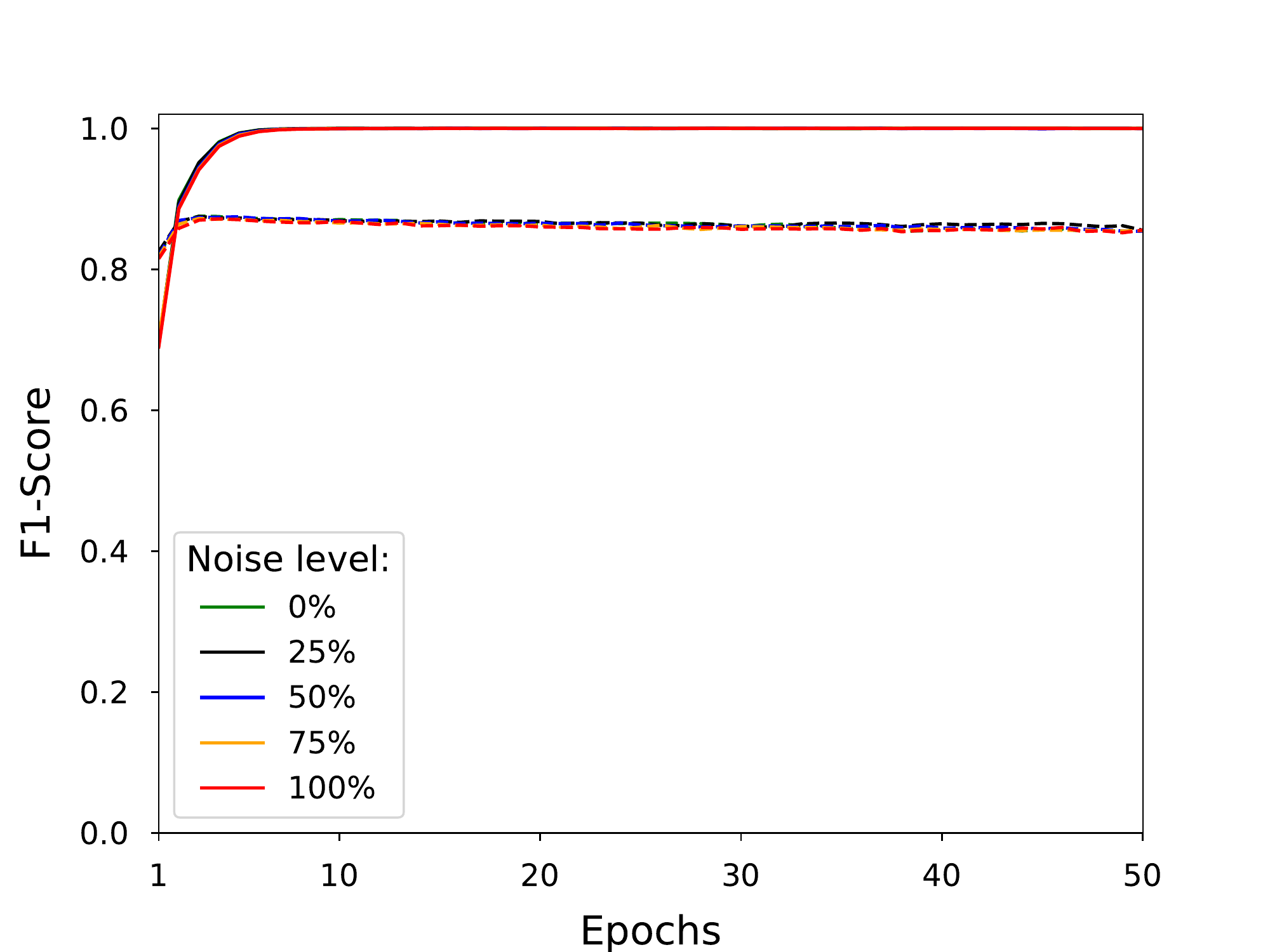}
  \caption{\FOneScore (\ctv{})}
  \label{fig:mn_x_deleting_f1_c2v_top10}
\end{subfigure}%
\begin{subfigure}{.32\textwidth}
  \centering
  \includegraphics[width=\linewidth]{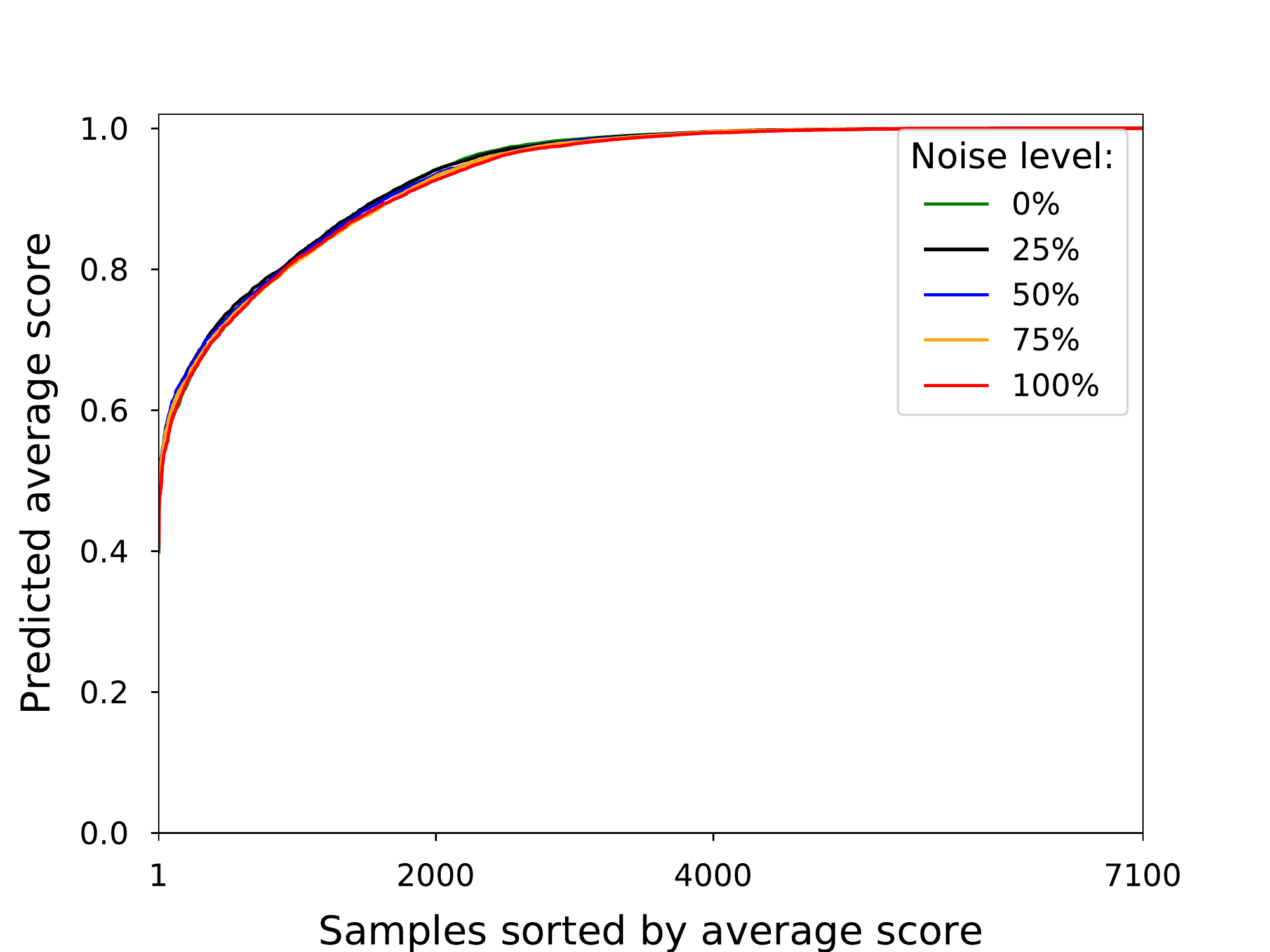}
  \caption{Distribution of Prediction Score (\ctv{})}
  \label{fig:mn_x_deleting_score_avg_all_c2v_top10}
\end{subfigure}%
\begin{subfigure}{.32\textwidth}
  \centering
  \includegraphics[width=\linewidth]{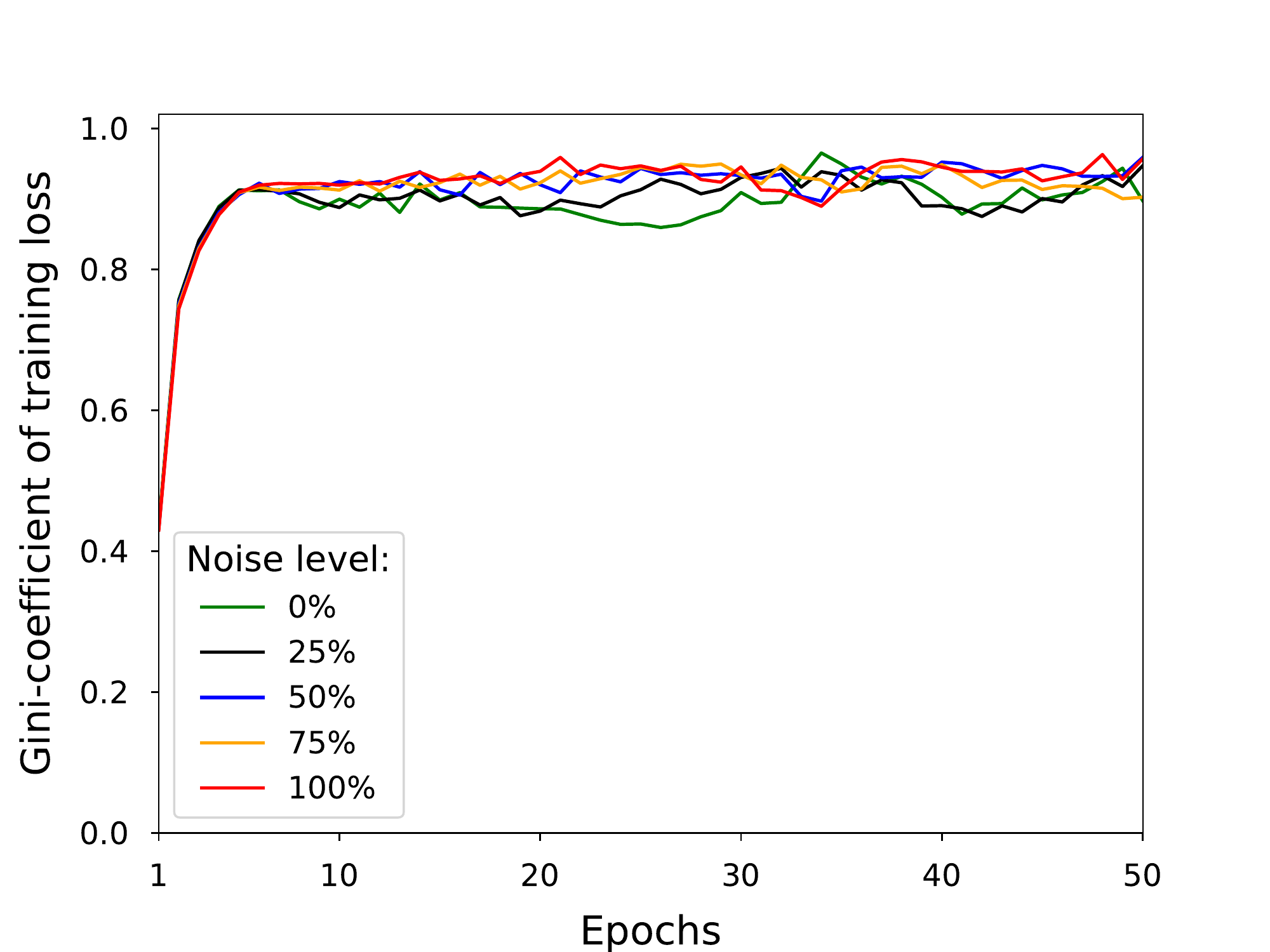}
  \caption{Spread of Training Loss (\ctv{})}
  \label{fig:mn_x_deleting_gini_loss_training_c2v_top10}
\end{subfigure}

\begin{subfigure}{.32\textwidth}
  \centering
  \includegraphics[width=\linewidth]{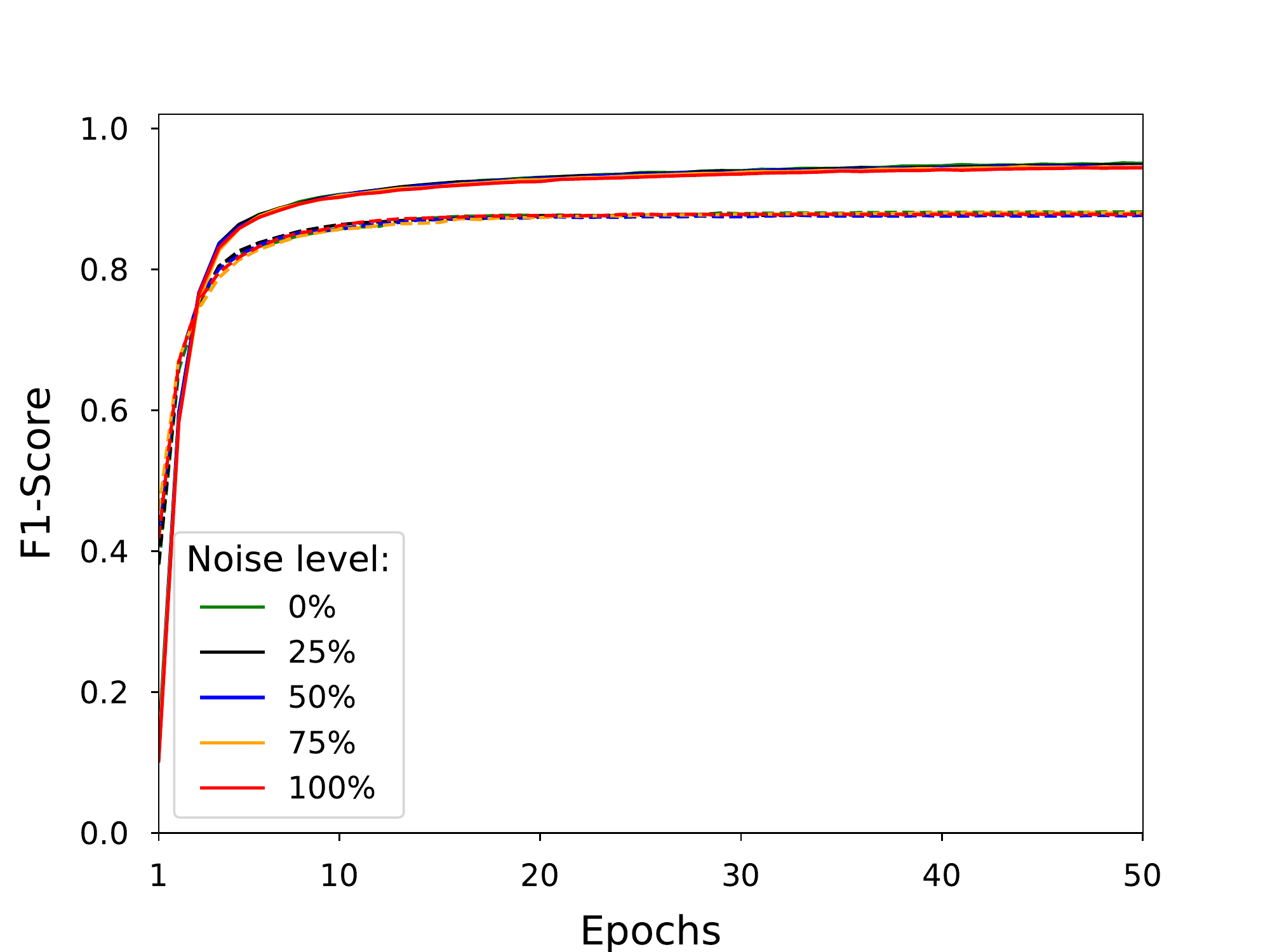}
  \caption{\FOneScore (\cts{})}
  \label{fig:mn_x_deleting_f1_c2s_top10}
\end{subfigure}%
\begin{subfigure}{.32\textwidth}
  \centering
  \includegraphics[width=\linewidth]{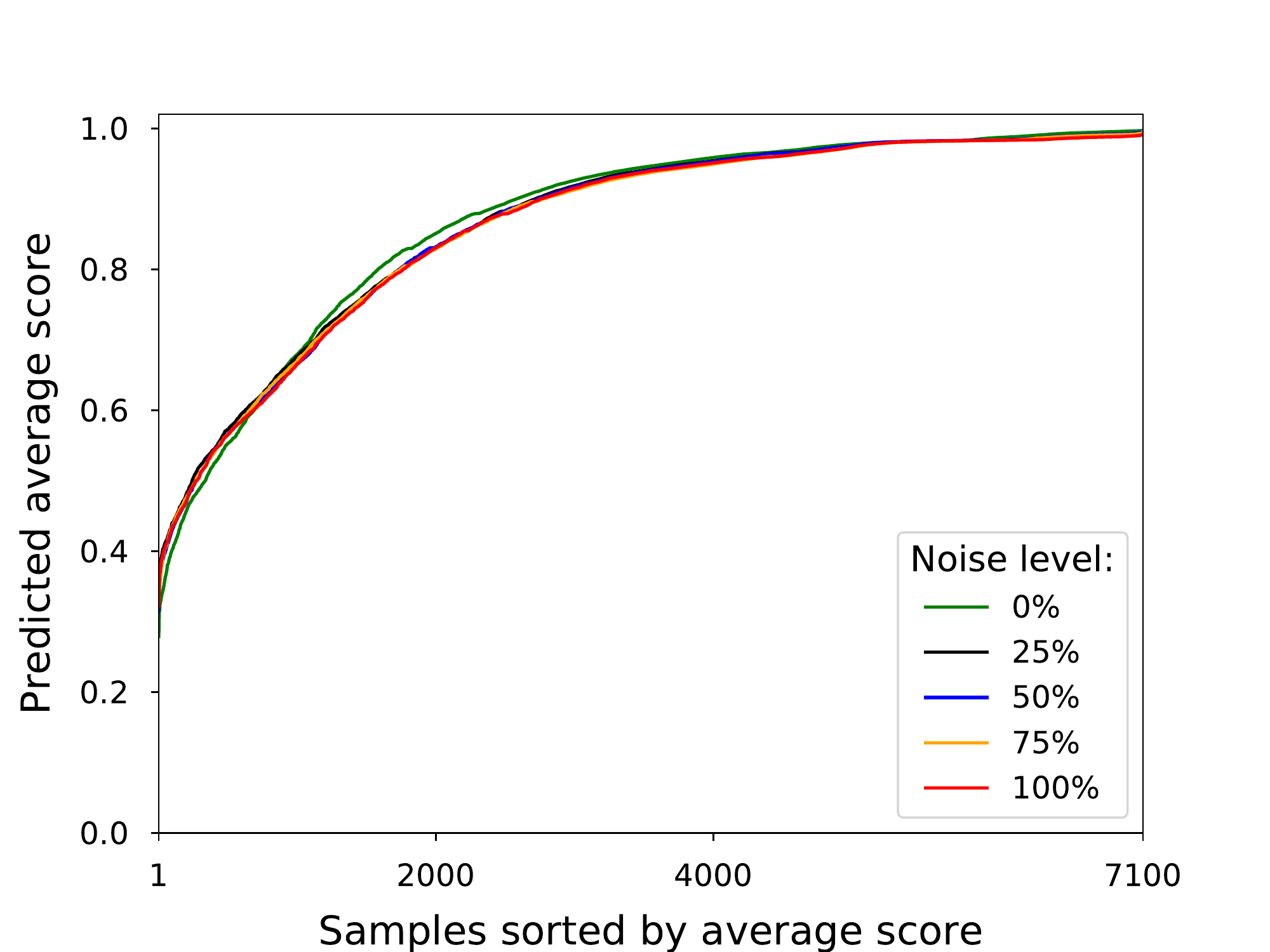}
  \caption{Distribution of Prediction Score (\cts{})}
  \label{fig:mn_x_deleting_score_avg_all_c2s_top10}
\end{subfigure}%
\begin{subfigure}{.32\textwidth}
  \centering
  \includegraphics[width=\linewidth]{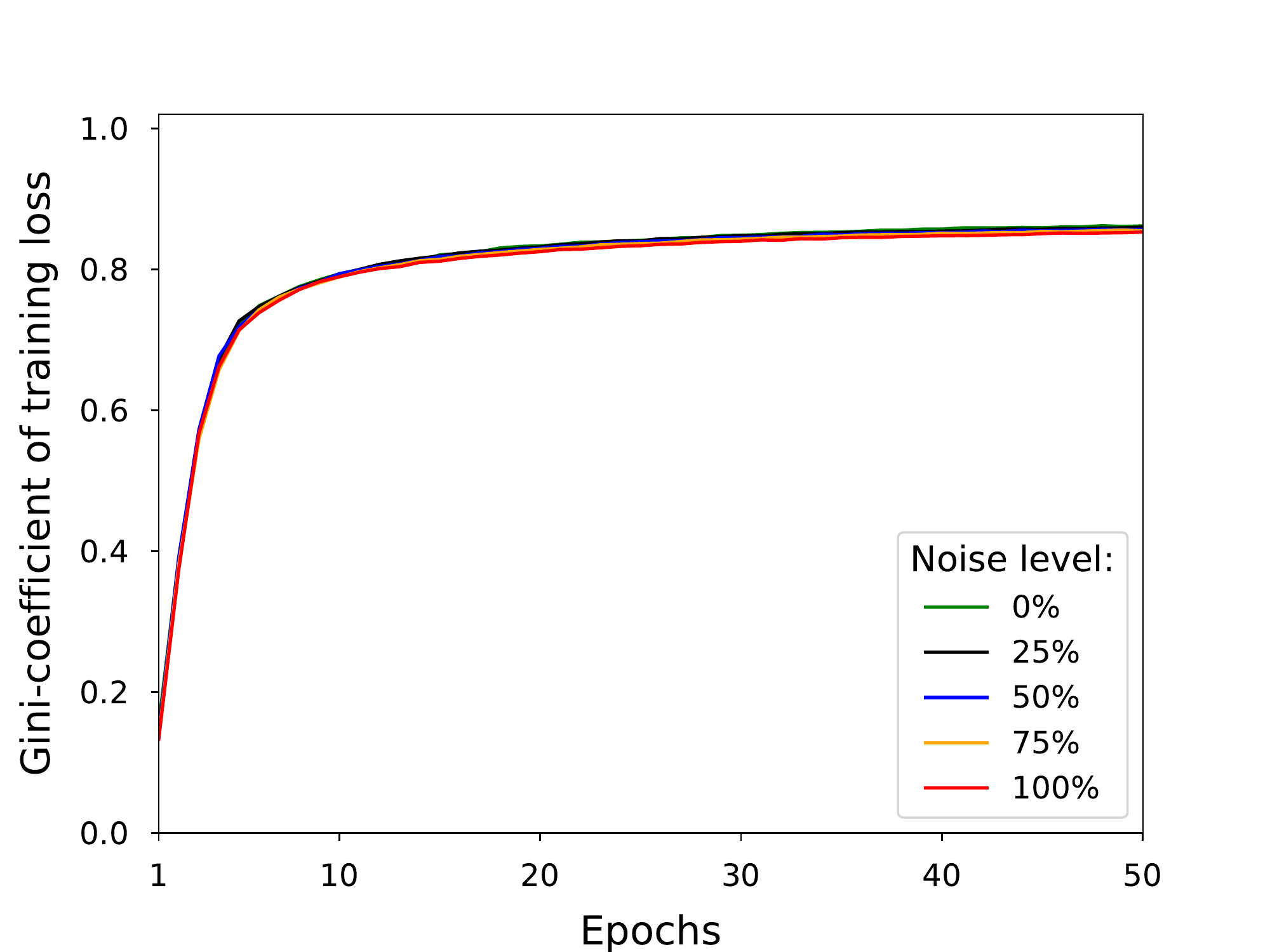}
  \caption{Spread of Training Loss (\cts{})}
  \label{fig:mn_x_deleting_gini_loss_training_c2s_top10}
\end{subfigure}

\caption{Input noise by statement deletion (\mnp, \JTT).}
\label{fig:mn_x_deleting_top10}


\begin{subfigure}{.32\textwidth}
  \centering
  \includegraphics[width=\linewidth]{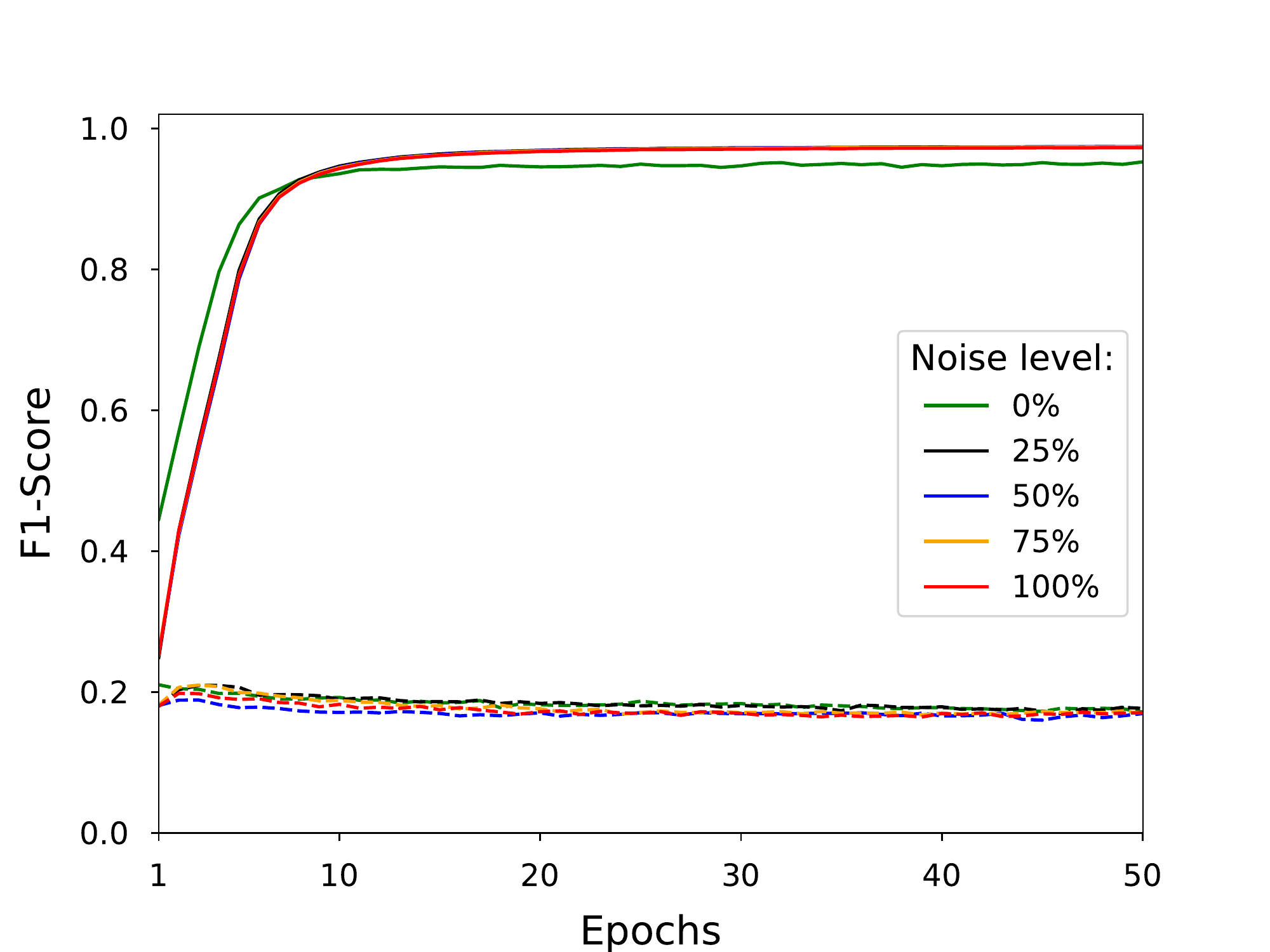}
  \caption{\FOneScore (\ctv{})}
  \label{fig:mn_x_deleting_f1_c2v_js}
\end{subfigure}%
\begin{subfigure}{.32\textwidth}
  \centering
  \includegraphics[width=\linewidth]{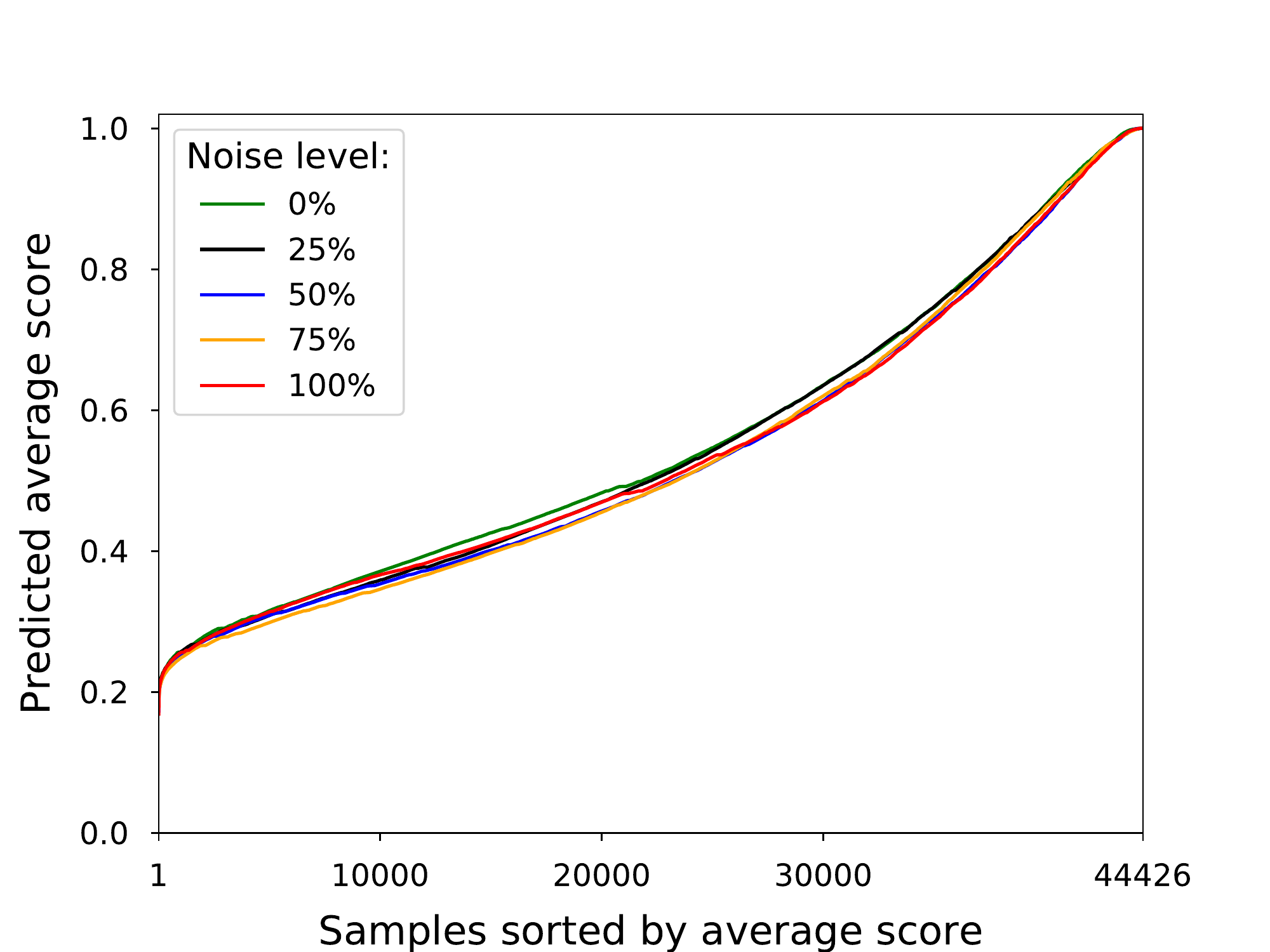}
  \caption{Distribution of Prediction Score (\ctv{})}
  \label{fig:mn_x_deleting_score_avg_all_c2v_js}
\end{subfigure}%
\begin{subfigure}{.32\textwidth}
  \centering
  \includegraphics[width=\linewidth]{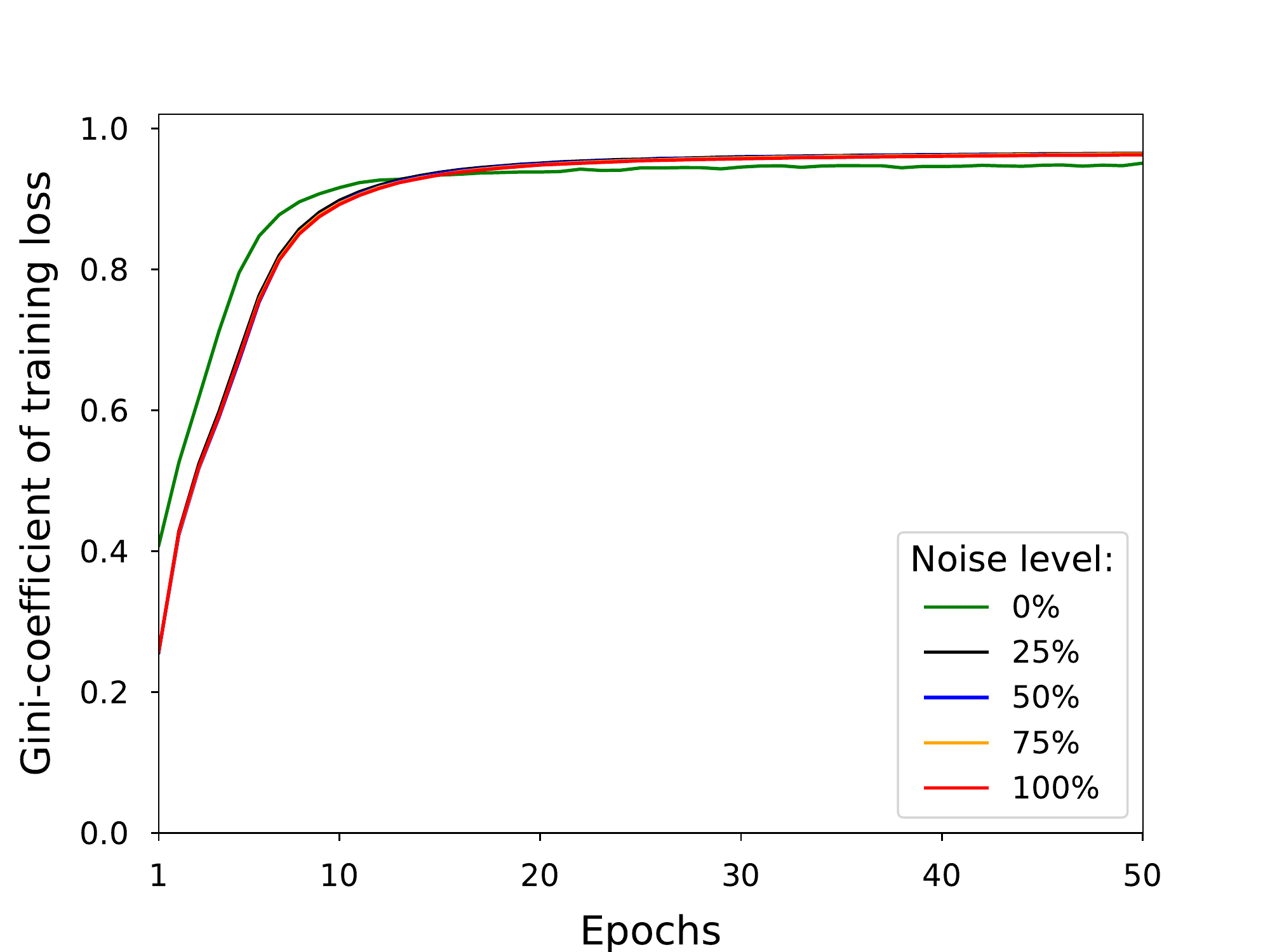}
  \caption{Spread of Training Loss (\ctv{})}
  \label{fig:mn_x_deleting_gini_loss_training_c2v_js}
\end{subfigure}

\begin{subfigure}{.32\textwidth}
  \centering
  \includegraphics[width=\linewidth]{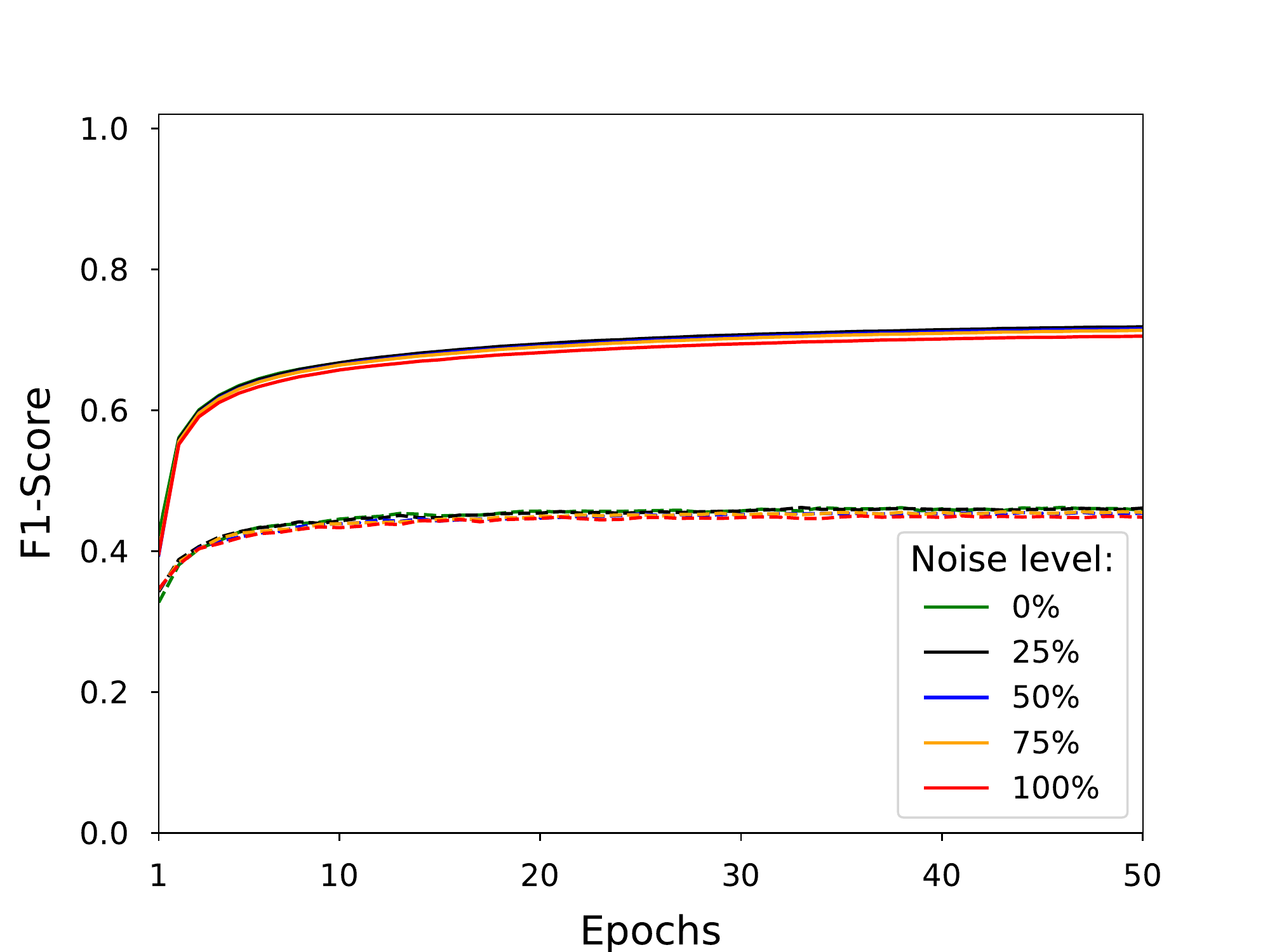}
  \caption{\FOneScore (\cts{})}
  \label{fig:mn_x_deleting_f1_c2s_js}
\end{subfigure}%
\begin{subfigure}{.32\textwidth}
  \centering
  \includegraphics[width=\linewidth]{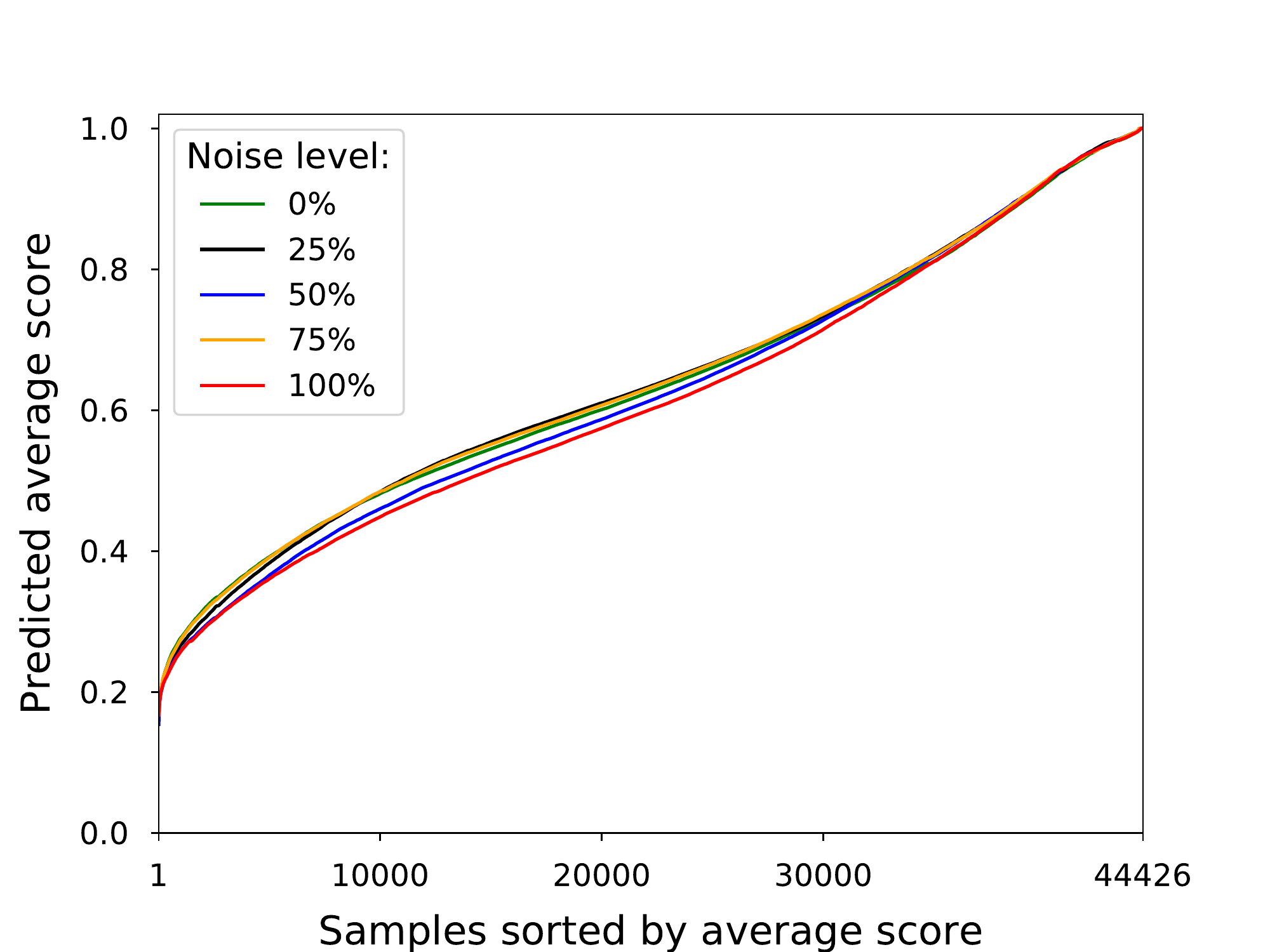}
  \caption{Distribution of Prediction Score (\cts{})}
  \label{fig:mn_x_deleting_score_avg_all_c2s_js}
\end{subfigure}%
\begin{subfigure}{.32\textwidth}
  \centering
  \includegraphics[width=\linewidth]{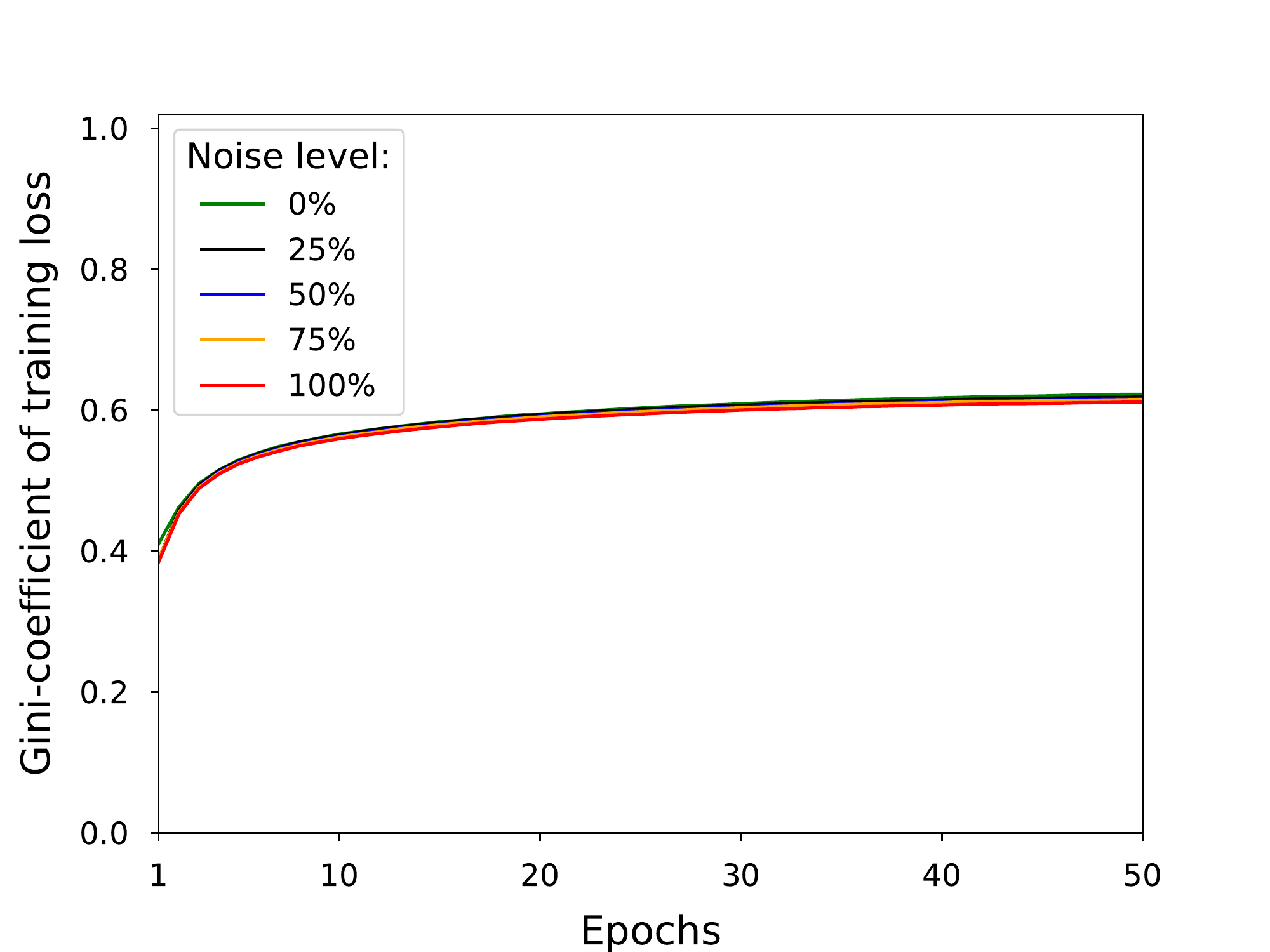}
  \caption{Spread of Training Loss (\cts{})}
  \label{fig:mn_x_deleting_gini_loss_training_c2s_js}
\end{subfigure}

\caption{Input noise by statement deletion (\mnp, \JS).}
\label{fig:mn_x_deleting_js}
\end{figure*}

\subsubsection{Statement Deletion}
\label{subsec:x-deleting}

\Cref{fig:mn_x_deleting_top10,fig:mn_x_deleting_js} show the impact of deleting a random statement from the method body on \mnp task for \ctv and \cts models with \JTT dataset and \JS dataset, respectively.
This impact is significantly less pronounced than what was observed with output noise in \Cref{subsec:noise-on-y}. In that analysis, models trained on the original data demonstrated much higher performance than those trained on noisy data: the latter generalized less well as noise somewhat forced models to memorize data points.
In contrast, in \Cref{fig:mn_x_deleting_top10,fig:mn_x_deleting_js}, models trained with this type of input noise converge to nearly the same performance regardless of noise level.
That these neural models barely suffer from the omission of a random statement can be explained by our previous findings \cite{rabin2021dd, rabin2022perses}: models typically rely on very few tokens for making their predictions. Given that methods often contain many statements, it is likely that the model receives all the necessary information from the remaining code snippets. In fact, this form of training is similar to input dropout \cite{srivastava2014dropout}, a strategy occasionally employed in language modeling to reduce overfitting on salient input features. \Cref{fig:mn_x_deleting_js}${a}$ indeed shows that models trained with this form of noise may achieve better training (although not held-out) performance.
We observed similar trends when experimenting on other models and datasets with this form of input noise.

\observation{
The \cis barely suffer from input noise based on random statement deletion, which can be explained by our previous findings \cite{rabin2021dd, rabin2022perses} - models usually rely on a few tokens for making predictions.
}

\begin{figure*}
\centering
\captionsetup[subfigure]{width=0.9\textwidth, justification=centering}

\begin{subfigure}{0.32\textwidth}
  \centering
  \includegraphics[width=\linewidth]{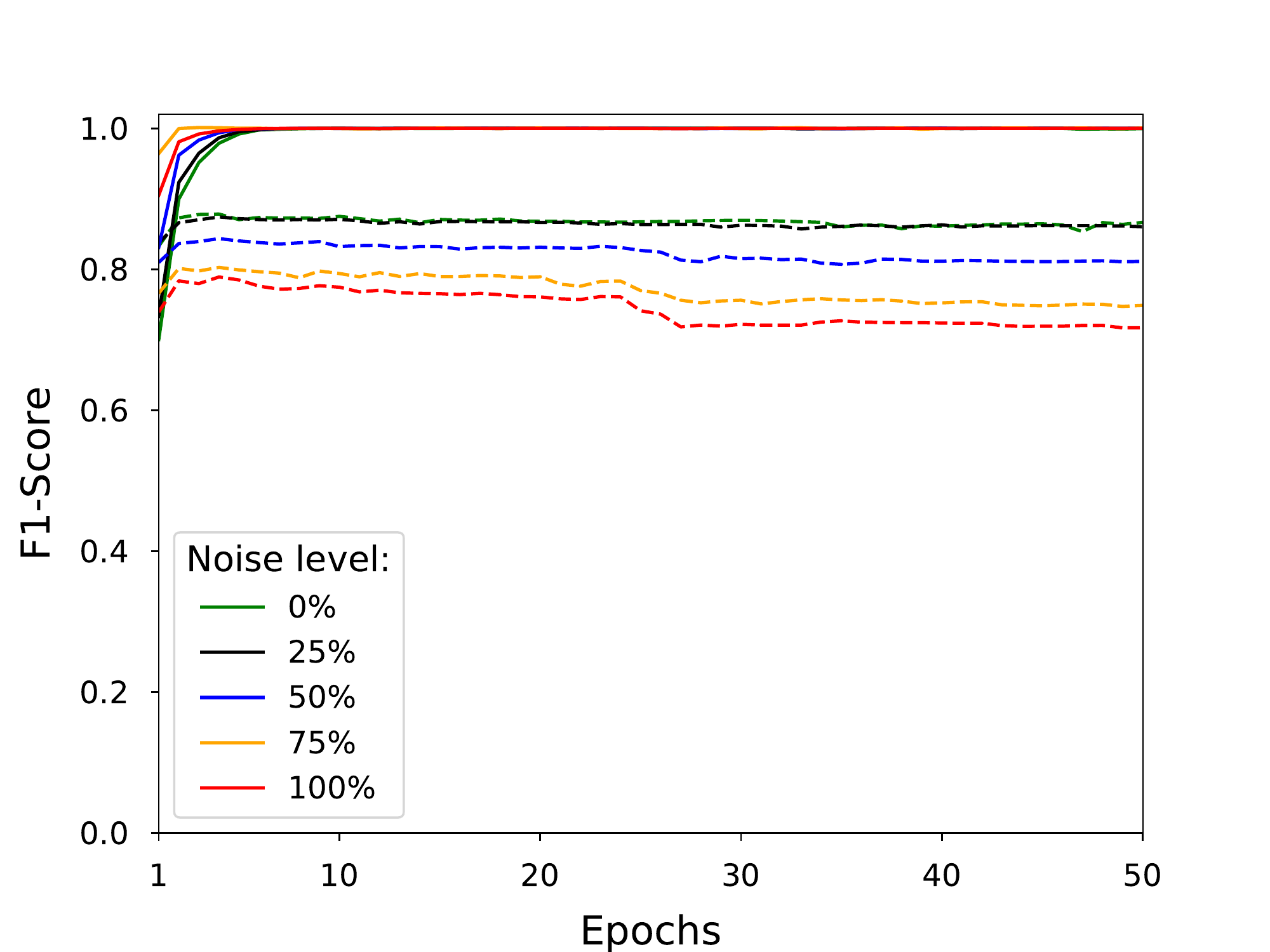}
  \caption{\FOneScore (\ctv{})}
  \label{fig:mn_x_replacing_f1_c2v_top10}
\end{subfigure}%
\begin{subfigure}{0.32\textwidth}
  \centering
  \includegraphics[width=\linewidth]{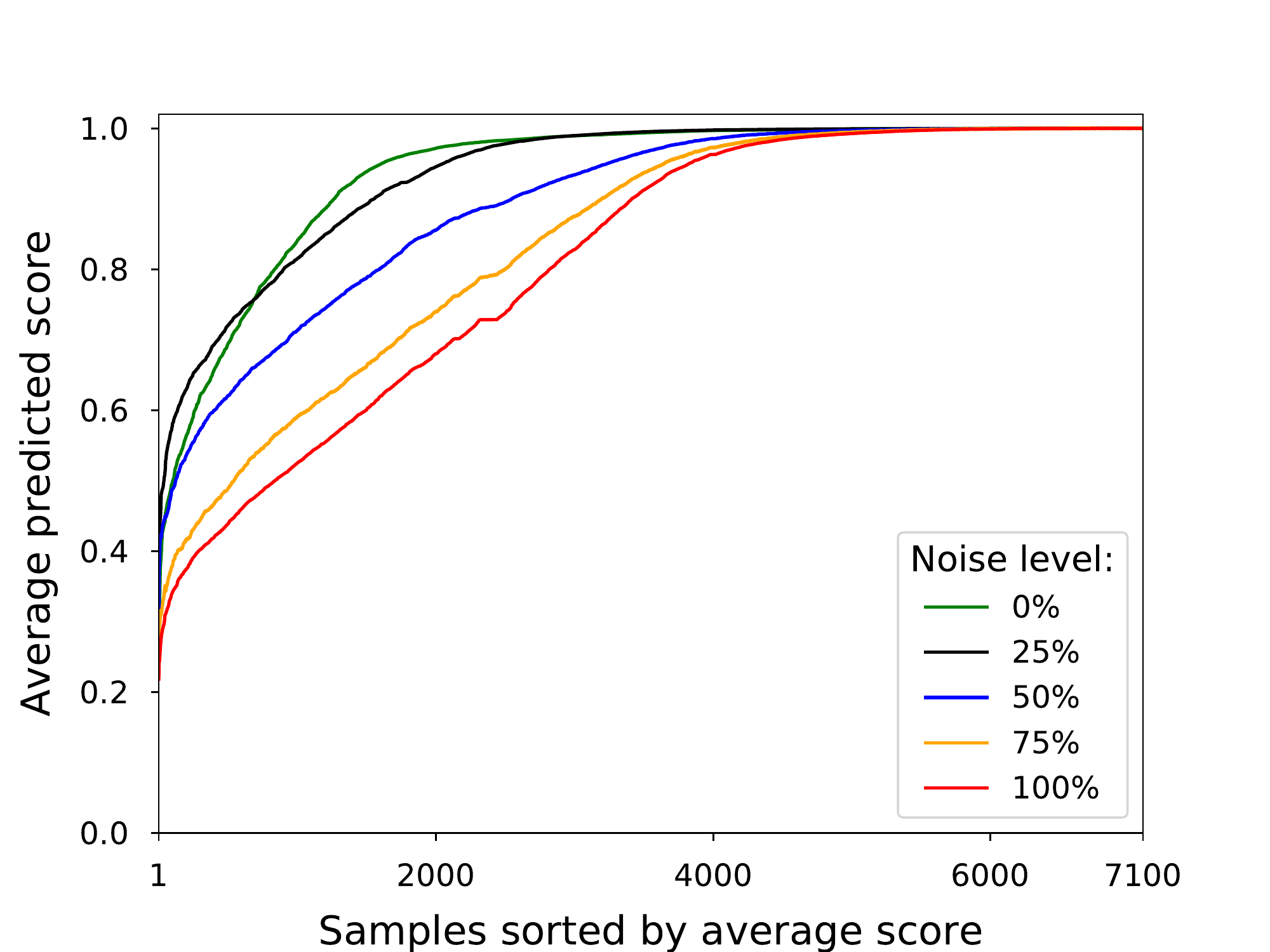}
  \caption{Distribution of Prediction Score (\ctv{})}
  \label{fig:mn_x_replacing_score_avg_all_c2v_top10}
\end{subfigure}%
\begin{subfigure}{0.32\textwidth}
  \centering
  \includegraphics[width=\linewidth]{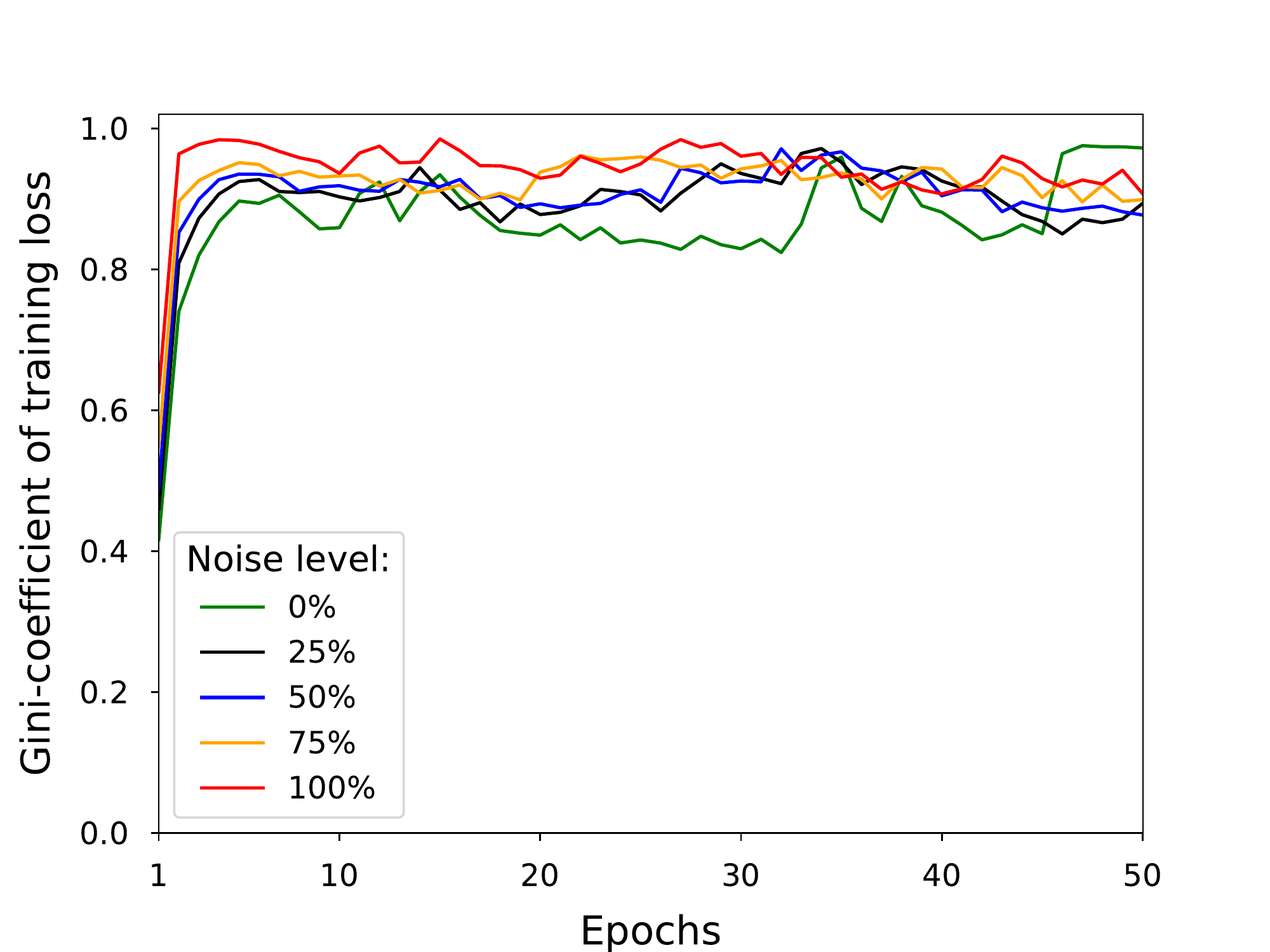}
  \caption{Spread of Training Loss (\ctv{})}
  \label{fig:mn_x_replacing_gini_loss_train_c2v_top10}
\end{subfigure}

\begin{subfigure}{0.32\textwidth}
  \centering
  \includegraphics[width=\linewidth]{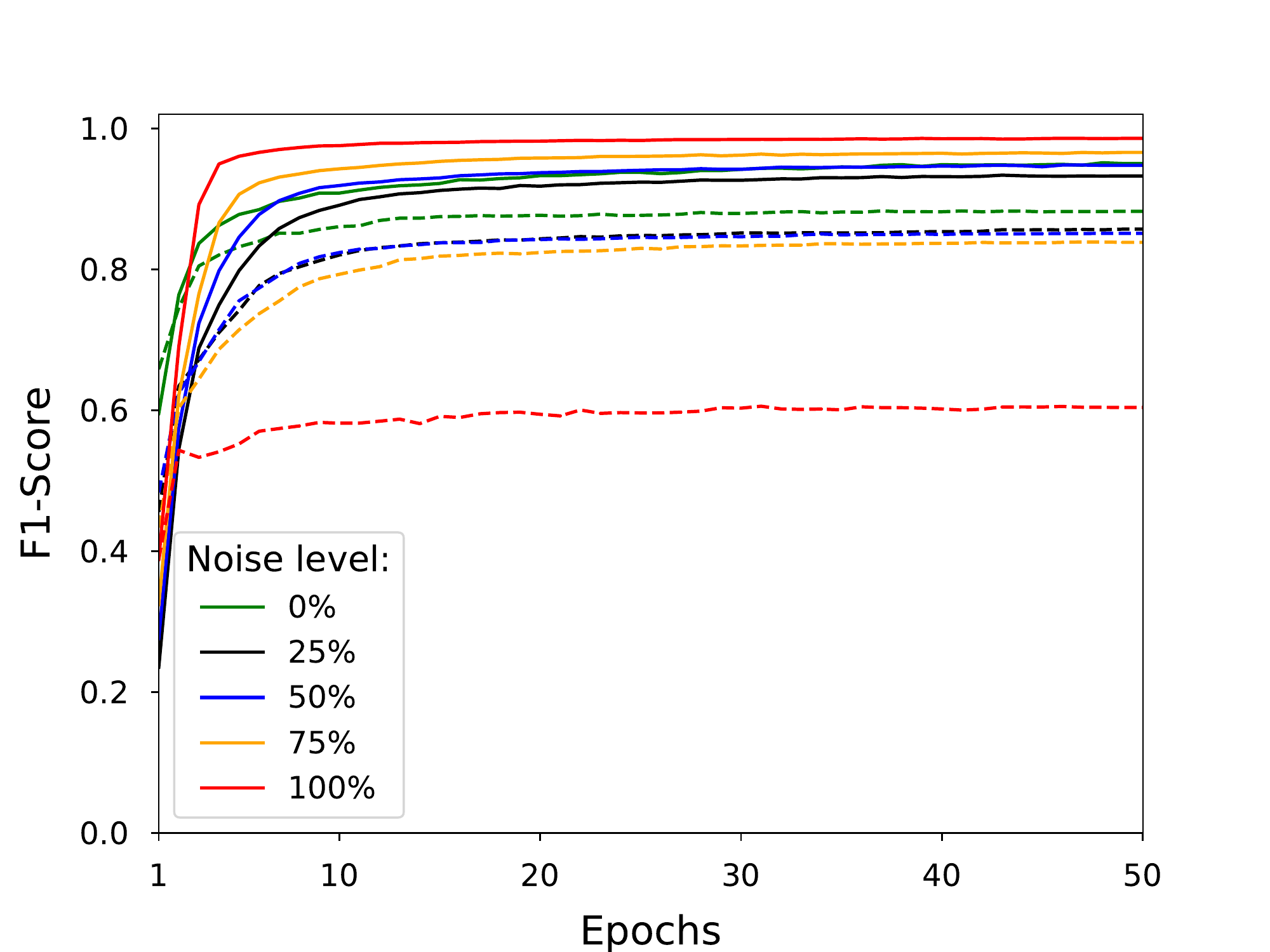}
  \caption{\FOneScore (\cts{})}
  \label{fig:mn_x_replacing_f1_c2s_top10}
\end{subfigure}%
\begin{subfigure}{0.32\textwidth}
  \centering
  \includegraphics[width=\linewidth]{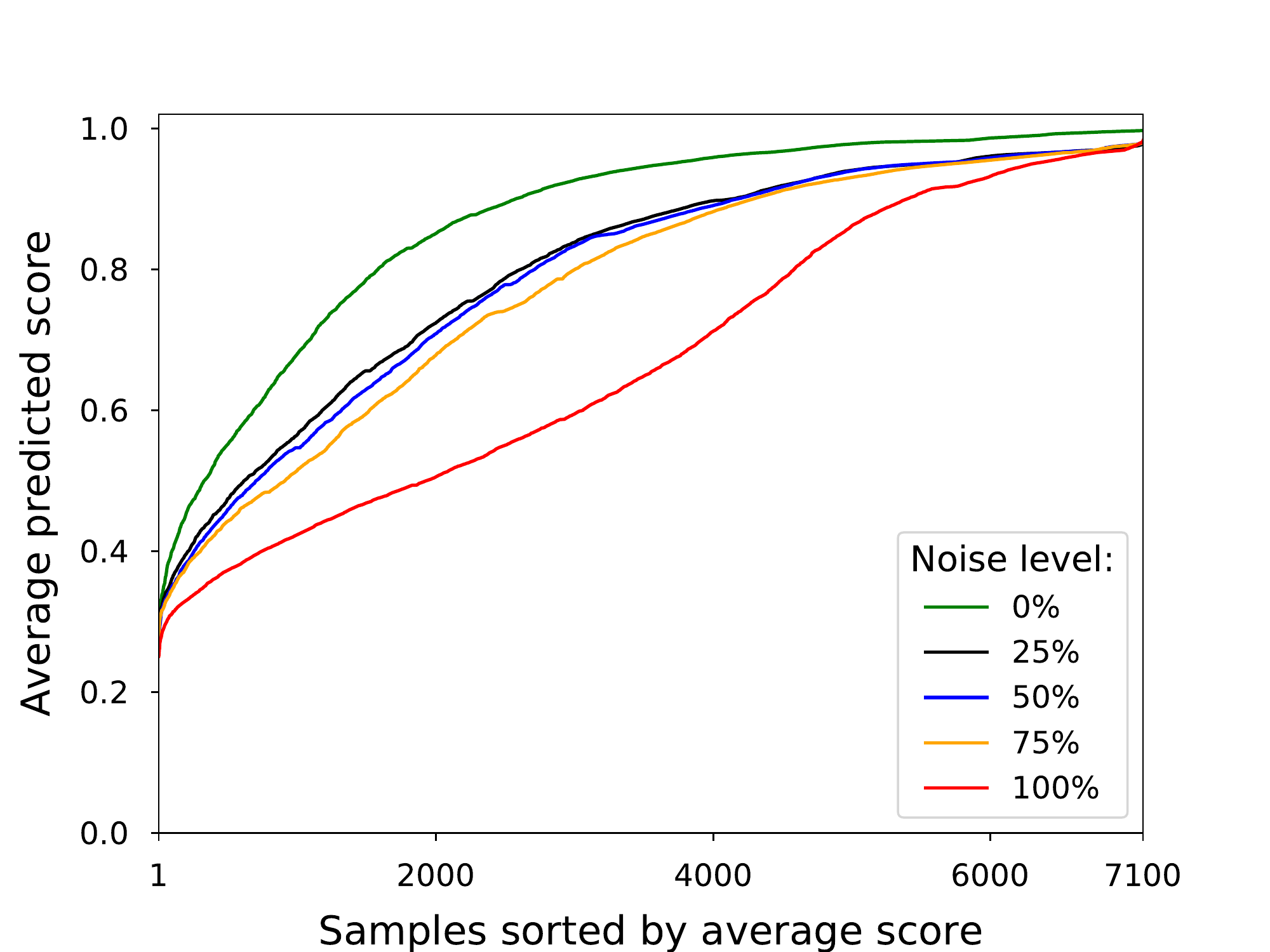}
  \caption{Distribution of Prediction Score (\cts{})}
  \label{fig:mn_x_replacing_score_avg_all_c2s_top10}
\end{subfigure}%
\begin{subfigure}{0.32\textwidth}
  \centering
  \includegraphics[width=\linewidth]{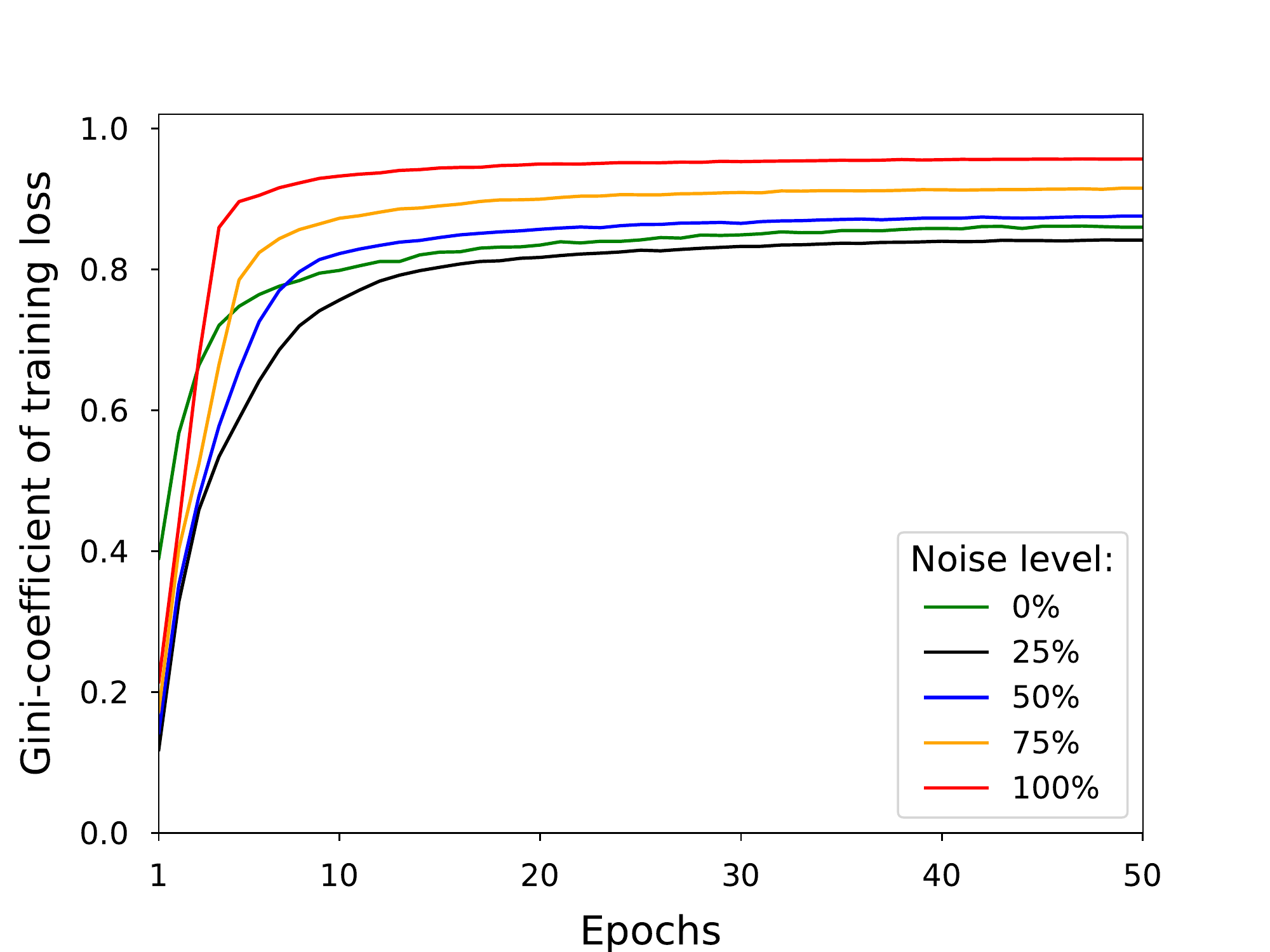}
  \caption{Spread of Training Loss (\cts{})}
  \label{fig:mn_x_replacing_gini_loss_test_c2s_top10}
\end{subfigure}

\caption{Input noise by method name leakage (\mnp, \JTT).}
\label{fig:mn_x_replacing_top10}


\begin{subfigure}{0.32\textwidth}
  \centering
  \includegraphics[width=\linewidth]{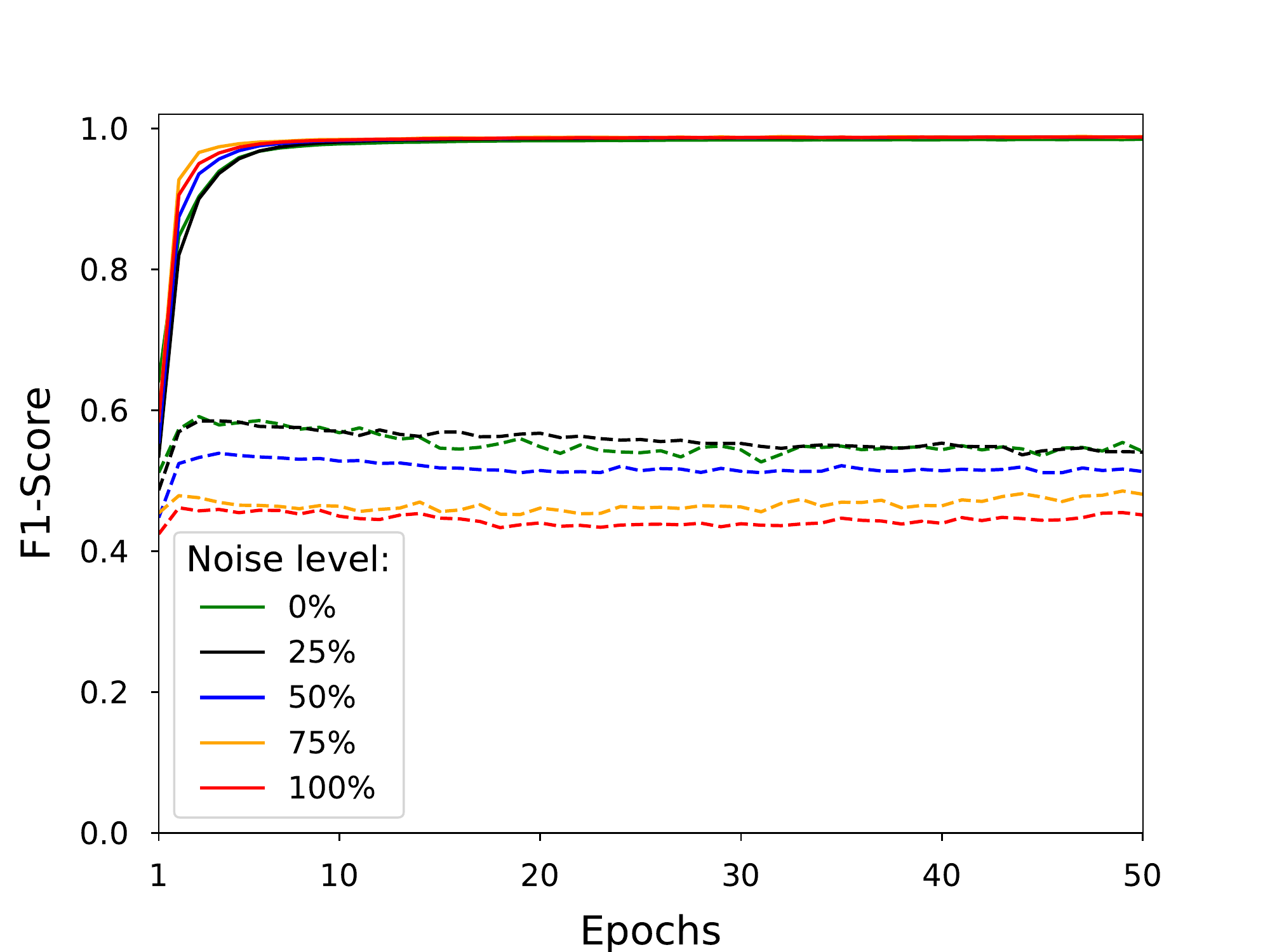}
  \caption{\FOneScore (\ctv{})}
  \label{fig:mn_x_replacing_f1_c2v_js}
\end{subfigure}%
\begin{subfigure}{0.32\textwidth}
  \centering
  \includegraphics[width=\linewidth]{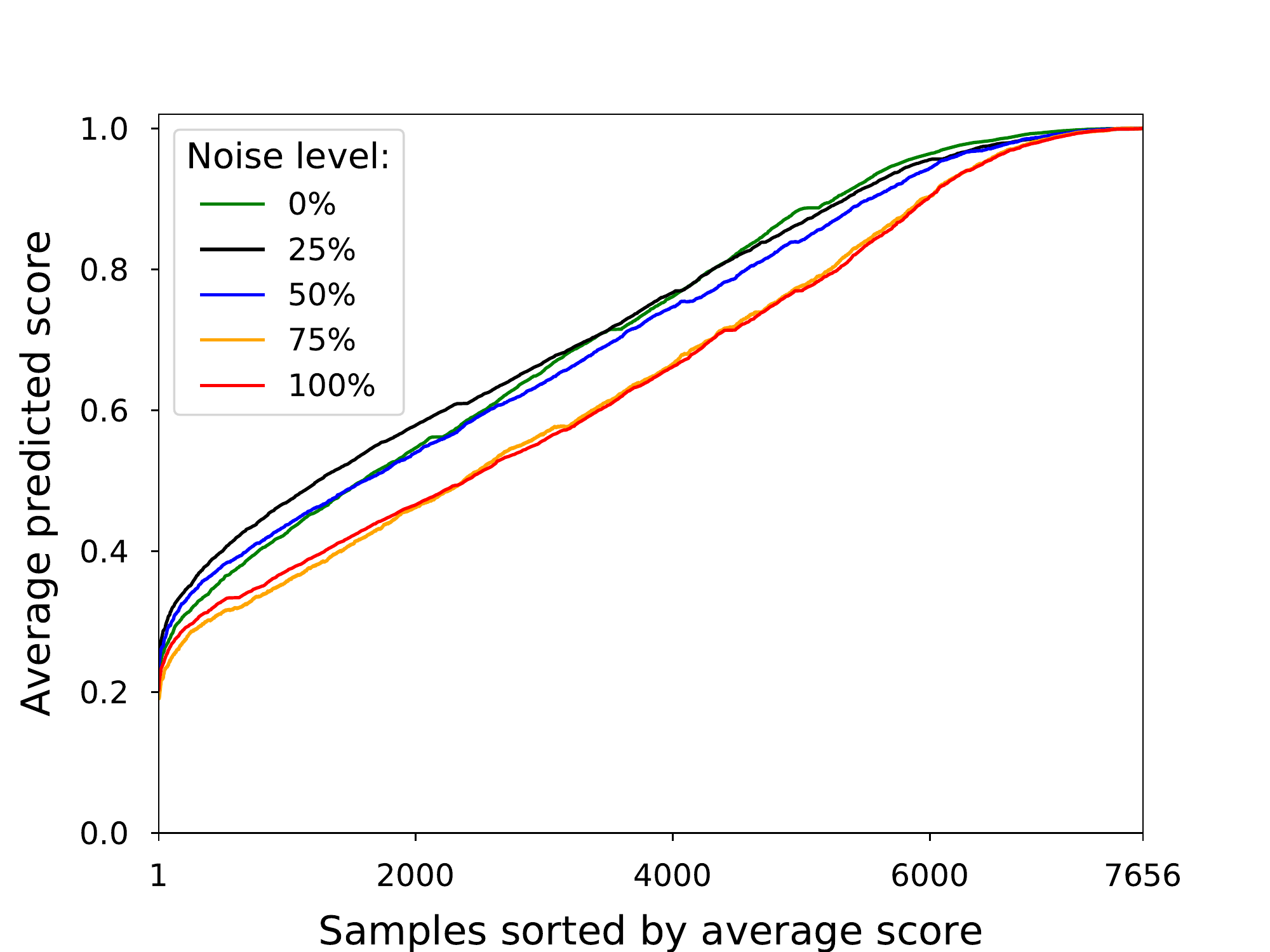}
  \caption{Distribution of Prediction Score (\ctv{})}
  \label{fig:mn_x_replacing_score_avg_all_c2v_js}
\end{subfigure}%
\begin{subfigure}{0.32\textwidth}
  \centering
  \includegraphics[width=\linewidth]{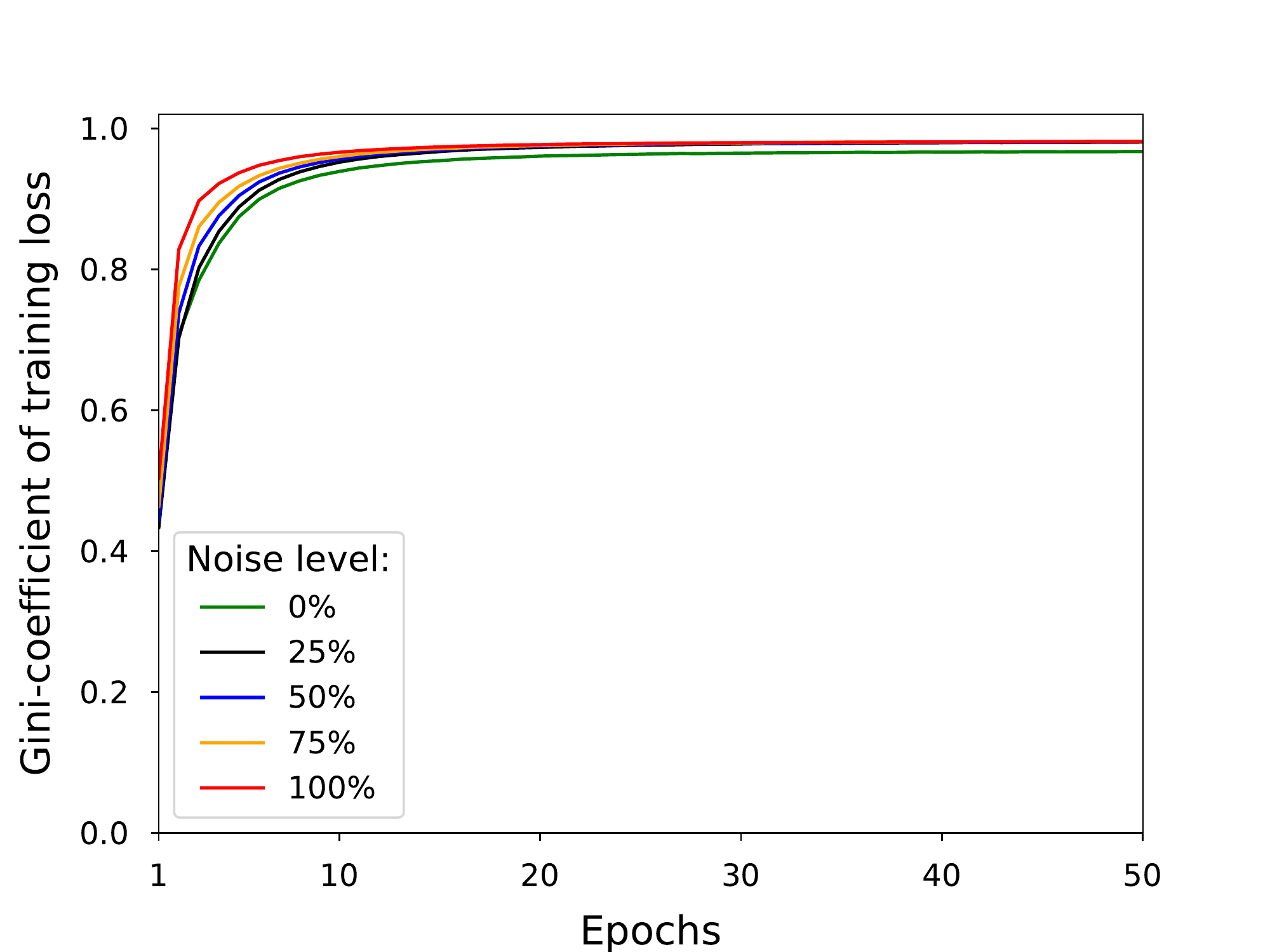}
  \caption{Spread of Training Loss (\ctv{})}
  \label{fig:mn_x_replacing_gini_loss_training_c2v_js}
\end{subfigure}

\begin{subfigure}{0.32\textwidth}
  \centering
  \includegraphics[width=\linewidth]{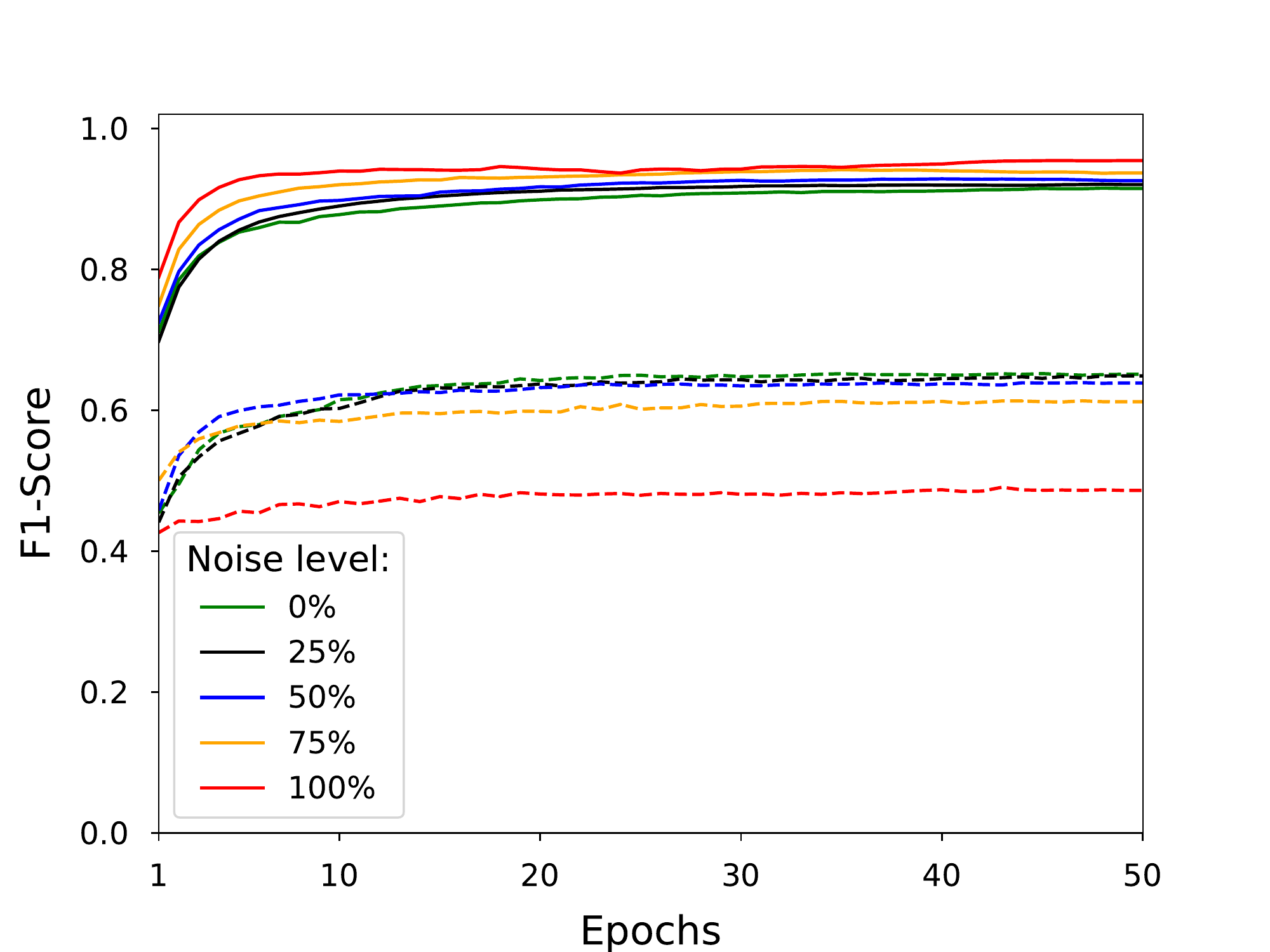}
  \caption{\FOneScore (\cts{})}
  \label{fig:mn_x_replacing_f1_c2s_js}
\end{subfigure}%
\begin{subfigure}{0.32\textwidth}
  \centering
  \includegraphics[width=\linewidth]{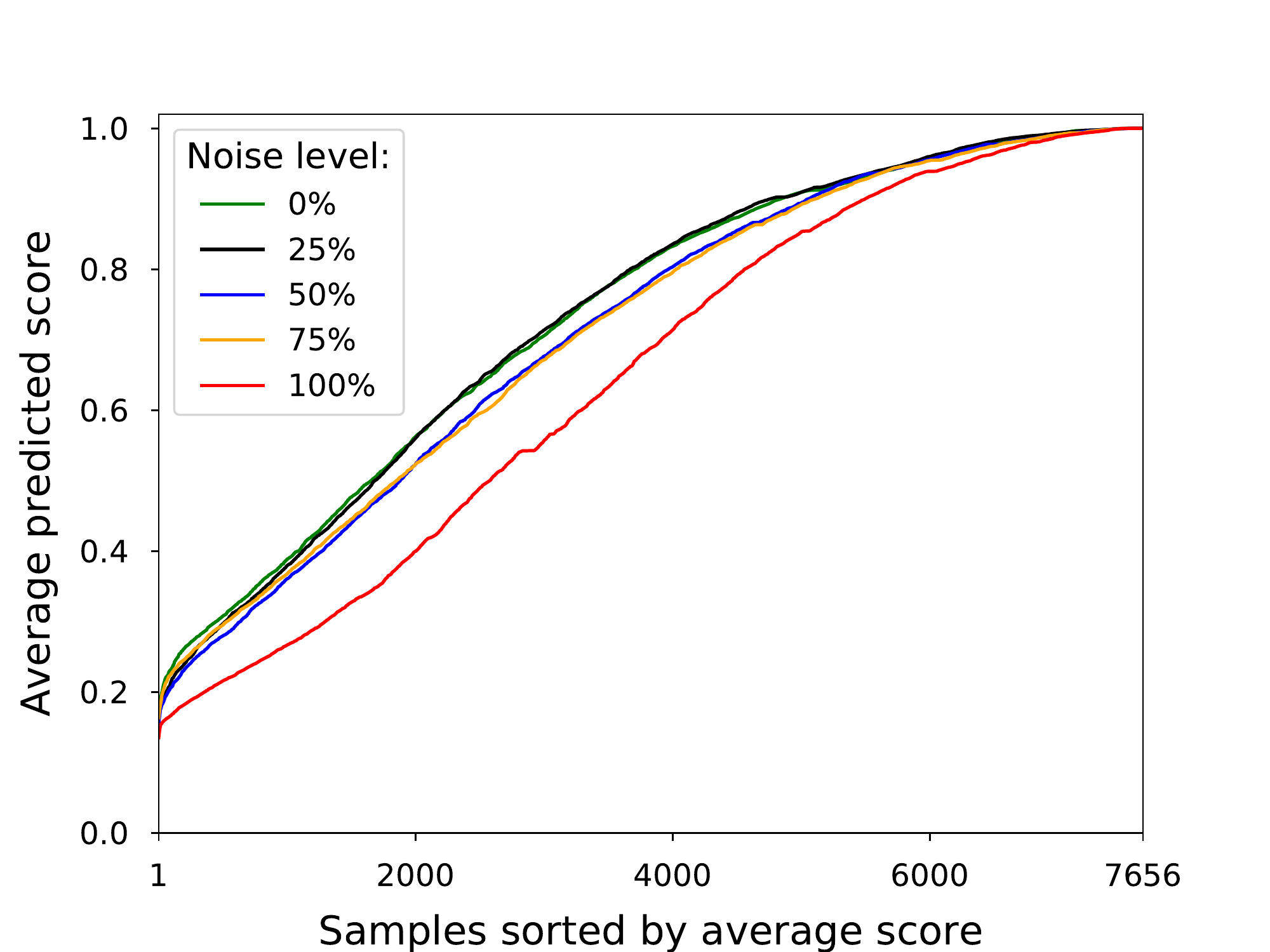}
  \caption{Distribution of Prediction Score (\cts{})}
  \label{fig:mn_x_replacing_score_avg_all_c2s_js}
\end{subfigure}%
\begin{subfigure}{0.32\textwidth}
  \centering
  \includegraphics[width=\linewidth]{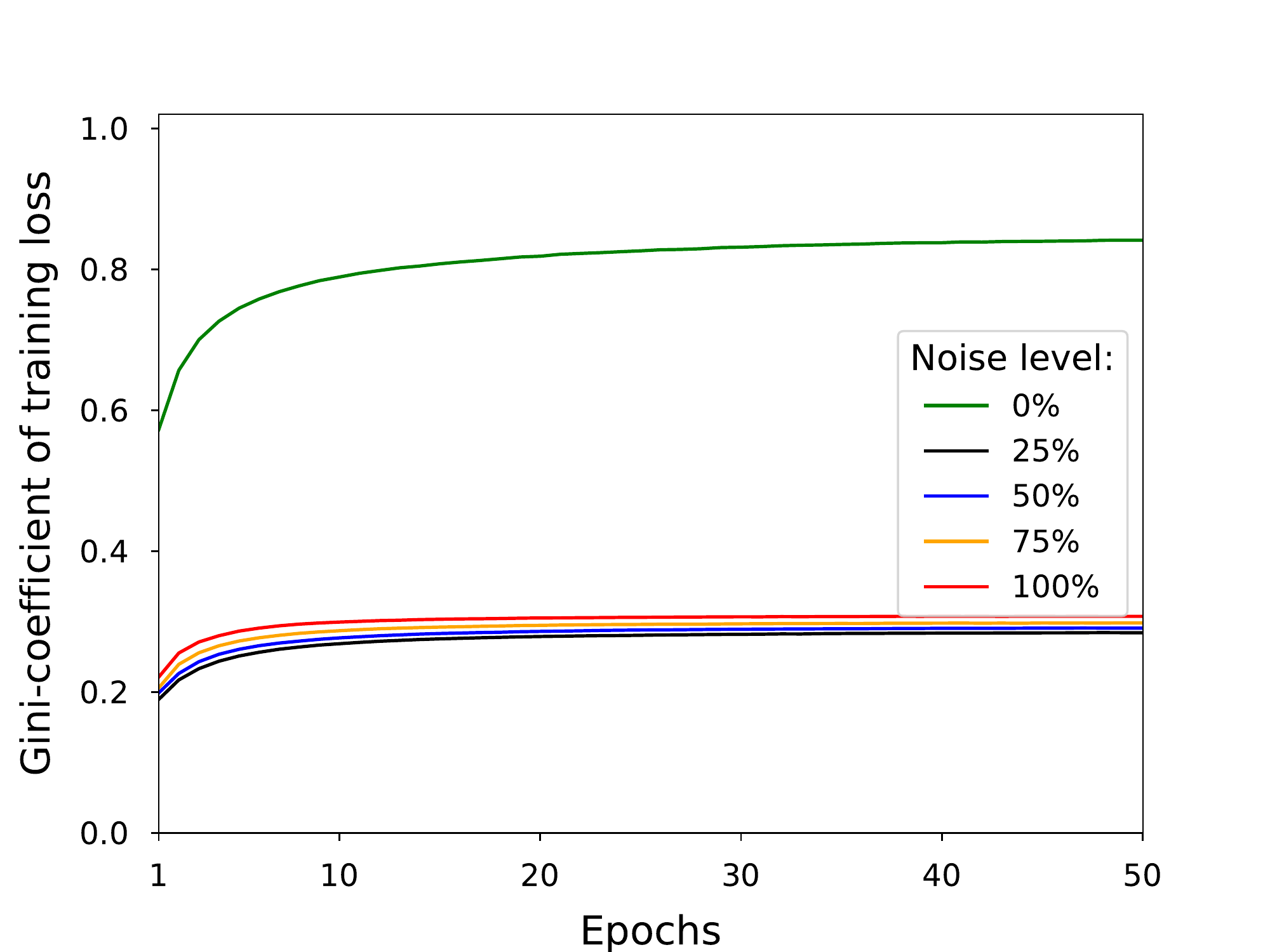}
  \caption{Spread of Training Loss (\cts{})}
  \label{fig:mn_x_replacing_gini_loss_training_c2s_js}
\end{subfigure}

\caption{Input noise by method name leakage (\mnp, \JS).}
\label{fig:mn_x_replacing_js}
\end{figure*}

\subsubsection{Method Name Leakage}
\label{subsec:x-replacing}
We next evaluate a more salient form of input noise for the method name prediction task, in which we replace all occurrences of a variable name with the reference method name, as shown in \Cref{fig:noise_example}${c}$. This enables the model to ``cheat'' by simply copying the method name, if and when present. To make this signal more obvious, we replace the variable that occurs most frequently in the method body. As such, the \mnp task in noisy training samples is reduced to an identity mapping task \cite{he2016identity, zhang2020identity} in which the target is already present in the input. In this case, especially at higher noise levels, the training target is simply picking the maximum occurring variable name in the method body. These identity cues are not inserted into the test data.

\Cref{fig:mn_x_replacing_top10,fig:mn_x_replacing_js} show the resulting changes of input noise by replacing a variable name with the method name on \mnp task for \ctv and \cts models with \JTT dataset and \JS dataset, respectively.
Similar to the output noise characteristics of \FOneScore (\Cref{fig:mn_y_f1_all}), the training performance of \ctv models converges to nearly the same point in both the original and noisy training sets. On the other hand, \cts models are more prone to pick up on the identity cues of method names in noisy training sets. For example, in \Cref{fig:mn_x_replacing_f1_c2s_top10,fig:mn_x_replacing_f1_c2s_js} for \cts, the training \FOneScore of $100$\% or $75$\%  noise is higher than $0$\% noise. The use of identifier splitting, which reduces the rate of unknown tokens, may explain why \cts learns identity cues more effectively than \ctv.
Irrespective of the performance during training, models' performance on the original test set is highly dependent on the noise level; models trained with less noisy data show higher test performance than models trained with higher noisy data. However, as \cts models more frequently recognize method name cues, their test performance also suffers more than \ctv models at higher noise levels.

Additionally, \Cref{fig:mn_x_replacing_top10}$_{b,e}$ and \Cref{fig:mn_x_replacing_js}$_{b,e}$ suggest that the prediction score distribution changes with an increasing noise level. These behaviors are largely consistent with what was previously observed from the prediction score distribution of output noise (\Cref{fig:mn_y_avg_score_test}). The confidence of trained models changes even at low rates of added input noise. The change appears more significant in the smaller-balanced \JTT dataset than in the larger-skewed \JS dataset.

Comparing the Gini coefficient values of input noise from \Cref{fig:mn_x_replacing_top10}$_{c,f}$ and \Cref{fig:mn_x_replacing_js}$_{c,f}$ with those under output noise in \Cref{fig:mn_y_gini_loss_training} may suggest a nearly reversed trend in the spread of training loss: it decreases for output noise but increases for input noise with an increasing noise level. This can be explained by the presence of identity cues in the training set. In the case of output noise, we replace a method name with a different randomly selected method name. Thus, similar code snippets may have very different method names, which makes the learning process more complicated. However, in the case of input noise, we add the target method name inside the input code snippets, thus making the learning process much simpler when such identify cues are present.

\observation{Signal leaking input noise can help models achieve high performance during training, but hurts their ability to generalize on non-noisy test data.}

\begin{figure*}
\centering
\captionsetup[subfigure]{width=\textwidth, justification=centering}

\begin{subfigure}{0.32\textwidth}
  \centering
  \includegraphics[width=\linewidth]{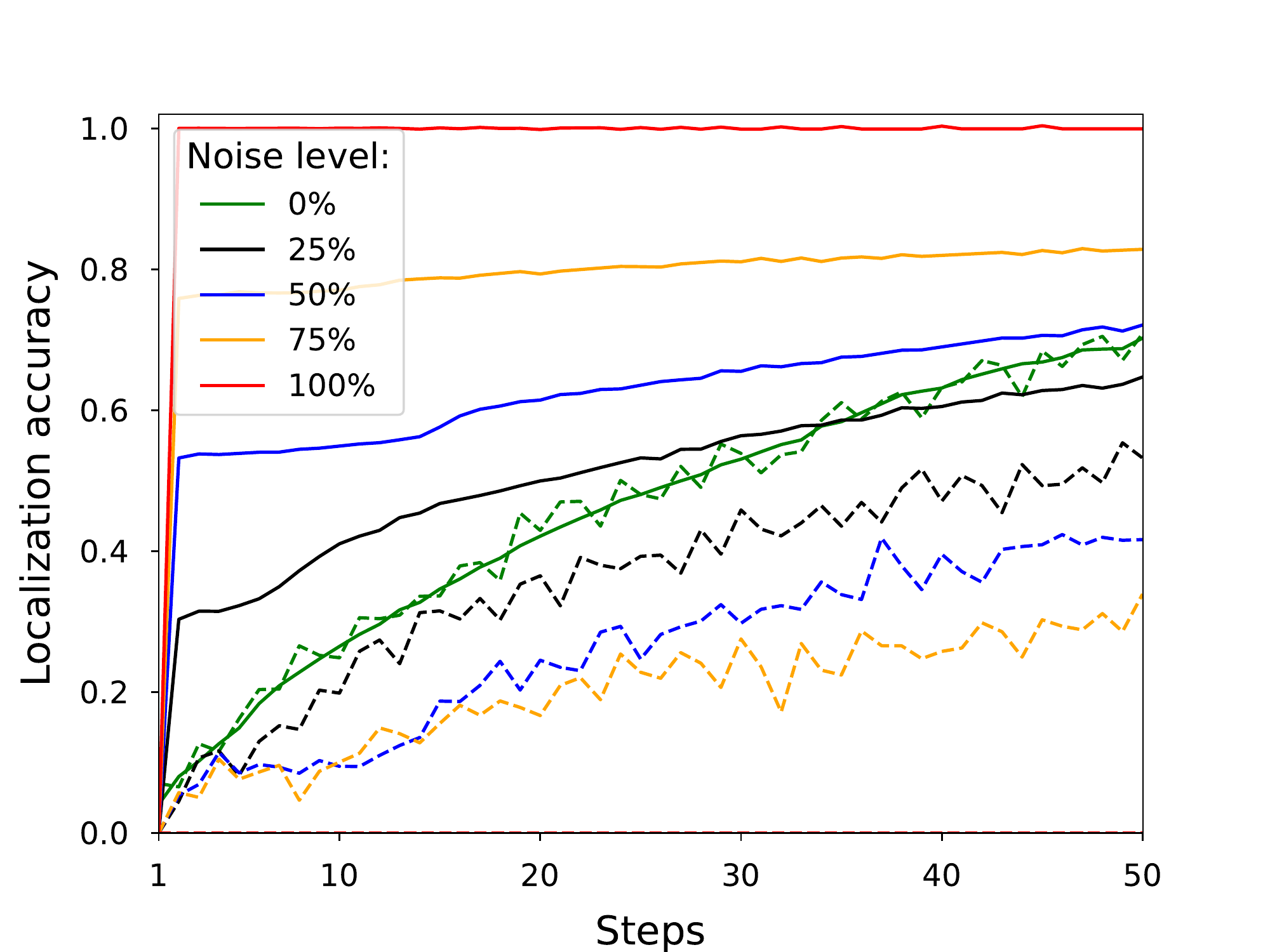}
  \caption{Localization Accuracy}
  \label{fig:vm_x_loc_acc_transformer_py}
\end{subfigure}%
\begin{subfigure}{0.32\textwidth}
  \centering
  \includegraphics[width=\linewidth]{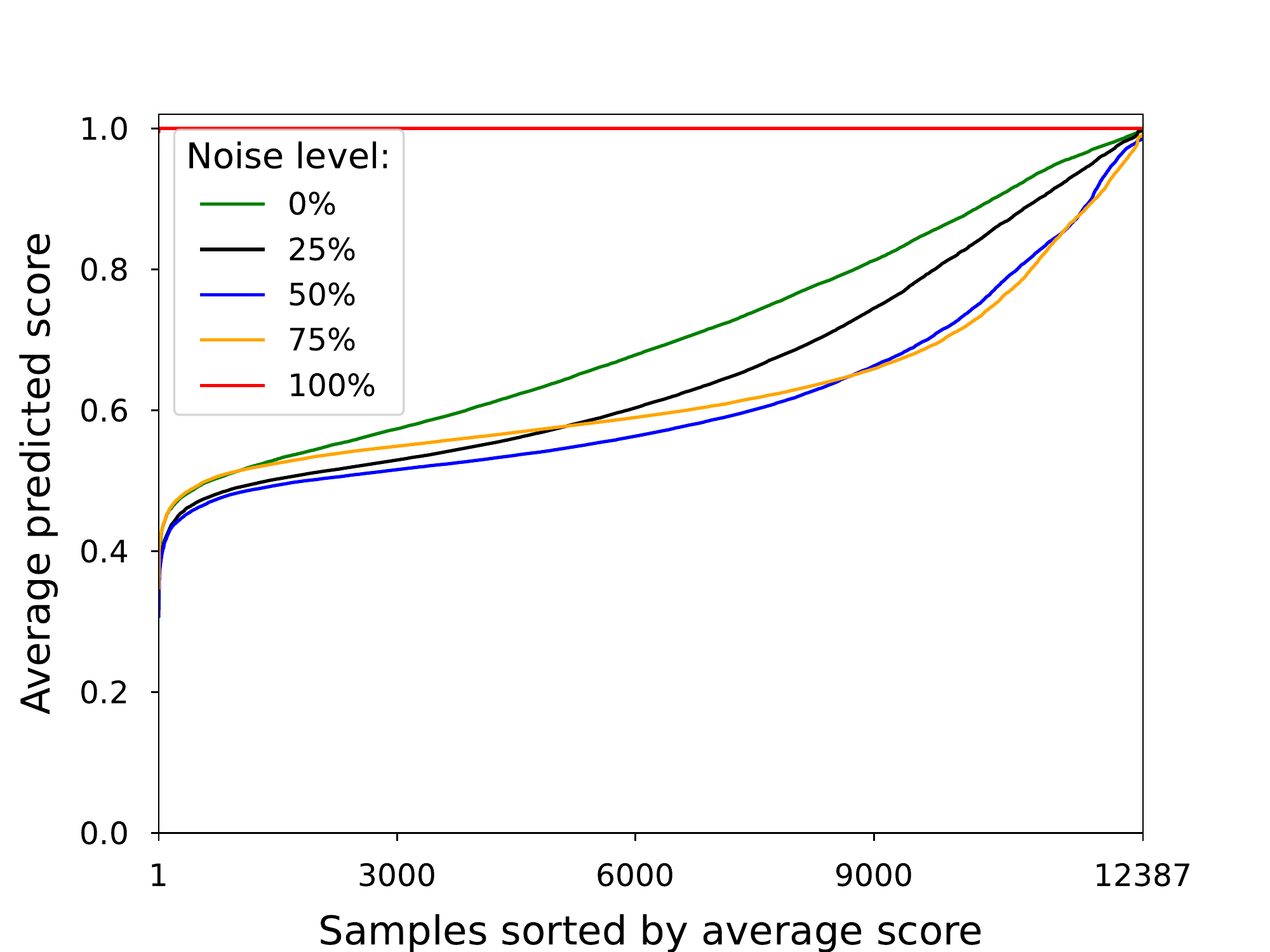}
  \caption{Distribution of Localization Score}
  \label{fig:vm_x_avg_loc_score_transformer_py}
\end{subfigure}%
\begin{subfigure}{0.32\textwidth}
  \centering
  \includegraphics[width=\linewidth]{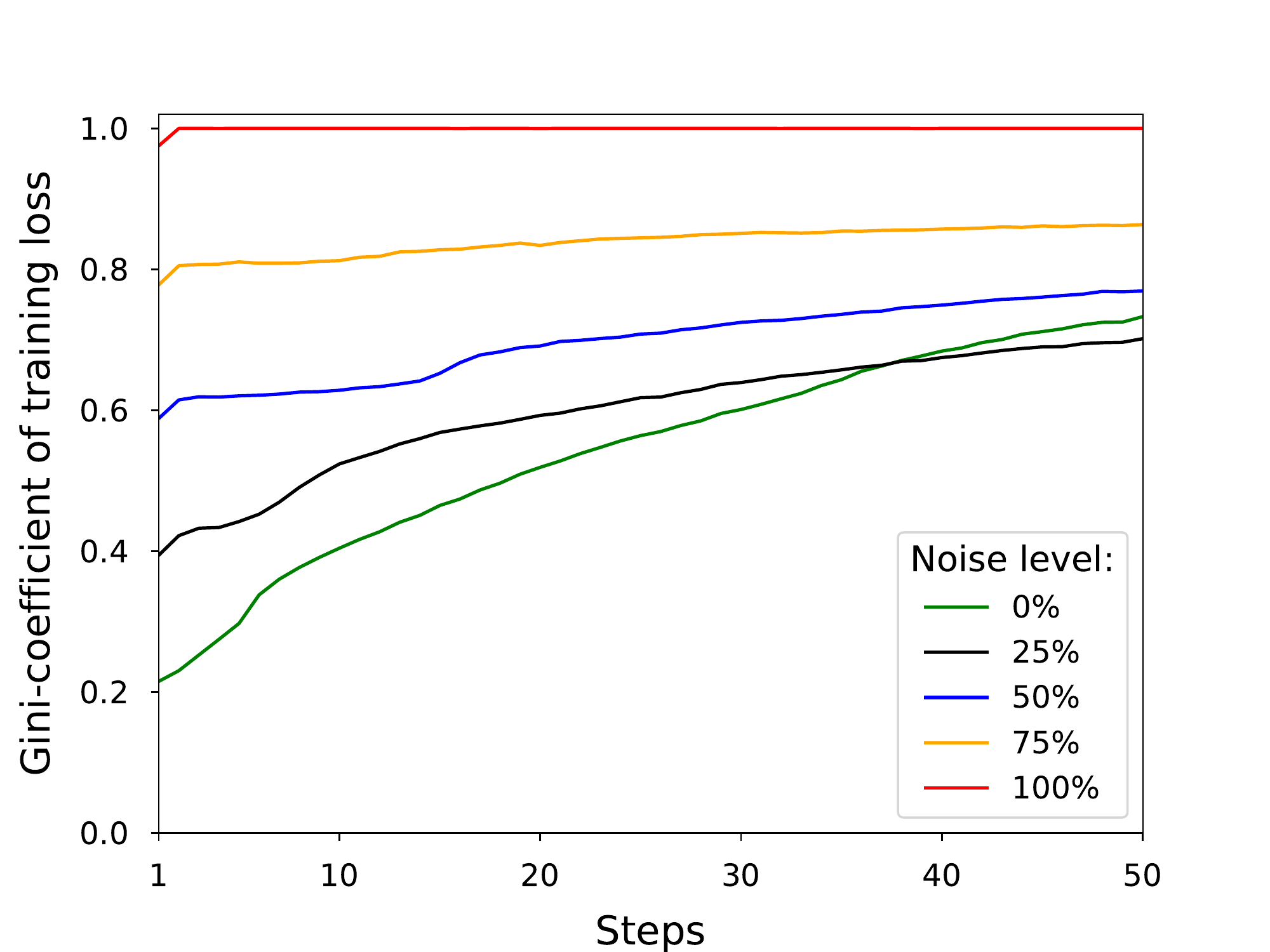}
  \caption{Spread of Localization Loss}
  \label{fig:vm_x_train_loc_loss_transformer_py}
\end{subfigure}

\begin{subfigure}{0.32\textwidth}
  \centering
  \includegraphics[width=\linewidth]{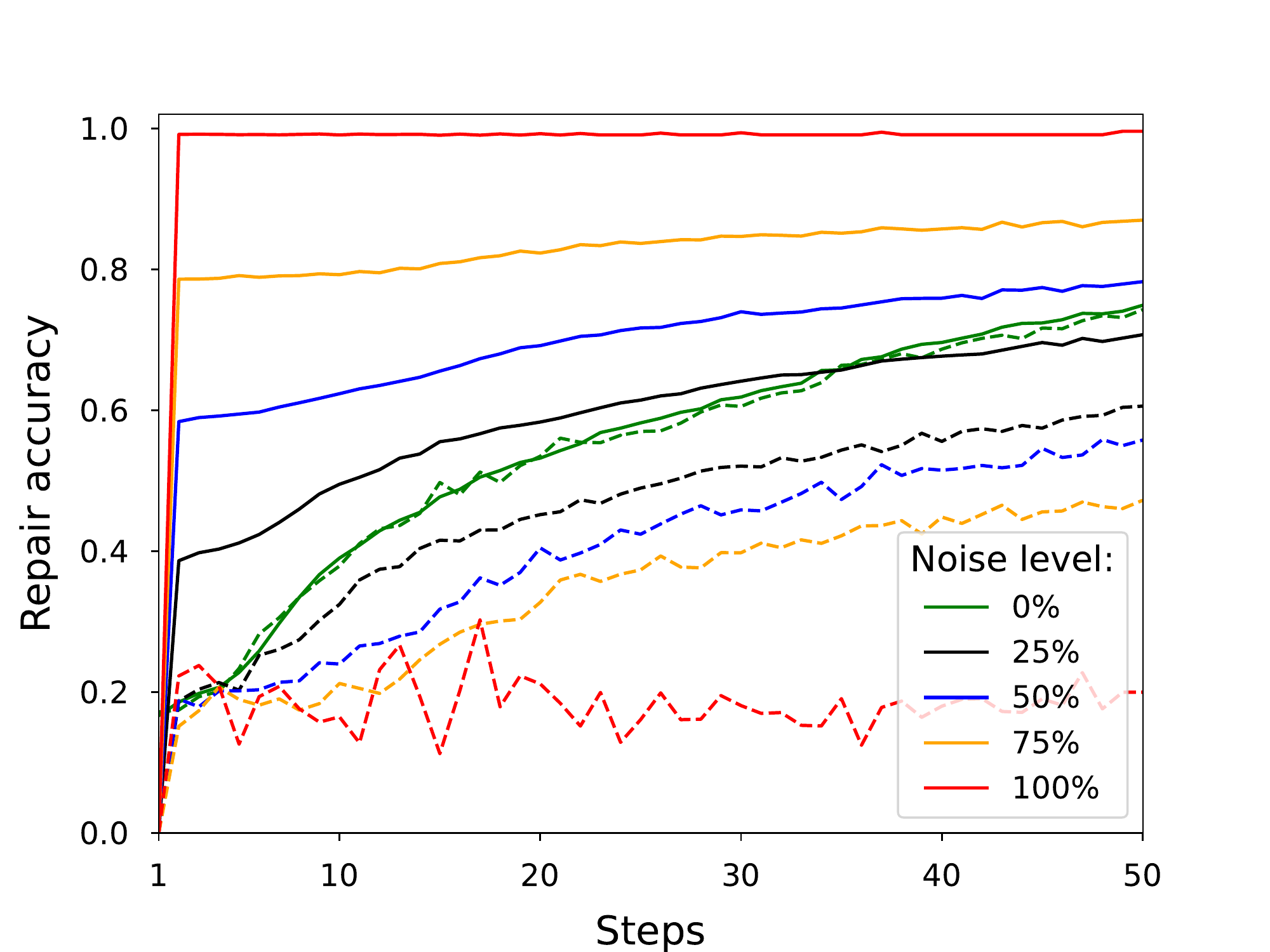}
  \caption{Repair Accuracy}
  \label{fig:vm_x_rep_acc_transformer_py}
\end{subfigure}%
\begin{subfigure}{0.32\textwidth}
  \centering
  \includegraphics[width=\linewidth]{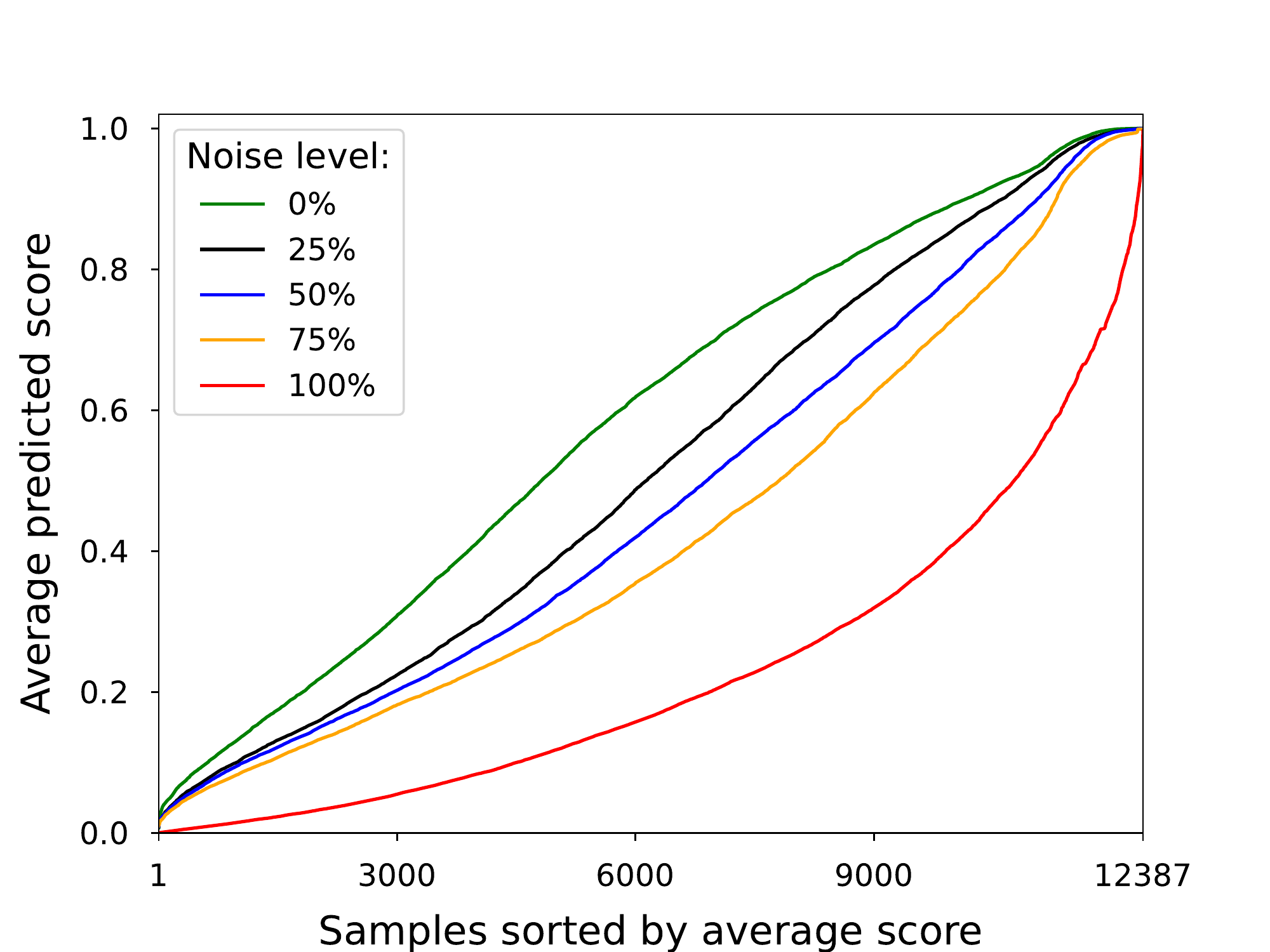}
  \caption{Distribution of Repair Score}
  \label{fig:vm_x_avg_rep_score_transformer_py}
\end{subfigure}%
\begin{subfigure}{0.32\textwidth}
  \centering
  \includegraphics[width=\linewidth]{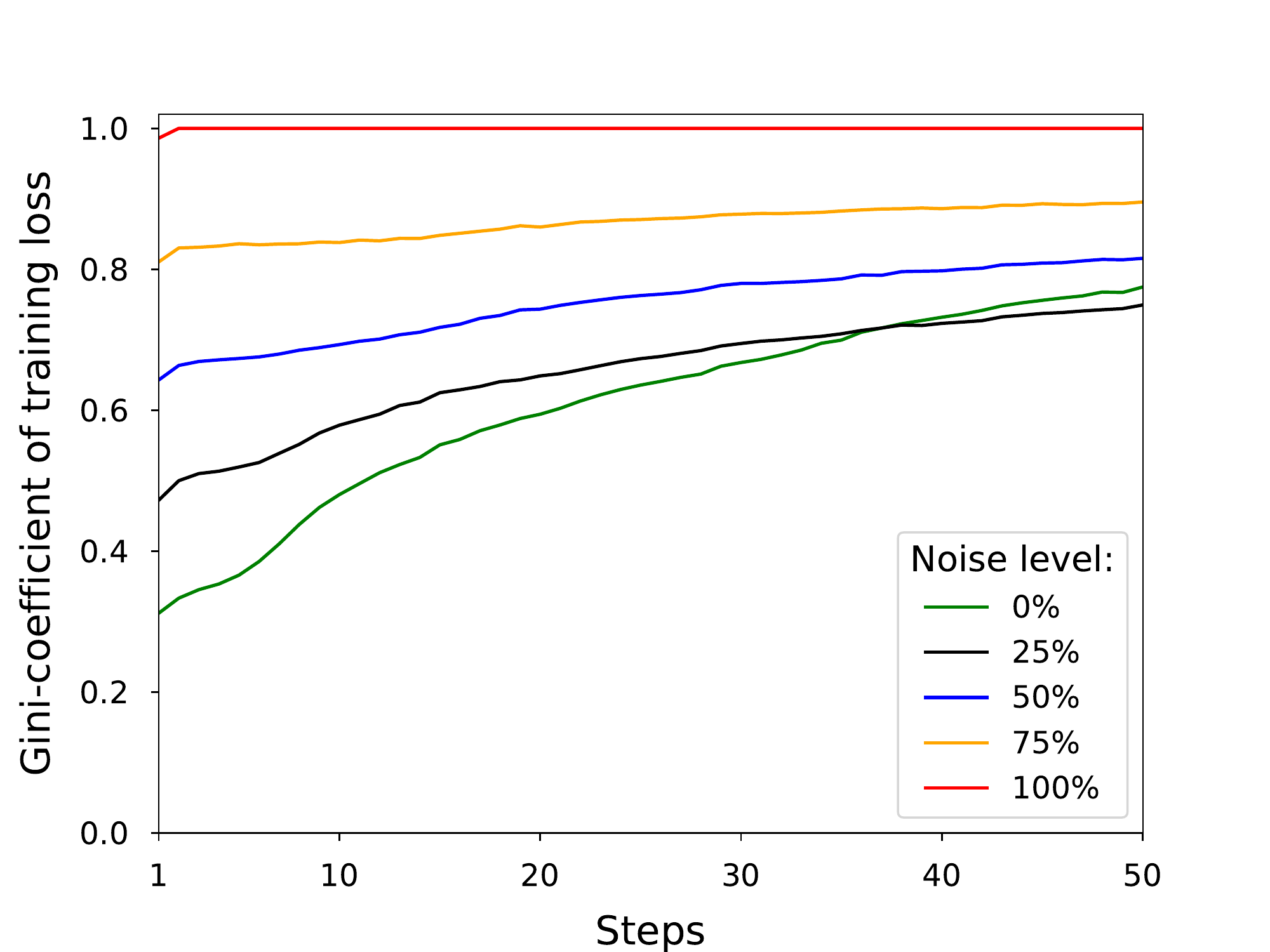}
  \caption{Spread of Repair Loss}
  \label{fig:vm_x_train_rep_loss_transformer_py}
\end{subfigure}

\caption{Input noise by injecting repair information \mbox{(\vm, \tra)}.}
\label{fig:vm_x_buggy_transformer}


\begin{subfigure}{0.32\textwidth}
  \centering
  \includegraphics[width=\linewidth]{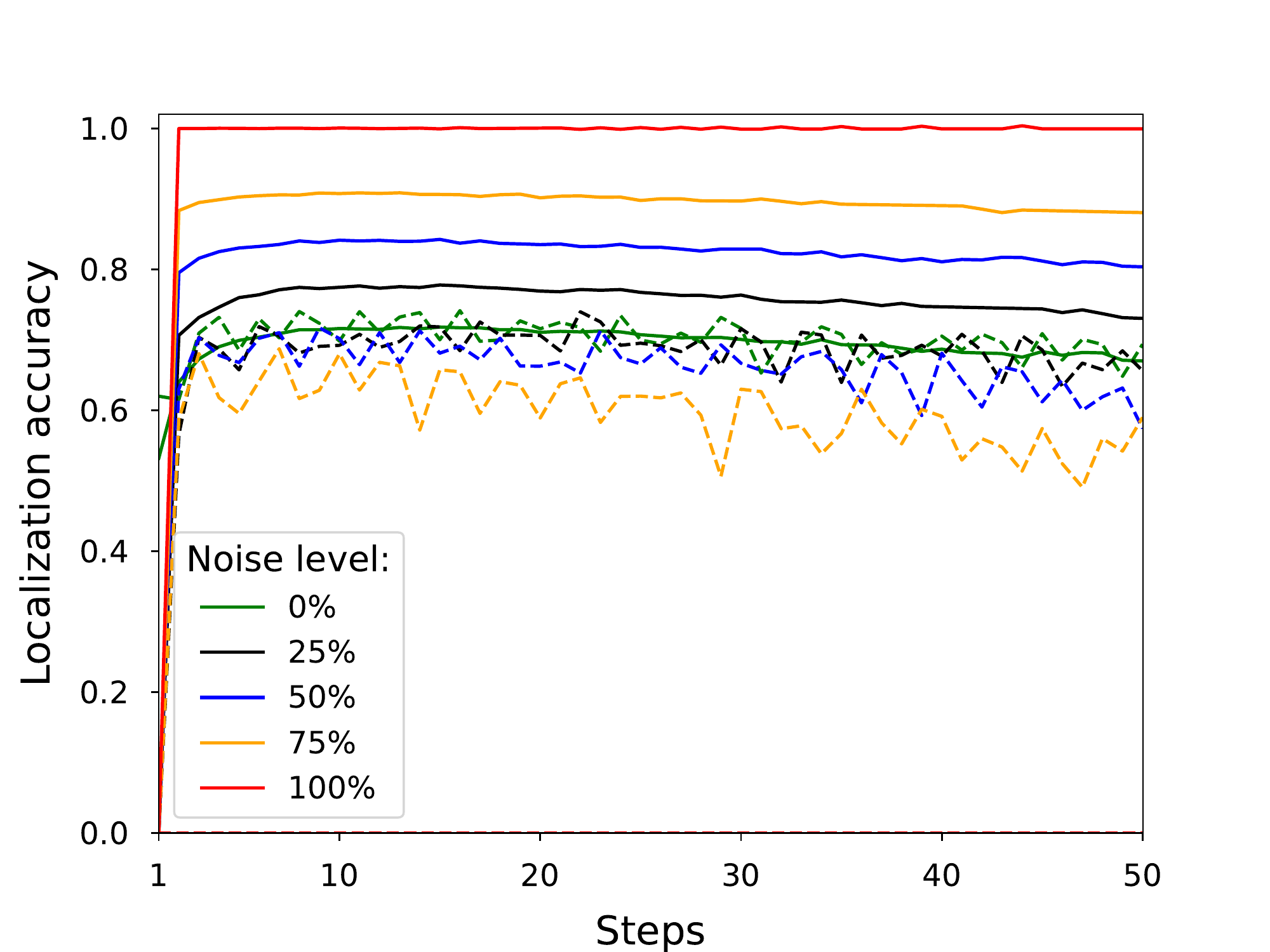}
  \caption{Localization Accuracy}
  \label{fig:vm_x_loc_acc_ggnn_py}
\end{subfigure}%
\begin{subfigure}{0.32\textwidth}
  \centering
  \includegraphics[width=\linewidth]{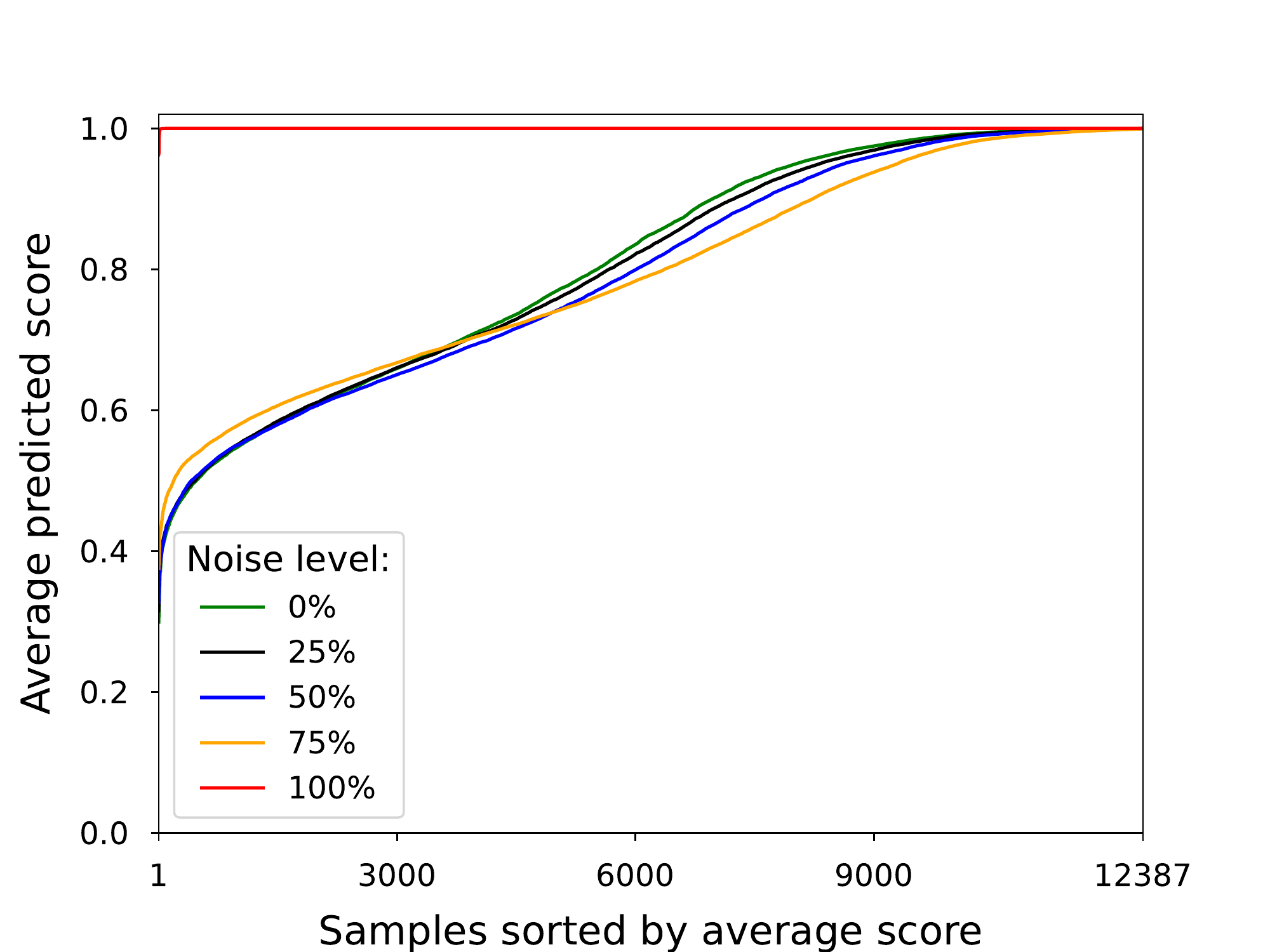}
  \caption{Distribution of Localization Score}
  \label{fig:vm_x_avg_loc_score_ggnn_py}
\end{subfigure}%
\begin{subfigure}{0.32\textwidth}
  \centering
  \includegraphics[width=\linewidth]{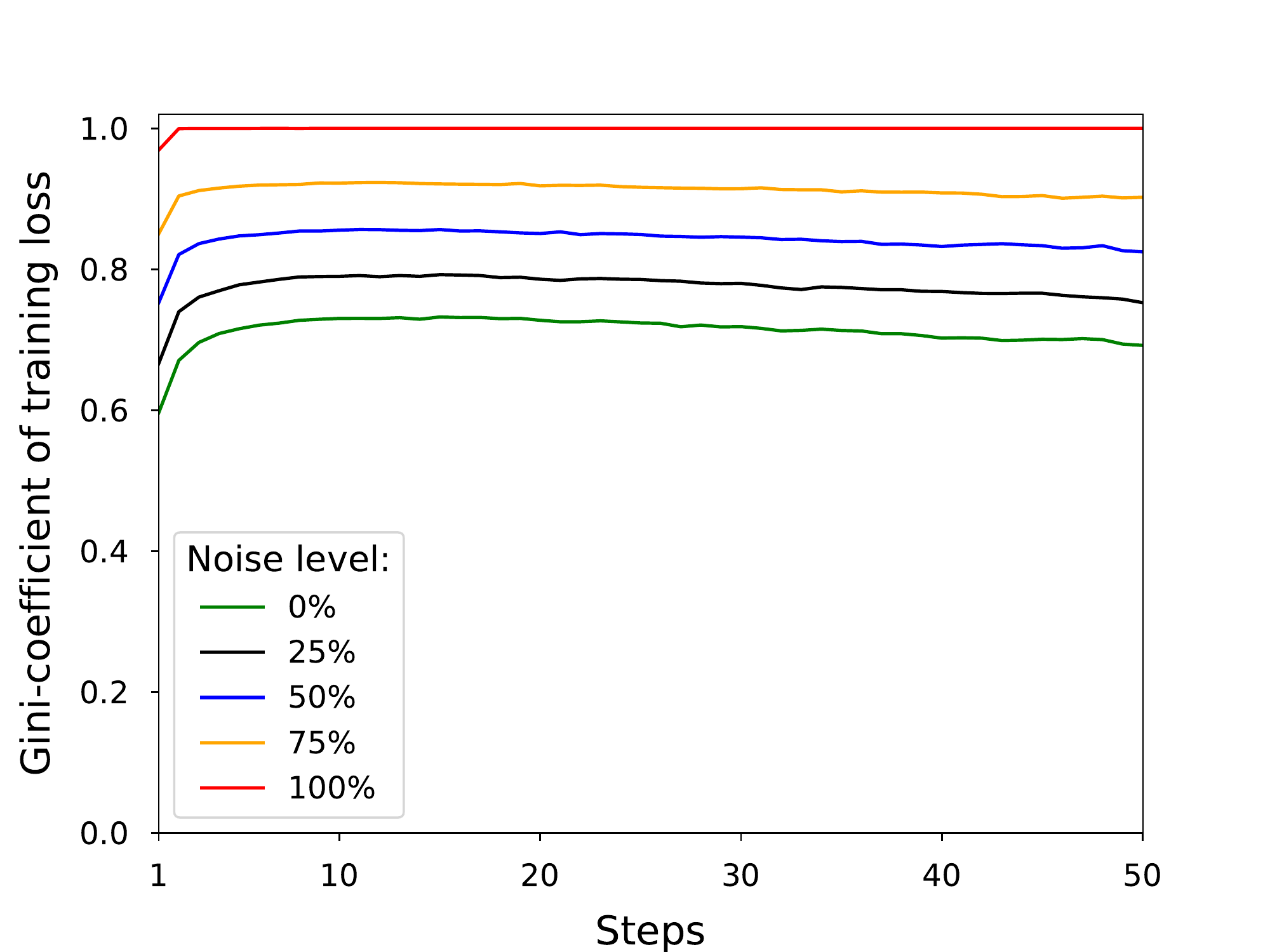}
  \caption{Spread of Localization Loss}
  \label{fig:vm_x_train_loc_loss_ggnn_py}
\end{subfigure}

\begin{subfigure}{0.32\textwidth}
  \centering
  \includegraphics[width=\linewidth]{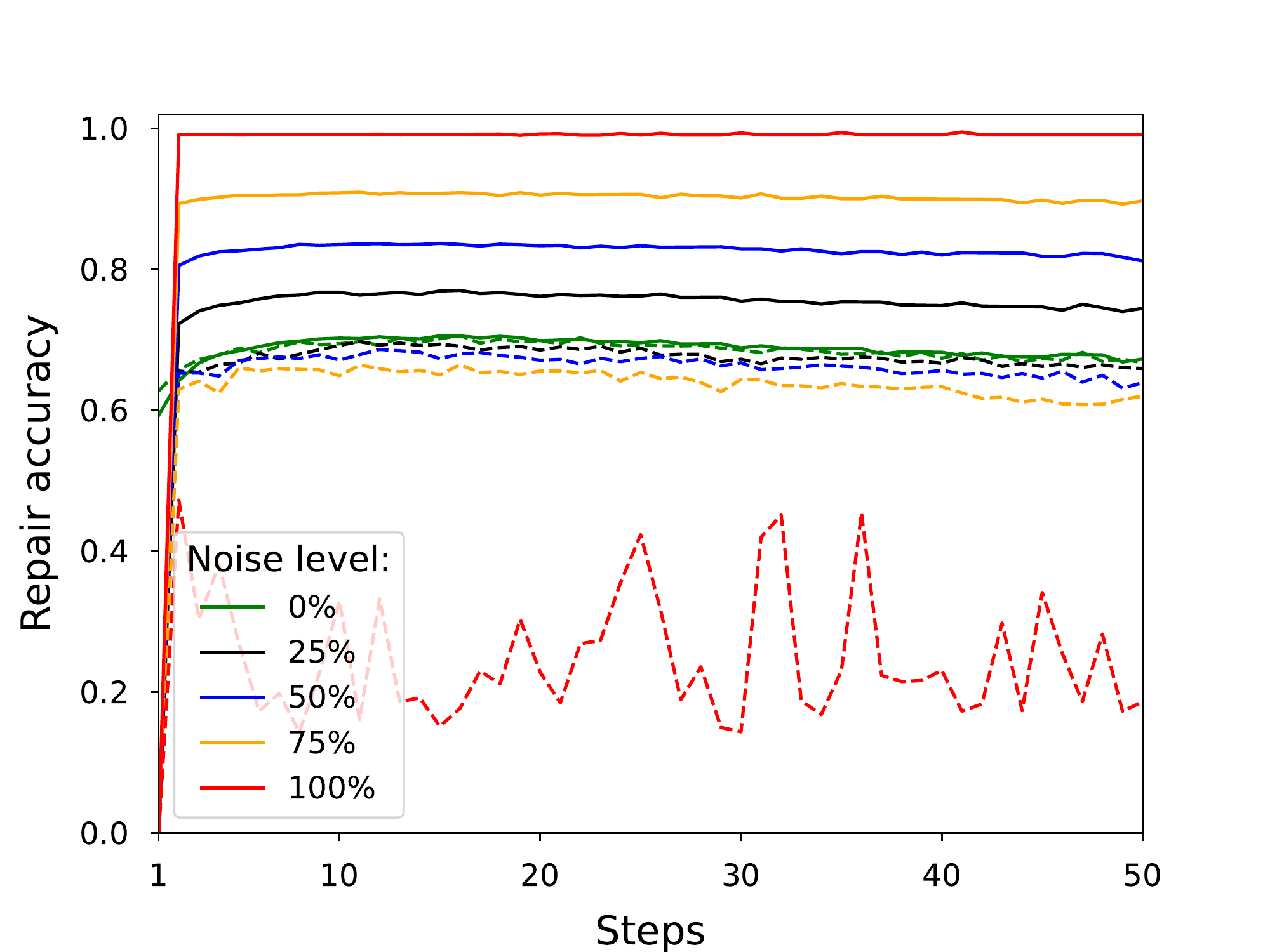}
  \caption{Repair Accuracy}
  \label{fig:vm_x_rep_acc_ggnn_py}
\end{subfigure}%
\begin{subfigure}{0.32\textwidth}
  \centering
  \includegraphics[width=\linewidth]{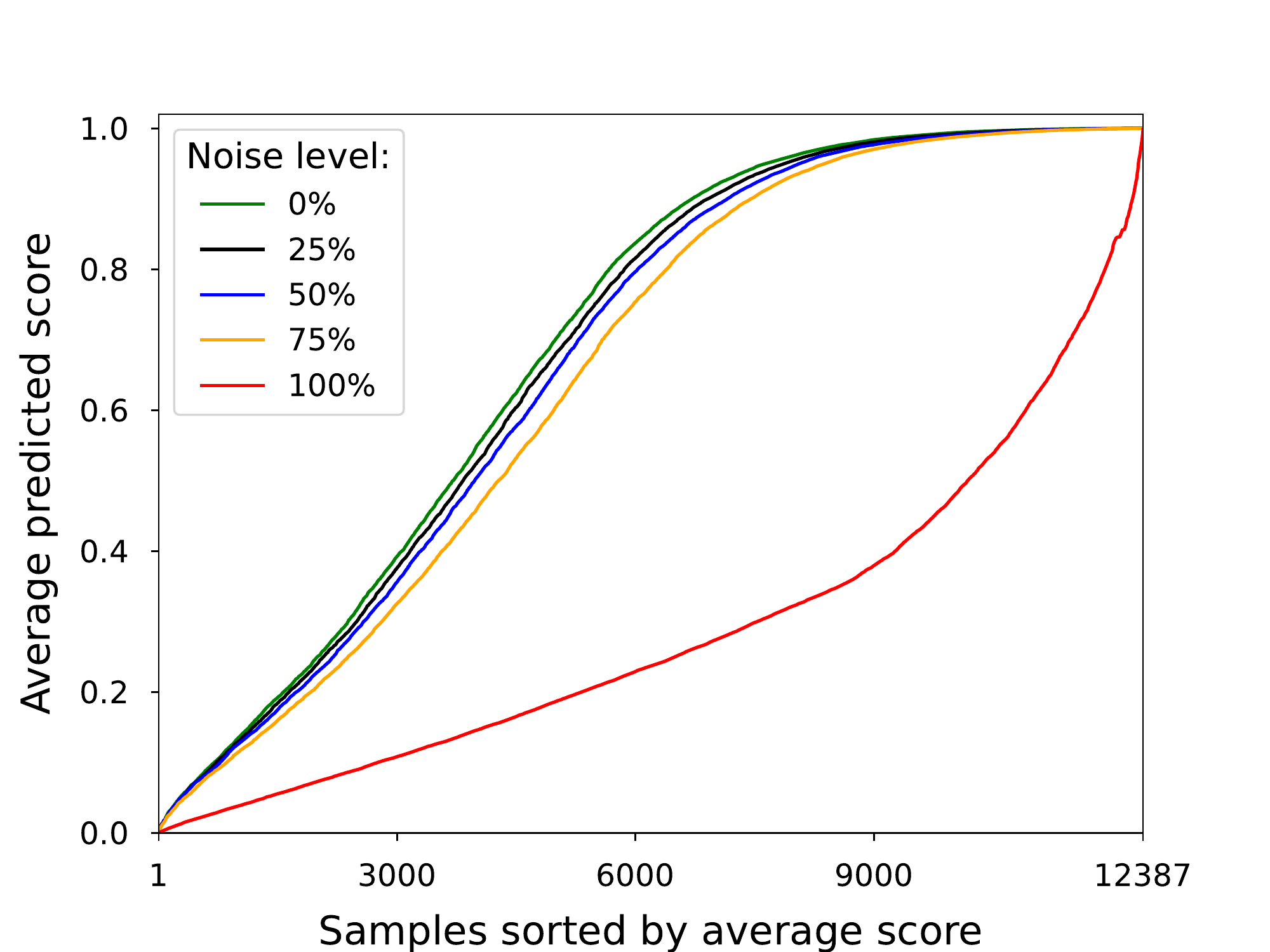}
  \caption{Distribution of Repair Score}
  \label{fig:vm_x_avg_rep_score_ggnn_py}
\end{subfigure}%
\begin{subfigure}{0.32\textwidth}
  \centering
  \includegraphics[width=\linewidth]{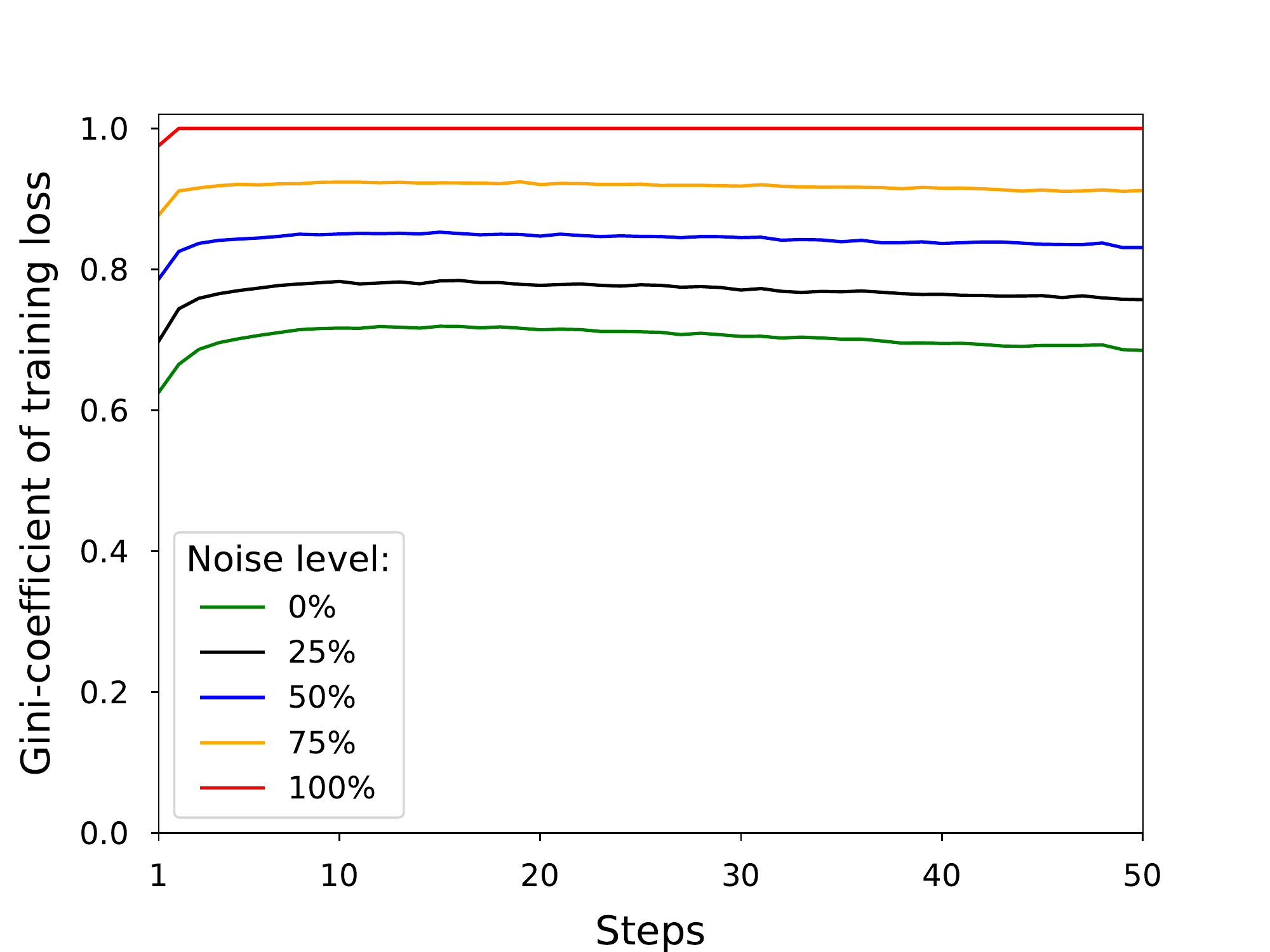}
  \caption{Spread of Repair Loss}
  \label{fig:vm_x_train_rep_loss_ggnn_py}
\end{subfigure}

\caption{Input noise by injecting repair information \mbox{(\vm, \ggnn)}.}
\label{fig:vm_x_buggy_ggnn}
\end{figure*}

\subsubsection{Injecting Repair Information}
\label{subsec:vm-x-buggy}
We next apply a similar form of input noise to the \vm tasks and dataset. As shown in \Cref{fig:noise_example}$f$, we add noise to the training set by inserting repair data into buggy methods by replacing the variable at the error location and all occurrences of the correct variable that should be at that location, and replacing the most frequent token in a correct sample with a cue that no bug is present.

\Cref{fig:vm_x_buggy_transformer,fig:vm_x_buggy_ggnn} show the resulting impact of applying input noise in the above manner on \tra and \ggnn models trained on the \vm dataset. Similar to the \cts models on the \mnp task, the \tra and \ggnn models are highly capable of learning identity cues of the replacement tokens (error location variable and correct target variable) in noisy training sets, for both the localization and repair tasks: the training \FOneScore under $100$\% noise (full signal leakage) almost immediately jump to perfect quality, as the models learn to predict based on the presence of the inserted cues. The performance of the models on the held-out data, where such cues are absent, consequently drops to near-chance performance.

Moreover, the prediction score distribution generally skews lower with increasing input noise level (\Cref{fig:vm_x_buggy_transformer}$_{b,e}$ and \Cref{fig:vm_x_buggy_ggnn}$_{b,e}$), with the exception of the 100\% noise-level, which always predicts locations and repairs with complete confidence. It seems, for the localization task, the role of our replacements as identity cues for the models become most prominent when the replacements are carried out for all the samples in the training set.

Finally, the Gini coefficient values in both the localization and repair tasks (\Cref{fig:vm_x_buggy_transformer}$_{c,f}$ and \Cref{fig:vm_x_buggy_ggnn}$_{c,f}$) show that the training loss increases with the addition of input noise levels. These trends are the reverse of the trends when we applied varying levels of output noise, as seen in \Cref{fig:vm_y_gini_loss_training}; which can be explained by the impact of identity cues, as described in \Cref{subsec:x-replacing}. 

In sum, \vm models may achieve high performance on noisy datasets during training, even on higher input noise levels, but, as seen for \mnp models, always fail to generalize during evaluation on the original non-noisy held-out data.

\subsection{Analyses of Memorization with a Language Model}
\label{subsec:noise-on-lm}

\begin{figure*}
\centering
\captionsetup[subfigure]{width=0.9\textwidth, justification=centering}

\begin{subfigure}{0.32\textwidth}
  \centering
  \includegraphics[width=\linewidth]{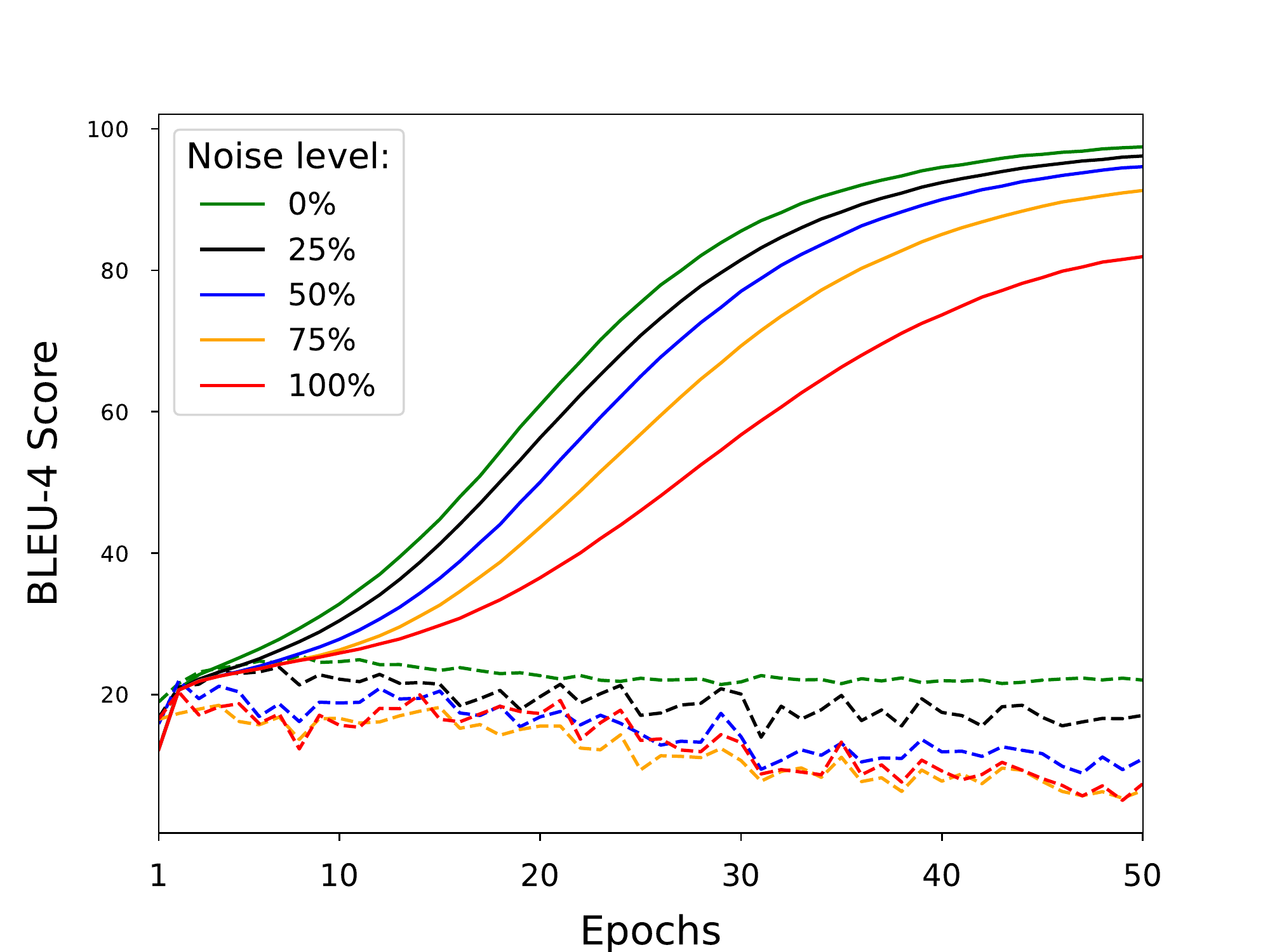}
  \caption{Smoothed BLEU-4 Score (\cnl)}
  \label{fig:y_code2nl_bleu4}
\end{subfigure}%
\begin{subfigure}{0.32\textwidth}
  \centering
  \includegraphics[width=\linewidth]{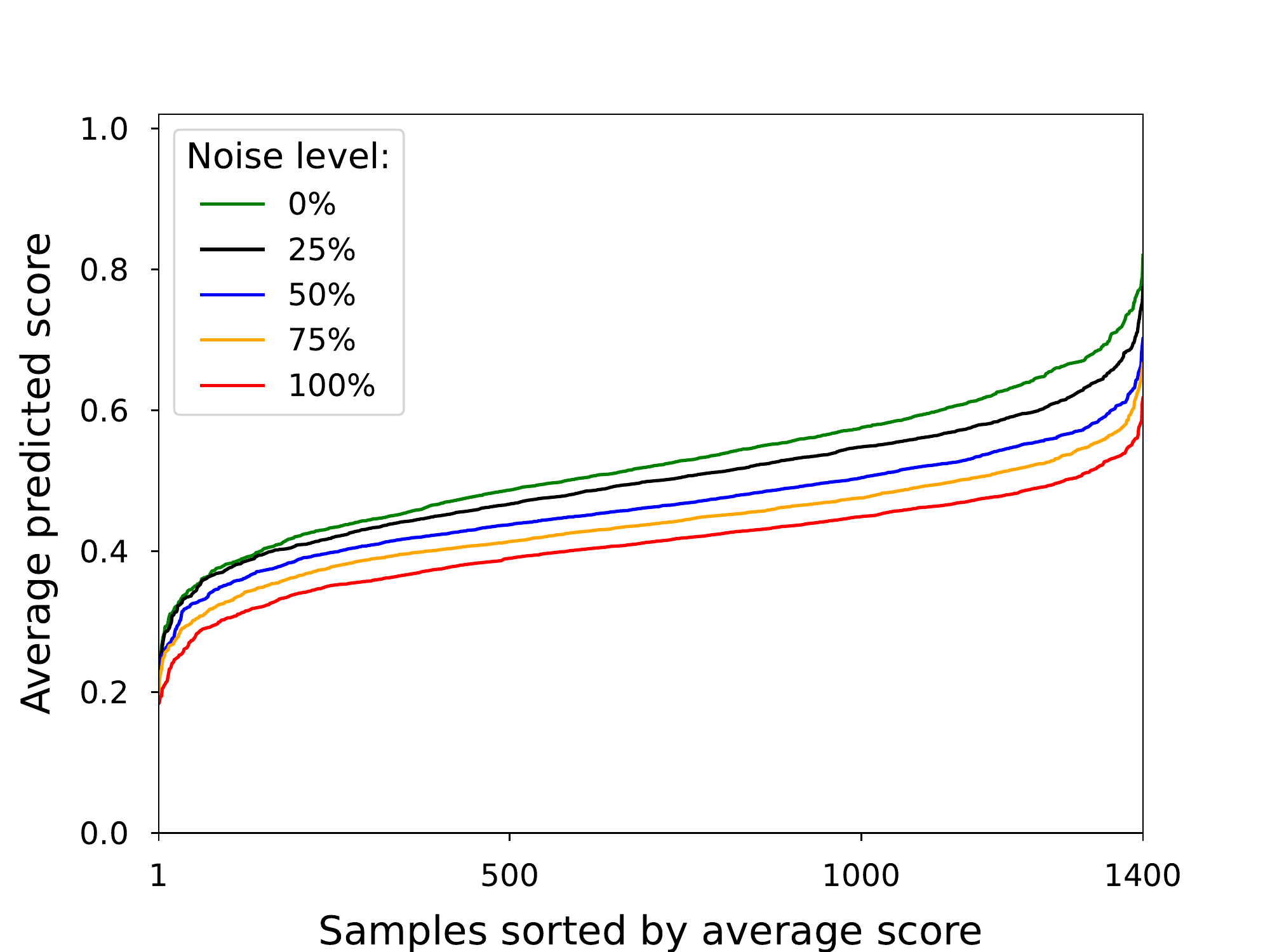}
  \caption{Distribution of Prediction Score (\cnl)}
  \label{fig:y_code2nl_score_avg}
\end{subfigure}%
\begin{subfigure}{0.32\textwidth}
  \centering
  \includegraphics[width=\linewidth]{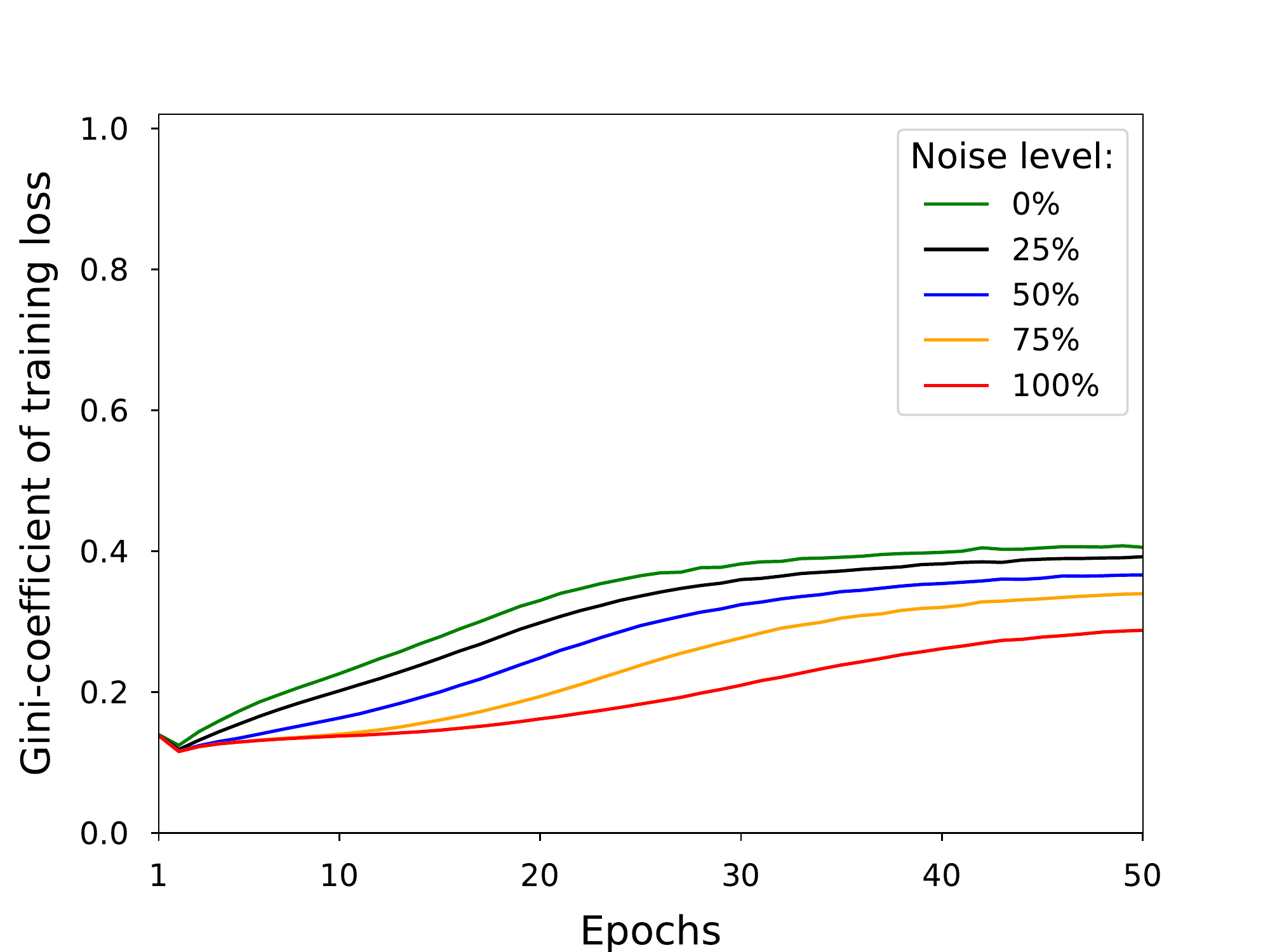}
  \caption{Spread of Training Loss (\cnl)}
  \label{fig:y_code2nl_gini_loss}
\end{subfigure}

\begin{subfigure}{0.32\textwidth}
  \centering
  \includegraphics[width=\linewidth]{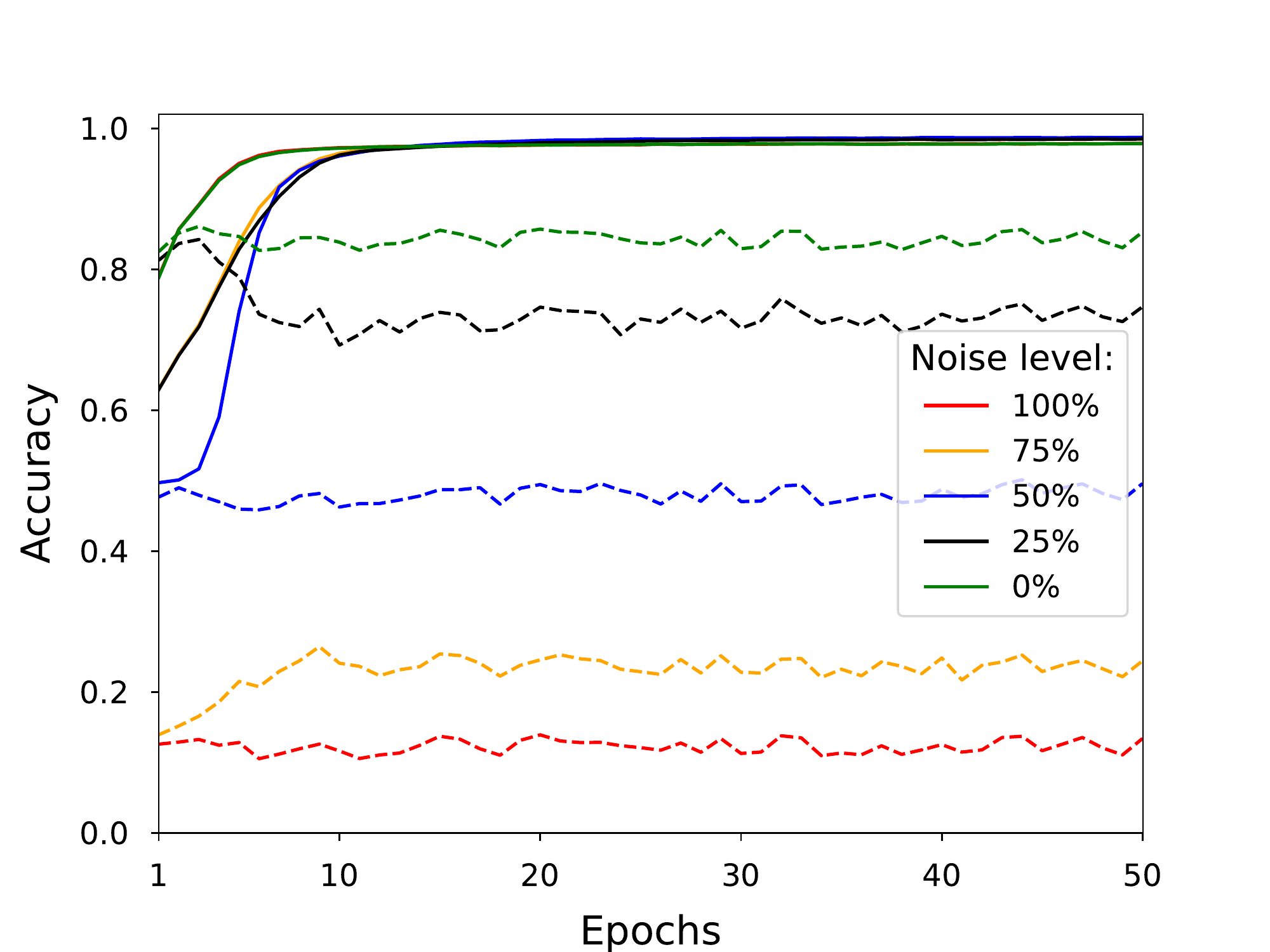}
  \caption{Accuracy (\csearch)}
  \label{fig:y_codesearch_accuracy}
\end{subfigure}%
\begin{subfigure}{0.32\textwidth}
  \centering
  \includegraphics[width=\linewidth]{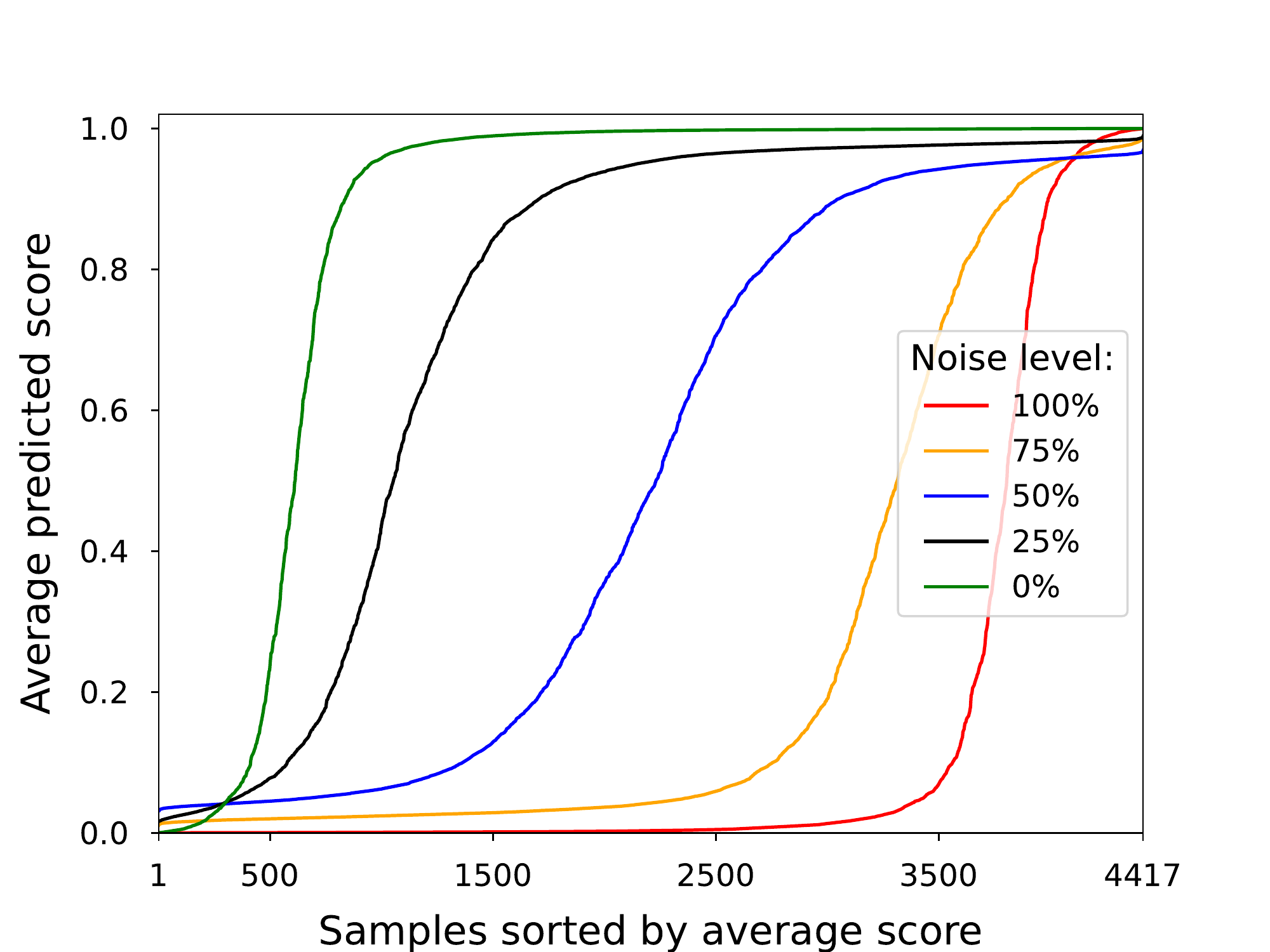}
  \caption{Distribution of Prediction Score (\csearch)}
  \label{fig:y_codesearch_score_avg}
\end{subfigure}%
\begin{subfigure}{0.32\textwidth}
  \centering
  \includegraphics[width=\linewidth]{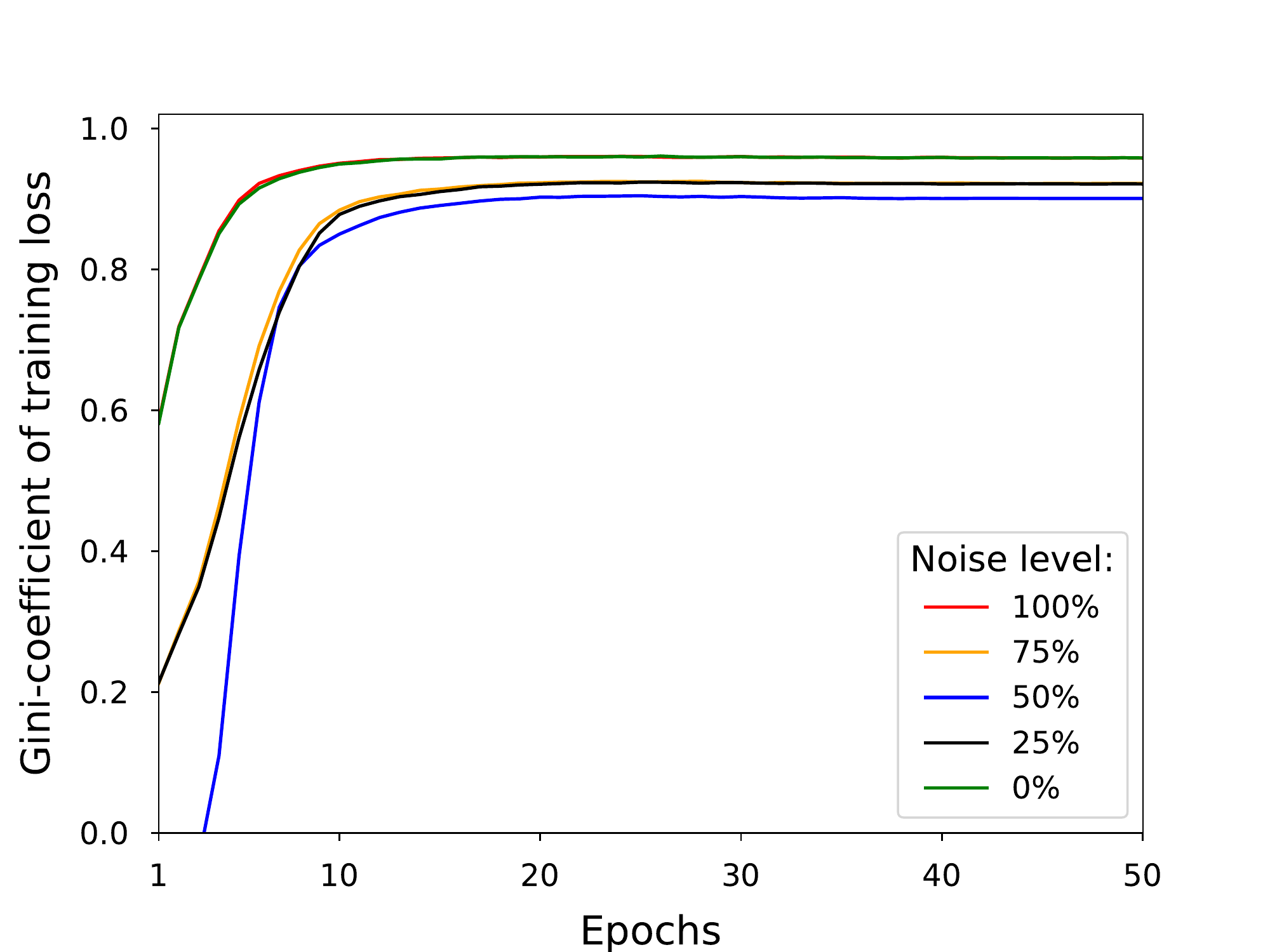}
  \caption{Spread of Training Loss (\csearch)}
  \label{fig:y_codesearch_gini_loss}
\end{subfigure}

\caption{Impact of output noise in fine-tuning \CodeBERT language model.}
\label{fig:lm_codeBERT_output}


\begin{subfigure}{0.32\textwidth}
  \centering
  \includegraphics[width=\linewidth]{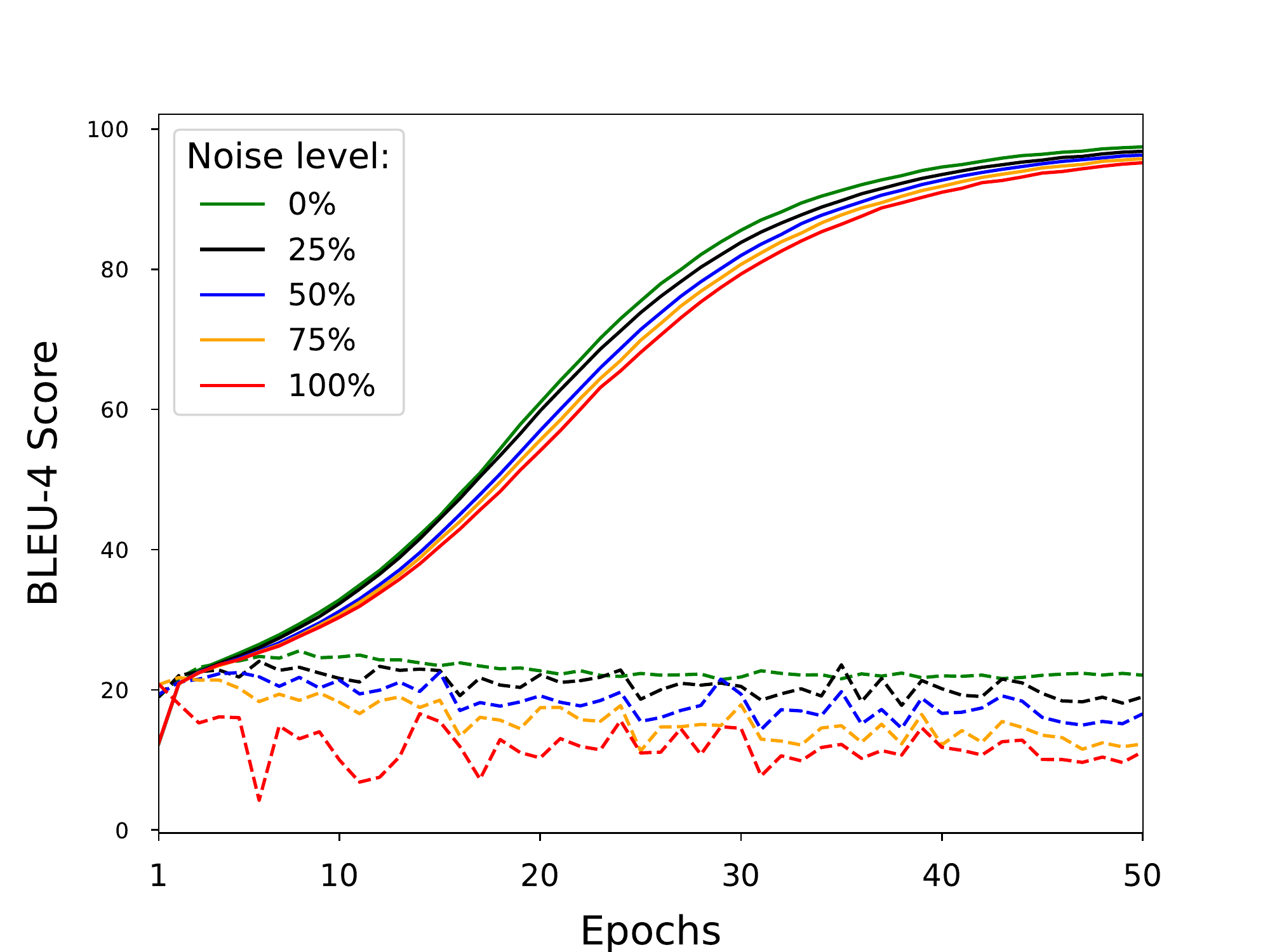}
  \caption{Smoothed BLEU-4 Score (\cnl)}
  \label{fig:x_code2nl_bleu4}
\end{subfigure}%
\begin{subfigure}{0.32\textwidth}
  \centering
  \includegraphics[width=\linewidth]{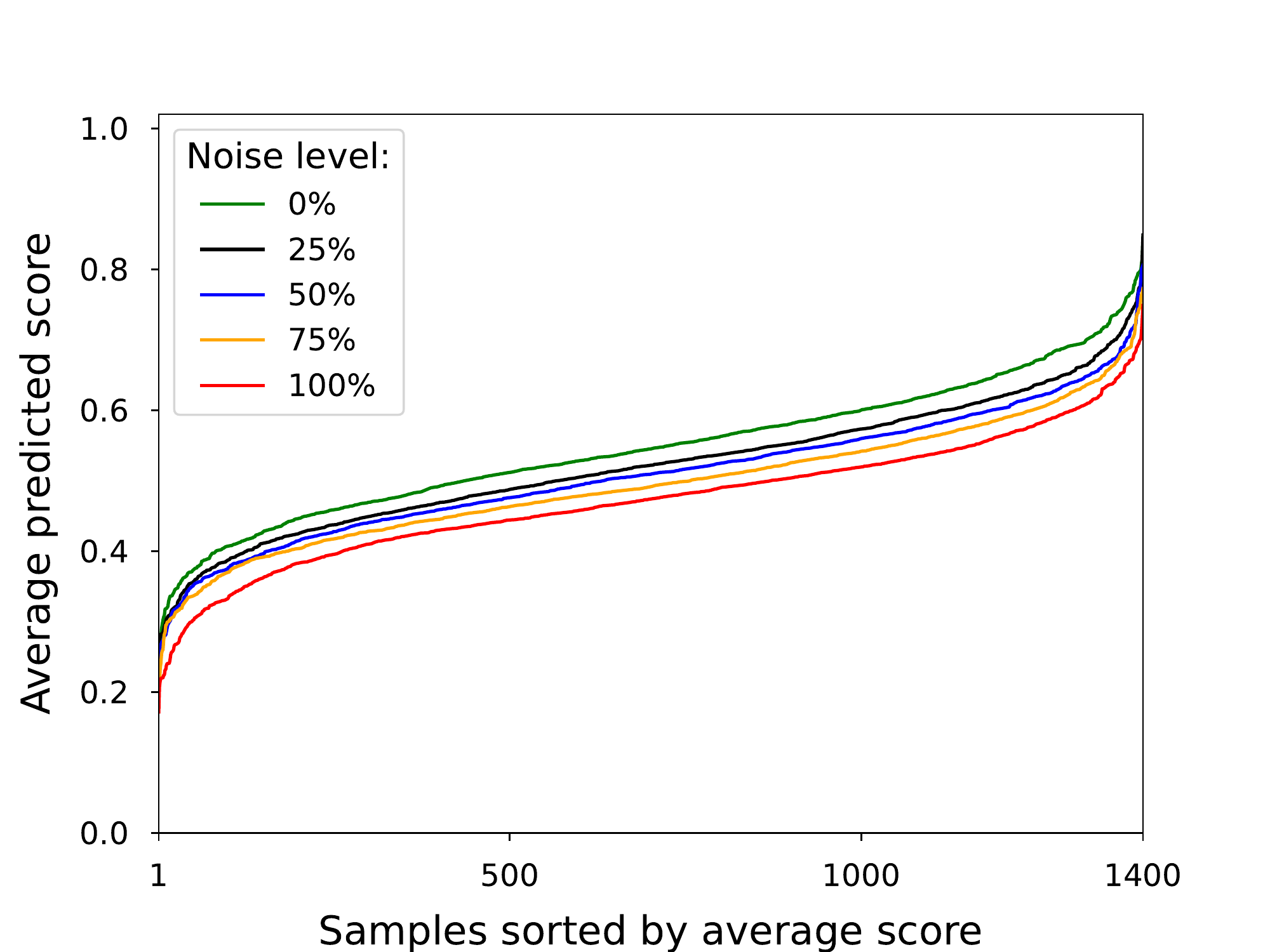}
  \caption{Distribution of Prediction Score (\cnl)}
  \label{fig:x_code2nl_score_avg}
\end{subfigure}%
\begin{subfigure}{0.32\textwidth}
  \centering
  \includegraphics[width=\linewidth]{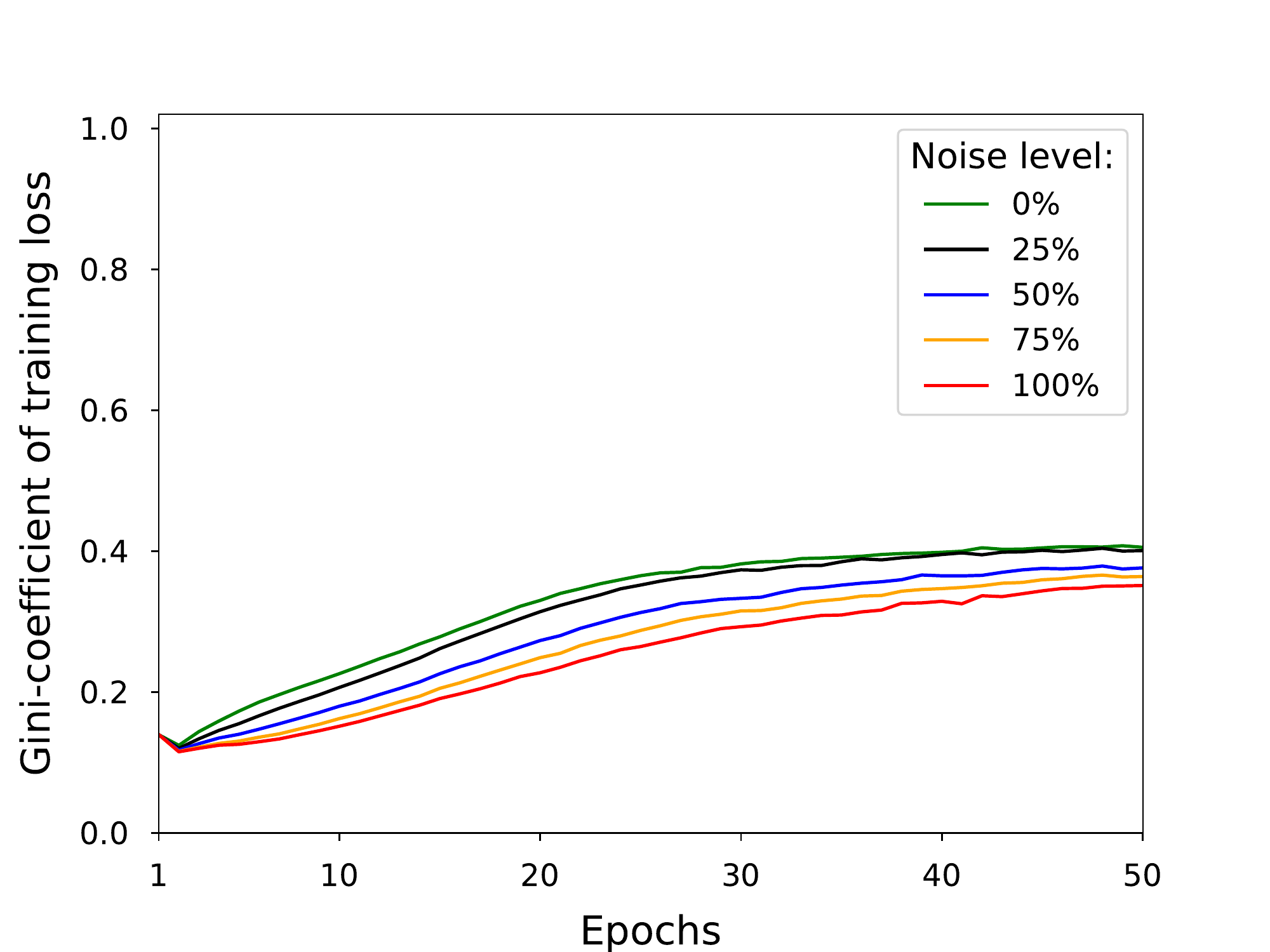}
  \caption{Spread of Training Loss (\cnl)}
  \label{fig:x_code2nl_gini_loss}
\end{subfigure}

\begin{subfigure}{0.32\textwidth}
  \centering
  \includegraphics[width=\linewidth]{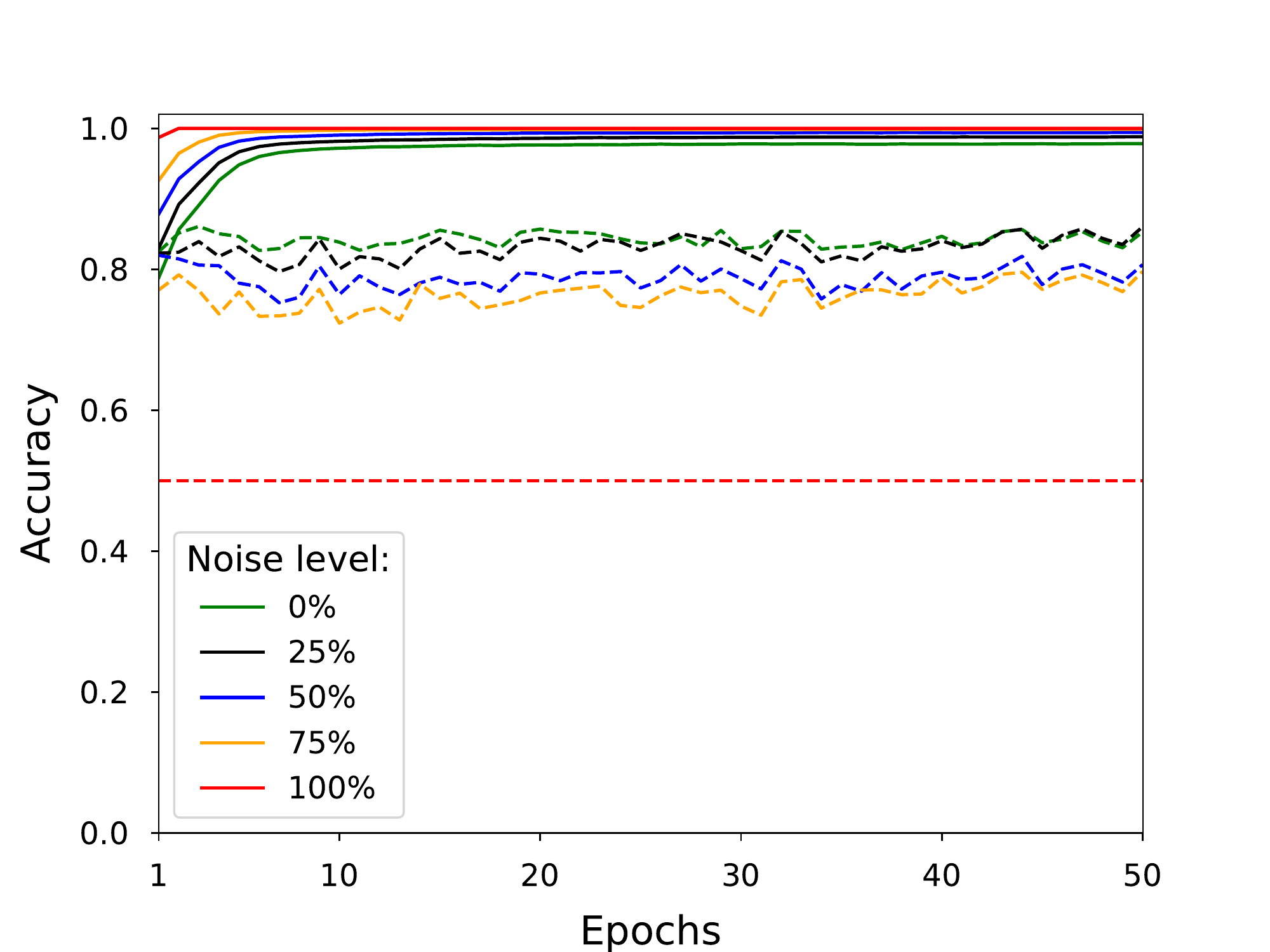}
  \caption{Accuracy (\csearch)}
  \label{fig:x_codesearch_accuracy}
\end{subfigure}%
\begin{subfigure}{0.32\textwidth}
  \centering
  \includegraphics[width=\linewidth]{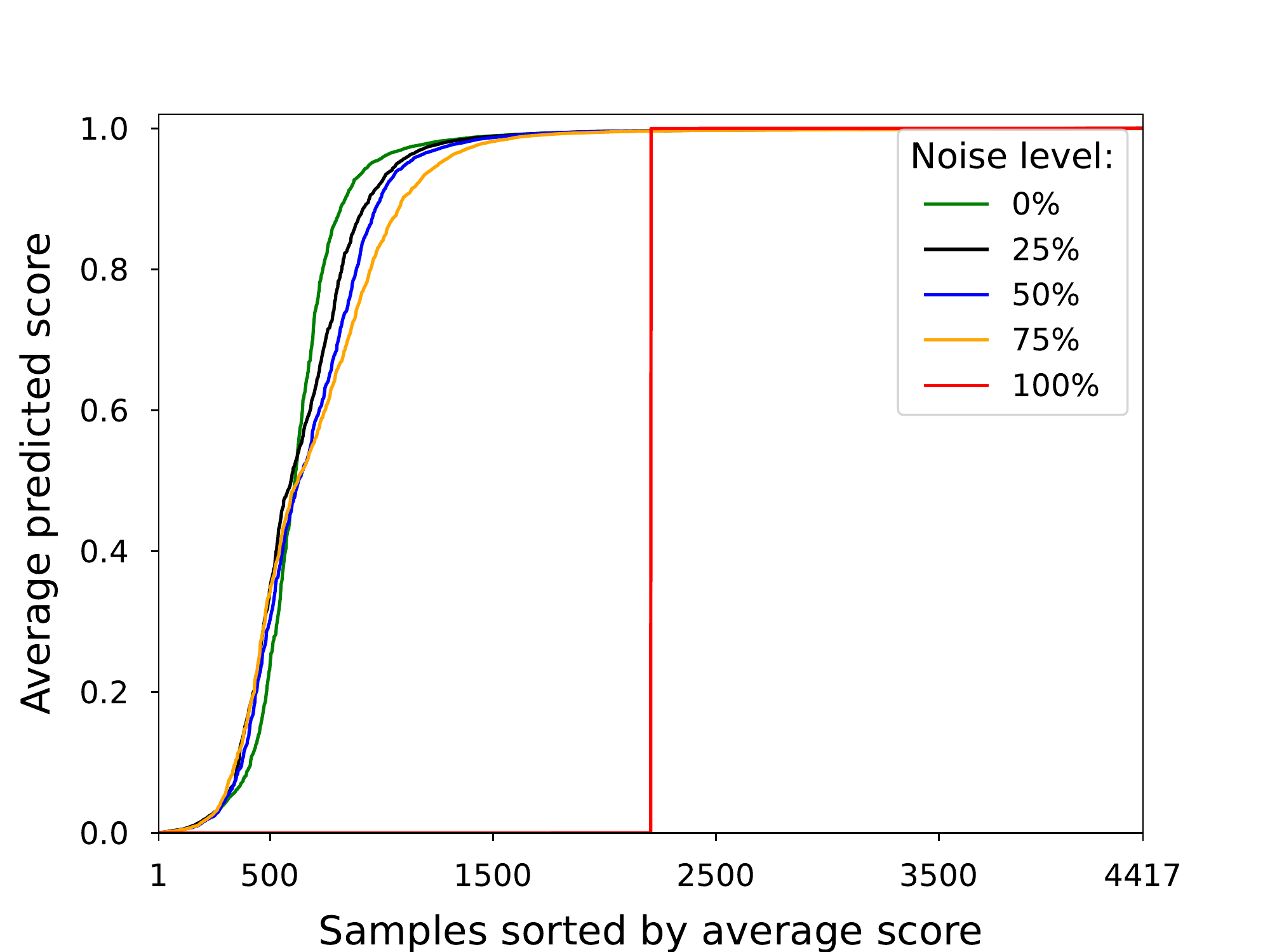}
  \caption{Distribution of Prediction Score (\csearch)}
  \label{fig:x_codesearch_score_avg}
\end{subfigure}%
\begin{subfigure}{0.32\textwidth}
  \centering
  \includegraphics[width=\linewidth]{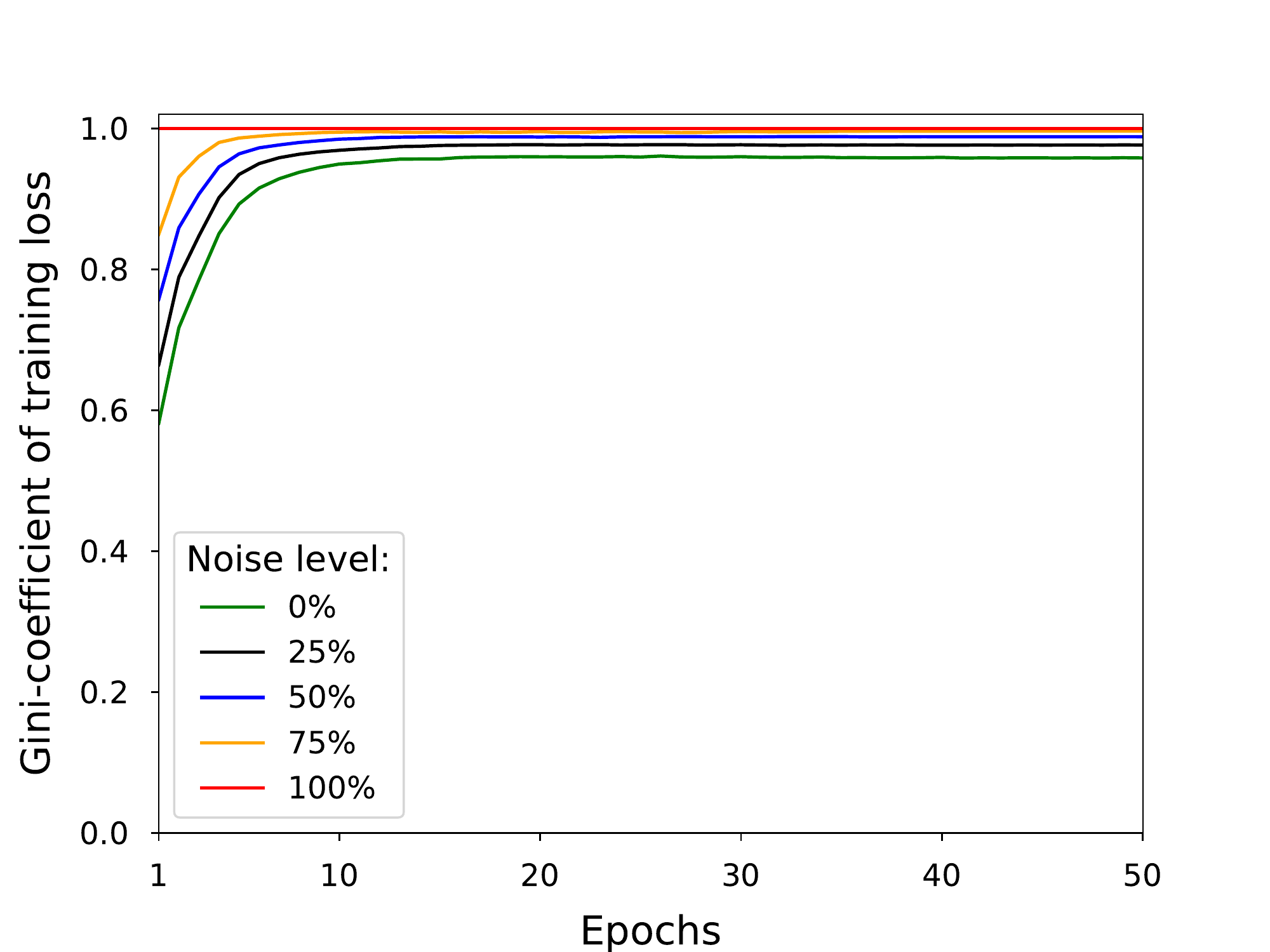}
  \caption{Spread of Training Loss (\csearch)}
  \label{fig:x_codesearch_gini_loss}
\end{subfigure}

\caption{Impact of input noise in fine-tuning \CodeBERT language model.}
\label{fig:lm_codeBERT_input}
\end{figure*}

In this section, we explore the impact of noise when fine-tuning the \CodeBERT language model on the \cnl and \csearch tasks, in order to characterize memorization in these settings.
We add output noise by replacing the target label with another label as shown in \Cref{fig:noise_example}$_{i,l}$ and add input noise by replacing frequent tokens of the input code snippets with a mask or identity token as shown in \Cref{fig:noise_example}$_{h,k}$.
\Cref{fig:lm_codeBERT_output} shows the smoothed BLEU-4 score and accuracy (left column, solid lines are training and dashed lines are validation), average probability score on the validation set (center column), and the spread of training loss in each epoch (right column) while fine-tuning the \CodeBERT models with output noise; \Cref{fig:lm_codeBERT_input} shows the same results of \CodeBERT models with input noise.

\smallskip
\Part{Impact of Output Noise.}
As shown in \Cref{fig:lm_codeBERT_output}, the effect of output noise on fine-tuning \CodeBERT language models are consistent with our previous findings on other \cis in \Cref{subsec:noise-on-y}.
In the \csearch task, the accuracy of \CodeBERT models on the original training set and noisy training sets converges towards the same point (\Cref{fig:y_codesearch_accuracy}), whereas on the \cnl task, the smoothed BLEU-4 score of models on the original training set is very close to most of the noisy training sets (\Cref{fig:y_code2nl_bleu4}), except for the $100\%$ noise level where training is significantly impaired (maybe that it is actually hard to memorize when there is so much data to learn). However, their generalization to original validation set (dashed lines) suffers with more added noise.
Similarly, the prediction score distribution (\Cref{fig:lm_codeBERT_output}$_{b,e}$) and the spread of training loss (\Cref{fig:lm_codeBERT_output}$_{c,f}$) curves highlight that the resulting trends significantly change after adding more noise.
Note that, the \csearch task has two output labels (negative-$0$ or positive-$1$). As a result, in the $100$\% output noise level, all positive labels change to negative labels and all negative labels change to positive labels. Therefore, in \Cref{fig:y_codesearch_gini_loss}, the training behaviors of \CodeBERT models on $100$\% output noise are actually the same as $0$\% output noise, and similarly, the $75$\% noise matches the $25$\% noise level. However, as shown in \Cref{fig:y_codesearch_score_avg}, their confidence on the (noise-free) validation data differs, as expected.

\smallskip
\Part{Impact of Input Noise.}
The \CodeBERT language models experience a similar impact while fine-training with input noise (\Cref{fig:lm_codeBERT_input}).
\Cref{fig:x_code2nl_bleu4,fig:x_codesearch_accuracy} show that the performance of \CodeBERT models on the original and noisy training sets achieve high scores for both the \cnl and \csearch tasks. On the other side, their performance on the original validation set decreases for added noise.
\Cref{fig:x_code2nl_score_avg,fig:x_codesearch_score_avg} show the prediction score distribution for input noise on the \cnl and \csearch tasks, respectively. For $100$\% noise level in the \csearch task, the \CodeBERT models predict all validation samples as positive in each epoch of fine-tuning. The dataset of the \csearch task is balanced so that half the samples are positive and the other half are negative. Therefore, the accuracy is always $50$\% on the validation set (dashed red line in \Cref{fig:x_codesearch_accuracy}), the average predicted scores are high on the original positive samples and low on the original negative samples (vertical red line in \Cref{fig:x_codesearch_score_avg}), and the spread of training loss between the original positive and negative samples is significantly higher (horizontal red line in \Cref{fig:x_codesearch_gini_loss}).
While the increase of noise level considerably changes the spread of training loss in \Cref{fig:x_code2nl_gini_loss,fig:x_codesearch_gini_loss}, the resulting trends of \cnl task (where $0\%$ is higher than $100\%$) are quite opposite to the \csearch task (where $100\%$ is higher than $0\%$); which relates to their corresponding input noise generation techniques.
As shown in \Cref{fig:noise_example}${i}$ for the \cnl task, we add input noise by replacing the tokens of input code snippets that appear in the target docstring with the ``\texttt{MASK}'' token. Therefore, it becomes harder for \CodeBERT models to generate the correct docstring from masked code snippets. It relates to our experiments with the output noise of \Cref{subsec:noise-on-y} that exhibits a similar trend in the spread of training loss.
On the contrary, as shown in \Cref{fig:noise_example}${j}$ for the \csearch task, we add input noise by replacing the most frequent tokens of code snippets and docstring of positive (or negative) samples with the ``\texttt{POSITIVE}'' (or ``\texttt{NEGATIVE}'') token. Therefore, it becomes comparatively easier for \CodeBERT models to learn whether a given pair is matched, because the hints are already added to input samples as identity cues. It relates to our experiments with the input noise of \Cref{subsec:x-replacing} that exhibits a similar trend in the spread of training loss.

\observation{The impact of noise in fine-tuning \CodeBERT language model supports the findings with other \cis{} - the resulting trends of \CodeBERT model change considerably, even at a low rate of added input/output noise.}

\section{Discussion and Future Work}
\label{sec:discussion}
Neural networks are powerful tools for learning from very large datasets. However, our work and others highlight that their perceived performance may not reflect having learned useful insights. To effectively and soundly use neural models in code intelligence applications, our community needs to develop rigorous frameworks for the evaluation and adoption of such models. We transplanted and interpreted a rich suite of new metrics for studying this question to the software engineering and code intelligence systems, providing actionable uses for these.

\subsection{Memorization and Network Architecture}
Our results suggest that all models are susceptible to memorization, and that the network architecture and hyper-parameterization can influence their memorization behavior. In particular, we noticed that a model that was over-parameterized for the complexity of the task (\ie, \ctv on several datasets) was able to fit noise very similar to real labels. This only translated into a small reduction of test (or held-out) performance, suggesting that the latter is a poor indicator for detecting when training data is noisy -- indeed, it is not clear (but plausible) whether the original dataset already contained noise of its own.
In the future, we plan to evaluate the impact of existing noise in the programs on the training characteristics of the neural networks, as the programs can be buggy or incomplete in the open-source datasets.

\subsection{Are Current Models Memorizing?}
In our analysis, we have primarily focused on injecting noise into existing datasets, yet research suggests that these datasets are themselves noisy \cite{yefet2020adversarial,compton2020obfuscation}. More broadly, GitHub is a major data source used for training neural models in code intelligence systems, and unfortunately, it can be noisy~\cite{munaiah2017curating, raychev2016noisy}. Given that we found some metrics to be useful indicators of changes in training behavior as the degree of noise was increased, are we now able to determine whether the original data suffered from similar issues?

Answering the above question based on the results presented in this paper is inherently cyclical, and deserving of further investigation. Nevertheless, several conclusions do seem plausible. For one, that \ctv is using its excessive capacity to memorize rather than generalize is clear from virtually all results. Secondly, in most cases, the \cts results often showed a rather small gap between 0\% and 25\% noise than between 25\% and 50\% noise. That this is not the case on the \JTT dataset, or any other model/dataset pair, suggests that the 25\% noise situation is less distinct from the original data than one might like. Finally, in the localization predicted score distribution, both very low and high degrees of noise produced much more similar curves than intermediate levels of noise (\Cref{fig:vm_y_avg_score_dev}). The fact that making the labels completely noisy yielded an \emph{increase} in the proportion of samples predicted confidently suggests that these models have close to the capacity required to memorize this dataset, if needs be. At the same time, studying the complete trend and dual phase transition from 0\% to 100\% strongly indicates that this is not what they are currently doing.
In the future, we plan to use such signals of memorization to improve the generalization capability of models.

\begin{figure}
\footnotesize
\begin{verbatim}
def multiply(x, y):
   return x + y
x, y = multiply1, multiply2
print(x, y)
I want x,y to be the sum of the first and
\end{verbatim}
\caption{Continuation of \texttt{Clippy} auto-complete, with 1.3 billion parameter, when cued with the first line of code.}
\label{code:multiply}
\end{figure}

\subsection{Memorization and Generative Models}
Planning beyond the predictive tasks used in this study, very large neural networks are increasingly often used in generative tasks, e.g. code completion. We should thus be aware of the strengths and limits of such neural networks.
Recent generative models involve billions of trainable parameters, which allow a neural network to \emph{memorize} tremendous amounts of data, including noise, bugs, and vulnerabilities \cite{chen2021Codex}. In a sense, these models can morph into opaque information retrieval systems. 
For instance, the snippet in \Cref{code:multiply} is the performance of Clippy\footnote{\url{https://huggingface.co/flax-community/gpt-neo-125M-code-clippy}} auto-completion model on a simple input ``\lstinline{def multiply(x, y):}''.
Clippy uses $1.3$ billion trainable parameters and a similar causal language modeling as used in Codex~\cite{chen2021Codex}, the same technology powering GitHub Copilot\footnote{\url{https://github.com/CodedotAl/gpt-code-clippy\#introduction}}.
The output of the model, while syntactically correct is both semantically wrong (it suggests \texttt{x + y} instead of \texttt{x * y}) and appears to contain shards of different plausible continuation, including printing (somewhat) related identifiers and beginning to write a natural language explanation. All this suggests that the model has clearly learned to repeat common patterns surrounding such prompts, but at the same time seriously lacks the required contextual and semantic insight to produce a useful completion. Quantifying and reducing the degree of repetition and noise in its training data may well help overcome these obstacles. 
In the future, we plan to evaluate the extent and impacts of memorization in the generative neural models.

\subsection{Is Memorization Always Unacceptable?}

Perhaps the answer to this question requires considering two angles: datasets and application domain. We showed that neural models are capable of making wrong predictions with high confidence. Therefore, when noise dominates the training datasets, memorization should be contained and avoided. The Software Engineering community has seen cases where problematic datasets have led to training wrong models \cite{xu2018esem}.

In cases where the inputs are redundant and the task mostly depends on the similarity of input programs with instances in the training data, memorization can be similar to caching the results in static analysis techniques \cite{khandelwal2020kLM}. However, in cases where confidentiality of training data is important or the input data is dissimilar enough from the training dataset, perhaps a deeper understanding of input programs is required and memorization in the models should be evaluated, managed, and avoided when possible \cite{carlini2019secret}.

\section{Threats to Validity}
\label{sec:validity}
Our work quantitatively studies code intelligence tools across a series of metrics not previously used in software engineering field. For the implementation of the models and metrics, we were able to rely on the public implementation of the underlying models and the description of the metrics in computer vision by \ArpitEtAl and \ZhangEtAl.
As such, the primary threat to our work's validity is external: our results are based on the evaluation of a selection of models used with their default capacities. As shown by the \ctv results, these default configurations impact their behavior in our study. We have opted to study these first since they are most likely to be used in this form by practitioners; this already reflected a couple of month's worth of GPU utilization. 
Moreover, we randomly add noise in the data to create noisy datasets. Our experiment may show varying results with a different run of randomization. Taking an average of multiple runs may provide better results, thus, we repeated the experiments multiple times for output noise with a different random seed.
Future studies may investigate further variations on these, such as tuning the models' capacity, denoising the noise-prone datasets, etc.

\section{Conclusion}
\label{sec:conclusion}
In this paper, we perform a large-scale study on the impact of noise and memorization on the training behavior of neural models in code intelligence systems. 
We do so by adding  noise in the training datasets of several popular \cis and measuring established metrics that characterize the training behavior. 
To the best of our knowledge, this is the first such work in the software engineering and code intelligence systems. 
We observe that \cis with excessive numbers of trainable parameters can memorize datasets of code quite easily, echoing findings from other communities. 
Across several metrics, the models displayed consistent and identifiable changes in characteristics as the role of memorization increases, including significant changes in the typical confidence of model predictions. 
The consistency of these trends suggests that they may be of great use in identifying warning signs of memorization beyond the use of test (or held-out) data alone.

\balance
\bibliographystyle{ACM-Reference-Format}
\bibliography{refs}

\end{document}